\def\isarxiv{1} 

\ifdefined\isarxiv
\documentclass[11pt]{article} 
\else

\documentclass{article}
\usepackage{hyperref}
\usepackage[accepted]{icml2023}

\fi

\usepackage{tabu}
\usepackage{array}
\usepackage{amsmath}
\usepackage{amsthm}
\usepackage{amssymb}

\usepackage{algorithm}
\usepackage{subfig}
\usepackage{dirtytalk}
\usepackage{color}
\usepackage[english]{babel}
\usepackage{graphicx}
\usepackage{grffile}
\usepackage{wrapfig,epsfig}
\usepackage{epstopdf}
\usepackage{enumitem}
\usepackage{url}
\usepackage{color}
\usepackage{epstopdf}
\usepackage{algpseudocode}
\usepackage[T1]{fontenc}
\usepackage{bbm}
\usepackage{comment}
\usepackage{dsfont}
\usepackage{mathtools}
\usepackage{enumitem}

\usepackage{tikz}

\ifdefined\isarxiv 
\usepackage{hyperref}
\hypersetup{colorlinks=true,citecolor=red,linkcolor=blue}
\else

\usepackage{microtype}
\fi 
\usetikzlibrary{arrows}
\ifdefined\isarxiv
\usepackage[margin=1in]{geometry} 
\else

\fi

\graphicspath{{./figs/}}

\usepackage[nameinlink,capitalize]{cleveref}
\usepackage{textcomp}
\usepackage{multirow}
\newtheorem{theorem}{Theorem}[section]
\newtheorem{lemma}[theorem]{Lemma}
\newtheorem{definition}[theorem]{Definition}

\newtheorem{proposition}[theorem]{Proposition}
\newtheorem{corollary}[theorem]{Corollary}

\newtheorem{assumption}[theorem]{Assumption}

\newtheorem{fact}[theorem]{Fact}
\newtheorem{remark}[theorem]{Remark}

\newcommand{\wh}{\widehat}
\newcommand{\wt}{\widetilde}
\newcommand{\ov}{\overline}

\newcommand{\N}{\mathcal{N}}

\newcommand{\R}{\mathbb{R}}

\renewcommand{\d}{\mathrm{d}}

\renewcommand{\tilde}{\wt}

\renewcommand{\d}{\mathrm{d}}
\newcommand{\ip}[2]{\langle {#1} , {#2} \rangle}

\DeclareMathOperator*{\E}{{\mathbb{E}}}

\DeclareMathOperator{\poly}{poly}

\DeclareMathOperator{\rank}{rank}

\setlist[itemize]{noitemsep, nolistsep}

\newcommand{\enc}{\mathsf{sk}}
\newcommand{\dec}{\mathsf{desk}}
\newcommand{\lo}{\mathrm{local}}
\newcommand{\gl}{\mathrm{global}}
\newcommand{\sketch}{\mathrm{sketch}}

\definecolor{mygreen}{RGB}{80,180,0}
\definecolor{b2}{RGB}{51,153,255}

\newcommand{\nc}{\newcommand}

\algblock{ParFor}{EndParFor}
\algnewcommand\algorithmicparfor{\textbf{parfor}}
\algnewcommand\algorithmicpardo{\textbf{do}}
\algnewcommand\algorithmicendparfor{\textbf{end\ parfor}}
\algrenewtext{ParFor}[1]{\algorithmicparfor\ #1\ \algorithmicpardo}
\algrenewtext{EndParFor}{\algorithmicendparfor}

\nc{\Ra}{\Rightarrow}
\nc{\zo}{\{0,1\}}	

\allowdisplaybreaks

\ifdefined\isarxiv
\else 
\icmltitlerunning{Sketching for First Order Method: Efficient Algorithm for Low-Bandwidth Channel and Vulnerability}
\fi

\begin{document}

\ifdefined\isarxiv
\title{Sketching for First Order Method: \\ Efficient Algorithm for Low-Bandwidth Channel and Vulnerability\thanks{A preliminary version of this paper appeared at ICML 2023.}}

\author{
Zhao Song\thanks{\texttt{zsong@adobe.com}. Adobe Research.} 
\and 
Yitan Wang\thanks{\texttt{yitan.wang@yale.edu}. Yale University. Supported by ONR Award N00014-20-1-2335.}
\and 
Zheng Yu\thanks{\texttt{yz388620@alibaba-inc.com}. Alibaba Inc.}
\and 
Lichen Zhang\thanks{\texttt{lichenz@mit.edu}. MIT. Supported by NSF grant No. CCF-1955217 and NSF grant No. DMS-2022448.}
}
\date{} 
\else

\twocolumn[

\icmltitle{Sketching for First Order Method: Efficient Algorithm for Low-Bandwidth Channel and Vulnerability}


\icmlsetsymbol{equal}{*}

\begin{icmlauthorlist}
\icmlauthor{Zhao Song}{adobe}
\icmlauthor{Yitan Wang}{yale}
\icmlauthor{Zheng Yu}{alibaba}
\icmlauthor{Lichen Zhang}{mit}
\end{icmlauthorlist}

\icmlaffiliation{adobe}{Adobe Research}
\icmlaffiliation{yale}{Yale University}
\icmlaffiliation{alibaba}{Alibaba Inc.}
\icmlaffiliation{mit}{MIT}

\icmlcorrespondingauthor{Zhao Song}{zsong@adobe.com}
\icmlcorrespondingauthor{Lichen Zhang}{lichenz@mit.edu}

\icmlkeywords{Machine Learning, ICML}

\vskip 0.3in
]

\printAffiliationsAndNotice{}
\fi

\ifdefined\isarxiv
\begin{titlepage}
  \maketitle
  \begin{abstract}
  Sketching is one of the most fundamental tools in large-scale machine learning. It enables runtime and memory saving via randomly compressing the original large problem into lower dimensions. In this paper, we propose a novel sketching scheme for the first order method in large-scale distributed learning setting, such that the communication cost per round between distributed agents is reduced, our bound on the total communication is no larger than for uncompressed distributed gradient descent, and the convergence of the algorithms is still guaranteed. Given gradient information in a high dimension $d$, the agent passes the compressed information processed by a sketching matrix $R\in \R^{s\times d}$ with $s\ll d$, and the receiver de-compressed via the de-sketching matrix $R^\top$ to ``recover'' the information in original dimension. Using such a framework, we develop algorithms for federated learning with lower per-round communication cost. However, such random sketching alone gives no formal privacy guarantee for the local data. We present a gradient-matching attack and prove that, under smoothness and non-degeneracy assumptions on the model, gradient descent on the data finds an input whose sketched gradient matches the observed one, at a linear rate. This input need not be the data, but an attacker that follows a client over several rounds with fresh sketches recovers the data with the same attack, under smoothness and non-degeneracy conditions, and no attack does so from the first $T<m/s$ rounds, where $m$ is the data dimension. As a remedy, we prove rigorously that the algorithm will be differentially private by adding additional random noises in gradient information. This noisy variant still sends only $s$ numbers per client in each round, and we prove that, for convex and strongly convex objectives, it converges at the rates of the noise-free algorithm, up to an additive error term that depends on the privacy level and, in our bounds, grows as the sketch size $s$ decreases. Our total communication guarantees are proved for the noise-free algorithm. Our sketching scheme can be further generalized to other learning settings and might be of independent interest itself.

  \end{abstract}
  \thispagestyle{empty}
  
\end{titlepage}

{\hypersetup{linkcolor=black}
\tableofcontents
}
\newpage
\else
  \begin{abstract}
  
  \end{abstract}
\fi

\section{Introduction}
Federated learning enables multiple parties to collaboratively train a machine learning model without directly exchanging training data. This has become particularly important in areas of artificial intelligence where users care about data privacy, security, and access rights, including healthcare~\cite{lgd+20,lmx+19}, internet of things~\cite{cys+20}, and fraud detection~\cite{zygw20}.

Given the importance and popularity of federated learning, two central aspects of this subject have been particularly studied: privacy and communication cost. The fundamental purpose of federated learning is to protect the data privacy of clients by only communicating the gradient information of a user. Unfortunately, recent studies~\cite{gbdm20,zlh19,wsz+19} have demonstrated that attackers can recover the input data from the communicated gradients. The reason why these attacks work is the gradients carry important information about the training data~\cite{ams+15,fjr15}. A very recent work~\cite{wll22} demonstrates that via computationally intense approach based on tensor decomposition, one can recover the training data from a single gradient and model parameters for over-parametrized networks.

Communication efficiency is also one of the core concerns. In a typical federated learning setting, the model is trained through gathering individual information from many clients who operate under a low bandwidth network. On the other hand, the size of the gradient is usually large due to the sheer parameter count of many modern machine learning models. This becomes even more problematic when conducting federated learning on mobile and edge devices, where the bandwidth of the network is further limited. Many works try to address this challenge through local optimization methods, such as local gradient descent (GD), local stochastic gradient descent (SGD)~\cite{kmyrsb16,mmr+17,stich2018local} and using classic data structures in streaming to compress the gradient~\cite{rpu20}. Despite of significant efforts on improving the communication cost of federated learning framework, none of these approaches by itself comes with a formal privacy guarantee, and as we will show, a gradient-matching attack still succeeds when the communicated gradients are sketched.

The above two concerns allude us to ask the following question:
\begin{center}
    {\it Is there an FL framework that protects the local privacy and has good performance even in low-bandwidth networks?}
\end{center}

In this paper, we address these goals by using tools from randomized linear algebra --- the linear sketches. Sketching matrices describe a distribution of random matrices $R:\R^{d}\rightarrow \R^{b_{\text{sketch}}}$ where $b_{\text{sketch}}\ll d$ and for vectors $x\in \R^d$ one has $\|Rx\|_2=(1\pm\epsilon)\|x\|_2$. While these random projections effectively reduce the dimension of the gradient, we still need to ``recover'' them to the original dimension for training purpose. To realize this goal, we apply the de-sketch matrix, which is essentially the transpose of $R$ as a decoder. Instead of running the gradient descent $w^{(t+1)}\leftarrow w^{(t)} - \eta\cdot g^{(t)}$ using true gradient $g^{(t)}\in\R^{d}$, we apply sketch and de-sketch to the gradient:
\begin{align*}
    w^{(t+1)}\leftarrow & ~ w^{(t)}-\eta\cdot R^\top \cdot R \cdot g^{(t)}.
\end{align*}
Here $R\in\R^{b_\text{sketch}\times d}$ denotes a sketching matrix that sketches the true gradient to a lower dimension and $R^\top\in\R^{d\times b_\text{sketch}}$ denotes the de-sketching process that maps the sketched gradient back to the true gradient dimension. To ensure that the gradient descent still has good convergence behavior under the linear map $x\mapsto R^\top R x$, we argue that it is enough for $R$ to satisfy the coordinate-wise embedding property~\cite{sy21}. This property states that $R^\top R g^{(t)}$ is an unbiased estimator of $g^{(t)}$ and has small second moment, and many of the popular sketching matrices satisfy this property. Hence, all clients will only communicate sketched gradients to the server, the server averages the sketched gradients and broadcasts them back to all clients. Finally, each client de-sketches the received gradients and performs local updates. Since the sketching dimension is always small compared to the original dimension, we save communication costs per iteration via sketching.

While the algorithm with sketch-and-de-sketch might seem simple and elegant, it is not enough to address the privacy challenge of federated learning. At the first glance, the sketching ``masks'' the communicated gradients, but it gives no formal privacy guarantee, and a malicious attacker can still run a gradient-matching attack on the sketched gradients. Specifically, we propose a highly-efficient attack algorithm such that the attacker only needs to observe the sketched gradient being communicated, the sketching matrix being used and the model parameters. Then, by running gradient descent on the data instead of the model parameters, the attacker finds an input whose sketched gradient matches the observed one, at a linear rate. This input need not be the private local data, since under the assumptions of our attack analysis, with $b_{\text{sketch}}<d$, a single sketched gradient does not determine the data (Remark~\ref{rem:sketch_non_identifiable}). Observing several rounds changes this. Algorithm~\ref{alg:fed_learn_kstep_dp} draws a fresh sketching matrix in every round while the local data stay the same. So an attacker that follows one client stacks its sketched gradients from $T$ rounds and runs the same attack on the sum of the per-round losses. It recovers the data at a linear rate when the model gradient and its Jacobian in the data are Lipschitz at the observed weights, the attacker keeps $m$ of the stacked coordinates, and the resulting square sketched Jacobian is uniformly non-singular (Theorem~\ref{thm:multi_square}). For models whose gradient is affine in a transform of the data, it can use all coordinates, provided the stacked sketched matrix has rank $m$ (Theorem~\ref{thm:multi_flat}). For such models and subsampled randomized Hadamard transform (SRHT) or uniform-sampling sketches, $Tb_{\rm sketch}=O(m\log(m/\delta))$ suffices with probability $1-\delta$ when $d\le m$ and, for every outcome of the sketches, the observed weights keep the matrices that map the transformed data to the gradients jointly as well conditioned over the observed rounds as their norms allow, up to constant factors (for $d<m$ no single round does this, so the weights must change across these rounds), by Proposition~\ref{prop:multi_random}, while no attack recovers the data from the first $T$ rounds when $Tb_{\rm sketch}<m$ (Proposition~\ref{prop:multi_necessary}). Sketching changes our iteration bound for the attack by at most a bounded factor: to reduce the respective loss by the same factor, our iteration bound for the attack on sketched gradients is at most $(\gamma_2/\gamma_1)^2$ times our bound for the attack without sketching, before rounding up to an integer, where $[\gamma_1,\gamma_2]$ contains the singular values of $R^\top$ (Remark~\ref{rem:attack_iteration_counts}). To the best of our knowledge, this is the first theoretical analysis of the convergence of the gradient-matching attack using simple and standard first-order methods that are widely-observed in practice~\cite{gbdm20,zlh19}. Moreover, compare to the tensor decomposition-based algorithm of~\cite{wll22}, our algorithm is much more computationally efficient and applies, from one round, to under-parametrized models, whose parameter dimension $d$ is at most the data dimension $m$, that meet the assumptions of Section~\ref{sec:attack_sketched_gradients}, and, from several rounds, to any model that meets the assumptions of Section~\ref{sec:attack_multi_round}, while~\cite{wll22} studies over-parametrized networks. On the other hand, the~\cite{wll22} algorithm produces stronger guarantees than ours, recovers the data from a single gradient and works for noisy gradients. Our attack and analysis apply not only to our sketching-based framework, but to any approach that communicates a linear sketch of the gradient with a full-row-rank sketching matrix known to the attacker, such as approaches building upon randomized data structures~\cite{rpu20}.

To circumvent this issue, we add Gaussian noise to every local stochastic gradient, so that the released messages are differentially private~\cite{dkmmn06}. Sketching alone gives no formal privacy guarantee, and gradient matching still works against sketched gradients. Differential privacy, in contrast, bounds what any attack can learn about one data point.

We summarize the contributions in this work as follows:

{\bf Our contributions:} We present our main technical contributions as follows:
    \begin{itemize}
        \item We introduce the sketch-and-de-sketch framework. Unlike the classical sketch-and-solve paradigm, our iterative sketch and de-sketch method can be combined with gradient-based methods and extended to broader optimization problems.
        
        \item We apply our sketch-and-de-sketch method to federated learning, obtaining an algorithm that only needs to communicate lower-dimensional vector, which is particularly useful in low-bandwidth networks.
        
        \item By adding Gaussian noise to every local gradient step, we obtain a variant of our algorithm that is differentially private and still sends only $b_\text{sketch}$ numbers per client in each round. For convex and strongly convex objectives, we prove that this variant converges at the rates of the noise-free algorithm, up to an additive error term that depends on the privacy level, grows as $b_\text{sketch}$ decreases (for $b_\text{sketch}\ll d$, our bounds on it scale like $d/\sqrt{b_\text{sketch}}$ for convex objectives and, up to a logarithmic factor, like $d^2/b_\text{sketch}$ for strongly convex ones), and does not decrease with the number of rounds. Our total communication guarantees are proved for the noise-free algorithm.
        
        \item We present a gradient-matching attack: from the sketched gradient $Rg$, the sketch $R$ and the weights $w$, gradient descent on the data finds an input whose sketched gradient matches $Rg$, at a linear rate under smoothness and non-degeneracy assumptions on the model. Under these assumptions and with $b_{\text{sketch}}<d$, a single sketched gradient does not determine the data, so this input need not be the local data. From several rounds with fresh sketches, the same attack recovers the data itself, under smoothness of the model and a non-degeneracy condition on the stacked sketched Jacobians. For SRHT or uniform-sampling sketches and models whose gradient is affine in a transform of the data, with $d\le m$ and observed weights that, for every outcome of the sketches, keep the matrices that map the transformed data to the gradients jointly as well conditioned over the observed rounds as their norms allow, up to constant factors (for $d<m$ the weights must change across them), $O((m/b_{\rm sketch})\log(m/\delta))$ observed rounds suffice with probability at least $1-\delta$. This is within a factor $O(\log(m/\delta))$ of the $m/b_{\rm sketch}$ rounds that any attack observing the first rounds needs, and no attack recovers the data from the first $T<m/b_{\rm sketch}$ rounds.
    \end{itemize}

\paragraph{Roadmap.} In section~\ref{sec:related}, we discuss related work and define common notations. 
In section~\ref{sec:problem_setup}, we describe the problem setting and assumptions. 
In section~\ref{sec:alg}, we present a federated learning framework with communication efficiency by leveraging sketching techniques. 
In section~\ref{sec:conv_main}, we analyze the convergence property of our proposed framework for smooth and convex objectives.
In section~\ref{sec:dp_main}, we discuss the privacy guarantee of our framework and the convergence of its private variant.
In section~\ref{sec:attack_sketched_gradients}, we discuss the feasibility of the gradient attacking when the framework shares sketched gradient information.
In section~\ref{sec:discuss}, we conclude the contribution and limitations of this paper.

\section{Related Work}
\label{sec:related}

\paragraph{Federated Learning.}
Federated learning (FL) is an emerging framework in distributed deep learning. FL allows multiple parties or clients collaboratively train a model without data sharing. In this learning paradigm, local clients perform most of the computation and a central sever update the model parameters through aggregation then transfers the parameters to local models~\cite{dcm+12, ss15,mmr+17}. In this way, the details of the data are not disclosed in between each party. 
Unlike the standard parallel setting, FL has three unique challenge~\cite{lsts20}, including communication cost, data heterogeneity and client robustness. In our work, we focus on the first two challenges. The training data are massively distributed over an incredibly large number of devices, and
the connection between the central server and a device is slow. A direct consequence is the slow communication, which motivated communication-efficient FL algorithm. Federated average (FedAvg)~\cite{mmr+17} firstly addressed the communication efficiency problem by introducing a global model to aggregate local stochastic gradient descent updates. Later, different variations and adaptations have arisen. This encompasses a myriad of possible approaches, including developing better optimization algorithms~\cite{wys+20}, generalizing model to heterogeneous clients under special assumptions~\cite{zls+18,kma+19,ljz+21} and utilizing succinct and randomized data structures~\cite{rpu20}. The work of \cite{lsy23} provides a provable guarantee federated learning algorithm for adversarial deep neural networks training.

\paragraph{Sketching.}

Sketching is a fundamental tool in many numerical linear algebra tasks, such as linear regression, low-rank approximation \cite{cw13,nn13,mm13,bw14,swz17,alszz18,makarychev2020improved}, distributed problems \cite{wz16,bwz16}, reinforcement learning \cite{wzd+20,ssx21}, tensor decomposition \cite{swz19_soda}, clustering \cite{emz21,dswy22}, convex programming \cite{lsz19,jswz21,sy21,jlsw20,qszz23}, gradient-based algorithm~\cite{xss21}, online optimization problems~\cite{rrs+21}, training neural networks~\cite{xzz18,bpsw21,syz21,szz21,gqsw22}, submodular maximization \cite{qsw23}, matrix sensing \cite{qsz23_sensing}, relational database \cite{qjs+22}, dynamic kernel estimation \cite{qrs+22}, and Kronecker product regression~\cite{rsz22}.

\paragraph{Gradient Leakage Attack.} A number of works \cite{zlh19,yin2021see,wll+20,rg20} have pointed out that the private information of local training data can be attacked using only the exchanged gradient information. Given the gradient of the neural network model with respect to the weights for a specific data, their method starts with a random generated dummy data and label, and its corresponding dummy gradients. By minimizing the difference between the true gradient and the dummy gradients using gradient descent, they show empirically that the dummy data and label will reveal the true data completely. The follow-up work \cite{zmb20} further discuss the case of classification task with cross-entropy loss, and observe that the true label can be recovered exactly. Therefore, they only need to minimize over the dummy data and have better empirical performance. Other attack methods include but not limited to membership inference and property inference
attacks \cite{ssss17,msds19}, training generative adversarial network (GAN) models \cite{hap17,gpmxw14} and other learning-based methods \cite{mss16,pmjfcs16}. Very recently,~\cite{wll22} uses tensor decomposition for gradient leakage attack on over-parametrized networks with provable guarantees. However, the tensor decomposition algorithm is inherently inefficient and their analysis is restricted to over-parametrized networks. 
\paragraph{Notations.}

For a positive integer $n$, we use $[n]$ to denote the set $\{1,2,\cdots, n\}$. 
We use $\E[\cdot]$ to denote expectation (if it exists), and use $\Pr[\cdot]$ to denote probability. For a vector $x$,  we use $\| x \|_2: = (\sum_{i=1}^n x_i^2)^{1/2}$ or $\|x\|$ to denote its $\ell_2$ norm. We denote $1_{\{x=l \}}$ for $l\in\R$ to be the indicator function which equals to 1 if $x=l$ and 0 otherwise. Let $f:A\rightarrow B$ and $g:C\rightarrow A$ be two functions, we use $f\circ g$ to denote the composition of functions $f$ and $g$, i.e., for any $x\in C$, $(f\circ g)(x)=f(g(x))$. We denote $I_{d}$ to be the identity mapping.

\section{Problem Setup}
\label{sec:problem_setup}

Consider a federated learning scenario with $N$ clients and corresponding local losses $f_c:\R^d\rightarrow \R$, our goal is to find
\begin{align}\label{eq:f_def}
    \min_{w\in\R^d} f(w) := \frac{1}{N}\sum_{c=1}^N f_c(w)
\end{align}

For the sake of discussion, we will be focusing on the classical convex and smooth setting for the objective function. Our paradigm will extends to non-convex objectives and we defer details to appendix~\ref{sec:app_kstep_non}.

\begin{assumption}\label{ass:f_ass}
Assume that the set of minimizers of~\eqref{eq:f_def} is nonempty. Each $f_c$ is $\mu$-strongly convex for $\mu\geq 0$ and $L$-smooth. That is, for all $x,y\in\R^d$,
\begin{align*}
    \frac{\mu}{2}\|y-x\|_2^2 
    \leq & ~ f_c(y) - f_c(x)-\ip{y-x}{\nabla f_c(x)} \\
    \leq & ~ \frac{L}{2}\|y-x\|_2^2.
\end{align*}
Note in the case $\mu=0$, this assumption reduces back to convexity and smoothness.
\end{assumption}
In addition to the above assumption, we allow local losses to have arbitrary heterogeneity. In other words, we allow $f_c$'s to vary between different clients. We measure this heterogeneity by $\sigma^2:=\frac{1}{N}\sum_{c=1}^N\|\nabla f_c(w^*)\|_2^2$, where $w^*$ is any minimizer of~\eqref{eq:f_def}. Under Assumption~\ref{ass:f_ass}, $\nabla f_c(w^*)$ does not depend on the choice of $w^*$.

Our results also contain an attack algorithm, which finds an input whose sketched gradient matches the communicated one by only inspecting the communicated sketched gradient, the sketching matrix and the model parameters. From several rounds with fresh sketching matrices, the same attack recovers the local data under the conditions of section~\ref{sec:attack_multi_round}. We defer those discussions to section~\ref{sec:attack_sketched_gradients}.

\section{Our Algorithm}
\label{sec:alg}
In this section, we propose a federated learning framework that addresses the communication efficiency issue. When the learning gradients are of high dimension, classical federated learning framework that communicates the exact gradient could incur a heavy communication cost per round. Sketching technique, which emerges as an effective way to reduce the dimension of vector while preserving significant amount of information~\cite{s06,w14}, is highly preferred in this setting. It enables us to compress the gradient vector into a lower dimension while preserving convergence rates, and greatly saves the communication cost per round.

\begin{algorithm}[!ht]\caption{Iterative sketching-based federated learning Algorithm with $K$ local steps}
\label{alg:fed_learn_kstep_dp}
\begin{algorithmic}[1]
\Procedure{\textsc{IterativeSketchingFL}}{}
\State{Each client initializes $w^0$ with the same seed}
\For{$t=1 \to T$} \Comment{$T$ denotes the total number of global steps}
\State{\color{blue} /* Client */}
\ParFor{$c=1\to N$} \Comment{$N$ denotes the total number of clients}
\If{$t=1$}
\State{$u_c^{t,0}\gets w^0$}
\Else
\State{$u_c^{t,0}\gets w^{t-1}+\dec_t(\Delta \wt{w}^{t-1})$} \Comment{$\dec_t: \R^{b_\text{sketch}}\rightarrow \R^d$ de-sketch the change}
\EndIf
\State {$w^t\gets u_c^{t,0}$}
\For{$k=1\to K$}
\State $u_c^{t,k}\gets u_c^{t,k-1}-\eta_{\mathrm{local}}\cdot \nabla f_c(u_c^{t,k-1})$
\EndFor
\State {$\Delta w_c(t)\gets u_c^{t,K}-w^t$} 
\State {Client $c$ sends $\enc_{t+1}(\Delta w_c(t))$ to server}\Comment{$\enc_{t+1}:\R^d\rightarrow \R^{b_\text{sketch}}$ sketch the change}
\EndParFor
\State {\color{blue} /* Server */}
\State{$\Delta\wt{w}^t\gets \eta_{\mathrm{global}}\cdot \frac{1}{N}\sum_{c=1}^N \enc_{t+1}(\Delta w_c(t))$} \Comment{$\Delta \wt{w}^t\in \R^{b_\text{sketch}}$}
\State {Server sends $\Delta\wt{w}^t$ to each client}
\EndFor
\EndProcedure
\end{algorithmic}
\end{algorithm}

Motivated by above discussion, we propose the iterative sketching-based federated learning algorithm, which builds upon vanilla local gradient descent: we start with a predetermined sequence of independent sketching matrices shared across all clients. In each round, local clients accumulate and sketch its change over $K$ local steps, then transmit the low-dimensional sketch to the server. Server then averages the sketches and transmits them back to all clients. Upon receiving, each client de-sketches to update the local model.

We highlight several distinct features of our algorithm:

\begin{itemize}
    \item {\bf Communication:} In each sync step, we only communicates a low-dimensional sketched gradients, indicating a smaller communication cost per round. This property is particularly valuable in a small-bandwidth setting.
    \item {\bf De-sketch:} We emphasize that unlike the classical sketch-and-solve paradigm that decreases the problem dimension, our algorithm applies sketching in each round, combined with a de-sketching process which recovers back to the true gradient dimension.
    \item {\bf Simpler server task:} Server only needs to do simple averaging, indicating no need of a trustworthy party as the server.
    \item {\bf Decentralization:} Our algorithm can be generalized to decentralized learning settings, where local clients can only communicate with neighboring nodes. In this case, it requires $O(\mathrm{diam})$ rounds to propagate the sketched local changes, where $\mathrm{diam}$ is the diameter of the network graph.
    \item {\bf Linearity:} Compared to the framework of~\cite{rpu20}, our de-sketching operator is linear, this adds flexibility to the analysis and further extensions to the framework.
\end{itemize}

\subsection{\texorpdfstring{$\enc/\dec$}{} via Coordinate-wise Embedding}\label{subsec:enc_dec_sketch}
In this section, we discuss the concrete realization of the $\enc_t/\dec_t$ operators in Algorithm~\ref{alg:fed_learn_kstep_dp} through random sketching matrices. Note we should require any processed gradient $\dec_t\circ \enc_t(g)$ to ``be close'' to the true gradient $g$ to avoid breaking the convergence property of the algorithm. To achieve this, we first introduce the following property for a broad family of sketching matrices, namely the \emph{coordinate-wise embedding}~\cite{sy21}, that naturally connects with $\enc_t/\dec_t$ operators.
\begin{definition}[$a$-coordinate-wise embedding]\label{def:cord_emb}
We say a randomized matrix $R \in \R^{b_\text{sketch} \times d}$ satisfying $a$-coordinate wise embedding if for any vector $g,h\in \R^d$, we have 
\begin{itemize}
    \item $\E_{ R \sim \Pi } [ h^\top R^\top R g ] = h^\top g$;
    \item $\E_{ R \sim \Pi } [ ( h^\top R^\top R g )^2 ] \leq (h^\top g)^2 + \frac{a}{b_\mathrm{sketch}} \|h\|_2^2\cdot \| g \|_2^2$.
\end{itemize}
\end{definition}

In general, well-known sketching matrices have their coordinate-wise embedding parameter $a$ being a small constant (See appendix~\ref{sec:sketching}). Note that if we choose $h$ to be one-hot vector $e_i$, then the above conditions translate to 
\begin{align*} 
\E_{R\sim \Pi} [R^\top R g] = g
\end{align*} 
and
\begin{align*} 
\E_{R\sim \Pi} [\|R^\top Rg\|_2^2] \leq (1+a\cdot \frac{d}{b_\mathrm{sketch}})\cdot \|g\|_2^2.
\end{align*}

This implies that by choosing 
\begin{align}\label{eq:define_sk_desk}
    \enc_t=  & ~ R_t\in\R^{b_\text{sketch}\times d}~(\text{sketching}), \notag\\
      \dec_t=  & ~ R_t^\top\in\R^{d\times b_\text{sketch}}~(\text{de-sketching})
\end{align}
for any iteration $t \geq 1$, where $R_t$'s are independent random matrices with sketching dimension $b_\text{sketch}$, we obtain an unbiased sketching/de-sketching scheme with bounded variance as state in the following Theorem~\ref{thm:enc_dec}. 
\begin{theorem}
\label{thm:enc_dec}
Let $\enc_t$ and $\dec_t$ be defined by Eq.~\eqref{eq:define_sk_desk} using a sequence of independent sketching matrices $R_t\in\R^{b_\text{sketch}\times d}$ satisfying $a$-coordinate wise embedding property~(Definition~\ref{def:cord_emb}). Then the following properties hold:
\begin{enumerate}
    \item Independence: Operators $(\enc_t, \dec_t)$'s are independent over different each iterations.
    \item Linearity: Both $\enc_t$ and $\dec_t$ are linear operators.
    \item Unbiased estimator: For any fixed vector $h\in \R^d$, it holds $\E[ \dec_t( \enc_t ( h ) ) ]  = h$.
    \item Bounded second moment: For any fixed vector $h\in \R^d$, it holds $\E[ \| \dec_t( \enc_t ( h ) )\|_2^2] \leq   (1+\alpha) \cdot \| h \|_2^2$, where $\alpha = a\cdot d/b_\text{sketch}$. The value of $\alpha>0$ is given in Table~\ref{tab:sketch_main} for common sketching matrices.
    \begin{table}[!ht]
    \centering
    \begin{tabular}{|c|c|c|c|}
    \hline
      \textbf{Reference} & 
      \textbf{Sketching matrix}  &  \textbf{Definition} & \textbf{Param} $\alpha$ \\
    \hline
      folklore & 
      Random Gaussian  & Def.~\ref{def:gaussian} & $3{d}/{b_\text{sketch}}$ \\
    \hline
      \cite{ldfu13} & 
      SRHT & Def.~\ref{def:srht} & $2{d}/{b_\text{sketch}}$\\
    \hline
      \cite{ams99} & 
      AMS sketch & Def.~\ref{def:ams} & $2{d}/{b_\text{sketch}}$ \\
    \hline
      \cite{ccf02} & 
      Count-sketch & Def.~\ref{def:cs} & $3{d}/{b_\text{sketch}}$ \\
     \hline
      \cite{nn13} & 
      Sparse embedding & Def.~\ref{def:sparse_1},\ref{def:sparse_2} & $2{d}/{b_\text{sketch}}$ \\
     \hline
    \end{tabular}
    \caption{Sketching matrices and their coordinate-wise embedding parameter $\alpha$.}
    \label{tab:sketch_main}
\end{table}
\end{enumerate}
\end{theorem}

\begin{proof}
Fix a vector $g\in \R^d$, note that condition 1 of Definition~\ref{def:cord_emb} implies that
\begin{align*}
    \E_{R\sim \Pi}[(R^\top Rg)_j] =  \E_{R\sim \Pi}[e_j^\top R^\top Rg] 
    =   g_j
\end{align*}
This means that in expectation, each coordinate of $R^\top Rg$ is equal to corresponding coordinate of $g$, therefore, we have
\begin{align*}
    \E_{R\sim\Pi}[R^\top Rg] = & ~ g
\end{align*}
This proves the unbiased property of Theorem~\ref{thm:enc_dec}. For the variance bound, note that using the second condition of coordinate-wise embedding, we have
\begin{align*}
     \E_{R\sim \Pi} \Big[ \sum_{j=1}^d (e_j^\top R^\top Rg)^2 \Big] = & ~ \E_{R\sim \Pi} \Big[ \sum_{j=1}^d (R^\top Rg)_j^2 \Big] \\
     = & ~ \E_{R\sim \Pi} [\|R^\top Rg\|_2^2 ] \\
     \leq & ~ \sum_{j=1}^d ((e_j^\top g)^2+\frac{a}{b_\text{sketch}}\cdot \|g\|_2^2) \\
     = & ~ (1+a\cdot \frac{d}{b_\text{sketch}})\cdot \|g\|_2^2
\end{align*}
Thus, we have proven that using $R^\top R$ as $\dec\circ \enc$, the variance parameter $\alpha$ is $a\cdot \frac{d}{b_\text{sketch}}$. By Table~\ref{tab:sketch_main}, $a$ is a small constant (2 or 3). Hence, we conclude that $\alpha=O(\frac{d}{b_\text{sketch}})$.

Note that the independence property can be satisfied via choosing independent sketching matrix $R $ at each iteration $t$, and linearity property is straightforward since $R$ is a linear transform.
\end{proof}

We will use the above property to instantiate the convergent proof and communication complexity in section~\ref{sec:conv_main}. 

\section{Convergence Analysis and Communication Complexity}
\label{sec:conv_main}
In this section, we analyze the convergence property of our proposed framework for smooth and convex objectives. Our analysis builds upon showing that the extra randomness introduced by sketching and de-sketching does not affect the convergence rate much. 

We first present our convergence result for strongly-convex objective.

\begin{theorem}[Informal version of Theorem~\ref{thm:kstep_kmr19_strcvx}]\label{thm:kstep_kmr19_strcvx_main}

If Assumption~\ref{ass:f_ass} holds with $\mu > 0$. If $\eta_\gl = 1$ and $\eta_\lo \leq \frac{1}{8(1+\alpha) L K }$,
\begin{align*}
    & ~ \E[f(w^{T+1}) - f(w^*)]\\
    \leq & ~\frac{L}{2}\E[\|w^0-w^*\|_2^2]e^{-\mu\eta_\lo T} + 4\eta_\lo^2 L^2 K^3\sigma^2/\mu.
\end{align*}
where $w^*$ is a minimizer of problem~\eqref{eq:f_def}.
\end{theorem}

Our iterative sketching method can be viewed as generating a stochastic gradient that has certain low-dimensional structure. Using the property of structured random matrices, our algorithm gives a better convergence rate than standard stochastic gradient descent. This is because in the standard stochastic gradient descent analysis, one only has an absolute upper bound on the second moment: 
\begin{align*} 
\E_{\wt g}[\|\wt g\|_2^2] \leq & ~ C^2
\end{align*}
for some parameter $C$, where $\wt g$ is the stochastic gradient with $\E_{\wt g}[\wt g]=g$. In contrast, coordinate-wise embedding guarantees that the second moment of our estimate is upper bounded \emph{multiplicatively in terms of $\|g\|_2^2$}: 
\begin{align*}
    \E_{R}[\|R^\top Rg\|_2^2] \leq & ~ \big(1+O(\frac{d}{b_{\rm sketch}}) \big)\cdot \|g\|_2^2,
\end{align*}
this nice property enables us to obtain a more refined analysis on the convergence.

We obtain the communication cost as a direct corollary:

\begin{corollary}[Informal version of Theorem~\ref{thm:communication_kstep_convex2}]\label{thm:communication_kstep_strongly_convex_main}
If Assumption~\ref{ass:f_ass} holds with $\mu > 0$ and $\eta_\gl = 1$. Then
 Algorithm~\ref{alg:fed_learn_kstep_dp} outputs an $\epsilon$-optimal solution ${w}^{T+1} \in \R^d$, which every client computes as $w^T+\dec_{T+1}(\Delta\wt w^T)$ from the last server message, satisfying $\E[f(w^{T+1}) - f(w^*)]\leq \epsilon$ by using
 \begin{align*}
     O( ({LN}/{\mu}) \max\{d, \sqrt{{\sigma^2}/{(\mu\epsilon)}}\}\log({L\E[\|w^0 - w^*\|_2^2]}/{\epsilon}))
 \end{align*} words (real numbers) of communication. The bound is attained at $K=1$ with $b_\text{sketch}=O(1)$, that is, $\alpha=\Theta(d)$.
\end{corollary}
We observe that compared to vanilla approaches, our method requires a step size shrinkage by a factor of $O(\alpha)$, thus enlarge the number of rounds approximately by a factor of $O(\alpha)$. Since the iterative sketching algorithm only communicates $O(b_\text{sketch}/d)$ as many words (real numbers) per round due to sketching, the total communication cost does not increase at all for commonly used sketching matrices, according to Theorem~\ref{thm:enc_dec}.

Theorem~\ref{thm:kstep_kmr19_strcvx_main} gives linear convergence down to its error floor $4\eta_\lo^2 L^2 K^3\sigma^2/\mu$. This floor comes from client drift during the $K$ local steps, not from sketching, whose error is multiplicative in the gradient and vanishes at $w^*$. At $K=1$ the drift vanishes (all clients start each round from the same point), and Theorem~\ref{thm:single_step_strong_convex} with $\eta_\gl=1$ gives $\E[f(w^{t+1})]-f(w^*)\leq (1-\mu\eta_\lo)^t(\E[f(w^0)]-f(w^*))$ for every $\eta_\lo\leq\frac{1}{(1+\alpha)L}$, with no floor. With $K>1$ local steps, local GD on heterogeneous smooth strongly convex losses is known to converge linearly only to a neighborhood of the solution~\cite{mkgcr20}.

We also have a similar observation for convergence in the convex losses case, as well as communication cost.
\begin{theorem}[Informal version of Theorem~\ref{thm:kstep_kmr19}]\label{thm:kstep_cvx}
If Assumption~\ref{ass:f_ass} holds with $\mu = 0$. If $\eta_\gl = 1$ and $\eta_\lo \leq \frac{1}{8(1+\alpha) L K }$,
\begin{align*}
    & ~ \E[f(\ov{w}^T) - f(w^*)] \\
    \leq & ~ \frac{4\E[\|w^0 - w^*\|_2^2]}{\eta_\lo K T } + 32\eta_\lo^2  L K^2 \sigma^2,
\end{align*}
where $\ov{u}^{t,k}=\frac{1}{N}\sum_{c=1}^N u_c^{t,k}$ and
\begin{align*}
\ov{w}^T = \frac{1}{KT}(\sum_{t=1}^{T}\sum_{k=0}^{K-1} \ov{u}^{t,k})
\end{align*}
is a virtual average of the parameters of Algorithm~\ref{alg:fed_learn_kstep_dp}, used only in the analysis. When $K>1$, no party forms it, since the averages $\ov{u}^{t,k}$ with $k\geq 1$ are never communicated.
\end{theorem}

\begin{corollary}[Informal version of Theorem~\ref{thm:communication_kstep_convex}]\label{thm:communication_kstep_convex_main}
If Assumption~\ref{ass:f_ass} holds with $\mu = 0$. Then  Algorithm~\ref{alg:fed_learn_kstep_dp} with $K=1$ and $\eta_\gl = 1$ outputs an $\epsilon$-optimal solution $\ov{w}^T = \frac{1}{T}\sum_{t=1}^T w^t \in \R^d$, which every client can compute, satisfying
\begin{align*}
\E[f(\ov{w}^T) - f(w^*)]\leq \epsilon
\end{align*}
by using
\begin{align*}
    O(\E[\|w^0-w^*\|_2^2]N \max\{{Ld}/{\epsilon}, {\sigma\sqrt{ L}}/{\epsilon^{3/2}}\})
\end{align*}
words (real numbers) of communication. The bound is attained at $K=1$ with $b_\text{sketch}=O(1)$, that is, $\alpha=\Theta(d)$.
\end{corollary}
We compare our communication cost with the work of \cite{khaled2019first}, which analyzes the local gradient descent using the same assumption and framework. The result of \cite{khaled2019first} shows a communication cost of 
\begin{align*} 
O\left(\E[\|w^0-w^*\|_2^2]Nd \cdot \max\{\frac{L}{\epsilon}, \frac{\sigma\sqrt{ L}}{\epsilon^{3/2}}\}\right),
\end{align*}
which is never smaller than our bound. This shows again our approach does not introduce extra overall communication cost.

\begin{remark}\label{rem:K1_sharper}
Both corollaries are attained at $K=1$, where Algorithm~\ref{alg:fed_learn_kstep_dp} is distributed gradient descent with sketched gradients. There the $\sigma$-dependent terms are artifacts of the $K$-step analysis. Theorems~\ref{thm:single_step_strong_convex} and~\ref{thm:single_step_convex}, with $\eta_\gl=1$, $\eta_\lo=\frac{1}{(1+\alpha)L}$ (resp.\ $\frac{1}{2(1+\alpha)L}$) and any $b_\text{sketch}\leq d$, give $O(\frac{LNd}{\mu}\log\frac{\E[f(w^0)]-f(w^*)}{\epsilon})$ (resp.\ $O(\frac{\E[\|w^0-w^*\|_2^2]LNd}{\epsilon})$) words, which never exceed the bounds of Corollaries~\ref{thm:communication_kstep_strongly_convex_main} and~\ref{thm:communication_kstep_convex_main}. Up to constants, these equal the cost of uncompressed distributed gradient descent ($R_t=I_d$, $\alpha=0$). So sketching lowers the per-round message from $d$ to $b_\text{sketch}$ words without increasing the total communication, and local steps $K>1$ improve none of the bounds of this section. The comparison with~\cite{khaled2019first} is a comparison of upper bounds. Uncompressed gradient descent also attains $O(\E[\|w^0-w^*\|_2^2]LNd/\epsilon)$.
\end{remark}

\section{Differential Privacy}
\label{sec:dp_main}

\begin{algorithm}[!ht]\caption{Private Iterative Sketching-based Federated Learning Algorithm with $K$ local steps}
\label{alg:fed_learn_kstep_dp_true}
\begin{algorithmic}[1]
\Procedure{\textsc{PrivateIterativeSketchingFL}}{}
\State{Each client initializes $w^0$ with the same seed}
\For{$t=1 \to T$} 
\State{\color{blue} /* Client */}
\ParFor{$c=1\to N$} 
\If{$t=1$}
\State{$u_c^{t,0}\gets w^0$}
\Else
\State{$u_c^{t,0}\gets w^{t-1}+\dec_t(\Delta \wt{w}^{t-1})$} 
\EndIf
\State {$w^t\gets u_c^{t,0}$}
\State $\sigma_{\mathrm{DP}}^2 \gets 8\log(1.25/\wh \delta)\ell_c^2/{\wh \epsilon}^2$
\For{$k=1\to K$}
\State {\color{red}$\xi_c^{t,k}\sim {\cal N}(0,\sigma_{\mathrm{DP}}^2\cdot I_{d})$ }
\State ${\mathcal{D}}_{c}^{t,k}\gets \textnormal{Random batch of local data}$
\State $u_c^{t,k}\gets u_c^{t,k-1}-\eta_{\mathrm{local}}\cdot (\frac{1}{|\mathcal{D}_c^{t,k}|}\cdot\sum_{z_i\in\mathcal{D}_{c}^{t,k}} \nabla f_{c}(u_c^{t,k-1}, z_i){\color{red}+\xi_c^{t,k}})$
\EndFor
\State {$\Delta w_c(t)\gets u_c^{t,K}-w^t$} 
\State {Client $c$ sends $\enc_{t+1}(\Delta w_c(t))$ to server}
\EndParFor
\State {\color{blue} /* Server */}
\State{$\Delta\wt{w}^t\gets \eta_{\mathrm{global}}\cdot \frac{1}{N}\sum_{c=1}^N \enc_{t+1}(\Delta w_c(t))$} 
\State {Server sends $\Delta\wt{w}^t$ to each client}
\EndFor
\EndProcedure
\end{algorithmic}
\end{algorithm}

In this section, we show that if each client adds a Gaussian noise corresponding to its local loss function, then the iterative sketching scheme is differentially private, and that it still converges, up to an additive error that depends on the privacy level.

To discuss the privacy guarantee of our proposed approach, we consider that each client $c$ trying to learn upon its local dataset $\mathcal{D}_c$ with corresponding local loss 
\begin{align*}
f_c(x) = \frac{1}{|\mathcal{D}_c|}\sum_{z_i\in \mathcal{D}_c}f_{c}(x, z_i),
\end{align*}
where we overload the notation $f_c$ to denote the local loss for notation simplicity. We assume that the per-example losses satisfy Assumption~\ref{ass:dp_diff} below. We also assume that the dataset for each client $c$ is disjoint. The batch $\mathcal{D}_c^{t,k}$ in Algorithm~\ref{alg:fed_learn_kstep_dp_true} is selected through a random set of indices drawn from a fixed distribution, independently of the data and of the noise vectors (for example, a uniformly random subset of a fixed size). In particular, it does not depend on the iterates. Two inputs of the algorithm are neighboring if they differ by replacing one data point of one client.

\begin{assumption}[Bounded gradient differences]\label{ass:dp_diff}
For every client $c\in[N]$, every $x\in\R^d$ and every two possible data points $z,z'$ of client $c$,
\begin{align*}
\|\nabla_x f_c(x,z)-\nabla_x f_c(x,z')\|_2\leq 2\ell_c.
\end{align*}
\end{assumption}
If every per-example loss is $\ell_c$-Lipschitz, that is, $\|\nabla_x f_c(x,z)\|_2\leq \ell_c$ for every $x\in\R^d$ and every possible data point $z$ of client $c$, then Assumption~\ref{ass:dp_diff} holds by the triangle inequality. Unlike Lipschitzness, Assumption~\ref{ass:dp_diff} is compatible with $\mu>0$ in Assumption~\ref{ass:f_ass}. For example, for $\mu>0$, the $\ell_2$-regularized logistic loss $f_c(x,(a,y))=\frac{\mu}{2}\|x\|_2^2+\log(1+e^{-y\cdot a^\top x})$ with $y\in\{-1,1\}$ and $\|a\|_2\leq\ell_c$ satisfies Assumption~\ref{ass:dp_diff}, and the resulting local losses satisfy Assumption~\ref{ass:f_ass} with $L=\mu+\max_{c\in[N]}\ell_c^2/4$.

To prove the final privacy guarantee of Algorithm~\ref{alg:fed_learn_kstep_dp_true}, we employ a localized analysis. Fix a data point, which belongs to exactly one client $c$. Once all randomness other than the noise vectors $\xi_c^{t,k}$ of client $c$ is fixed, everything the algorithm releases is a deterministic function of the $TK$ noisy gradients computed by client $c$. Each noisy gradient adds Gaussian noise to a minibatch gradient whose $\ell_2$ sensitivity is at most $2\ell_c$, so it is a Gaussian mechanism. We then apply advanced composition once, over all $TK$ noisy steps of client $c$. We defer the privacy proofs to appendix~\ref{sec:dp_thm} and the convergence proofs to appendix~\ref{sec:dp_conv}.

\begin{lemma}[Informal version of Lemma~\ref{lem:app:single_agent_dp}]\label{lem:single_agent_dp}
Let $\wh \epsilon,\wh \delta\in (0,1)$, $\delta'\in (0,1]$ and $c\in [N]$. For client $c$, the $K$ noisy local steps of one round of Algorithm~\ref{alg:fed_learn_kstep_dp_true}, started from a fixed point, are
\begin{align*}
\Big(\sqrt{2K\log(1/\delta')}\cdot \wh \epsilon+2K\wh \epsilon^2,~\delta'+K\cdot \wh \delta\Big) \mathrm{-DP}.
\end{align*}
\end{lemma}

\begin{theorem}[Informal version of Theorem~\ref{thm:app:alg_dp}]\label{thm:alg_dp}
Suppose Assumption~\ref{ass:dp_diff} holds and the batch indices are drawn from a fixed distribution, independently of the data and of the noise vectors. Let $\wh \epsilon,\wh \delta\in (0,1)$ and $\delta'\in (0,1]$. Then, Algorithm~\ref{alg:fed_learn_kstep_dp_true}, whose output consists of the sketching matrices and all messages sent by the clients and by the server, is $(\epsilon_{\mathrm{DP}},\delta_{\mathrm{DP}})$-DP, with
\begin{align*}
    \epsilon_{\mathrm{DP}} = \sqrt{2TK\log(1/\delta')}\cdot \wh \epsilon+2TK\wh \epsilon^2, & ~~~\delta_{\mathrm{DP}} = \delta'+TK\cdot \wh \delta.
\end{align*}
If $\wh \epsilon\leq \sqrt{\log(1/\delta')/(2TK)}$, then the second term of $\epsilon_{\mathrm{DP}}$ is at most the first.
\end{theorem}
\begin{proof}[Proof Sketch]
Fix a data point, which belongs to one client $c$. Condition on all randomness other than the noise vectors $\xi_c^{t,k}$ of client $c$: the initialization $w^0$, the sketching matrices $R_t$, the batches and the noise of the other clients, and the batch indices of client $c$. Given these, the sketching matrices and every released message, and hence any model computed from them, are a deterministic function of the sequence of $TK$ noisy gradients of client $c$, that is, of the minibatch gradients plus the noise $\xi_c^{t,k}$. By post-processing (Lemma~\ref{lem:app:post_process}), it suffices to show that this sequence is private. Given the earlier noisy gradients, the current iterate and the batch are fixed. Then the minibatch gradient has $\ell_2$ sensitivity at most $2\ell_c$ (Lemma~\ref{lem:app:l2_gradient}), and $\sigma_{\mathrm{DP}}^2=2\log(1.25/\wh \delta)(2\ell_c)^2/\wh \epsilon^2$. So each noisy gradient is $(\wh \epsilon,\wh \delta)$-DP by the Gaussian mechanism (Lemma~\ref{lem:app:gaussian_DP}). Adaptive advanced composition (Lemma~\ref{lem:app:advanced_comp}) over the $TK$ noisy steps gives $(\epsilon_{\mathrm{DP}},\delta_{\mathrm{DP}})$-DP for every fixed value of the conditioned randomness. Averaging the DP inequality over that randomness, whose distribution does not depend on the data, gives the bound for Algorithm~\ref{alg:fed_learn_kstep_dp_true}. Finally, $2TK\wh \epsilon^2\leq \sqrt{2TK\log(1/\delta')}\cdot \wh \epsilon$ holds if and only if $\wh \epsilon\leq \sqrt{2TK\log(1/\delta')}/(2TK)=\sqrt{\log(1/\delta')/(2TK)}$.
\end{proof}

Compared to an iterative sketching framework we described without Gaussian noises, Algorithm~\ref{alg:fed_learn_kstep_dp_true} injects extra noises at each local step for each local client. It also performs sub-sampling. Each client still sends only $b_\text{sketch}$ numbers per round. The convergence analysis in section~\ref{sec:conv_main} is for Algorithm~\ref{alg:fed_learn_kstep_dp}, which uses exact local gradients and no noise. Theorem~\ref{thm:dp_conv_main} below extends it to Algorithm~\ref{alg:fed_learn_kstep_dp_true}. Moreover, at a fixed privacy budget the noise grows with the number $TK$ of noisy steps. Indeed, $\epsilon_{\mathrm{DP}}$ in Theorem~\ref{thm:alg_dp} is at least $\sqrt{2TK\log(1/\delta')}\cdot \wh \epsilon$. Keeping $\epsilon_{\mathrm{DP}}$ fixed therefore forces $\wh \epsilon\leq \epsilon_{\mathrm{DP}}/\sqrt{2TK\log(1/\delta')}$, and hence
\begin{align*}
    \sigma_{\mathrm{DP}}^2 = \frac{8\log(1.25/\wh \delta)\ell_c^2}{\wh \epsilon^2} \geq \frac{16TK\log(1/\delta')\log(1.25/\wh \delta)\ell_c^2}{\epsilon_{\mathrm{DP}}^2}.
\end{align*}

\begin{theorem}[Informal version of Corollary~\ref{cor:app:dp_util_k1}]\label{thm:dp_conv_main}
Suppose Assumptions~\ref{ass:f_ass} and~\ref{ass:dp_diff} hold, $\enc_t$ and $\dec_t$ are as in Theorem~\ref{thm:enc_dec}, $\ov\ell^2:=\frac{1}{N}\sum_{c=1}^N\ell_c^2>0$, and every batch $\mathcal{D}_c^{t,k}$ is a uniformly random subset of $\mathcal{D}_c$ of a fixed size, and $w^0$, the sketching matrices, the batch index sets and the noise vectors are mutually independent, with $\E[\|w^0\|_2^2]<\infty$. Let $\epsilon\in(0,1]$, $\delta\in(0,1)$ and $K=1$, and run Algorithm~\ref{alg:fed_learn_kstep_dp_true} with $\wh\delta=\delta/(2T)$ and $\wh\epsilon=\epsilon/(2\sqrt{2T\log(2/\delta)})$. Then Algorithm~\ref{alg:fed_learn_kstep_dp_true} is $(\epsilon,\delta)$-DP, by Theorem~\ref{thm:alg_dp} with $\delta'=\delta/2$. Let $w^*$ be a minimizer of $f$, $D_0:=\E[\|w^0-w^*\|_2^2]$, $\Delta_0:=\E[f(w^0)]-f(w^*)$ and $\Lambda_T:=\log(2/\delta)\log(2.5T/\delta)$, and let $D>0$ and $\Delta$ be numbers that do not depend on the data, with $D\geq D_0$ and $\Delta\geq\Delta_0$. With the step size $\eta_\gl\cdot\eta_\lo$ of Corollary~\ref{cor:app:dp_util_k1}, the average $\ov w^T=\frac{1}{T}\sum_{t=1}^T w^t$, which every client can compute, satisfies
\begin{align*}
\E[f(\ov w^T)]-f(w^*)\leq \frac{2(1+\alpha)LD}{T}+\frac{17\,\ov\ell}{\epsilon}\sqrt{\frac{(1+\alpha)d\,D\Lambda_T}{N}}.
\end{align*}
If moreover $\mu>0$ and $\Upsilon:=\frac{36(1+\alpha)Ld\Lambda_T\ov\ell^2}{\mu N\epsilon^2}<\mu\Delta$, then with the other step size of Corollary~\ref{cor:app:dp_util_k1},
\begin{align*}
\E[f(w^{T+1})]-f(w^*)\leq \Delta e^{-\mu T/((1+\alpha)L)}+\frac{\Upsilon}{\mu}\Big(1+\log\frac{\mu\Delta}{\Upsilon}\Big).
\end{align*}
\end{theorem}

The first terms in Theorem~\ref{thm:dp_conv_main} are the noise-free bounds of Theorems~\ref{thm:single_step_convex} and~\ref{thm:single_step_strong_convex} at the largest step sizes allowed there (the second after $(1-x)^T\leq e^{-xT}$), with $D_0$ and $\Delta_0$ replaced by the data-independent upper bounds $D$ and $\Delta$, which the step sizes use so that they do not depend on the data. The second terms come from the sub-sampling and the noise, which enter the analysis only as an unbiased perturbation of each local step with bounded variance. They do not decrease as $T$ grows, because at a fixed privacy budget the noise variance grows with the number of noisy steps, and they grow with $(1+\alpha)d$. They decrease with the number $N$ of clients, as $1/\sqrt N$ and, up to a logarithmic factor, $1/N$, but not with the size of the local datasets, because the noise of every step is calibrated to the sensitivity bound $2\ell_c$, which does not use the batch size. Unlike standard bounds for private empirical risk minimization, which decrease with the number of data points, they are therefore small only when $N$ is large compared with $(1+\alpha)d\Lambda_T/\epsilon^2$ (times $L/\mu$ in the strongly convex case). So for Algorithm~\ref{alg:fed_learn_kstep_dp_true} a smaller sketch dimension $b_\text{sketch}$ lowers the communication per round but raises this error term, and the total communication bounds of section~\ref{sec:conv_main} are for Algorithm~\ref{alg:fed_learn_kstep_dp}. Together with Theorem~\ref{thm:alg_dp}, Theorem~\ref{thm:dp_conv_main} gives an algorithm that only communicates low-dimensional information, has a differential privacy guarantee and converges at the rates that section~\ref{sec:conv_main} proves for $K=1$, up to an additive error that depends on the privacy level. Theorem~\ref{thm:app:dp_conv_k} and Corollary~\ref{cor:app:dp_util_k} in appendix~\ref{sec:dp_conv} give bounds of the same form for $K\geq 1$ local steps.

We would also like to point out via more advanced techniques in differential privacy such as moment account or gradient clipping, the privacy-utility trade-off of Algorithm~\ref{alg:fed_learn_kstep_dp_true} can be improved. We do not aim to optimize over these perspectives in this paper since our purpose is to motivate the use of differential privacy techniques. As we will show in section~\ref{sec:attack_sketched_gradients}, for $K=1$ and without the Gaussian noise, there exists a simple, iterative algorithm that uses the communicated sketched gradient, the sketching matrix and the model parameter to find, at a linear rate, an input whose sketched gradient matches the communicated one, for models and sketching matrices that satisfy the assumptions of Corollary~\ref{cor:sketch_attack_rate}. Such an input need not be the local data (Remark~\ref{rem:sketch_non_identifiable}). From several rounds with fresh sketches, the same attack recovers the local data itself, under the conditions of Section~\ref{sec:attack_multi_round}.

\section{Attack Sketched Gradients}\label{sec:attack_sketched_gradients}

To complement our algorithmic contribution, we show that under certain conditions on the loss functions $f_c$'s for $c\in [N]$ and the local step $K=1$, the messages of Algorithm~\ref{alg:fed_learn_kstep_dp} \emph{without the additive Gaussian noise} are open to a gradient-matching attack. To achieve this goal, we present an attacking algorithm that runs gradient descent on the data and finds, at a linear rate, an input whose (sketched) gradient matches the communicated one. Such an input need not be the local data (see the discussion after Corollary~\ref{cor:attack_rate} and Remark~\ref{rem:sketch_non_identifiable}). From one sketched gradient with $b_{\rm sketch}<m$ the data cannot be determined, but from several rounds with fresh sketches the same attack recovers the local data, under the conditions of Section~\ref{sec:attack_multi_round}.

\subsection{Warm-up: Attacking Algorithm without Sketching}

To start off, we describe an attacking algorithm without sketching being applied. We denote the loss function of the model by $F(w,x)$, where $x\in\mathbb{R}^m$ is the input and $w \in \mathbb{R}^d$ is the model parameter. We do not constrain $F(w,x)$ to be the loss of any specific model or task. $F(w,x)$ can be an $\ell_2$ loss of linear regression model, a cross-entropy loss of a neural network, or any function that the clients in the training system want to minimize. Our guarantees below apply to those $F$ that satisfy the hypotheses of Theorem~\ref{thm:F_L_smooth_criticl}, and the $\ell_2$ loss $F(w,x)=(w^\top x-y)^2$ of linear regression, with features $x$ and label $y$, does not satisfy them when $w\neq 0$, since its gradient $\nabla_w F(w,x)=2(w^\top x-y)x$ is quadratic in $x$ and hence not Lipschitz in $x$. In our federated learning scenario, the local loss of client $c$ is $f_c(w)=F(w,\tilde x^{(c)})$, where $\tilde x^{(c)}$ is its local data, and the training objective is $\frac{1}{N}\sum_{c=1}^N f_c(w)$. With $K=1$, client $c$ sends $R(-\eta_{\rm local}\nabla f_c(w))$, from which $R\nabla_w F(w,\tilde x^{(c)})$ is obtained by rescaling with the known $\eta_{\rm local}$. Note that one can view the local loss function $f_c$ being associated with the local dataset that can only be accessed by client $c$. During training, client $c$ will send the gradient computed with the local training data $\nabla_w F(w,\tilde{x}^{(c)})$ to the server where $\tilde x^{(c)}$ denotes the local data.

The attacker can observe the gradient information shared in the algorithm. For client $c$, the attacker could observe $g = \nabla_w F(w, \tilde{x}^{(c)})$ and $w$. Intuitively, one can view attacker has hijacked one of the client and hence gaining access to the model parameter. Local data $\tilde x^{(c)}$ will not be revealed to the attacker.

The attacker also has access to a gradient oracle, meaning that it can generate arbitrary data $x\in \R^m$ and feed into the oracle, and the oracle will return the gradient with respect to $x$ and parameter $w$. The attacker will then try to find $x$ that minimizes 
\begin{align*}
L(x)= \|\nabla_w F(w,x) - g\|^2
\end{align*}
by running gradient descent. The attacker will start with random initialization $x_0$, and iterates as
\begin{align*}
x_{t+1} = x_t - \eta\cdot\nabla L(x_t)
\end{align*}
where $\eta>0$ is the step size chosen by the attacker.

To formalize the analysis, we introduce some key definitions. 
Given a function $F:\R^d\times \R^m\rightarrow \R$, a data point $x\in \R^m$, a fixed (gradient) vector $g\in \R^d$ and a fixed (weight) vector $w \in \R^d$, we define the function $L$ as follows:
\begin{align*}
L(x) := \|\nabla_w F(w,x) - g\|^2.
\end{align*}

In this subsection and the next, we consider the regime where $d\leq m$, i.e., the model is \emph{under-parametrized}. The several-round results of Section~\ref{sec:attack_multi_round} do not need $d\le m$. The over-parametrized setting is studied in a recent work~\cite{wll22} that uses tensor decomposition to recover the data from gradients. In contrast, our approach simply applies gradient descent, therefore it can easily get stuck in a local minima, which is often the case in over-parametrized setting. However, our algorithm is notably simpler and computationally efficient.

To better illustrate properties we want on $L$, we define the matrix $K$ as follows:
\begin{definition}\label{def:Phi_K}
Let $F:\R^d\times \R^m\rightarrow \R$, suppose $F$ is differentiable on both $x$ and $w$, then we define pseudo-Hessian mapping $\Phi : \R^{m} \times \R^d \rightarrow \R^{m \times d}$ as follows
\begin{align*}
    \Phi(x,w) = \nabla_x \nabla_w F(x,w).
\end{align*}
Correspondingly, we define a pseudo-kernel $K : \R^m\times\R^d \rightarrow \R^{d \times d}$ with respect to $\nabla_x F(w,x)$ as:
\begin{align*}
    K(x, w) = \Phi(x,w)^\top \Phi(x,w).
\end{align*}
Note the weight vector $w$ is fixed in our setting, we write $K(x) = K(x,w)$ for simplicity.

\end{definition}
For a regular Hessian matrix, one considers taking second derivative with respect to a single variable. Here, our input is $\nabla_w F(w,x)$ and we need to take gradient of the input with respect to $x$, hence, it is instructive to study the structure of $\nabla_x \nabla_w F(w,x)$. 

We additionally introduce several key definitions that can be implied through some basic assumptions we will make. The first is a generalization of smoothness to the notion of \emph{semi-smoothness}.
\begin{definition}[Semi-smoothness]\label{def:semi_smooth}
For any $p \in [0,1]$ and any set $S\subseteq \R^m$, we say $L:\R^m\rightarrow \R$ is $(a,b,p)$-semi-smooth on $S$ if for any $x\in S$ and any $y\in \R^m$, we have
\begin{align*}
    L(y) \le & ~ L(x) + \langle \nabla L(x), y-x \rangle \\
    & ~ + b\|y-x\|^2
    + a \|x-y\|^{2-2p}L(x)^{p}.
\end{align*}
We say $L$ is $(a,b,p)$-semi-smooth if this holds with $S=\R^m$. When $a>0$ and $p>0$, the term $L(x)^p$ presumes $L\geq 0$ on $S$.
\end{definition}

For example, the following functions satisfy the inequality of Definition~\ref{def:semi_smooth} with $a=0$. The function $L(x)=\|x\|^2$ is $(0,1,p)$-semi-smooth for every $p$, since $L(y)=L(x)+\langle 2x,y-x\rangle+\|y-x\|^2$. If $L(x)=h(w^\top x+b)$ for a twice differentiable $h:\R\rightarrow\R$ with bounded $h''$, as for $L(x)=\ln(1+\exp(w^\top x))$, $L(x)=\tanh(w^\top x+b)$ and $L(x)=\text{sigmoid}(w^\top x+b)$, then Taylor's theorem gives
\begin{align*}
    L(y)\le & ~ L(x)+\langle\nabla L(x),y-x\rangle \\
    & ~ + \frac{\|w\|^2}{2}\sup_{t\in\R}|h''(t)|\cdot\|y-x\|^2 .
\end{align*}
A concave function satisfies the inequality with $a=b=0$ for all $x,y$ in its convex domain, for example $L(x)=\sqrt{w^\top x+b}$ on $\{x: w^\top x+b>0\}$ and $L(x)=\log(w^\top x)$ on $\{x: w^\top x>0\}$. Since $a=0$ in all these examples, the term $L(x)^p$ plays no role, and the examples may take negative values (as $\tanh$ and $\log$ do).

\begin{definition}[Non-critical point]\label{def:no_critical_point}
We say $L:\R^m\rightarrow \R$ is $(\theta_1,\theta_2)$-non-critical point if 
\begin{align*} 
\theta_1^2 \cdot L(x) \leq \| \nabla L(x) \|^2 \leq \theta_2^2 \cdot L(x) .
\end{align*}
\end{definition}

The intuition for non-critical point property is that if $L(x)$ is large enough, then gradient descent can still make progress because $\|\nabla L(x)\|^2$ is lower bounded by $\theta_1^2\cdot L(x)$. Suppose $\nabla_w F(w,x)$ is Lipschitz in $x$, the pseudo-Hessian $\Phi(x,w)$ is Lipschitz in $x$, and the pseudo-kernel is non-degenerate. Then the corresponding $L$ is semi-smooth on its sublevel sets and non-critical point:

\begin{theorem}\label{thm:F_L_smooth_criticl}
If $F$ satisfies the following properties: $\forall x,y\in \R^m$,
\begin{itemize}
    \item $\nabla_w F(w,x)$ is $\beta$-Lipschitz w.r.t. $x$,
    \item $\Phi$ is $M$-Lipschitz w.r.t. $x$ in spectral norm, i.e., $\|\Phi(x,w)-\Phi(y,w)\|\le M\|x-y\|$,
    \item $K$'s eigenvalues can be bounded by
    \begin{align*}
    0 < \theta_1^2 \le \lambda^2_1(x) \le \cdots \le \lambda^2_{d}(x) \le \theta_2^2 ,
    \end{align*}
\end{itemize}
then we have
\begin{itemize}
    \item for all $x,y\in\R^m$,
    \begin{align*}
        L(y)\le & ~ L(x)+\langle\nabla L(x),y-x\rangle \\
        & ~ +\big(\beta^2+M L(x)^{1/2}\big)\|y-x\|^2 ,
    \end{align*}
    so for every $x_0\in\R^m$, $L$ is $(0,\beta^2+M L(x_0)^{1/2},1/2)$-semi-smooth (Def.~\ref{def:semi_smooth}) on $S_0:=\{x\in\R^m: L(x)\le L(x_0)\}$,
    \item $L$ satisfies $(2\theta_1,2\theta_2)$-non-critical point (Def.~\ref{def:no_critical_point}).
\end{itemize}
\end{theorem}

The proof is in Lemmas~\ref{lem:what_F_implies_L_semi_smoothness:app} and~\ref{lem:what_F_implies_L_no_critical_point:app} in Appendix~\ref{sec:from_F_to_L:app}. The lower bound $\theta_1>0$ requires $d\le m$, because $K(x)\in\R^{d\times d}$ has rank at most $\min(d,m)$. Since $\Phi(x,w)^\top$ is the Jacobian of the $\beta$-Lipschitz map $x\mapsto\nabla_w F(w,x)$, we have $\|\Phi(x,w)\|\le\beta$. Hence $\theta_1\le\lambda_1(x)\le\|\Phi(x,w)\|\le\beta$, and the upper bound always holds with $\theta_2=\beta$. For example, $F(w,x)=w^\top Ax$ with $A\in\R^{d\times m}$ of full row rank meets all three conditions with $\beta=\theta_2=\sigma_{\max}(A)$, $\theta_1=\sigma_{\min}(A)$ and $M=0$, since $\nabla_w F(w,x)=Ax$ and $\Phi(x,w)=A^\top$.

We state Theorem~\ref{thm:cost_convergence} here and the proof is provided in appendix \ref{sec:cost_1}.
\begin{theorem}\label{thm:cost_convergence}
Let $L:\R^m\rightarrow\R$ satisfy $L(x)\ge 0$ for all $x$, let $x^*$ be a point with $L(x^*)=0$, let $x_0\in\R^m$ and $S_0:=\{x\in\R^m: L(x)\le L(x_0)\}$. Let $a\ge 0$, $b>0$, and
\begin{itemize}
    \item $\theta_1^2 > 2a\cdot\theta_2$,
    \item $0<\eta \le 1/(2b)$,
    \item $\gamma = \eta(\theta_1^2-2a\cdot \theta_2)/2$.
\end{itemize}

Suppose we run gradient descent algorithm to update $x_{t+1}$ in each iteration as
\begin{align*}
    x_{t+1} = & ~x_t - \eta \cdot \nabla L(x) |_{x = x_t}.
\end{align*}

If we assume that $L$ is $(a,b,1/2)$-semi-smooth on the set $S_0$ (Def.~\ref{def:semi_smooth}) and that $L$ is $(\theta_1,\theta_2)$-non-critical point (Def.~\ref{def:no_critical_point}),
then every $x_t$ lies in $S_0$ and
\begin{align*}
    L(x_{t+1}) - L(x^*) \leq (1-\gamma ) (L(x_t) - L(x^*)).
\end{align*}
\end{theorem}

With $a=0$, Theorem~\ref{thm:F_L_smooth_criticl} and Theorem~\ref{thm:cost_convergence} give a linear rate for the attack.
\begin{corollary}\label{cor:attack_rate}
Suppose $F$ satisfies the conditions of Theorem~\ref{thm:F_L_smooth_criticl}. Let $x_0\in\R^m$, and run gradient descent $x_{t+1}=x_t-\eta\cdot\nabla L(x_t)$ with $0<\eta\le 1/(2(\beta^2+M L(x_0)^{1/2}))$. Then for every $t\ge 0$,
\begin{align*}
    L(x_{t+1})\le(1-2\eta\theta_1^2)\cdot L(x_t),
\end{align*}
so $L(x_T)\le(1-2\eta\theta_1^2)^T\cdot L(x_0)$ for every $T\ge 0$, and $0\le 1-2\eta\theta_1^2<1$. Moreover, $x_t$ converges to a point $x_\infty$ with $\nabla_w F(w,x_\infty)=g$, and $\|x_t-x_\infty\|\le\frac{2\eta\theta_2\,q^t}{1-q}L(x_0)^{1/2}$ for every $t\ge 0$, where $q:=(1-2\eta\theta_1^2)^{1/2}<1$.
\end{corollary}
\begin{proof}
We have $L\ge 0$ and $L(\tilde x^{(c)})=0$, since $g=\nabla_w F(w,\tilde x^{(c)})$. Let $b_0:=\beta^2+M L(x_0)^{1/2}$, and note $b_0\ge\beta^2\ge\theta_1^2>0$. By Theorem~\ref{thm:F_L_smooth_criticl}, $L$ is $(0,b_0,1/2)$-semi-smooth on $S_0$ and $(2\theta_1,2\theta_2)$-non-critical point. We apply Theorem~\ref{thm:cost_convergence} with $x^*=\tilde x^{(c)}$, $a=0$, $b=b_0$, and $2\theta_1,2\theta_2$ in place of $\theta_1,\theta_2$. Its first condition reads $4\theta_1^2>0$, its second is our choice of $\eta$, and $\gamma=\eta(2\theta_1)^2/2=2\eta\theta_1^2$. This gives $L(x_{t+1})\le(1-2\eta\theta_1^2)L(x_t)$, and induction on $T$ gives the bound on $L(x_T)$. Finally, $2\eta\theta_1^2>0$, and $\theta_1\le\beta$ gives
\begin{align*}
2\eta\theta_1^2\le\frac{\theta_1^2}{\beta^2+M L(x_0)^{1/2}}\le\frac{\theta_1^2}{\beta^2}\le 1 .
\end{align*}
For the last claim, the contraction gives $L(x_t)^{1/2}\le q^t L(x_0)^{1/2}$, and the upper bound of the non-critical point condition gives $\|x_{t+1}-x_t\|=\eta\|\nabla L(x_t)\|\le 2\eta\theta_2 q^t L(x_0)^{1/2}$. These step lengths are summable, so $(x_t)$ is a Cauchy sequence with a limit $x_\infty$. Continuity of $L$ gives $L(x_\infty)=0$, and summing the step lengths from $t$ on gives the bound.
\end{proof}
Given $\beta$ and $M$, the attacker can compute this step size, because it observes $g$ and can therefore evaluate $L(x_0)$.

Theorem~\ref{thm:cost_convergence} and Corollary~\ref{cor:attack_rate} state that a gradient descent that starts with a dummy data point $x_0$ can \emph{converge} in the sense that it generates a point $x_T$ whose gradient is close to the gradient $g$ of the local data $\tilde x^{(c)}$. They do not say that $x_T$ is close to $\tilde x^{(c)}$. That would follow if $F$ had the property that similar gradients imply similar data points. Under the assumption on $K$ in Theorem~\ref{thm:F_L_smooth_criticl}, this property can hold only when $d=m$. If $d<m$, the inputs whose gradient equals $g$ form a submanifold of dimension $m-d\geq 1$ through $\tilde x^{(c)}$ (Remark~\ref{rem:sketch_non_identifiable} with $R=I_d$). If $d=m$, the map $G(x):=\nabla_w F(w,x)$ sends $\R^m$ to $\R^m$. It is continuously differentiable, since its Jacobian $\Phi(x,w)^\top$ is $M$-Lipschitz in $x$. The singular values of the square matrix $\Phi(x,w)$ are at least $\theta_1$, because the eigenvalues of $K(x)=\Phi(x,w)^\top\Phi(x,w)$ are at least $\theta_1^2$, so $\Phi(x,w)^\top$ is invertible and $\|(\Phi(x,w)^\top)^{-1}\|\leq 1/\theta_1$ for every $x$. By Hadamard's global inverse function theorem (in the form of Hadamard and L\'evy~\cite{p74}, which asks for an invertible Jacobian whose inverse is bounded on all of $\R^m$), $G$ is a bijection of $\R^m$ onto $\R^m$ with a continuously differentiable inverse. The Jacobian of $G^{-1}$ at $G(x)$ is $(\Phi(x,w)^\top)^{-1}$, so the mean value inequality shows that $G^{-1}$ is $(1/\theta_1)$-Lipschitz. Since $g=G(\tilde x^{(c)})$, Corollary~\ref{cor:attack_rate} then gives
\begin{align*}
    \|x_T-\tilde x^{(c)}\| = & ~ \|G^{-1}(G(x_T))-G^{-1}(g)\| \\
    \leq & ~ \frac{\|G(x_T)-g\|}{\theta_1} = \frac{L(x_T)^{1/2}}{\theta_1} \\
    \leq & ~ \frac{(1-2\eta\theta_1^2)^{T/2}L(x_0)^{1/2}}{\theta_1} ,
\end{align*}
so in this case the attack recovers the data at a linear rate. Recovery of the data by gradient matching has been widely observed in practice~\cite{zlh19,yin2021see,zmb20}.

\subsection{Attacking Gradients under Sketching}
Now we consider the setting where \emph{sketched} gradients are shared instead of the true gradient. Let $R:\R^d\rightarrow \R^{b_{\text{sketch}}}$ be a sketching operator. As for the sketching matrices in this paper, we take $R$ to be linear, that is, a matrix $R\in\R^{b_{\text{sketch}}\times d}$ with $R(u)=Ru$, and $\|R\|$ denotes its spectral norm. Then the gradient we observe becomes $R(\nabla_w F(w,x))$. Additionally, we can also observe the sketching matrix $R$ and model parameter $w$. In this setting, the objective function we consider becomes
\begin{align*}
     L_{R}(x) :=  \|R(\nabla_w F(w,x))-R(g)\|^2.
\end{align*}

It is reasonable to assume the attacker has access to $R$, since frameworks that make use of sketching~\cite{rpu20} do so by sharing the sketching matrix across all nodes.

Lemma~\ref{lem:sketching_semi_smooth} and Lemma~\ref{lem:sketching_non_critical_point} show that with reasonable assumptions about $R$, which are typical properties of popular sketch matrices, $L_R$ still satisfies the semi-smoothness condition on its sublevel sets and the non-critical-point condition. We defer all the proofs to appendix~\ref{sec:sketching:app}.

\begin{lemma}\label{lem:sketching_semi_smooth}
If the sketching operator $R$ satisfies
$\|R(u)-R(v)\| \le \tau\|u-v\|$ for all $u,v\in\R^d$ and $\|R\|\leq \gamma_2$,
and $F$ satisfies the conditions as in Theorem~\ref{thm:F_L_smooth_criticl},
then for all $x,y\in\R^m$,
\begin{align*}
    L_R(y)\le & ~ L_R(x)+\langle\nabla L_R(x),y-x\rangle \\
    & ~ +\big(\tau^2\beta^2+\gamma_2 M L_R(x)^{1/2}\big)\|y-x\|^2 .
\end{align*}
Hence for every $x_0\in\R^m$, $L_R$ is $(0,B,1/2)$-semi-smooth on $S_0^R:=\{x\in\R^m: L_R(x)\le L_R(x_0)\}$, where $B = \tau^2\beta^2+\gamma_2 M L_R(x_0)^{1/2}$.
\end{lemma}

\begin{lemma}\label{lem:sketching_non_critical_point}
If the sketching operator $R$ satisfies that the singular values of $R^\top$ lie in $[\gamma_1,\gamma_2]$ with $\gamma_1>0$, and $F$ satisfies conditions as in Theorem~\ref{thm:F_L_smooth_criticl},
then $L_R(x)$ is $(2\theta_1\gamma_1,2\theta_2\gamma_2)$-non-critical-point.
\end{lemma}

While $R$ itself is a short and fat matrix and is impossible to have nonzero smallest singular value, our singular value assumption is imposed on $R^\top \in \R^{d\times b_{\rm sketch}}$, hence reasonable. Moreover, for many sketching matrices $R$ (such as each entry being i.i.d. Gaussian random variables), the matrix $R^\top$ is full rank almost surely.
Combining Lemma~\ref{lem:sketching_semi_smooth} and Lemma~\ref{lem:sketching_non_critical_point}, Theorem~\ref{thm:sketching_convergence_of_L} and Corollary~\ref{cor:sketch_attack_rate} show that gradient matching still succeeds for sketched gradients: gradient descent on $L_R$ finds an input whose sketched gradient matches $R(g)$, at a linear rate.

\begin{theorem}\label{thm:sketching_convergence_of_L}
If the sketching operator $R$ satisfies
\begin{itemize}
    \item $\|R(u)-R(v)\| \le \tau\|u-v\|$ for all $u,v\in\R^d$,
    \item $0<\gamma_1\leq \sigma_1(R^\top)\leq \ldots\leq \sigma_{b_{\rm sketch}}(R^\top)\leq \gamma_2$,
\end{itemize}
$F$ satisfies the conditions in Theorem~\ref{thm:F_L_smooth_criticl}, and $x_0\in\R^m$,
then $L_R(x)$ is
\begin{itemize}
    \item $(0,\tau^2\beta^2+\gamma_2 M L_R(x_0)^{1/2},1/2)$-semi-smooth on $S_0^R=\{x\in\R^m: L_R(x)\le L_R(x_0)\}$,
    \item $(2\theta_1\gamma_1,2\theta_2\gamma_2)$-non-critical-point.
\end{itemize}
\end{theorem}

\begin{corollary}\label{cor:sketch_attack_rate}
Suppose $R$ and $F$ satisfy the conditions of Theorem~\ref{thm:sketching_convergence_of_L}. Let $x_0\in\R^m$, let $B=\tau^2\beta^2+\gamma_2 M L_R(x_0)^{1/2}$, and run gradient descent $x_{t+1}=x_t-\eta\cdot\nabla L_R(x_t)$ with $0<\eta\le 1/(2B)$. Then for every $t\ge 0$,
\begin{align*}
    L_R(x_{t+1})\le(1-2\eta\theta_1^2\gamma_1^2)\cdot L_R(x_t),
\end{align*}
and $0\le 1-2\eta\theta_1^2\gamma_1^2<1$. Moreover, $x_t$ converges to a point $x_\infty$ with $R\nabla_w F(w,x_\infty)=R g$, and $\|x_t-x_\infty\|\le\frac{2\eta\theta_2\gamma_2\,q^t}{1-q}L_R(x_0)^{1/2}$ for every $t\ge 0$, where $q:=(1-2\eta\theta_1^2\gamma_1^2)^{1/2}<1$.
\end{corollary}
\begin{proof}
We have $L_R\ge 0$ and $L_R(\tilde x^{(c)})=0$, since $g=\nabla_w F(w,\tilde x^{(c)})$. By Theorem~\ref{thm:sketching_convergence_of_L}, $L_R$ is $(0,B,1/2)$-semi-smooth on $S_0^R$ and $(2\theta_1\gamma_1,2\theta_2\gamma_2)$-non-critical point. We apply Theorem~\ref{thm:cost_convergence} to $L_R$ with $x^*=\tilde x^{(c)}$, $a=0$, $b=B$, and $2\theta_1\gamma_1,2\theta_2\gamma_2$ in place of $\theta_1,\theta_2$. Its first condition reads $4\theta_1^2\gamma_1^2>0$, its second is our choice of $\eta$, and $\gamma=\eta(2\theta_1\gamma_1)^2/2=2\eta\theta_1^2\gamma_1^2$. This gives the contraction. Here $\|Ru\|=\|R(u)-R(0)\|\le\tau\|u\|$ gives $\|R\|\le\tau$, so $\gamma_1\le\sigma_1(R^\top)\le\|R^\top\|=\|R\|\le\tau$. Together with $\theta_1\le\beta$, this shows $B\ge\tau^2\beta^2\ge\gamma_1^2\theta_1^2>0$, as Theorem~\ref{thm:cost_convergence} requires. For the range of the factor, $2\eta\theta_1^2\gamma_1^2>0$ and
\begin{align*}
2\eta\theta_1^2\gamma_1^2\le\frac{\theta_1^2\gamma_1^2}{B}\le\frac{\theta_1^2\gamma_1^2}{\tau^2\beta^2}\le 1 .
\end{align*}
For the last claim, the contraction gives $L_R(x_t)^{1/2}\le q^t L_R(x_0)^{1/2}$, and the upper bound of the non-critical point condition gives $\|x_{t+1}-x_t\|=\eta\|\nabla L_R(x_t)\|\le 2\eta\theta_2\gamma_2 q^t L_R(x_0)^{1/2}$. These step lengths are summable, so $(x_t)$ is a Cauchy sequence with a limit $x_\infty$. Continuity of $L_R$ gives $L_R(x_\infty)=0$, and summing the step lengths from $t$ on gives the bound.
\end{proof}

\begin{remark}\label{rem:attack_iteration_counts}
We compare our bounds on the number of iterations that the two attacks need, started from the same $x_0$, to reduce their own losses by the same factor $\epsilon\in(0,1)$: the loss $L$ without sketching, and the loss $L_R$ with sketching. Let $b_0:=\beta^2+M L(x_0)^{1/2}$ and take $\eta=1/(2b_0)$ in Corollary~\ref{cor:attack_rate}. Since $0\le 1-u\le e^{-u}$ for $u\in[0,1]$,
\begin{align*}
    L(x_T)\le(1-\theta_1^2/b_0)^T L(x_0)\le e^{-T\theta_1^2/b_0}L(x_0),
\end{align*}
so $L(x_T)\le\epsilon L(x_0)$ once $T\ge T_0:=\lceil b_0\ln(1/\epsilon)/\theta_1^2\rceil$. For the sketched attack, take $\tau=\|R\|$, which is allowed since $\|R(u)-R(v)\|\le\|R\|\cdot\|u-v\|$, and note $\|R\|=\sigma_{\max}(R^\top)\le\gamma_2$. Take $\eta=1/(2B)$ in Corollary~\ref{cor:sketch_attack_rate}. In the same way, $L_R(x_T)\le e^{-T\theta_1^2\gamma_1^2/B}L_R(x_0)$, so $L_R(x_T)\le\epsilon L_R(x_0)$ once $T\ge T_R:=\lceil B\ln(1/\epsilon)/(\theta_1^2\gamma_1^2)\rceil$. Since $L_R(x_0)=\|R(\nabla_w F(w,x_0)-g)\|^2\le\|R\|^2 L(x_0)\le\gamma_2^2 L(x_0)$,
\begin{align*}
    B = & ~ \|R\|^2\beta^2+\gamma_2 M L_R(x_0)^{1/2} \\
    \le & ~ \gamma_2^2\beta^2+\gamma_2^2 M L(x_0)^{1/2} = \gamma_2^2 b_0 ,
\end{align*}
and therefore
\begin{align*}
    T_R\le\left\lceil(\gamma_2/\gamma_1)^2\cdot b_0\ln(1/\epsilon)/\theta_1^2\right\rceil .
\end{align*}
So the bound on $T_R$ is the formula for $T_0$ with an extra factor $(\gamma_2/\gamma_1)^2$ inside the ceiling.
\end{remark}

\begin{remark}\label{rem:sketch_non_identifiable}
Corollary~\ref{cor:sketch_attack_rate} gives an input $x_\infty$ whose sketched gradient equals $R(g)$. It does not give the local data $\tilde x^{(c)}$, and when $b_{\rm sketch}<m$ (in particular when $b_{\rm sketch}<d$, as in sketching) one sketched gradient cannot determine it. Suppose $R$ and $F$ satisfy the conditions of Theorem~\ref{thm:sketching_convergence_of_L}, and write $G(x):=\nabla_w F(w,x)$. Since $\gamma_1>0$, all $b_{\rm sketch}$ singular values of $R^\top\in\R^{d\times b_{\rm sketch}}$ are positive, so $R$ has full row rank $b_{\rm sketch}$ and $b_{\rm sketch}\leq d$. Since $K(x)=\Phi(x,w)^\top\Phi(x,w)$ is positive definite, $\Phi(x,w)^\top\in\R^{d\times m}$ has rank $d$. The map $x\mapsto RG(x)$ is continuously differentiable with Jacobian $R\Phi(x,w)^\top$, which is continuous because $\Phi$ is $M$-Lipschitz in $x$. This Jacobian maps $\R^m$ onto $\R^{b_{\rm sketch}}$, because $\Phi(x,w)^\top$ maps $\R^m$ onto $\R^d$ and $R$ maps $\R^d$ onto $\R^{b_{\rm sketch}}$. So every point is regular, and by the regular value theorem (a consequence of the implicit function theorem) the set
\begin{align*}
    Z_R:=\{x\in\R^m: R\nabla_w F(w,x)=R g\}
\end{align*}
is a continuously differentiable submanifold of $\R^m$ of dimension $m-b_{\rm sketch}$. It contains $\tilde x^{(c)}$, and $L_R=0$ on $Z_R$. When $b_{\rm sketch}<m$, this dimension is at least $1$, so $Z_R$ has infinitely many points. In particular, this holds when $b_{\rm sketch}<d$, as in sketching, since then $m-b_{\rm sketch}\geq d-b_{\rm sketch}\geq 1$. Every point of $Z_R$, taken as the local data, produces the same observation $(R,w,R g)$. So this observation cannot determine $\tilde x^{(c)}$, whatever method the attacker uses. The same argument with $R=I_d$ (so $b_{\rm sketch}=d$ and $\gamma_1=\gamma_2=1$) shows that without sketching the inputs whose gradient equals $g$ form a submanifold of dimension $m-d$. Sketching enlarges this set, which contains the unsketched one, to dimension $m-b_{\rm sketch}$, so one observation is consistent with more inputs.

For example, let $d=m=2$, $\beta>0$, $F(w,x)=\beta\langle x,w\rangle$ and $R=[1~~0]$. Then $\nabla_w F(w,x)=\beta x$ and $\Phi(x,w)=\beta I_2$, so the conditions of Theorem~\ref{thm:F_L_smooth_criticl} hold with $\theta_1=\theta_2=\beta$ and $M=0$, and $R$ meets those of Theorem~\ref{thm:sketching_convergence_of_L} with $\tau=\gamma_1=\gamma_2=1$. Write an input as $x=(u,v)$ and the local data as $\tilde x^{(c)}=(\tilde u,\tilde v)$. Then
\begin{align*}
    L_R(x)=\beta^2(u-\tilde u)^2, \quad \nabla L_R(x)=(2\beta^2(u-\tilde u),0),
\end{align*}
so gradient descent on $L_R$ never changes the second coordinate $v$ of its iterate. Without sketching, $L(x)=\beta^2\|x-\tilde x^{(c)}\|^2$, and gradient descent with $\eta=1/(2\beta^2)$ reaches $\tilde x^{(c)}$ in one step.
\end{remark}

\subsection{Recovering the Data from Several Rounds}\label{sec:attack_multi_round}

By Remark~\ref{rem:sketch_non_identifiable}, one sketched gradient with $b_{\rm sketch}<m$ cannot determine the local data. Algorithm~\ref{alg:fed_learn_kstep_dp}, however, draws a fresh sketching matrix in every round, and the weights change from round to round, while the local data stay the same. We show that, under the conditions of Theorems~\ref{thm:multi_square} and~\ref{thm:multi_flat}, an attacker that follows one client over several rounds recovers its data with the same gradient-matching attack, run on the sum of the per-round losses.

\paragraph{Setting.} Fix a client $c$ and write $\tilde x:=\tilde x^{(c)}\in\R^m$. Run Algorithm~\ref{alg:fed_learn_kstep_dp} with $K=1$ and without additive noise. The attacker observes $T$ rounds $t_1<\cdots<t_T$, fixed in advance. The attacker is the server, or an eavesdropper on client $c$'s channel: it sees client $c$'s own message in each observed round, not only the average $\Delta\wt w^t$ that the clients receive. In this subsection $T$ is the number of observed rounds, not the total number of rounds of Algorithm~\ref{alg:fed_learn_kstep_dp}. Likewise, $\alpha$, $\mu$, $\kappa$, $a$ and $\varepsilon$ denote constants local to this subsection and to Appendix~\ref{sec:multi_round:app}. In round $t_j$ it sees the weights $w^{t_j}$, the sketching matrix $\bar R_j:=\enc_{t_j+1}=R_{t_j+1}\in\R^{b_j\times d}$ that client $c$ applies (in Algorithm~\ref{alg:fed_learn_kstep_dp}, $b_j=b_{\rm sketch}$), and the message $\bar R_j(-\eta_{\rm local}\nabla f_c(w^{t_j}))$. Rescaling by the known $\eta_{\rm local}$ gives $y_j:=\bar R_j\nabla_wF(w^{t_j},\tilde x)$. Let $n:=\sum_{j=1}^T b_j$ and $\Gamma:=\big(\sum_{j=1}^T\|\bar R_j\|^2\big)^{1/2}$, and define $H:\R^m\rightarrow\R^n$ and $J(x)\in\R^{n\times m}$ by
\begin{align*}
H(x):= & ~ \begin{bmatrix}\bar R_1\nabla_wF(w^{t_1},x)\\ \vdots\\ \bar R_T\nabla_wF(w^{t_T},x)\end{bmatrix}, \\
J(x):= & ~ \begin{bmatrix}\bar R_1\Phi(x,w^{t_1})^\top\\ \vdots\\ \bar R_T\Phi(x,w^{t_T})^\top\end{bmatrix},
\end{align*}
so that $J(x)$ is the Jacobian of $H$ at $x$ (Definition~\ref{def:Phi_K}), and $y:=(y_1,\ldots,y_T)=H(\tilde x)$. The attacker runs gradient descent $x_{k+1}=x_k-\eta\nabla\mathcal L(x_k)$ on
\begin{align*}
\mathcal L(x):= & ~ \|H(x)-y\|^2 \\
= & ~ \sum_{j=1}^T\|\bar R_j\nabla_wF(w^{t_j},x)-y_j\|^2 ,
\end{align*}
which is the sum of the single-round losses $L_{\bar R_j}$ of the previous subsection, each with $w=w^{t_j}$. The attacker can evaluate $\mathcal L(x_0)$, so it can compute the step sizes below from the model constants.

\begin{assumption}\label{ass:multi_round}
For every $j\in[T]$ and all $x,z\in\R^m$, $\|\nabla_wF(w^{t_j},x)-\nabla_wF(w^{t_j},z)\|\le\beta\|x-z\|$ and $\|\Phi(x,w^{t_j})-\Phi(z,w^{t_j})\|\le M\|x-z\|$.
\end{assumption}
These are the first two conditions of Theorem~\ref{thm:F_L_smooth_criticl} at each observed weight. The condition on $K$, which forces $d\le m$, is not needed.

\begin{theorem}[Recovery from several rounds]\label{thm:multi_square}
Suppose Assumption~\ref{ass:multi_round} holds, $n=m$, and $\sigma_{\min}(J(x))\ge\mu>0$ for every $x\in\R^m$. Let $x_0\in\R^m$, $B:=\Gamma^2\beta^2+\Gamma M\,\mathcal L(x_0)^{1/2}$ and $0<\eta\le 1/(2B)$. Then $0\le 1-2\eta\mu^2<1$, and for every $k\ge 0$
\begin{align*}
\mathcal L(x_{k+1})\le & ~ (1-2\eta\mu^2)\,\mathcal L(x_k), \\
\|x_k-\tilde x\|\le & ~ \frac{\mathcal L(x_k)^{1/2}}{\mu}\le\frac{(1-2\eta\mu^2)^{k/2}\,\mathcal L(x_0)^{1/2}}{\mu}.
\end{align*}
In particular, $\tilde x$ is the only $x\in\R^m$ with $H(x)=y$, and the attack recovers it at a linear rate. If $n>m$, the same holds when the attacker keeps $m$ of the $n$ observed coordinates (rows of the $\bar R_j$) for which the hypothesis holds, with $H$, $J$, $\Gamma$ and $\mathcal L$ computed from the kept rows.
\end{theorem}

Proposition~\ref{prop:multi_necessary}(iii) below shows why Theorem~\ref{thm:multi_square} asks for a square system: with more rows than unknowns, a uniformly full-rank Jacobian does not make the data the only consistent input. To use all rows, the next theorem assumes more structure.

\begin{theorem}[All rows, gradients affine in a transform of the data]\label{thm:multi_flat}
Suppose that for every $j\in[T]$ and $x\in\R^m$,
\begin{align*}
\nabla_wF(w^{t_j},x)=A_j\psi(x)+q_j+e_j(x),
\end{align*}
where $A_j\in\R^{d\times m}$ and $q_j\in\R^d$. Here $\psi:\R^m\rightarrow\R^m$ is differentiable, $D\psi$ is $M_\psi$-Lipschitz, and $\sigma_{\min}(D\psi(x))\ge\nu>0$ and $\|D\psi(x)\|\le\ell$ for all $x$. Each $e_j:\R^m\rightarrow\R^d$ is differentiable with $\|De_j(x)\|\le\varepsilon_0$ for all $x$, and $De_j$ is $M_e$-Lipschitz. Let
\begin{align*}
\mathcal A:=\begin{bmatrix}\bar R_1A_1\\ \vdots\\ \bar R_TA_T\end{bmatrix}\in\R^{n\times m},
\end{align*}
assume $\alpha:=\sigma_{\min}(\mathcal A):=\min_{\|u\|=1}\|\mathcal Au\|>0$ (so $n\ge m$), and let $\varepsilon:=\Gamma\varepsilon_0$ and $x_0\in\R^m$.
\begin{enumerate}
\item[(a)] If $\varepsilon_0=0$, let $B:=\|\mathcal A\|^2\ell^2+\|\mathcal A\|M_\psi\,\mathcal L(x_0)^{1/2}$ and $0<\eta\le1/(2B)$. Then $0\le1-2\eta\alpha^2\nu^2<1$, and for every $k\ge0$
\begin{align*}
\mathcal L(x_{k+1})\le & ~ (1-2\eta\alpha^2\nu^2)\,\mathcal L(x_k), \\
\|x_k-\tilde x\|\le & ~ \frac{(1-2\eta\alpha^2\nu^2)^{k/2}\,\mathcal L(x_0)^{1/2}}{\alpha\nu}.
\end{align*}
\item[(b)] If $\varepsilon\le\nu^2\alpha^2/(4\ell\|\mathcal A\|)$, let $B:=(\ell\|\mathcal A\|+\varepsilon)^2+(\|\mathcal A\|M_\psi+\Gamma M_e)\,\mathcal L(x_0)^{1/2}$ and $0<\eta\le1/(2B)$. Then $0\le1-2\eta\alpha^2\nu^2/9<1$, and for every $k\ge0$
\begin{align*}
\mathcal L(x_{k+1})\le & ~ \Big(1-\frac{2\eta\alpha^2\nu^2}{9}\Big)\mathcal L(x_k), \\
\|x_k-\tilde x\|\le & ~ \frac{4}{3\alpha\nu}\Big(1-\frac{2\eta\alpha^2\nu^2}{9}\Big)^{k/2}\mathcal L(x_0)^{1/2}.
\end{align*}
\end{enumerate}
In both cases, $\tilde x$ is the only $x\in\R^m$ with $H(x)=y$.
\end{theorem}

\paragraph{Example.} Consider label inference: the attacker knows the inputs $z_1,\ldots,z_m$ of the client's examples, and the private data are their labels $x\in\R^m$. Take $F(w,x)=\frac12\|x-p(w)\|^2$ for a continuously differentiable prediction map $p:\R^d\rightarrow\R^m$, for example $p(w)=(h(w,z_1),\ldots,h(w,z_m))$ for a network $h$. Then $\nabla_wF(w,x)=-Dp(w)^\top x+Dp(w)^\top p(w)$. So Theorem~\ref{thm:multi_flat}(a) applies with $\psi={\rm id}$, $\nu=\ell=1$, $M_\psi=0$, $A_j=-Dp(w^{t_j})^\top$ and $e_j=0$, whenever the stacked matrix with blocks $\bar R_jDp(w^{t_j})^\top$ has rank $m$. At every $w$ with $\rank Dp(w)=d$, this $F$ meets the conditions of Theorem~\ref{thm:F_L_smooth_criticl} with $\Phi(x,w)=-Dp(w)$, $\beta=\theta_2=\|Dp(w)\|$, $\theta_1=\sigma_{\min}(Dp(w))$ and $M=0$. Replacing $x$ by $\psi(x)$ in $F$, for a map $\psi$ as in Theorem~\ref{thm:multi_flat}, gives a model that is also nonlinear in the data.

\begin{proposition}[Random sketches, adaptive weights]\label{prop:multi_random}
Let $\mathcal G_0\subseteq\mathcal G_1\subseteq\cdots\subseteq\mathcal G_T$ be $\sigma$-algebras. For every $j\in[T]$, assume the following.
\begin{itemize}
\item $\bar R_j$ is $\mathcal G_j$-measurable and independent of $\mathcal G_{j-1}$.
\item $\E[\bar R_j^\top\bar R_j]=I_d$ and $\|\bar R_j^\top\bar R_j\|\le\rho$ almost surely.
\item $A_j$ is $\mathcal G_{j-1}$-measurable with $\|A_j\|\le a$ almost surely.
\end{itemize}
Let $\mathcal K_T:=\sum_{j=1}^TA_j^\top A_j$ and let $\mathcal A$ be as in Theorem~\ref{thm:multi_flat}. Then for all $0<\kappa\le\bar K$,
\begin{align*}
& ~ \Pr\Big[\kappa I_m\preceq\mathcal K_T\preceq\bar KI_m\ \text{ and }\ \sigma_{\min}(\mathcal A)^2<\kappa/2\Big] \\
\le & ~ m\exp\Big(-\frac{\kappa^2}{10\rho a^2\bar K}\Big).
\end{align*}
\end{proposition}
Algorithm~\ref{alg:fed_learn_kstep_dp} meets the measurability and independence conditions when $w^0$ is independent of the sketching matrices and $A_j=A(w^{t_j})$ for a measurable map $A$ (as in the example), and it meets the bound on $A_j$ when, in addition, $\|A(w)\|\le a$ for all $w$. Take $\mathcal G_{j-1}:=\sigma(w^0,R_2,\ldots,R_{t_j})$ (that is, $\sigma(w^0)$ when $t_j=1$) for $j\in[T]$ and $\mathcal G_T:=\sigma(w^0,R_2,\ldots,R_{t_T+1})$. The SRHT and uniform sampling matrices (Definitions~\ref{def:srht} and~\ref{def:unif_sample}) have $\E[R^\top R]=I_d$ and $\|R^\top R\|=d/b_{\rm sketch}$, so $\rho=d/b_{\rm sketch}$.

If $\kappa^2\ge10\rho a^2\bar K\ln(m/\delta)$, then with probability at least $1-\delta$ one of two things holds. Either $\mathcal K_T$ violates $\kappa I_m\preceq\mathcal K_T\preceq\bar KI_m$, or $\sigma_{\min}(\mathcal A)^2\ge\kappa/2$, so that Theorem~\ref{thm:multi_flat}(a) applies with $\alpha^2\ge\kappa/2$ to a model as in that theorem with these $A_j$ and $\varepsilon_0=0$.

How many rounds does this need? If $\kappa I_m\preceq\mathcal K_T$ can hold at all, then $\kappa\le{\rm tr}(\mathcal K_T)/m\le T\min(d,m)a^2/m$, since each $A_j^\top A_j$ has rank at most $\min(d,m)$ and norm at most $a^2$. Since $\kappa\le\bar K$, the condition above gives $\kappa\ge10\rho a^2\ln(m/\delta)$. With $\rho=d/b_{\rm sketch}$, it can therefore hold only if $Tb_{\rm sketch}\ge10\,m\ln(m/\delta)$ when $d\le m$, and only if $Tb_{\rm sketch}\ge10\,d\ln(m/\delta)$ when $d>m$. When $d\le m$ and the weights keep $\mathcal K_T$ as well conditioned as the trace allows, that is $\kappa=\bar K=Tda^2/m$, the condition is exactly $Tb_{\rm sketch}\ge10\,m\ln(m/\delta)$. More generally, if $\kappa=Tda^2/(c_1m)$ and $\bar K=c_2Tda^2/m$ for some $c_1,c_2\ge1$, so that $\mathcal K_T$ is well conditioned up to these factors, the condition reads $Tb_{\rm sketch}\ge10c_1^2c_2\,m\ln(m/\delta)$. When $c_1$ and $c_2$ are constants, this is within a logarithmic factor of $Tb_{\rm sketch}\ge m$, which any attack needs when it observes the first $T$ rounds (Proposition~\ref{prop:multi_necessary}).

If $d=m$, $\sigma_{\min}(A(w))\ge\theta>0$ and $\|A(w)\|\le a$ for all $w$, then $\kappa:=T\theta^2$ and $\bar K:=Ta^2$ satisfy $\kappa I_m\preceq\mathcal K_T\preceq\bar KI_m$ surely, and the condition reads $T\ge10(d/b_{\rm sketch})(a/\theta)^4\ln(m/\delta)$. So this many observed rounds give $\sigma_{\min}(\mathcal A)^2\ge T\theta^2/2$ with probability at least $1-\delta$. For the example with $d=m$, these $a$ and $\theta$ are the constants $\beta$ and $\theta_1$ of Theorem~\ref{thm:F_L_smooth_criticl}, when those hold uniformly in $w$.

\begin{proposition}[What cannot be dropped]\label{prop:multi_necessary}
Run Algorithm~\ref{alg:fed_learn_kstep_dp} without additive noise, with any $K$. Fix $w^0$, all sketching matrices and the data of the other clients, and assume that $\nabla_wF$ is continuous on $\R^d\times\R^m$. The attacker's view of the first $T$ rounds consists of all weights, sketching matrices and messages of these rounds.
\begin{enumerate}
\item[(i)] If $Tb_{\rm sketch}<m$, there are two distinct values of $\tilde x$ that give the same view of the first $T$ rounds. The view of any set of rounds that ends with round $T$ is part of the view of the first $T$ rounds. So no attack recovers every data point from rounds that all come before round $\lceil m/b_{\rm sketch}\rceil$, and the first round alone never suffices when $b_{\rm sketch}<m$.
\item[(ii)] Suppose $\Phi(x,w)$ is continuous and does not depend on $w$, as for $F(w,x)=\langle w,Ax\rangle$, and $d<m$. Then for every $T$ there are two distinct values of $\tilde x$ that give the same view of the first $T$ rounds.
\item[(iii)] For $n>m$, Assumption~\ref{ass:multi_round} together with $\sigma_{\min}(J(x))\ge\mu>0$ for all $x$ does not make $\tilde x$ the only solution of $H(x)=y$. Let $m=d=1$, $F(w,x)=\cos(x-w)$, $T=2$, $\bar R_1=\bar R_2=1$, $w^{t_1}=0$ and $w^{t_2}=\pi/2$. Then Assumption~\ref{ass:multi_round} holds with $\beta=M=1$, and $\sigma_{\min}(J(x))=1$ for all $x$. But $H(\tilde x+2\pi)=H(\tilde x)$, and $\mathcal L(x)=2-2\cos(x-\tilde x)$ has a stationary point at $\tilde x+\pi$, where $\mathcal L=4$.
\end{enumerate}
\end{proposition}

\begin{remark}[One round]\label{rem:multi_one_round}
Let $T=1$ and $R:=\bar R_1$, and suppose $F$ and $R$ satisfy the conditions of Theorem~\ref{thm:sketching_convergence_of_L}. If $b_{\rm sketch}<m$, the data are not determined (Remark~\ref{rem:sketch_non_identifiable}). Otherwise $b_{\rm sketch}\le d\le m\le b_{\rm sketch}$ gives $b_{\rm sketch}=d=m$. Then $R$ and $\Phi(x,w)$ are square and $\sigma_{\min}(R\Phi(x,w)^\top)\ge\gamma_1\theta_1$. Theorem~\ref{thm:multi_square}, with $\mu=\gamma_1\theta_1$ and $\Gamma=\|R\|$, applies to the iterates of Corollary~\ref{cor:sketch_attack_rate}, whose step size is admissible since $\|R\|\le\min(\tau,\gamma_2)$. It gives $\|x_k-\tilde x\|\le(1-2\eta\gamma_1^2\theta_1^2)^{k/2}L_R(x_0)^{1/2}/(\gamma_1\theta_1)$. So under these conditions, one sketched gradient determines the data if and only if $b_{\rm sketch}=d=m$, that is, only without compression.
\end{remark}

\begin{remark}[Gaussian sketches: the exact round count]\label{rem:multi_gaussian}
Consider the setting of Theorem~\ref{thm:multi_flat} with $\varepsilon_0=0$ and $n=m$, where the attacker keeps only the rows it needs in the last observed round.
\begin{enumerate}
\item[(1)] Suppose $A_1,\ldots,A_T$ do not depend on the sketching matrices, and $\bar R_1,\ldots,\bar R_T$ are independent with densities (for example random Gaussian matrices, Definition~\ref{def:gaussian}). Then $\mathcal A$ is invertible almost surely if and only if there are subspaces $V_j\subseteq{\rm row}(A_j)$ with $\dim V_j=b_j$ and $V_1\oplus\cdots\oplus V_T=\R^m$.
\item[(2)] Suppose $d=m$, $A_j=A(w^{t_j})$ for a measurable map $A$ with $A(w)$ invertible for every $w$, and the $R_t$ of Algorithm~\ref{alg:fed_learn_kstep_dp} are random Gaussian matrices, independent of each other and of $w^0$. Then $\mathcal A$ is invertible almost surely, although the weights depend on earlier sketching matrices.
\end{enumerate}
In case (2), and in case (1) whenever the subspace condition holds, Theorem~\ref{thm:multi_flat}(a) applies almost surely with $n=m$. Taking $b_j=b_{\rm sketch}$ for $j<T$ and keeping $m-(T-1)b_{\rm sketch}$ rows in the last round, this needs only $T=\lceil m/b_{\rm sketch}\rceil$ observed rounds. When these are the first $T$ rounds, this is the smallest number allowed by Proposition~\ref{prop:multi_necessary}(i). Since $\sigma_{\min}(\mathcal A)$ can be small, this gives no rate that holds with high probability. Proposition~\ref{prop:multi_random} supplies such a rate for bounded sketches.
\end{remark}

The proofs of the results in this subsection are in Appendix~\ref{sec:multi_round:app}. Together they say that a sketch with $b_{\rm sketch}<m$ rows hides the local data in one sketched gradient, but, for models and sketches that meet the conditions above, not from an attacker that observes enough rounds. No attack recovers the data from the first $T$ rounds when $T<m/b_{\rm sketch}$. For SRHT or uniform-sampling sketches and models as in Theorem~\ref{thm:multi_flat}(a), $O((m/b_{\rm sketch})\log(m/\delta))$ observed rounds suffice with probability at least $1-\delta$ when $d\le m$ and $\mathcal K_T$ is well conditioned in the sense above for every outcome of the sketches. In general, except on an event of probability at most $\delta$, either $\mathcal K_T$ is not well conditioned or these rounds suffice.

\subsection{Discussion}

As popularized in~\cite{rpu20}, in federated learning, sketching can be applied to gradient vectors efficiently while squashing down the dimension of vectors being communicated. However, as indicated by Corollary~\ref{cor:sketch_attack_rate}, as soon as the attacker has access to the sketching operator, gradient matching still succeeds for sketched gradients. By Remark~\ref{rem:attack_iteration_counts}, to reduce the respective loss by the same factor, our iteration bound for the attack on sketched gradients is at most $(\gamma_2/\gamma_1)^2$ times our bound without sketching, before rounding up to an integer. By Remark~\ref{rem:sketch_non_identifiable}, sketching enlarges the set of inputs consistent with one observation. But by Section~\ref{sec:attack_multi_round} above, an attacker that follows a client over enough rounds recovers its data under the conditions stated there, so sketching gives no formal privacy guarantee. This motivates mechanisms that ``encode'' the gradients with a formal privacy guarantee. One can adapt a cryptography-based algorithms at the expense of higher computation cost~\cite{bik+17}, or alternatively, as we have shown in this paper, using differential privacy. We ``mask'' the gradient via Gaussian noises. Then Theorem~\ref{thm:alg_dp} bounds how much the released messages, and hence the output of any attack that sees only them, can depend on a single data point. Instead of injecting noises directly onto the gradient, one can also add noises \emph{after} applying the sketching~\cite{kkmm12,n22}. We believe this approach will also lead to interesting privacy guarantees.

\section{Conclusion}
\label{sec:discuss}
In this work, we propose the iterative sketch-based federated learning framework, which only communicates sketched updates and has a noisy, differentially private variant. Such a framework enjoys the benefits of both better privacy and lower communication cost per round. We also rigorously prove that, for the noise-free algorithm, the randomness from sketching will not introduce extra overall communication cost. For convex and strongly convex objectives, our analysis of the private algorithm also covers local stochastic gradient descent with sketched updates and unbiased minibatches of bounded variance, which is the case of zero noise (Theorems~\ref{thm:app:dp_conv_k1} and~\ref{thm:app:dp_conv_k}). We expect our approach to extend to other gradient-based optimization algorithms, such as gradient descent with momentum, because for any fixed vector $h$ the sketched and de-sketched vector $R^\top R h$ is an unbiased estimator of $h$ with second moment at most $(1+\alpha)\|h\|_2^2$. For stochastic or noisy gradients, such an analysis must also track an additive variance term, as our analysis of the private algorithm does.

By a simple modification to our algorithm with additive Gaussian noises on the gradients, we can also prove the differential privacy of our learning system by ``hiding'' the most important component in the system for guarding safety and privacy. The noisy variant still communicates only $b_\text{sketch}$ numbers per client in each round. The noise does not make the local updates biased, and its variance enters our convergence analysis as an additive term. So, for convex and strongly convex objectives, the noisy variant converges at the rates of the noise-free algorithm, up to an additive error term that depends on the privacy level and grows with $(1+\alpha)d$, so a smaller sketch raises it. At a fixed privacy budget, the noise level of the noisy variant grows with the total number $TK$ of noisy local steps, so this error does not decrease with the number of rounds. Our total communication guarantees are proved for the noise-free algorithm.

To complement our algorithmic result, we also present a gradient-matching attack. Our attack algorithm is essentially that of gradient descent, but instead of optimizing over the model parameters, it optimizes over the input data. Even though the FL algorithm tries to ``hide'' information via random projections or data structures, we show that as long as the attacker has access to the sketching operator, gradient descent on the data finds an input whose sketched gradient matches the communicated one, at a linear rate under smoothness and non-degeneracy assumptions on the model. Our attack algorithm is also computationally efficient. The input it finds need not be the private data, since under these assumptions and with $b_{\text{sketch}}<d$ a single sketched gradient does not determine the data. Observing several rounds makes recovery possible. From $T$ rounds of the noise-free algorithm with $K=1$ and independent sketching matrices, the same attack run on the sum of the per-round losses recovers the data at a linear rate, under smoothness of the model and a non-degeneracy condition on the stacked sketched Jacobians, and no attack recovers the data from the first $T$ rounds when $Tb_{\rm sketch}<m$ (Section~\ref{sec:attack_multi_round}). So, under these conditions, sketching alone delays, but does not prevent, recovery of the data.

\section*{Acknowledgement}

The authors would like to thank Lianke Qin for many helpful discussions. Yitan Wang gratefully acknowledges support from ONR Award N00014-20-1-2335. Lichen Zhang is supported by NSF grant No. CCF-1955217 and NSF grant No. DMS-2022448. The corrections in this version were found and written with the help of Claude Opus 5.5. The authors have verified the corrections and take full responsibility for the content of this paper. 

\ifdefined\isarxiv 
\bibliography{ref}
\bibliographystyle{alpha}
\else 

\bibliography{ref}
\bibliographystyle{icml2023}
\fi 

\appendix
\onecolumn
\ifdefined\isarxiv
\else


\fi
\section*{Appendix}
\paragraph{Roadmap.} We organize the appendix as follows. In section~\ref{sec:app_preli}, we introduce some notations and definitions that will be used across the appendix. In section~\ref{sec:app_prob}, we study several probability tools we will be using in the proof of cretain properties of various sketching matrices. In section~\ref{sec:app_optimzation_bg}, we lay out some key assumptions on local objective function $f_c$ and global objective function $f$, in order to proceed our discussion of convergence theory. 
In section~\ref{sec:sketching}, we discuss the $(\alpha,\beta,\delta)$-coordinate wise embedding property of~\cite{sy21} through several commonly used sketching matrices.
In section~\ref{sec:app_1step}, we give complete proofs for single-step scheme. We dedicate sections~\ref{sec:app_kstep_convex} and~\ref{sec:app_kstep_non} to illustrate formal analysis of the convergence results of Algorithm~\ref{alg:fed_learn_kstep_dp} under $k$ local steps, given different assumptions of objective function $f$. In section~\ref{sec:dp}, we introduce additive noise to make our gradients differentially private, and conclude that an SGD version of our algorithm is indeed differentially private. We also prove that, for convex and strongly convex objectives, this version converges, up to an additive error caused by the noise and the sub-sampling (section~\ref{sec:dp_conv}).
In section~\ref{sec:conditions:app}, we provide some preliminary definitions on gradient attack and elementary lemmas.
 In section~\ref{sec:from_F_to_L:app}, we show what conditions of $F$ would imply semi-smoothness and non-critical point of $L$. In section~\ref{sec:solution}, we prove with proper assumptions, $x_t$ converges to the unique optimal solution $x^*$. For the attack losses with $F$ as in Theorem~\ref{thm:F_L_smooth_criticl}, these assumptions can hold only with $d=m$ and no compression ($b_{\rm sketch}=d$). In section~\ref{sec:cost:app}, we prove $L(x_t)$ converges to $L(x^*)$ under proper conditions. In section~\ref{sec:sketching:app}, we extend the discussion by considering sketching and show what conditions of sketching would imply proper conditions of $L$. In section~\ref{sec:multi_round:app}, we prove that, under the conditions of section~\ref{sec:attack_multi_round}, observing several rounds with fresh sketches lets the attack recover the data, and that no attack recovers the data from the first $T$ rounds when $Tb_{\rm sketch}<m$.

\section{Preliminary}
\label{sec:app_preli}

For a positive integer $n$, we use $[n]$ to denote the set $\{1,2,\cdots, n\}$. 
We use $\E[\cdot]$ to denote expectation (if it exists), and use $\Pr[\cdot]$ to denote probability. For a function $f$, we use $\wt{O}(f)$ to denote $O(f \poly \log f)$. For a vector $x$,  For a vector $x$, we use $\| x \|_1:=\sum_{i} |x_i|$ to denote its $\ell_1$ norm, we use $\| x \|_2: = (\sum_{i=1}^n x_i^2)^{1/2}$ to denote its $\ell_2$ norm, we use $\| x\|_{\infty}:= \max_{i \in [n]} |x_i|$ to denote its $\ell_{\infty}$ norm. For a matrix $A$ and a vector $x$, we define $\| x \|_{A} := \sqrt{ x^\top A x }$. For a full rank square matrix $A$, we use $A^{-1}$ to denote its true inverse. For a matrix $A$, we use $A^\dagger$ to denote its pseudo-inverse. For a matrix $A$, we use $\| A \|$ to denote its spectral norm. We use $\| A \|_F:= ( \sum_{i,j} A_{i,j}^2 )^{1/2}$ to denote its Frobenius norm. We use $A^\top$ to denote the transpose of $A$. We denote $1_{\{x=l \}}$ for $l\in\R$ to be the indicator function which equals to 1 if $x=l$ and 0 otherwise. Let $f:A\rightarrow B$ and $g:C\rightarrow A$ be two functions, we use $f\circ g$ to denote the composition of functions $f$ and $g$, i.e., for any $x\in C$, $(f\circ g)(x)=f(g(x))$. Given a real symmetric matrix $A\in \R^{d\times d}$, we use $\lambda_1(A),\ldots,\lambda_d(A)$ denote its smallest to largest eigenvalues. Given a real matrix $A$, we use $\sigma_{\min}(A)$ and $\sigma_{\max}(A)$ to denote its smallest and largest singular values. 
\section{Probability}
\label{sec:app_prob}
\begin{lemma}[Chernoff bound \cite{c52}]\label{lem:chernoff}
Let $Y = \sum_{i=1}^n Y_i$, where $Y_i=1$ with probability $p_i$ and $Y_i = 0$ with probability $1-p_i$, and all $Y_i$ are independent. Let $\mu = \E[Y] = \sum_{i=1}^n p_i$. Then \\
1. $ \Pr[ Y \geq (1+\delta) \mu ] \leq \exp ( - \delta^2 \mu / 3 ) $, for all $0 < \delta \leq 1$ ; \\
2. $ \Pr[ Y \leq (1-\delta) \mu ] \leq \exp ( - \delta^2 \mu / 2 ) $, for all $ 0 < \delta < 1$. 
\end{lemma}

\begin{lemma}[Hoeffding bound \cite{h63}]\label{lem:hoeffding}
Let $Z_1, \cdots, Z_n$ denote $n$ independent bounded variables in $[a_i,b_i]$. Let $Z= \sum_{i=1}^n Z_i$, then we have
\begin{align*}
\Pr[ | Z - \E[Z] | \geq t ] \leq 2\exp \left( - \frac{2t^2}{ \sum_{i=1}^n (b_i - a_i)^2 } \right).
\end{align*}
\end{lemma}

\begin{lemma}[Bernstein inequality \cite{b24}]\label{lem:bernstein}
Let $W_1, \cdots, W_n$ be independent zero-mean random variables. Suppose that $|W_i| \leq M$ almost surely, for all $i$. Then, for all positive $t$,
\begin{align*}
\Pr \left[ \sum_{i=1}^n W_i > t \right] \leq \exp \left( - \frac{ t^2/2 }{ \sum_{j=1}^n \E[ W_j^2]  + M t /3 } \right).
\end{align*}
\end{lemma}

\begin{lemma}[Khintchine's inequality, \cite{k23,h81}]
Let $\sigma_1, \cdots, \sigma_n$ be i.i.d. sign random variables, and let $z_1, \cdots, z_n$ be real numbers, not all zero. Then for all $t > 0$
\begin{align*}
\Pr \Big[ \Big| \sum_{i=1}^n z_i \sigma_i \Big| \geq  t \| z \|_2 \Big] \leq 2\exp(-t^2/2).
\end{align*}
\end{lemma}

\begin{lemma}[Hason-wright inequality \cite{hv71,rv13}]\label{lem:hason_wright}
Let $z \in \R^n$ denote a random vector with independent entries $z_i$ with $\E[z_i]=0$ and $|z_i| \leq K$. Let $B$ be an $n \times n$ matrix. Then, for every $t \geq 0$,
\begin{align*}
    \Pr[ | z^\top B z - \E[ z^\top B z ] | > t ]\leq 2 \cdot \exp ( -c \min \{ t^2 / ( K^4 \| B \|_F^2 ) , t / ( K^2 \| B \| )  \} ).
\end{align*}
\end{lemma}

We state a well-known Lemma (see Lemma 1 on page 1325 in \cite{lm00}).
\begin{lemma}[Laurent and Massart \cite{lm00}]\label{lem:chi_square_tail}
Let $Z \sim {\cal X}_k^2$ be a chi-squared distributed random variable with $k$ degrees of freedom. Each one has zero mean and $\sigma^2$ variance. Then
\begin{align*}
\Pr[ Z - k \sigma^2 \geq ( 2 \sqrt{kt} + 2t ) \sigma^2 ] \leq \exp (-t), \\
\Pr[ k \sigma^2 - Z \geq 2 \sqrt{k t} \sigma^2 ] \leq \exp(-t).
\end{align*}
\end{lemma}

\begin{lemma}[Tail bound for sub-exponential distribution \cite{sds11}]\label{lem:sub_exp} 
We say that a random variable $X$ with mean $\mu=\E[X]$ is in $\mathrm{SE}(\sigma^2,\alpha)$ with parameters $\sigma > 0, \alpha>0$ if:
\begin{align*}
\E[e^{\lambda (X-\mu)}]\leq \exp( \lambda^2\sigma^2/2) , \quad \forall |\lambda| < 1/\alpha .
\end{align*}
Let $X \in \mathrm{SE}(\sigma^2,\alpha)$ and $\E[X]=\mu$, then for all $t\geq 0$:
\begin{align*}
\Pr[ |X-\mu|\geq t ] \leq 2\exp( - 0.5 \min\{t^2/\sigma^2,t/\alpha\} ).
\end{align*}
\end{lemma}

\begin{lemma}[Matrix Chernoff bound \cite{t11,ldfu13}]\label{lem:matrix_chernoff}
Let $\mathcal{X}$ be a finite set of positive-semidefinite matrices with dimension $d \times d$, and suppose that
\begin{align*}
	\max_{X\in\mathcal{X}} \lambda_{\max}(X) \leq B.
\end{align*}
Sample $\{X_1, \cdots, X_n \}$ uniformly at random from $\mathcal{X}$ without replacement. We define $\mu_{\min}$ and $\mu_{\max}$ as follows:
\begin{align*}
	\mu_{\min}:= n \cdot\lambda_{\min}(\E_{X \sim {\cal X}}[X])~~\mathrm{and}~~\mu_{\max}:=n \cdot\lambda_{\max}(\E_{X \sim {\cal X}}[X]).
\end{align*}
Then 
\begin{align*}
	&\Pr\Big[\lambda_{\min}(\sum_{i=1}^n X_i) \leq (1 - \delta) \mu_{\min}\Big] \leq d \cdot \exp (-\delta^2 \mu_{\min} / (2B))~\mathrm{for}~ \delta \in [0,1), \\
	&\Pr\Big[\lambda_{\max}(\sum_{i=1}^n X_i) \geq (1 + \delta) \mu_{\max}\Big] \leq d \cdot \exp{(-\delta^2 \mu_{\max} / (3B))} ~\mathrm{for}~ \delta \in [0,1].
\end{align*}
\end{lemma} 
\section{Optimization Backgrounds}
\label{sec:app_optimzation_bg}

\begin{definition}
\label{def:L_smooth}
Let $f:\R^d\rightarrow \R$ be a function, we say $f$ is \emph{$L$-smooth} if for any $x,y\in \R^d$, we have
\begin{align*}
    \|\nabla f(x)-\nabla f(y)\|_2 & \leq L\|x-y\|_2
\end{align*}
This implies that for any $x,y \in \R^d$, we have
\begin{align*}
    f(y) & \leq f(x)+\ip{y-x}{\nabla f(x)}+\frac{L}{2}\|y-x\|_2^2
\end{align*}
\end{definition}

\begin{definition}\label{def:convex}
Let $f:\R^d\rightarrow \R$ be a function, we say $f$ is convex if for any $x,y\in \R^d$, we have
\begin{align*}
    f(x) \geq & ~ f(y)+\ip{x-y}{\nabla f(y)}
\end{align*}

\end{definition}
\begin{definition}\label{def:mu_strong_convex}
Let $f:\R^d\rightarrow \R$ be a function, we say $f$ is \emph{$\mu$-strongly-convex} if for any $x,y\in \R^d$, we have
\begin{align*}
    f(y) & \geq f(x)+\ip{y-x}{\nabla f(x)}+\frac{\mu}{2}\|y-x\|_2^2
\end{align*}
This implies that for any $x,y \in \R^d$, we have
\begin{align}
    \|\nabla f(x)-\nabla f(y)\|_2 & \geq \mu \|x-y\|_2\notag
\end{align}
To see this, add the inequality above to the same inequality with $x$ and $y$ swapped, which gives $\ip{x-y}{\nabla f(x)-\nabla f(y)}\geq \mu\|x-y\|_2^2$, and apply the Cauchy-Schwarz inequality.
\end{definition}
\begin{fact}\label{fact:L_smooth_fact}
Let $f:\R^d\rightarrow \R$ be an $L$-smooth and convex function, then for any $x,y\in \R^d$, we have
\begin{align*}
    f(y)-f(x) \geq & ~ \ip{y-x}{\nabla f(x)}+\frac{1}{2L}\cdot \|\nabla f(y)-\nabla f(x)\|_2^2
\end{align*}
\end{fact}
\begin{fact}[Inequality 4.12 in~\cite{bcn18}]\label{fact:mu_upper_bound}
Let $f:\R^d\rightarrow \R$ be a $\mu$-strongly convex function. Let $x^*$ be the minimizer of $f$.  
Then for any $x\in \R^d$, we have
\begin{align*}
    f(x)-f(x^*) & \leq \frac{1}{2\mu} \|\nabla f(x)\|_2^2
\end{align*}
\end{fact} 
\section{Sketching Matrices as Coordinate-wise Embedding}\label{sec:sketching}
In this section, we discuss the $(\alpha,\beta,\delta)$-coordinate wise embedding property of~\cite{sy21} through several commonly used sketching matrices.

We consider several standard sketching matrices:
\begin{enumerate}
	\item Random Gaussian matrices.
	\item Subsampled randomized Hadamard/Fourier transform matrices \cite{ldfu13}.
  \item AMS sketch matrices \cite{ams99}, random $\{-1,+1\}$ per entry.
	\item Count-Sketch matrices \cite{ccf02}, each column only has one non-zero entry, and is $-1,+1$ half probability each.
	\item Sparse embedding matrices \cite{nn13}, each column only has $s$ non-zero entries, and each entry is $- \frac{1}{\sqrt{s}}, + \frac{1}{\sqrt{s}}$ half probability each.
	\item Uniform sampling matrices.
\end{enumerate}

\subsection{Definition}

\begin{definition}[$k$-wise independence]
$\mathcal{H} = \{h:[n]\to[l]\}$ is a $k$-wise independent hash family if for all pairwise distinct $i_1,\ldots,i_k\in[n]$ and $\forall j_1,\cdots, j_k\in[l]$, 
\begin{align*}
  \Pr_{h\in\mathcal{H}}[h(i_1)=j_1\land\cdots\land h(i_k)=j_k] = \frac{1}{l^k}.
\end{align*}
\end{definition}

\begin{definition}[Random Gaussian matrix]\label{def:gaussian}
We say $R\in\R^{b\times n}$ is a random Gaussian matrix if all entries are sampled from $\N(0,1/b)$ independently.
\end{definition}

\begin{definition}[Subsampled randomized Hadamard/Fourier transform matrix \cite{ldfu13}]\label{def:srht}
We say $R\in\R^{b\times n}$ is a subsampled randomized Hadamard transform matrix\footnote{In this case, we require $\log{n}$ to be an integer.} if it is of the form $R = \sqrt{n/b}SHD$, where $S\in\R^{b\times n}$ is a random matrix whose rows are $b$ uniform samples (without replacement) from the standard basis of $\R^n$, $H\in\R^{n\times n}$ is a normalized Walsh-Hadamard matrix, and $D\in\R^{n\times n}$ is a diagonal matrix whose diagonal elements are i.i.d. Rademacher random variables.
\end{definition}

\begin{definition}[AMS sketch matrix \cite{ams99}]\label{def:ams}
Let $h_1,h_2,\cdots,h_b$ be $b$ random hash functions picking from a 4-wise independent hash family $\mathcal{H} = \{h:[n]\to\{-\frac{1}{\sqrt{b}},+\frac{1}{\sqrt{b}}\}\}$. Then $R\in\R^{b\times n}$ is a AMS sketch matrix if we set $R_{i,j} = h_i(j)$.
\end{definition}

\begin{definition}[Count-sketch matrix \cite{ccf02}]\label{def:cs}
Let $h : [n] \rightarrow [b]$  be a random $2$-wise independent hash function and $\sigma : [n] \rightarrow \{-1,+1\}$ be a random $4$-wise independent hash function. Then $R\in\R^{b\times n}$ is a count-sketch matrix if we set $R_{h(i),i} = \sigma(i)$ for all $i\in[n]$ and other entries to zero.
\end{definition}

\begin{definition}[Sparse embedding matrix I \cite{nn13}]\label{def:sparse_1}
We say $R\in\R^{b\times n}$ is a sparse embedding matrix with parameter $s$ if each column has exactly $s$ non-zero elements being $\pm 1/\sqrt{s}$ uniformly at random, whose locations are picked uniformly at random without replacement (and independent across columns)
\footnote{For our purposes the signs need only be $O(\log d)$-wise independent, and each column can be specified by a $O(\log d)$-wise
independent permutation, and the seeds specifying the permutations in different columns need only
be $O(\log d)$-wise independent.}.
\end{definition}

\begin{definition}[Sparse embedding matrix II \cite{nn13}]\label{def:sparse_2}
Let  $h : [n]\times[s] \rightarrow [b/s]$ be a a ramdom 2-wise independent hash function and $\sigma:[n]\times[s]\to \{-1,1\}$ be a 4-wise independent. Then $R\in\R^{b\times n}$ is a sparse embedding matrix II with parameter $s$ if we set $R_{(j-1)b/s+h(i,j),i} = \sigma(i,j)/\sqrt{s}$ for all $(i,j)\in[n]\times [s]$ and all other entries to zero.\footnote{This definition has the same behavior as sparse embedding matrix I for our purpose.}
\end{definition}

\begin{definition}[Uniform sampling matrix]\label{def:unif_sample}
We say $R\in\R^{b\times n}$ is a uniform sampling matrix if it is of the form $R = \sqrt{n/b}SD$, where $S\in\R^{b\times n}$ is a random matrix whose rows are $b$ uniform samples (without replacement) from the standard basis of $\R^n$, and $D\in\R^{n\times n}$ is a diagonal matrix whose diagonal elements are i.i.d. Rademacher random variables.
\end{definition}

\subsection{Coordinate-wise Embedding}

We define coordinate-wise embedding as follows
\begin{definition}[$(\alpha,\beta,\delta)$-coordinate-wise embedding]
We say a randomized matrix $R \in \R^{b \times n}$ satisfying $(\alpha,\beta,\delta)$-coordinate wise embedding if
\begin{align*}
1. & ~ \E_{ R \sim \Pi } [ g^\top R^\top R h ] = g^\top h, \\
2. & ~ \E_{ R \sim \Pi } [ ( g^\top R^\top R h )^2 ] \leq (g^\top h)^2 + \frac{\alpha}{b} \|g\|_2^2 \| h \|_2^2, \\
3. & ~ \Pr_{ R \sim \Pi } \left[ | g^\top R^\top R h  - g^\top h  | \geq \frac{\beta}{ \sqrt{b} } \| g \|_2 \| h \|_2 \right] \leq \delta.
\end{align*}
\end{definition}
\begin{remark}
Given a randomized matrix $R \in \R^{b \times n}$ satisfying $(\alpha,\beta,\delta)$-coordinate wise embedding and any orthogonal projection $P\in\R^{n\times n}$, above definition implies
\begin{align*}
1. & ~ \E_{ R \sim \Pi } [ P R^\top R h ] = Ph, \\
2. & ~ \E_{ R \sim \Pi } [ ( P R^\top R h )_i^2 ] \leq (P h)_i^2 + \frac{\alpha}{b} \| h \|_2^2, \\
3. & ~ \Pr_{ R \sim \Pi } \left[ | ( P R^\top R h )_i - ( P h )_i | \geq \frac{\beta}{ \sqrt{b} } \| h \|_2 \right] \leq \delta.
\end{align*}
since $\|P\|_2 \leq 1$ implies $\|P_{i,:}\|_2 \leq 1$ for all $i\in[n]$.
\end{remark}

\subsection{Expectation and Variance}

\begin{lemma}\label{lem:expectation}
Let $R \in \R^{b \times n}$ denote any of the random matrix in Definition~\ref{def:gaussian}, \ref{def:srht}, \ref{def:ams}, \ref{def:sparse_1}, \ref{def:sparse_2},~\ref{def:unif_sample}. 
Then for any fixed vector $h\in \R^n$ and any fixed vector $g \in \R^{n}$, the following properties hold:
\begin{align*}
& \E_{R \sim \Pi}[ g^\top R^\top R h ] = g^\top h
\end{align*}
\end{lemma}

\begin{proof}
\begin{align*}
\E_{R \sim \Pi}[ g^\top R^\top R h ]
= g^\top \E_{R \sim \Pi}[R^{\top}R] h
= g^\top I h
= g^\top h.
\end{align*}
\end{proof}

\begin{lemma}\label{lem:var_srht_ams}
Let $R \in \R^{b \times n}$ denote a subsampled randomized Hadamard transform or AMS sketch matrix as in Definition~\ref{def:srht},~\ref{def:ams}.
Then for any fixed vector $h\in \R^n$ and any fixed vector $g \in \R^{n}$, the following properties hold:
\begin{align*}
& \E_{R \sim \Pi}[ ( g^\top R^\top R h )^2 ] \leq (g^\top h)^2+ \frac{2}{b}\|g\|_2^2 \cdot \|h\|_2^2.
\end{align*}
\end{lemma}

\begin{proof}

If $\E_a [a] = b$, it is easy to see that
\begin{align*}
\E_a[ (a - b)^2] = \E_a[ a^2 - 2 a b + b^2 ] = \E_a[ a^2 - b^2 ] 
\end{align*}

We can rewrite it as follows:
\begin{align*}
 \E_{R \sim \Pi}[ ( g^\top R^\top R h )^2 - ( g^\top h )^2] = \E_{R \sim \Pi}[( g^\top ( R^\top R - I ) h )^2 ],
\end{align*}

It can be bounded as follows:
\begin{align*}
& ~ \E_{R \sim \Pi}[( g^\top (R^\top R-I) h )^2]\\
= & ~ \E_{R \sim \Pi}\left[\left( \sum_{k=1}^b  ( R g )_k  (R h)_k -  g^\top h \right)^2\right] \\
= & ~ \E_{R \sim \Pi}\left[\left( \sum_{k=1}^b \sum_{i=1}^n R_{k,i} g_i \cdot  \sum_{j \in [n] \backslash \{ i \} } R_{k,j} h_j \right)^2\right] \\
= & ~ \E_{R \sim \Pi}\left[\left( \sum_{k=1}^b \sum_{i=1}^n R_{k,i} g_{i} \cdot \sum_{j \in [n] \backslash \{ i \} } R_{k,j} h_j \right) \cdot \left( \sum_{k'=1}^b \sum_{i'=1}^n R_{k',i'} g_{i'} \cdot \sum_{j'\in [n] \backslash \{ i' \} } R_{k',j'} h_{j'}\right)\right]\\
= & ~ \sum_{k=1}^b \sum_{k'=1}^b \sum_{i \neq j} \sum_{i' \neq j'} g_i h_j g_{i'} h_{j'} \cdot \E_{R \sim \Pi}[ R_{k,i} R_{k,j} R_{k',i'} R_{k',j'} ] \\
= & ~ \Big( b\cdot \frac{1}{b^2} + b(b-1)\kappa \Big) \sum_{i \neq j} ( g_{i}^2 h_j^2 + g_{i} h_{i} g_j h_j ) \\
\leq & ~ \frac{1}{b} \sum_{i \neq j} ( g_{i}^2 h_j^2 + g_{i} h_{i} g_j h_j ) \\
\leq & ~ \frac{2}{b} \| g \|_2^2 \|h\|_2^2,
\end{align*}
where $\kappa$ is defined below. The second step follows from $R_{k,i}^2=1/b$, $\forall k,i \in [b] \times [n]$, so that the terms with $i=j$ sum to $g^\top h$.

For the fifth step, fix $i \neq j$ and $i' \neq j'$. The expectation $\E[ R_{k,i} R_{k,j} R_{k',i'} R_{k',j'} ]$ vanishes unless $(i',j')=(i,j)$ or $(i',j')=(j,i)$. For SRHT, this holds because the signs $D_{l,l}$ are independent of each other and of $S$, so a sign that appears only once has mean zero. For AMS with $k=k'$, it holds by the 4-wise independence of $h_k$. For AMS with $k\neq k'$, the expectation is always zero, because the rows are independent and $\E[R_{k,i}R_{k,j}]=0$ for $i\neq j$. When $(i',j')\in\{(i,j),(j,i)\}$, the product equals $R_{k,i}R_{k,j}R_{k',i}R_{k',j}$. If $k=k'$, it equals $R_{k,i}^2R_{k,j}^2=1/b^2$. If $k\neq k'$, we denote its expectation by $\kappa$. For AMS, $\kappa=0$. For SRHT, write $R_{k,i}=\sqrt{n/b}\,H_{s_k,i}D_{i,i}$, where $s_1,\ldots,s_b$ are the sampled rows. Then $R_{k,i}R_{k,j}R_{k',i}R_{k',j}=(n/b)^2 H_{s_k,i}H_{s_k,j}H_{s_{k'},i}H_{s_{k'},j}$. Since the rows are sampled without replacement, $(s_k,s_{k'})$ is uniform over the $n(n-1)$ ordered pairs of distinct rows. The columns of $H$ are orthonormal, so $\sum_{r=1}^n H_{r,i}H_{r,j}=0$ for $i\neq j$, and $H_{r,i}^2=1/n$. Hence
\begin{align*}
\kappa = & ~ \frac{n^2}{b^2}\cdot\frac{1}{n(n-1)}\sum_{r\neq r'} H_{r,i}H_{r,j}H_{r',i}H_{r',j} \\
= & ~ \frac{n^2}{b^2}\cdot\frac{1}{n(n-1)}\Big( \Big(\sum_{r=1}^n H_{r,i}H_{r,j}\Big)^2 - \sum_{r=1}^n H_{r,i}^2H_{r,j}^2 \Big) \\
= & ~ \frac{n^2}{b^2}\cdot\frac{1}{n(n-1)}\Big( 0 - n\cdot\frac{1}{n^2} \Big)
= -\frac{1}{b^2(n-1)}.
\end{align*}
Summing over the $b$ pairs with $k=k'$ and the $b(b-1)$ pairs with $k\neq k'$ gives the coefficient $b\cdot\frac{1}{b^2}+b(b-1)\kappa$ in the fifth step. It equals $\frac{1}{b}$ for AMS, and for SRHT it equals
\begin{align*}
\frac{1}{b}-\frac{b-1}{b(n-1)} = \frac{n-b}{b(n-1)}.
\end{align*}
Thus the variance of the SRHT estimate is exactly $\frac{n-b}{b(n-1)}\sum_{i \neq j} ( g_{i}^2 h_j^2 + g_{i} h_{i} g_j h_j )$. In both cases the coefficient lies in $[0,\frac{1}{b}]$, since $1\leq b\leq n$. The sixth step uses this together with $\sum_{i \neq j} ( g_{i}^2 h_j^2 + g_{i} h_{i} g_j h_j ) = \frac{1}{2}\sum_{i\neq j}(g_ih_j+g_jh_i)^2\geq 0$. The last step follows from $\sum_{i\neq j}g_i^2h_j^2\leq \|g\|_2^2\|h\|_2^2$ and $\sum_{i\neq j}g_ih_ig_jh_j=(g^\top h)^2-\sum_{i=1}^n g_i^2h_i^2\leq (g^\top h)^2\leq \|g\|_2^2\|h\|_2^2$, by the Cauchy-Schwartz inequality.

Therefore, 
\begin{align*}
 \E_{R \sim \Pi}[ ( g^\top R^\top R h )^2 - (g^\top h)^2] 
 =  \E_{R \sim \Pi}[( g^\top (R^\top R-I) h )^2]
\leq  \frac{2}{b}\| g \|_2^2 \|h\|_2^2.
\end{align*}
\end{proof}

\begin{lemma}\label{lem:var_gaussian}
Let $R \in \R^{b \times n}$ denote a random Gaussian matrix as in Definition~\ref{def:gaussian}.
Then for any fixed vector $h\in \R^n$ and any fixed vector $g \in \R^{n}$, the following properties hold:
\begin{align*}
& \E_{R \sim \Pi}[ ( g^\top R^\top R h )^2 ] \leq (g^\top h)^2+ \frac{3}{b}\|g\|_2^2 \cdot \|h\|_2^2.
\end{align*}
\end{lemma}
\begin{proof}
Note
\begin{align*}
& ~ \E_{R \sim \Pi}[( g^\top R^\top R h )^2]\\
= & ~ \E_{R \sim \Pi}\left[\left( \sum_{k=1}^b \sum_{i=1}^n R_{k,i} g_i \cdot \sum_{j=1}^n R_{k,j} h_j\right)^2\right] \\
= & ~ \E_{R \sim \Pi}\left[\left( \sum_{k=1}^b \sum_{i=1}^n R_{k,i} g_{i} \cdot \sum_{j=1}^n R_{k,j} h_j \right) \cdot \left( \sum_{k'=1}^b \sum_{i'=1}^n R_{k',i'} g_{i'} \cdot \sum_{j'=1}^n R_{k',j'} h_{j'}\right)\right]\\
= & ~ \E_{R \sim \Pi}\Big[
\left(\sum_{k=1}^b\sum_{k'\in [b] \backslash \{k\}}\sum_{i=1}^n \sum_{i'=1}^n R_{k,i}^2 R_{k',i'}^2 g_i h_i g_{i'} h_{i'}\right)
+ \left(\sum_{k=1}^b \sum_{i=1}^n R_{k,i}^4 g_i^2 h_i^2\right) \\
&~ + \left( \sum_{k=1}^b \sum_{i=1}^n \sum_{j \in [n] \backslash \{ i \} } R_{k,i}^2 R_{k,j}^2 g_{i}^2 h_j^2\right)
+ \left(\sum_{k=1}^b\sum_{i=1}^n\sum_{i'\in [n] \backslash \{i\}}^n R_{k,i}^2 R_{k,i'}^2 g_i h_i g_{i'} h_{i'}\right) \\
&~ +  \left( \sum_{k=1}^b \sum_{i=1}^n \sum_{j \in [n] \backslash \{ i \} } R_{k,i}^2 R_{k,j}^2 g_{i} h_{j} g_j h_i\right) \Big]\\
= & ~ \frac{b-1}{b}\sum_{i=1}^n \sum_{i'=1}^n g_i h_i g_{i'} h_{i'} 
+ \frac{3}{b}\sum_{i=1}^n g_i^2 h_i^2 \\
& ~ + \frac{1}{b}  \sum_{i =1}^n \sum_{j \in [n] \backslash [i]} g_{i}^2 h_j^2 
+ \frac{1}{b} \sum_{i =1}^n \sum_{i' \in [n] \backslash [i]}  g_{i} h_{i} g_{i'} h_{i'}  
+ \frac{1}{b} \sum_{i =1}^n \sum_{j \in [n] \backslash [i]} g_{i} h_{j} g_j h_i \\
\leq & ~ (g^\top h)^2 + \frac{3}{b} \| g \|_2^2 \|h\|_2^2,
\end{align*}
where the third step follows from that for independent entries of a random Gaussian matrix, $\E[ R_{k,i} R_{k,j} R_{k',i'} R_{k',j'} ] \neq 0$ only if 1. $k \neq k'$, $i=j$, $i'=j'$, or 2. $k=k'$, $i=i'=j=j'$, or 3. $k=k'$, $i=i' \neq j=j'$, or 4. $k=k'$, $i=j \neq i'=j'$, or 5. $k=k'$, $i=j' \neq i' = j$, the fourth step follows from $\E[R_{k,i}^2]=1/b$ and $\E[R_{k,i}^4]=3/b^2$, and the last step follows from Cauchy-Schwartz inequality.
\end{proof}

\begin{lemma}\label{lem:var_count_sketch}
Let $R \in \R^{b \times n}$ denote a count-sketch matrix as in Definition~\ref{def:cs}.
Then for any fixed vector $h\in \R^n$ and any fixed vector $g \in \R^{n}$, the following properties hold:
\begin{align*}
& \E_{R \sim \Pi}[ ( g^\top R^\top R h )^2 ] \leq (g^\top h)^2 + \frac{3}{b} \| g \|_2^2 \|h\|_2^2.
\end{align*}
\end{lemma}
\begin{proof}
Note
\begin{align*}
& ~ \E_{R \sim \Pi}[( g^\top R^\top R h )^2]\\
= & ~ \E_{R \sim \Pi}\left[\left( \sum_{k=1}^b \sum_{i=1}^n R_{k,i} g_i \sum_{j=1}^n R_{k,j} h_j\right)^2\right] \\
= & ~ \E_{R \sim \Pi}\left[\left( \sum_{k=1}^b \sum_{i=1}^n R_{k,i} g_{i}  \sum_{j=1}^n R_{k,j} h_j \right) \cdot \left( \sum_{k'=1}^b \sum_{i'=1}^n R_{k',i'} g_{i'} \sum_{j'=1}^n R_{k',j'} h_{j'}\right)\right]\\
= & ~ \E_{R \sim \Pi}\Big[
\left(\sum_{k=1}^b\sum_{k'\in [b] \backslash \{k\}}\sum_{i=1}^n \sum_{i' \in [n] \backslash \{i\}}^n R_{k,i}^2 R_{k',i'}^2 g_i h_i g_{i'} h_{i'}\right)
+ \left(\sum_{k=1}^b \sum_{i=1}^n R_{k,i}^4 g_i^2 h_i^2\right) \\
&~ + \left( \sum_{k=1}^b \sum_{i=1}^n \sum_{j \in [n] \backslash \{ i \} } R_{k,i}^2 R_{k,j}^2 g_{i}^2 h_j^2\right)
+ \left(\sum_{k=1}^b\sum_{i=1}^n\sum_{i'\in [n] \backslash \{i\}}^n R_{k,i}^2 R_{k,i'}^2 g_i h_i g_{i'} h_{i'}\right) \\
&~ +  \left( \sum_{k=1}^b \sum_{i=1}^n \sum_{j \in [n] \backslash \{ i \} } R_{k,i}^2 R_{k,j}^2 g_{i} h_{j} g_j h_i\right) \Big]\\
= & ~ \frac{b-1}{b}\sum_{i=1}^n\sum_{i' \in [n] \backslash i}g_i h_i g_{i'} h_{i'} 
+ \sum_{i=1}^n g_i^2 h_i^2 \\
& ~ + \frac{1}{b} \sum_{i =1}^n \sum_{j \in [n] \backslash \{i\}} g_{i}^2 h_j^2 
+ \frac{1}{b} \sum_{i =1}^n \sum_{i' \in [n] \backslash \{i\}} g_{i} h_{i} g_{i'} h_{i'} 
+ \frac{1}{b} \sum_{i =1}^n \sum_{j \in [n] \backslash \{i\}} g_{i} h_{j} g_j h_i \\
\leq & ~ (g^\top h)^2 + \frac{3}{b} \| g \|_2^2 \|h\|_2^2,
\end{align*}
where in the third step we are again considering what values of $k,k',i,i',j,j'$ that makes \\$\E[ R_{k,i} R_{k,j} R_{k',i'} R_{k',j'} ] \neq 0$. Since the hash function $\sigma(\cdot)$ of the count-sketch matrix is 4-wise independent, $\forall k,k'$, when $i\neq i'\neq j\neq j'$, or $i= i' = j \neq j'$ (and the other 3 symmetric cases), we have that $\E[ R_{k,i} R_{k,j} R_{k',i'} R_{k',j'} ] = 0$. Since the count-sketch matrix has only one non-zero entry in every column, when $k\neq k'$, if $i=i'$ or $i=j'$ or $j=i'$ or $j=j'$, we also have $\E[ R_{k,i} R_{k,j} R_{k',i'} R_{k',j'} ] = 0$. Thus we only need to consider the cases: 1. $k \neq k'$, $i=j \neq i'=j'$, or 2. $k=k'$, $i=i'=j=j'$, or 3. $k=k'$, $i=i' \neq j=j'$, or 4. $k=k'$, $i=j \neq i'=j'$, or 5. $k=k'$, $i=j' \neq i' = j$. And the fourth step follows from $\E[R_{k,i}^2]=1/b$ and $\E[R_{k,i}^4]=1/b$, and the last step follows from Cauchy-Schwartz inequality.
\end{proof}

\begin{lemma}\label{lem:var_sparse_embedding}
Let $R \in \R^{b \times n}$ denote a sparse embedding matrix as in Definition~\ref{def:sparse_1},~\ref{def:sparse_2}.
Then for any fixed vector $h\in \R^n$ and any fixed vector $g \in \R^{n}$, the following properties hold:
\begin{align*}
2. & \E_{R \sim \Pi}[ ( g^\top R^\top R h )^2 ] \leq (g^\top h)^2+ \frac{2}{b}\|g\|_2^2 \cdot \|h\|_2^2.
\end{align*}
\end{lemma}
\begin{proof}
Note
\begin{align*}
& ~ \E_{R \sim \Pi}[( g^\top R^\top R h )^2]\\
= & ~ \E_{R \sim \Pi}\left[\left( \sum_{k=1}^b \sum_{i=1}^n R_{k,i} g_i \sum_{j=1}^n R_{k,j} h_j\right)^2\right] \\
= & ~ \E_{R \sim \Pi}\left[\left( \sum_{k=1}^b \sum_{i=1}^n R_{k,i} g_{i}  \sum_{j=1}^n R_{k,j} h_j \right) \cdot \left( \sum_{k'=1}^b \sum_{i'=1}^n R_{k',i'} g_{i'} \sum_{j'=1}^n R_{k',j'} h_{j'}\right)\right]\\
= & ~ \E_{R \sim \Pi}\Big[\left( \sum_{k=1}^b \sum_{i=1}^n R_{k,i}^2 g_{i}^2 \sum_{j \in [n] \backslash \{ i \} } R_{k,j}^2 h_j^2\right) + \left( \sum_{k=1}^b \sum_{i=1}^n R_{k,i}^2 g_{i} h_{i} \sum_{j \in [n] \backslash \{ i \} } R_{k,j}^2 g_j h_j\right) \\
& ~ + \left(\sum_{k}\sum_{i\neq i'}R_{k,i}^2 R_{k,i'}^2 g_i h_i g_{i'} h_{i'}\right) + \left(\sum_{k}\sum_{i} R_{k,i}^4 g_i^2 h_i^2\right) + \left(\sum_{k\neq k'}\sum_{i\neq i'} R_{k,i}^2 R_{k',i'}^2 g_i h_i g_{i'} h_{i'}\right) \\
& ~ + \left( \sum_{k\neq k'}\sum_{i} R_{k,i}^2 R_{k',i}^2 g_i^2 h_i^2 \right) \Big]\\
= & ~ \frac{1}{b}  \sum_{i \neq j} g_{i}^2 h_j^2 + \frac{1}{b} \sum_{i\neq j} g_{i} h_{i} g_j h_j + \frac{1}{b} \sum_{i\neq i'} g_{i} h_{i} g_{i'} h_{i'} + \frac{1}{s}\sum_{i}g_i^2 h_i^2 + \frac{b-1}{b}\sum_{i\neq i'}g_i h_i g_{i'} h_{i'} + \frac{s-1}{s}\sum_{i}g_i^2 h_i^2 \\
\leq & ~ (g^\top h)^2 + \frac{2}{b} \| g \|_2^2 \|h\|_2^2,
\end{align*}
where the third step follows from the fact that the sparse embedding matrix has independent columns and $s$ non-zero entry in every column, the fourth step follows from $\E[R_{k,i}^2]=1/b$, $\E[R_{k,i}^4]=1/(bs)$, and $\E[R_{k,i}^2 R_{k',i}^2] = \frac{s(s-1)}{b(b-1)}\cdot \frac{1}{s^2}, \forall k \neq k'$ and the last step follows from Cauchy-Schwartz inequality.
\end{proof}

\begin{lemma}\label{lem:var_uniform_sample}
Let $R \in \R^{b \times n}$ denote a uniform sampling matrix as in Definition~\ref{def:unif_sample}.
Then for any fixed vector $h\in \R^n$ and any fixed vector $g \in \R^{n}$, the following properties hold:
\begin{align*}
2. & \E_{R \sim \Pi}[ ( g^\top R^\top R h )^2 ] \leq (g^\top h)^2 + \frac{n}{b} \| g \|_2^2 \|h\|_2^2.
\end{align*}
\end{lemma}
\begin{proof}
Note
\begin{align*}
& ~ \E_{R \sim \Pi}[( g^\top R^\top R h )^2]\\
= & ~ \E_{R \sim \Pi}\left[\left( \sum_{k=1}^b \sum_{i=1}^n R_{k,i} g_i \sum_{j=1}^n R_{k,j} h_j\right)^2\right] \\
= & ~ \E_{R \sim \Pi}\left[\left( \sum_{k=1}^b \sum_{i=1}^n R_{k,i} g_{i}  \sum_{j=1}^n R_{k,j} h_j \right) \cdot \left( \sum_{k'=1}^b \sum_{i'=1}^n R_{k',i'} g_{i'} \sum_{j'=1}^n R_{k',j'} h_{j'}\right)\right]\\
= & ~ \E_{R \sim \Pi} \left[ \left(\sum_{k}\sum_{i} R_{k,i}^4 g_i^2 h_i^2\right) + \left(\sum_{k\neq k'}\sum_{i\neq i'} R_{k,i}^2 R_{k',i'}^2 g_i h_i g_{i'} h_{i'}\right) \right]\\
= & ~ \frac{n}{b}\sum_{i}g_i^2 h_i^2 + \frac{(b-1)n}{(n-1)b}\sum_{i\neq i'}g_i h_i g_{i'} h_{i'} \\
\leq & ~ (g^\top h)^2 + \frac{n}{b} \| g \|_2^2 \|h\|_2^2,
\end{align*}
where the third step follows from the fact that the random sampling matrix has one non-zero entry in every row, the fourth step follows from $\E[R_{k,i}^2R_{k',i'}^2]=n/ ( (n-1)b^2 )$ for $k\neq k',~i\neq i'$ and $\E[R_{k,i}^4]=n/b^2$.
\end{proof}
\begin{remark}
Lemma~\ref{lem:var_uniform_sample} indicates that uniform sampling fails in bounding variance in some sense, since the upper bound give here involves $n$.
\end{remark}

\subsection{Bounding Inner Product}

\begin{lemma}[Gaussian]\label{lem:gaussian_ip}
Let $n\geq 2$, $\delta\in(0,1)$, and let $R \in \R^{b\times n}$ be a random Gaussian matrix (Definition~\ref{def:gaussian}). There are absolute constants $C,C'>0$ such that, if $b\geq C'\log(n/\delta)$, then we have:
\begin{align*}
 \Pr \Big[ \max_{i \neq j} |\langle {R}_{*,i},{R}_{*,j} \rangle| \geq C\sqrt{\frac{\log(n/\delta)}{b}} \Big] \leq O(\delta).
\end{align*}
\end{lemma}
\begin{proof}

Note for $i \neq j$, ${R}_{*,i}, {R}_{*,j} \sim \N( 0, \frac{1}{b}I_b )$ are two independent Gaussian vectors. Let $z_k={R}_{k,i} {R}_{k,j} $ and $z = \langle {R}_{*,i} , {R}_{*,j}  \rangle$. Then we have for any $|\lambda|\leq b/2$,
\begin{align*}
\E[ e^{\lambda z_k} ] = \frac{ 1 }{ \sqrt{ 1 - \lambda^2 / b^2 } } \leq \exp( \lambda^2 / b^2 ),
\end{align*}
where the first step follows from $z_k = \frac{1}{4} (R_{k,i} + R_{k,j})^2 - \frac{1}{4} (R_{k,i} - R_{k,j})^2 = \frac{1}{2b} (Q_1-Q_2)$, where $Q_1 = \frac{b}{2}(R_{k,i} + R_{k,j})^2$ and $Q_2 = \frac{b}{2}(R_{k,i} - R_{k,j})^2$ are independent $\chi_1^2$ variables (the Gaussian variables $R_{k,i} \pm R_{k,j}$ are uncorrelated with variance $2/b$), and $\E[e^{\lambda Q}] = \frac{1}{\sqrt{1-2\lambda}}$ for any $Q\sim \chi_1^2$ and $\lambda<1/2$, so that $\E[ e^{\lambda z_k} ] = (1-\lambda/b)^{-1/2}(1+\lambda/b)^{-1/2}$. The second step follows from $-\frac{1}{2}\log(1-x)\leq x$ for $x=\lambda^2/b^2\in[0,1/4]$.

Since $\E[z_k]=0$ and $\exp(\lambda^2/b^2)=\exp(\lambda^2\cdot\frac{2}{b^2}/2)$, this implies $z_k \in \text{SE}( 2/b^2 , 2/b )$ is a sub-exponential random variable. Since $z_1,\ldots,z_b$ are independent, $\E[e^{\lambda z}]=\prod_{k=1}^b\E[e^{\lambda z_k}]\leq \exp(\lambda^2\cdot\frac{2}{b}/2)$ for $|\lambda|<b/2$, that is, $z = \sum_{k=1}^b z_k\in \text{SE}(2/b, 2/b)$. By sub-exponential concentration Lemma~\ref{lem:sub_exp} with $\sigma^2=\alpha=2/b$, we have
\begin{align*}
\Pr [ |z| \geq t ] \leq 2\exp\Big( -\frac{1}{2}\min\Big\{\frac{bt^2}{2},\frac{bt}{2}\Big\} \Big) = 2\exp( -b t^2 / 4 )
\end{align*}
for $0<t\leq 1$. Picking $t = 2\sqrt{  \log(2n^2/\delta) / b }$, we have $bt^2/4=\log(2n^2/\delta)$, so
\begin{align*}
\Pr [ |z| \geq t ] \leq 2\exp(-\log(2n^2/\delta)) = \delta / n^2,
\end{align*}
provided that $t\leq 1$. Since $n\geq 2$ and $\delta<1$, we have $2n^2/\delta\leq n^3/\delta^3$, hence $\log(2n^2/\delta)\leq 3\log(n/\delta)$ and $t\leq 2\sqrt{3\log(n/\delta)/b}$. Take $C=2\sqrt{3}$ and $C'=12$. Then $b\geq C'\log(n/\delta)$ gives $t\leq 1$, and $t\leq C\sqrt{\log(n/\delta)/b}$, so
\begin{align*}
\Pr \Big[ |\langle {R}_{*,i} , {R}_{*,j}\rangle| \geq  C\sqrt{\frac{\log(n/\delta)}{b}} \Big] \leq \Pr [ |z| \geq t ] \leq \delta / n^2.
\end{align*}
Taking the union bound over the fewer than $n^2$ pairs $(i,j) \in [n] \times [n]$ with $i\neq j$, the probability in the lemma is at most $\delta$, which completes the proof.
\end{proof}

\begin{lemma}[SRHT]\label{lem:srht_ip} 
Let $n\geq 2$, $\delta\in(0,1)$, and let $R \in \R^{b \times n}$ be a subsample randomized Hadamard transform (Definition~\ref{def:srht}). There is an absolute constant $C>0$ such that we have:
\begin{align*}
\Pr \Big[ \max_{i\neq j}|\langle {R}_{*,i} , {R}_{*,j} \rangle| \geq  C\sqrt{\frac{ \log(n/\delta) }{b}} \Big] \leq O(\delta).
\end{align*}
\end{lemma}
\begin{proof}
For fixed $i \neq j$, let $X = [R_{*,i},R_{*,j}] \in \R^{b \times 2}$. Then $X^\top X = \sum_{k=1}^b G_k$, where
\begin{align*}
G_k = [R_{k,i},R_{k,j}]^\top [R_{k,i},R_{k,j}] =
\begin{bmatrix}
\frac{1}{b} & R_{k,i}R_{k,j} \\
R_{k,i}R_{k,j} & \frac{1}{b}
\end{bmatrix}.
\end{align*}
Note the eigenvalues of $G_k$ are $0$ and $\frac{2}{b}$, since $|R_{k,i}R_{k,j}|=\frac{1}{b}$. Conditioned on $D$, the matrices $G_1,\ldots,G_b$ are $b$ samples without replacement from the set $\{x_rx_r^\top : r\in[n]\}$, where $x_r=\sqrt{n/b}\,((HD)_{r,i},(HD)_{r,j})^\top$. The average of this set is $\frac{1}{b}I_2$, because the columns of $HD$ are orthonormal, so $\E[X^{\top} X] = b \cdot \E[G_k] = I_2$ for all $k\in[b]$. Thus, we can apply matrix Chernoff bound~\ref{lem:matrix_chernoff} to $X^\top X$ with dimension $2$, $B=\frac{2}{b}$ and $\mu_{\min}=\mu_{\max}=1$, and we have
\begin{align*}
  &\Pr\Big[\lambda_{\min}(X^\top X) \leq 1-t\Big] \leq 2\exp\Big(-\frac{t^2}{2\cdot 2/b}\Big) = 2\exp{(-{t^2b}/{4})}~\mathrm{for}~t\in[0,1),~\mathrm{and}\\
  &\Pr\Big[\lambda_{\max}(X^\top X) \geq 1+t\Big] \leq 2\exp\Big(-\frac{t^2}{3\cdot 2/b}\Big) = 2\exp{(-{t^2b}/{6})} ~\mathrm{for}~t\in[0,1].
\end{align*}
which implies, for $t\in[0,1)$, that the eigenvalues of $X^\top X$ are between $[1-t,1+t]$ with probability at least $1-2\exp{(-\frac{t^2b}{4})}-2\exp{(-\frac{t^2b}{6})}\geq 1-4\exp{(-\frac{t^2b}{6})}$. So the eigenvalues of $X^\top X - I_2$ are between $[-t,t]$ with the same probability. Note
\begin{align*}
 X^\top X-I_2 = \begin{bmatrix} 0 & \langle {R}_{*,i} , {R}_{*,j} \rangle \\
 \langle {R}_{*,i} , {R}_{*,j} \rangle & 0 \end{bmatrix},
\end{align*}
since $\|R_{*,i}\|_2^2=b\cdot\frac{1}{b}=1$. Its spectral norm is $|\langle {R}_{*,i} , {R}_{*,j} \rangle|$. Thus, for $t\in[0,1)$, we have
\begin{align*}
    \Pr\Big[|\langle {R}_{*,i},{R}_{*,j}\rangle| \geq t\Big] \leq 4\exp{(-{t^2b}/{6})}.
\end{align*}
Take $C=5$. Since $n\geq 2$ and $\delta<1$, we have $4n^2/\delta\leq n^4/\delta^4$, hence $\log(4n^2/\delta)\leq 4\log(n/\delta)$. We consider two cases.

If $b> 24\log(n/\delta)$, pick $t = \sqrt{6\log(4n^2/\delta)/b}$. Then $t\leq\sqrt{24\log(n/\delta)/b}<1$, $t\leq C\sqrt{\log(n/\delta)/b}$ and $4\exp(-t^2b/6)=\delta/n^2$. So
\begin{align*}
    \Pr\Big[|\langle {R}_{*,i},{R}_{*,j}\rangle| \geq C\sqrt{\frac{\log(n/\delta)}{b}}\Big] \leq \Pr\Big[|\langle {R}_{*,i},{R}_{*,j}\rangle| \geq t\Big] \leq \delta/n^2.
\end{align*}

If $b\leq 24\log(n/\delta)$, then $C\sqrt{\log(n/\delta)/b}\geq 5/\sqrt{24}>1\geq \|R_{*,i}\|_2\|R_{*,j}\|_2\geq |\langle {R}_{*,i},{R}_{*,j}\rangle|$, so the event above has probability $0$.

Taking a union bound over the fewer than $n^2$ pairs $(i,j) \in [n] \times [n]$ with $i \neq j$, the probability in the lemma is at most $\delta$, which completes the proof.
\end{proof}

\begin{lemma}[AMS]\label{lem:ams_ip}Let $R\in\R^{b\times n}$ be a random AMS matrix (Definition~\ref{def:ams}). Let $\{\sigma_i,~i\in[n]\}$ be independent Rademacher random variables and $\ov{R}\in\R^{b\times n}$ with $\ov{R}_{*,i} = \sigma_i R_{*,i},~\forall i\in[n]$. Then we have:
\begin{align*}
\Pr \Big[ \max_{i \neq j} | \langle \ov{R}_{*,i} , \ov{R}_{*,j} \rangle | \geq C\sqrt{\frac{ \log(n/\delta) }{b}} \Big] \leq O(\delta),
\end{align*}
where $n\geq 2$, $\delta\in(0,1)$ and $C>0$ is an absolute constant.
\end{lemma}
\begin{proof}
Note for any fixed $i\neq j$, $\ov{R}_{*,i}$ and $\ov{R}_{*,j}$ are independent. By Hoeffding inequality (Lemma~\ref{lem:hoeffding}), we have
\begin{align*}
  \Pr \Big[ |\langle \ov{R}_{*,i} , \ov{R}_{*,j}\rangle| \geq t \Big] \leq 2\exp \Big( - \frac{2t^2}{ \sum_{i=1}^b (\frac{1}{b} - (-\frac{1}{b}))^2 } \Big) \leq 2e^{-t^2b/2}
\end{align*}
Choosing $t = \sqrt{2\log(2n^2/\delta)}/\sqrt{b}$, we have
\begin{align*}
  \Pr \Big[ |\langle \ov{R}_{*,i} , \ov{R}_{*,j}\rangle| \geq \sqrt{2\log(2n^2/\delta)}/\sqrt{b} \Big] \leq \frac{\delta}{n^2}.
\end{align*}
Since $n\geq 2$ and $\delta<1$, we have $2n^2/\delta\leq n^3/\delta^3$, hence $\log(2n^2/\delta)\leq 3\log(n/\delta)$ and
\begin{align*}
\sqrt{2\log(2n^2/\delta)}/\sqrt{b}\leq C\sqrt{\log(n/\delta)/b}~~~\text{with}~C=\sqrt{6}.
\end{align*}
Taking a union bound over the fewer than $n^2$ pairs $(i,j) \in [n] \times [n]$ with $i \neq j$, the probability in the lemma is at most $\delta$, which completes the proof.
\end{proof}

\begin{lemma}[Count-Sketch]\label{lem:cs_ip}
Let $R \in \R^{b\times n}$ be a count-sketch matrix (Definition~\ref{def:cs}). Let $\{\sigma_i,~i\in[n]\}$ be independent Rademacher random variables and $\ov{R}\in\R^{b\times n}$ with $\ov{R}_{*,i} = \sigma_i R_{*,i},~\forall i\in[n]$. Then we have:
\begin{align*}
  \max_{i\neq j} |\langle \ov{R}_{*,i} , \ov{R}_{*,j}\rangle| \leq 1.
\end{align*}
\end{lemma}
\begin{proof}
Directly follow the definition of count-sketch matrices.
\end{proof}

\begin{lemma}[Sparse embedding]\label{lem:sparse_ip}
Let $R\in\R^{b\times n}$ be a sparse embedding matrix with parameter $s$ (Definition~\ref{def:sparse_1} and \ref{def:sparse_2}) with fully independent signs, that is, the signs of all nonzero entries of $R$ are independent of each other and of their positions. Let $\{\sigma_i,~i\in[n]\}$ be independent Rademacher random variables and $\ov{R}\in\R^{b\times n}$ with $\ov{R}_{*,i} = \sigma_i R_{*,i},~\forall i\in[n]$. Then we have:
\begin{align*}
\Pr \Big[ \max_{i\neq j} |\langle \ov{R}_{*,i} , \ov{R}_{*,j}\rangle| \geq  \frac{c\sqrt{ \log(n/\delta) }}{\sqrt{s}} \Big] \leq O(\delta),
\end{align*}
where $n\geq 2$, $\delta\in(0,1)$ and $c>0$ is an absolute constant.
\end{lemma}
\begin{proof}

Assume ${R}_{*,i}$ and ${R}_{*,j}$ has $u$ non-zero elements at the same positions, where $0\leq u\leq s$. Conditioned on the positions, the $u$ products of the overlapping entries are independent and take the values $\pm\frac{1}{s}$ with probability $\frac{1}{2}$ each, because the signs are fully independent. Then by Hoeffding inequality (Lemma~\ref{lem:hoeffding}), we have
\begin{align}\label{eq:crucial}
  \Pr[|\langle \ov{R}_{*,i} , \ov{R}_{*,j} \rangle|\geq t] \leq 2\exp \left( - \frac{2t^2}{ \sum_{i=1}^u (\frac{1}{s} - (-\frac{1}{s}))^2 } \right) \leq 2\exp (- t^2 s^2 / ( 2u ) )
\end{align}
Let $t = \sqrt{(2u/s^2)\log(2n^2/\delta)}$, we have
\begin{align}\label{eq:crucial2}
  \Pr \Big[ |\langle \ov{R}_{*,i} , \ov{R}_{*,j} \rangle|\geq \sqrt{ 2s^{-1} \log( 2n^2/\delta)} \Big] \notag
  \leq & ~ \Pr \Big[ | \langle {R}_{*,i} , {R}_{*,j} \rangle | \geq \sqrt{ 2us^{-2} \log( 2n^2/\delta)} \Big] \\
  \leq & ~ \delta/n^2
\end{align}
since $u\leq s$ (if $u=0$, the inner product is $0$). Since $n\geq 2$ and $\delta<1$, we have $\log(2n^2/\delta)\leq 3\log(n/\delta)$, so the threshold $\sqrt{ 2s^{-1} \log( 2n^2/\delta)}$ is at most $c\sqrt{\log(n/\delta)/s}$ with $c=\sqrt{6}$. By taking a union bound over the fewer than $n^2$ pairs $(i,j) \in [n] \times [n]$ with $i \neq j$, the probability in the lemma is at most $\delta$, which completes the proof.
\end{proof}

\subsection{Infinite Norm Bound}

\begin{lemma}[SRHT and AMS]\label{lem:srht_ams_inf}
Let $n\geq 2$ and $\delta\in(0,1)$. Let $R \in \R^{b \times n}$ denote a subsample randomized Hadamard transform (Definition~\ref{def:srht}) or AMS sketching matrix (Definition~\ref{def:ams}). In the AMS case, assume in addition fully independent signs, that is, all entries of $R$ are independent.
Then there is an absolute constant $C>0$ such that, for any fixed vector $h\in \R^n$ and any fixed vector $g \in \R^{n}$, the following properties hold:
\begin{align*}
\Pr_{R \sim \Pi}\Big[ | (g^\top R^\top R h) - (g^\top h) | >  \frac{ C\log^{1.5}(n/\delta) }{\sqrt{b}} \|g \|_2 \|h\|_2 \Big] \leq O(\delta).
\end{align*}
\end{lemma}

\begin{proof}
We can rewrite $(g^\top R^{\top}R h) -(g^\top h)$ as follows:,
\begin{align*}
    (g^\top R^{\top}R h) -(g^\top h)  
    = &~\sum_{i=1}^n \sum_{j \in [n]\backslash i}^n g_i h_j \langle R_{*,i}, R_{*,j} \rangle + \sum_{i=1}^n g_i h_i ( \| R_{*,i} \|_2^2 - 1)\\
    = &~\sum_{i=1}^n \sum_{j \in [n]\backslash i}^n g_i h_j \langle \sigma_i \ov{R}_{*,i}, \sigma_j \ov{R}_{*,j}\rangle.
\end{align*}
where $\sigma_i$'s are independent Rademacher random variables and $\ov{R}_{*,i} = \sigma_i R_{*,i},~\forall i\in[n]$, and the second step follows from $\| R_{*,i} \|_2^2 = 1, \forall i \in [n]$. We choose $\sigma$ independent of $R$, and then $\sigma$ is also independent of $\ov{R}$. For SRHT, $\ov{R}=\sqrt{n/b}SHD'$, where the diagonal matrix $D'$ has entries $D'_{i,i}=\sigma_iD_{i,i}$, which are i.i.d. Rademacher and independent of $\sigma$. For AMS with fully independent signs, given $\sigma$ the entries of $\ov{R}$ are again independent and uniform on $\{-\frac{1}{\sqrt{b}},+\frac{1}{\sqrt{b}}\}$. In both cases $\ov{R}$ has the same distribution as $R$ and is independent of $\sigma$, so we may apply the Hanson-Wright inequality over $\sigma$ conditioned on $\ov{R}$.

We define matrix $A \in \R^{n \times n}$ and $B \in \R^{n \times n}$ as follows:
\begin{align*}
    A_{i,j} = & ~ g_i h_j \cdot \langle \ov{R}_{*,i}, \ov{R}_{*,j}\rangle, &  \forall i \in [n], j \in [n] \\
    B_{i,j} = & ~ g_i h_j \cdot \max_{i'\neq j'}  | \langle \ov{R}_{*,i'}, \ov{R}_{*,j'}\rangle | & \forall i \in [n], j \in [n]
\end{align*}
We define $A^{\circ} \in \R^{n \times n}$ to be the matrix $A \in \R^{n \times n}$ with removing diagonal entries, applying Hason-wright inequality (Lemma~\ref{lem:hason_wright}), we have
\begin{align*}
    \Pr_{\sigma} [ |\sigma^\top A^{\circ} \sigma | \geq \tau ] \leq 2 \cdot \exp ( -c \min\{ \tau^2/\| A^{\circ} \|_F^2, \tau / \| A^{\circ} \| \} )
\end{align*}
We can upper bound $\| A^{\circ} \|$ and $\| A^{\circ} \|_F$. 
\begin{align*}
   \| A^{\circ} \| 
   \leq & ~ \| A^{\circ} \|_F \\
   \leq & ~ \| B \|_F \\
   = & ~ \| g \|_2 \cdot \| h \|_2 \cdot \max_{i \neq j} |\langle \ov{R}_{*,i} , \ov{R}_{*,j} \rangle|,
\end{align*}
where the second step holds since $A^{\circ}_{i,i}=0$ and $|A^{\circ}_{i,j}|\leq |B_{i,j}|$ for $i\neq j$, and the last step holds since $B$ is rank-$1$.

Since $|\langle \ov{R}_{*,i},\ov{R}_{*,j} \rangle|=|\langle R_{*,i},R_{*,j} \rangle|$, by Lemma~\ref{lem:srht_ip} (for AMS, we use Lemma~\ref{lem:ams_ip}) with probability at least $1-O(\delta)$, we have :
\begin{align*}
 \max_{i \neq j} |\langle \ov{R}_{*,i},\ov{R}_{*,j} \rangle| \leq C_1\sqrt{\frac{\log(n/\delta)}{b}},
\end{align*}
where $C_1$ is the absolute constant of that lemma.
Conditioning on the above event holds, which depends only on $\ov{R}$. Then $\| A^{\circ} \| \leq \| A^{\circ} \|_F \leq M$, where $M := C_1\| g \|_2 \| h \|_2 \sqrt{\log(n/\delta)/b}$. If $g=0$ or $h=0$, there is nothing to prove, so assume $M>0$. Let $c$ be the absolute constant of the Hanson-Wright inequality, and take
\begin{align*}
C = C_1\cdot\max\{1/c,\,1/\log 2\},~~~ \tau =  C\| g \|_2 \cdot \| h \|_2 \cdot \log^{1.5}(n/\delta) / \sqrt{b}.
\end{align*}
Then $r:=\tau/M = (C/C_1)\log(n/\delta)$. Since $n\geq 2$, we have $\log(n/\delta)\geq \log(2/\delta)\geq\log 2$, so $r\geq 1$ and $r\geq \log(2/\delta)/c$. As $r\geq 1$, we have $\min\{ \tau^2/\| A^{\circ} \|_F^2, \tau / \| A^{\circ} \| \}\geq \min\{r^2,r\}=r$, hence
\begin{align*}
    \Pr_{\sigma} [ |\sigma^\top A^{\circ} \sigma | \geq \tau ] \leq 2\exp(-cr) \leq 2\exp(-\log(2/\delta)) = \delta.
\end{align*}
Adding the failure probability $O(\delta)$ of the event above, we can show that
\begin{align*}
   \Pr \left[  \Big| (g^\top R^{\top} R h) - (g^\top h) \Big| \geq \| g \|_2 \cdot \| h \|_2 \frac{ C\log^{1.5} (n/\delta) }{ \sqrt{b} } \right] \leq O(\delta).
\end{align*}
Thus, we complete the proof.
\end{proof}

\begin{lemma}[Random Gaussian]\label{lem:gaussian_inf}
Let $n\geq 2$, $\delta\in(0,1)$, and let $R \in \R^{b \times n}$ denote a random Gaussian matrix (Definition~\ref{def:gaussian}). There are absolute constants $C,C'>0$ such that, if $b\geq C'\log(n/\delta)$, then for any fixed vector $h\in \R^n$ and any fixed vector $g \in \R^{n}$, the following properties hold:
\begin{align*}
\Pr_{R \sim \Pi}\Big[ | (g^\top R^\top R h) - (g^\top h) | >  \frac{ C\log^{1.5}(n/\delta) }{\sqrt{b}} \|g \|_2 \|h\|_2 \Big] \leq O(\delta).
\end{align*}
\end{lemma}
\begin{proof}
We follow the same procedure as proving Lemma~\ref{lem:srht_ams_inf}.

We can rewrite $(g^\top R^{\top}R h) -(g^\top h)$ as follows:,
\begin{align}\label{eq:gp3}
    (g^\top R^{\top}R h) -(g^\top h)  
    = &~\sum_{i=1}^n \sum_{j \in [n]\backslash i}^n g_i h_j \langle R_{*,i}, R_{*,j} \rangle + \sum_{i=1}^n g_i h_i ( \| R_{*,i} \|_2^2 - 1)\notag\\
    = &~\sum_{i=1}^n \sum_{j \in [n]\backslash i}^n g_i h_j \langle \sigma_i \ov{R}_{*,i}, \sigma_j \ov{R}_{*,j}\rangle + \sum_{i=1}^n g_i h_i ( \| R_{*,i} \|_2^2 - 1).
\end{align}
where $\sigma_i$'s are independent Rademacher random variables, independent of $R$, and $\ov{R}_{*,i}=\sigma_iR_{*,i}$. Since the entries of $R$ are independent and symmetric, $\ov{R}$ has the same distribution as $R$ and is independent of $\sigma$.

To bound the first term $\sum_{i=1}^n \sum_{j \in [n]\backslash i}^n g_i h_j \langle \sigma_i \ov{R}_{*,i}, \sigma_j \ov{R}_{*,j}\rangle$, we define matrix $A \in \R^{n \times n}$ and $B \in \R^{n \times n}$ as follows:
\begin{align*}
    A_{i,j} = & ~ g_i h_j \cdot \langle \ov{R}_{*,i}, \ov{R}_{*,j}\rangle, &  \forall i \in [n], j \in [n] \\
    B_{i,j} = & ~ g_i h_j \cdot \max_{i'\neq j'}  | \langle \ov{R}_{*,i'}, \ov{R}_{*,j'}\rangle | & \forall i \in [n], j \in [n]
\end{align*}
We define $A^{\circ} \in \R^{n \times n}$ to be the matrix $A \in \R^{n \times n}$ with removing diagonal entries, applying Hason-wright inequality (Lemma~\ref{lem:hason_wright}), we have
\begin{align*}
    \Pr_{\sigma} [ |\sigma^\top A^{\circ} \sigma | \geq \tau ] \leq 2 \cdot \exp ( -c \min\{ \tau^2/\| A^{\circ} \|_F^2, \tau / \| A^{\circ} \| \} )
\end{align*}
We can upper bound $\| A^{\circ} \|$ and $\| A^{\circ} \|_F$. 
\begin{align*}
   \| A^{\circ} \| 
   \leq & ~ \| A^{\circ} \|_F \\
   \leq & ~ \| B \|_F \\
   = & ~ \| g \|_2 \cdot \| h \|_2 \cdot \max_{i \neq j} |\langle \ov{R}_{*,i} , \ov{R}_{*,j} \rangle|,
\end{align*}
where the second step holds since $A^{\circ}_{i,i}=0$ and $|A^{\circ}_{i,j}|\leq |B_{i,j}|$ for $i\neq j$, and the last step holds since $B$ is rank-$1$.

Let $C'$ be the constant of Lemma~\ref{lem:gaussian_ip}, where we may assume $C'\geq 2$ (its proof gives $C'=12$), and let $C_1$ be the constant $C$ of that lemma. Since $b\geq C'\log(n/\delta)$, Lemma~\ref{lem:gaussian_ip} shows that with probability at least $1-O(\delta)$, we have :
\begin{align*}
 \max_{i \neq j} | \langle \ov{R}_{*,i},\ov{R}_{*,j} \rangle | \leq C_1\sqrt{\frac{\log(n/\delta)}{b}}.
\end{align*}
Conditioning on the above event holds, which depends only on $\ov{R}$.

Let $c$ be the absolute constant of the Hanson-Wright inequality, and choose
\begin{align*}
C_2=C_1\cdot\max\{1/c,\,1/\log 2\},~~~\tau =  C_2\| g \|_2 \cdot \| h \|_2 \cdot \log^{1.5}(n/\delta) / \sqrt{b}.
\end{align*}
Then the calculation in the proof of Lemma~\ref{lem:srht_ams_inf} (with $C_1$ the constant of Lemma~\ref{lem:gaussian_ip}) gives $\Pr_{\sigma} [ |\sigma^\top A^{\circ} \sigma | \geq \tau ]\leq\delta$. Adding the failure probability of the event above, we can show that
\begin{align}\label{eq:gb1}
   \Pr \left[ \Big| \sum_{i=1}^n \sum_{j \in [n] \backslash i}^n g_i h_j \langle \sigma_i \ov{R}_{*,i}, \sigma_j \ov{R}_{*,j}\rangle \Big| \geq \| g \|_2 \cdot \| h \|_2 \frac{ C_2\log^{1.5} (n/\delta) }{ \sqrt{b} } \right] \leq O(\delta).
\end{align}

To bound the second term $\sum_{i=1}^n g_i h_i ( \| R_{*,i} \|_2^2 - 1)$, note that $\| R_{*,i} \|_2^2$ is a sum of $b$ squares of independent ${\cal N}(0,1/b)$ variables for every $i\in[n]$. Applying Lemma~\ref{lem:chi_square_tail} with $k=b$ and $\sigma^2=1/b$, we have, for every $t>0$,
\begin{align*}
  \Pr \left[ \| R_{*,i} \|_2^2 - 1 \geq 2\sqrt{t/b}+2t/b \right] \leq e^{-t}, ~~~ \Pr \left[ 1-\| R_{*,i} \|_2^2 \geq 2\sqrt{t/b} \right] \leq e^{-t}.
\end{align*}
Take $t=\log(2n/\delta)$. Since $n\geq 2$ and $\delta<1$, we have $2n/\delta\leq n^2/\delta^2$, hence $t\leq 2\log(n/\delta)\leq C'\log(n/\delta)\leq b$. So $2t/b\leq 2\sqrt{t/b}$, and
\begin{align*}
  \Pr \left[ \Big| \| R_{*,i} \|_2^2 - 1 \Big| \geq 4\sqrt{\frac{\log(2n/\delta)}{b}} \right] \leq 2e^{-t} = \delta / n.
\end{align*}
By a union bound over $i\in[n]$, with probability at least $1-\delta$ this holds for no $i\in[n]$, and then
\begin{align}\label{eq:gb2}
  \Big| \sum_{i=1}^n g_i h_i ( \| R_{*,i} \|_2^2 - 1 ) \Big| \leq & ~ \max_{i\in[n]}\Big| \| R_{*,i} \|_2^2 - 1 \Big|\cdot \sum_{i=1}^n |g_i h_i| \notag \\
  \leq & ~ 4\sqrt{\frac{\log(2n/\delta)}{b}}\|g\|_2\|h\|_2 \notag \\
  \leq & ~ \frac{4\sqrt{2}}{\log 2}\cdot\frac{\log^{1.5}(n/\delta)}{\sqrt{b}}\|g\|_2\|h\|_2,
\end{align}
where the second step uses the Cauchy-Schwartz inequality, and the last step uses $\log(2n/\delta)\leq 2\log(n/\delta)$ and $\sqrt{x}\leq x^{1.5}/\log 2$ for $x=\log(n/\delta)\geq\log 2$.
Plugging the bounds Eq.~\eqref{eq:gb1} and~\eqref{eq:gb2} back to Eq.~\eqref{eq:gp3}, the lemma holds with $C=C_2+4\sqrt{2}/\log 2$ and failure probability $O(\delta)+\delta=O(\delta)$, which completes the proof.
\end{proof}

\begin{lemma}[Count-sketch]\label{lem:cs_inf}
Let $\delta\in(0,1)$, and let $R \in \R^{b \times n}$ denote a count-sketch matrix (Definition~\ref{def:cs}) with fully independent signs, that is, $\sigma(1),\ldots,\sigma(n)$ are independent of each other and of the positions of the nonzero entries. Then there is an absolute constant $C>0$ such that, for any fixed vector $h\in \R^n$ and any fixed vector $g \in \R^{n}$, the following properties hold:
\begin{align*}
\Pr_{R \sim \Pi} \Big[  | (g^\top R^{\top} R h) - (g^\top h) | \geq C\log ( 1 / \delta ) \| g \|_2 \| h \|_2  \Big] \leq O(\delta).
\end{align*}
\end{lemma}
\begin{proof}
We follow the identical procedure as proving Lemma~\ref{lem:srht_ams_inf} to apply Hason-wright inequality (Lemma~\ref{lem:hason_wright}). Every column of $R$ has exactly one nonzero entry $\pm 1$, so $\|R_{*,i}\|_2^2=1$ and the same decomposition holds. The auxiliary Rademacher variables of that proof are chosen independent of $R$. Since the signs of $R$ are fully independent, the signs of $\ov{R}$ are again independent and uniform given the auxiliary variables, so $\ov{R}$ has the same distribution as $R$ and is independent of them.

Then note Lemma~\ref{lem:cs_ip} shows
\begin{align*}
  \max_{i \neq j} |\langle \ov{R}_{*,i} , \ov{R}_{*,j} \rangle| \leq 1,
\end{align*}
so $\| A^{\circ} \| \leq \| A^{\circ} \|_F \leq M := \| g \|_2 \| h \|_2$. If $\delta>1/2$, the claimed bound holds trivially with failure probability at most $1<2\delta$. Let $\delta\leq 1/2$, so that $\log(1/\delta)\geq\log 2$ and $\log(2/\delta)\leq 2\log(1/\delta)$. We may assume $M>0$. Let $c$ be the absolute constant of the Hanson-Wright inequality and take $C=\max\{2/c,\,1/\log 2\}$. Thus, choosing $\tau =  C \| g \|_2 \cdot \| h \|_2 \cdot \log(1/\delta)$, we have $r:=\tau/M=C\log(1/\delta)\geq 1$ and $cr\geq 2\log(1/\delta)\geq\log(2/\delta)$, so that $\min\{ \tau^2/\| A^{\circ} \|_F^2, \tau / \| A^{\circ} \| \}\geq r$ and we can show that
\begin{align*}
   \Pr \left[  | (g^\top R^{\top} R h) - (g^\top h) | \geq C \| g \|_2 \cdot \| h \|_2  \log ( 1 / \delta )  \right] \leq 2\exp(-cr) \leq \delta.
\end{align*}
which completes the proof.
\end{proof}

\begin{lemma}[Count-sketch 2]\label{lem:cs_inf_2}
Let $R \in \R^{b \times n}$ denote a count-sketch matrix (Definition~\ref{def:cs}). Then for any fixed vector $h\in \R^n$ and any fixed vector $g \in \R^{n}$, the following properties hold:
\begin{align*}
\Pr_{R \sim \Pi} \Big[  | (g^\top R^{\top} R h) - (g^\top h) | \geq \frac{1}{\sqrt{b\delta}} \| g \|_2 \| h \|_2  \Big] \leq O(\delta).
\end{align*}
\end{lemma}
\begin{proof}
It is known that a count-sketch matrix with $b=\epsilon^{-2} \delta^{-1}$ rows satisfies the $(\epsilon, \delta, 2)$-JL moment property (see e.g. Theorem~14 of \cite{w14}). Using Markov's inequality, $(\epsilon, \delta, 2)$-JL moment property implies 
\begin{align*}
\Pr_{R \sim \Pi} \Big[  | (g^\top R^{\top} R h) - (g^\top h) | \geq \epsilon \| g \|_2 \| h \|_2  \Big] \leq O(\delta),
\end{align*}
where $\epsilon = \frac{1}{\sqrt{b\delta}}$.
\end{proof}

\begin{lemma}[Sparse embedding]\label{lem:sparse_inf}
Let $n\geq 2$, $\delta\in(0,1)$, and let $R \in \R^{b \times n}$ denote a sparse-embedding matrix (Definition~\ref{def:sparse_1} and \ref{def:sparse_2}) with fully independent signs, that is, the signs of all nonzero entries of $R$ are independent of each other and of their positions.
Then there is an absolute constant $C>0$ such that, for any fixed vector $h\in \R^n$ and any fixed vector $g \in \R^{n}$, the following properties hold:
\begin{align*}
3. & \Pr_{R \sim \Pi}\Big[ | (g^\top R^\top R h) - (g^\top h) | >  \frac{ C\log^{1.5}(n/\delta) }{\sqrt{s}} \|g \|_2 \|h\|_2 \Big] \leq O(\delta).
\end{align*}
\end{lemma}

\begin{proof}
We follow the identical procedure as proving Lemma~\ref{lem:srht_ams_inf} to apply Hason-wright inequality (Lemma~\ref{lem:hason_wright}). Every column of $R$ has exactly $s$ nonzero entries $\pm 1/\sqrt{s}$, so $\|R_{*,i}\|_2^2=1$ and the same decomposition holds. The auxiliary Rademacher variables of that proof are chosen independent of $R$. Since the signs of $R$ are fully independent, the signs of $\ov{R}$ are again independent and uniform given the auxiliary variables, so $\ov{R}$ has the same distribution as $R$ and is independent of them.

Then note Lemma~\ref{lem:sparse_ip} shows with probability at least $1-O(\delta)$ we have
\begin{align*}
 \max_{i \neq j} |\langle \ov{R}_{*,i},\ov{R}_{*,j} \rangle| \leq \frac{c_1\sqrt{\log(n/\delta)}}{\sqrt{s}},
\end{align*}
where $c_1$ is the absolute constant of that lemma. Conditioning on the above event holds, we have $\| A^{\circ} \| \leq \| A^{\circ} \|_F \leq M:=c_1\| g \|_2 \| h \|_2 \sqrt{\log(n/\delta)/s}$, and we may assume $M>0$. Let $c$ be the absolute constant of the Hanson-Wright inequality and take $C=c_1\cdot\max\{1/c,\,1/\log 2\}$. Choosing $\tau =  C \| g \|_2 \cdot \| h \|_2 \cdot \log^{1.5}(n/\delta)/\sqrt{s}$, we get $r:=\tau/M=(C/c_1)\log(n/\delta)\geq\max\{1,\log(2/\delta)/c\}$, since $\log(n/\delta)\geq\log(2/\delta)\geq\log 2$. As in the proof of Lemma~\ref{lem:srht_ams_inf}, this gives $\Pr_{\sigma} [ |\sigma^\top A^{\circ} \sigma | \geq \tau ]\leq 2\exp(-cr)\leq\delta$. Adding the failure probability of the event above, we can show that
\begin{align*}
   \Pr \left[  | (g^\top R^{\top} R h) - (g^\top h) | \geq \frac{ C \log^{1.5} ( n / \delta ) }{ \sqrt{s} } \| g \|_2 \cdot \| h \|_2 \right] \leq O(\delta).
\end{align*}
Thus, we complete the proof.
\end{proof}

\begin{lemma}[Uniform sampling]\label{lem:uniform_inf}
Let $R \in \R^{b \times n}$ denote a uniform sampling matrix (Definition~\ref{def:unif_sample}). Then for any fixed vector $h\in \R^n$ and any fixed vector $g \in \R^{n}$, the following properties hold:
\begin{align*}
3. |(g^\top R^{\top}R h) -(g^\top h)|  
    \leq (1+\frac{n}{b}) \| g \|_2 \| h \|_2 
\end{align*}
where $I\subset [n]$ be the subset of indexes chosen by the uniform sampling matrix.
\end{lemma}
\begin{proof}
We can rewrite $(g^\top R^{\top}R h) -(g^\top h)$ as follows:,
\begin{align*}
    (g^\top R^{\top}R h) -(g^\top h)  
    = &~\sum_{i=1}^n \sum_{j \in [n]\backslash i}^n g_i h_j \langle R_{*,i}, R_{*,j} \rangle + \sum_{i=1}^n g_i h_i ( \| R_{*,i} \|_2^2 - 1)\\
    = &~\frac{n}{b}\sum_{i\in I} g_i h_i - \sum_{i=1}^n g_i h_i.
\end{align*}
where the second step follows from the uniform sampling matrix has only one nonzero entry in each row.

Let $I\subset [n]$ be the subset chosen by the uniform sampling matrix, then $\| R_{*,i} \|_2^2 = n/b$ for $i\in I$ and $\| R_{*,i} \|_2^2 = 0$ for $i\in[n]\setminus I$. So we have
\begin{align*}
    |(g^\top R^{\top}R h) -(g^\top h)|  
    = & ~ \Big| \sum_{i\in I} g_i h_i (\frac{n}{b} - 1) - \sum_{i\in [n]\setminus I} g_i h_i \Big| \\
    \leq & ~ (1+\frac{n}{b}) \| g \|_2 \| h \|_2.
\end{align*}
\end{proof}
\section{Analysis of Convergence: Single-step Scheme}
\label{sec:app_1step}

\subsection{Preliminary}\label{sec:notation}
Throughout the proof of convergence, we will use ${\cal F}_t$ to denote the $\sigma$-algebra generated by $w^0$, $w^1$, $\ldots$, $w^t$. Also, we use $\eta$ as a shorthand for $\eta_{\mathrm{global}}\cdot \eta_{\mathrm{local}}$.

In this section, we analyze Algorithm~\ref{alg:fed_learn_kstep_dp} with $K=1$ local step. It sets $w^1=w^0$, and for every $t\geq 1$, round $t$ starts from $w^t$ and ends with
\begin{align*}
    w^{t+1} = w^t+\dec_{t+1}(\Delta \wt{w}^t) = w^t-\eta\cdot \dec_{t+1}(\enc_{t+1}(\nabla f(w^t))),
\end{align*}
since $\Delta w_c(t)=-\eta_{\mathrm{local}}\cdot \nabla f_c(w^t)$, $\Delta \wt{w}^t = \eta_{\mathrm{global}}\cdot\frac{1}{N}\sum_{c=1}^N\enc_{t+1}(\Delta w_c(t))$, and $\enc_{t+1},\dec_{t+1}$ are linear. Hence ${\cal F}_t$ is contained in the $\sigma$-algebra generated by $w^0$ and $R_2,\ldots,R_t$. As the sketching matrices are independent of each other and of $w^0$, $R_{t+1}$ is independent of ${\cal F}_t$.

\subsection{Strongly-convex \texorpdfstring{$f$}{} Convergence Analysis}
\begin{theorem}\label{thm:single_step_strong_convex}
Let $f:\R^d\rightarrow \R$ satisfying Assumption~\ref{ass:f_ass} with $\mu>0$. Let $w^*\in \R^d$ be the optimal solution to $f$ and assume $\enc/\dec$ functions satisfying Theorem~\ref{thm:enc_dec}. Suppose $\eta:=\eta_{\mathrm{global}}\cdot \eta_{\mathrm{local}}$ has the property that $\eta\leq \frac{1}{(1+\alpha)L}$, then
\begin{align*}
    \E[f(w^{t+1})]-f(w^*) & \leq (1-\mu\eta)^t\cdot (f(w^0)-f(w^*))
\end{align*}
\end{theorem}

\begin{proof}
Fix $t\geq 1$. We shall first bound $f(w^{t+1})-f(w^t)$:
\begin{align*}
   f(w^{t+1})-f(w^t)  \leq & \ip{w^{t+1}-w^t}{\nabla f(w^t)}+\frac{L}{2}\|w^{t+1}-w^t\|_2^2 \\
    = &  \ip{\dec_{t+1}(\Delta \wt{w}^t)}{\nabla f(w^t)}+\frac{L}{2}\|\dec_{t+1}(\Delta \wt{w}^t)\|_2^2 \\
    = & -\ip{\eta_{\mathrm{global}}\cdot \dec_{t+1}(\frac{1}{N}\sum_{c=1}^N\enc_{t+1}(\eta_{\mathrm{local}}\cdot \nabla f_c(w^t)))}{\nabla f(w^t)}\\
    & +\frac{L}{2}\|\eta_{\mathrm{global}}\cdot \dec_{t+1}(\frac{1}{N}\sum_{c=1}^N\enc_{t+1}(\eta_{\mathrm{local}}\cdot \nabla f_c(w^t))) \|_2^2 \\
    = & -\eta_{\mathrm{global}}\cdot \eta_{\mathrm{local}}\cdot \ip{\dec_{t+1}(\enc_{t+1}(\nabla f(w^t)))}{\nabla f(w^t)} \\
    & + \frac{L}{2}\cdot(\eta_{\mathrm{global}}\cdot \eta_{\mathrm{local}})^2\cdot \|\dec_{t+1}(\enc_{t+1}(\nabla f(w^t))) \|_2^2
\end{align*}
where the first step uses the $L$-smoothness condition of $f$, and the last step uses the linearity property of $\enc/\dec$ functions.

Taking expectation conditioned on $\mathcal{F}_t$, and using that $w^t$ is ${\cal F}_t$-measurable while $R_{t+1}$ is independent of ${\cal F}_t$, we get
\begin{align}\label{eq:strong_convex_bound}
    & ~ \E[f(w^{t+1})-f(w^t)\mid\mathcal{F}_t] \notag\\
    \leq & ~ -\eta\cdot\ip{\E[\dec_{t+1}(\enc_{t+1}(\nabla f(w^t)))\mid \mathcal{F}_t]}{\nabla f(w^t)}+\frac{L\eta^2}{2}\E[\|\dec_{t+1}(\enc_{t+1}(\nabla f(w^t))) \|_2^2\mid \mathcal{F}_t] \notag\\
    \leq & ~ -\eta\cdot\ip{\nabla f(w^t)}{\nabla f(w^t)}+\frac{L\eta^2}{2}(1+\alpha)\cdot \|\nabla f(w^t)\|_2^2 \notag\\
    \leq & ~ -\frac{\eta}{2}\cdot \|\nabla f(w^t)\|_2^2 \notag\\
    \leq & ~ -\mu\eta\cdot(f(w^t)-f(w^*))
\end{align}
where the second step comes from the fact that $\dec_{t+1}(\enc_{t+1}(h))$ is an unbiased estimator for any fixed $h\in \R^d$ and the bound on its variance, applied to $h=\nabla f(w^t)$, which is fixed given ${\cal F}_t$, the third step comes from $\eta \leq \frac{1}{(1+\alpha)L}$, and the last step comes from Fact~\ref{fact:mu_upper_bound}.

Upon rearranging and subtracting both sides by $f(w^*)$, we get
\begin{align}\label{eq:f_wt+1}
    \E[f(w^{t+1})-f(w^*)\mid \mathcal{F}_t] & \leq (1-\mu\eta)\cdot (f(w^t)-f(w^*))
\end{align}
Note that if we apply expectation over $\mathcal{F}_t$ on both sides of Eq.~\eqref{eq:f_wt+1} we can get
\begin{align}
    \E[f(w^{t+1})]-f(w^*) & \leq (1-\mu\eta)\cdot (\E[f(w^t)]-f(w^*))
\end{align}
for every $t\geq 1$. Notice $0\leq 1-\mu\eta< 1$, since $\mu\leq L$ and $\eta\leq \frac{1}{(1+\alpha)L}$, so this is a contraction map. If we iterate this recurrence relation from $t$ down to $1$ and use $w^1=w^0$, we will finally get
\begin{align}\label{eq:f_wt_f*_gap}
    \E[f(w^{t+1})-f(w^*)] & \leq (1-\mu \eta)^t\cdot (f(w^0)-f(w^*)) .
\end{align}
For $t=0$, this holds with equality, since $w^1=w^0$.
\end{proof}
\subsection{Convex \texorpdfstring{$f$}{} Convergence Analysis}
Assume $f$ is a convex function, we obtain a convergence bound in terms of the average of all parameters.
\begin{theorem}\label{thm:single_step_convex}
Let $f:\R^d\rightarrow \R$ satisfying Assumption~\ref{ass:f_ass} with $\mu=0$. Suppose $\enc/\dec$ functions satisfying Theorem~\ref{thm:enc_dec}. If $\eta:=\eta_{\mathrm{global}}\cdot \eta_{\mathrm{local}} \leq \frac{1}{2(1+\alpha)L}$, then for every $T\geq 1$,
\begin{align*}
    \E[f(\ov{w}^T)-f(w^*)] \leq & ~ \frac{\E[\|w^0-w^*\|_2^2]}{\eta\cdot T}
\end{align*}
where $\ov{w}^T:=\frac{1}{T}\sum_{t=1}^T w^t$ and $w^*\in\R^d$ is the optimal solution.
\end{theorem}
\begin{proof}
Fix $t\geq 1$. We shall first compute the gap between $w^{t+1}$ and $w^*$:
\begin{align}\label{eq:convex_gap_1}
    & ~ \|w^{t+1}-w^*\|_2^2 \notag \\
    = & ~ \|w^t+\dec_{t+1}(\Delta \wt{w}^t)-w^*\|_2^2 \notag\\
    = & ~ \|w^t-\eta\cdot \dec_{t+1}(\enc_{t+1}(\nabla f(w^t)))-w^*\|_2^2\notag \\
    = & ~ \|w^t-w^*\|_2^2+\eta^2\cdot \|\dec_{t+1}(\enc_{t+1}(\nabla f(w^t)))\|_2^2-2\eta\cdot \ip{w^t-w^*}{\dec_{t+1}(\enc_{t+1}(\nabla f(w^t)))}
\end{align}
Since $w^t$ is ${\cal F}_t$-measurable and $R_{t+1}$ is independent of ${\cal F}_t$, the unbiasedness of $\dec_{t+1}\circ \enc_{t+1}$ gives
\begin{align}\label{eq:convex_ip_1}
    \E[\ip{w^t-w^*}{\dec_{t+1}(\enc_{t+1}(\nabla f(w^t)))}\mid {\cal F}_t] = & ~ \E[\ip{w^t-w^*}{\nabla f(w^t)}\mid {\cal F}_t]
\end{align}
Taking expectation of Eq.~\eqref{eq:convex_gap_1} conditioned on ${\cal F}_t$ and plug in Eq.~\eqref{eq:convex_ip_1}, we get
\begin{align}\label{eq:convex_gap_2}
     & ~ \E[\|w^{t+1}-w^*\|_2^2\mid {\cal F}_t] \notag \\
    = & ~ \E[\|w^t-w^*\|_2^2\mid {\cal F}_t]+\eta^2\cdot \E[\|\dec_{t+1}(\enc_{t+1}(\nabla f(w^t)))\|_2^2\mid {\cal F}_t]-2\eta\cdot \E[\ip{w^t-w^*}{\nabla f(w^t)}\mid {\cal F}_t]\notag \\
    \leq & ~ \E[\|w^t-w^*\|_2^2\mid {\cal F}_t]+\eta^2\cdot (1+\alpha)\cdot \E[\|\nabla f(w^t)\|_2^2\mid {\cal F}_t]+2\eta\cdot \E[\ip{w^*-w^t}{\nabla f(w^t)}\mid {\cal F}_t]\notag \\
    \leq & ~ \E[\|w^t-w^*\|_2^2\mid {\cal F}_t]+\eta^2\cdot (1+\alpha)\cdot \E[\|\nabla f(w^t)\|_2^2\mid {\cal F}_t]+2\eta\cdot \E[f(w^*)-f(w^t)\mid {\cal F}_t]
\end{align}
where the second step follows from the variance of $\dec_{t+1}\circ \enc_{t+1}$, and the last step follows from the convexity of $f$.

Taking the expectation over $\mathcal{F}_t$ and re-organizing the above equation, we can get
\begin{align*}
    2\eta\cdot \E[f(w^t)-f(w^*)] \leq & ~ \E[\|w^t-w^*\|_2^2]-\E[\|w^{t+1}-w^*\|_2^2]+\eta^2\cdot (1+\alpha)\cdot \E[\|\nabla f(w^t)\|_2^2]\\
    \leq & ~ \E[\|w^t-w^*\|_2^2]-\E[\|w^{t+1}-w^*\|_2^2]+\eta^2\cdot (1+\alpha)\cdot 2L\cdot \E[f(w^t) - f(w^*)]
\end{align*}
where the second step follows from the convexity and $L$-smoothness of $f$. Rearrange the above inequality, we have
\begin{align*}
    (2\eta -\eta^2\cdot (1+\alpha)\cdot 2L)\cdot \E[f(w^t)-f(w^*)]\leq \E[\|w^t-w^*\|_2^2]-\E[\|w^{t+1}-w^*\|_2^2]
\end{align*}
Note $\eta \leq \frac{1}{2(1+\alpha)L}$, we have
\begin{align*}
    \eta\cdot \E[f(w^t)-f(w^*)]\leq \E[\|w^t-w^*\|_2^2]-\E[\|w^{t+1}-w^*\|_2^2]
\end{align*}
Sum over the $T$ rounds $t=1,\ldots,T$, and use $w^1=w^0$, we arrive at
\begin{align}
    \eta\cdot \sum_{t=1}^T \E[f(w^t)-f(w^*)] \leq &  \E[\|w^1-w^*\|_2^2]-\E[\|w^{T+1}-w^*\|_2^2] \leq  \E[\|w^0-w^*\|_2^2]
\end{align}
Let $\ov{w}^T=\frac{1}{T}\sum_{t=1}^T w^t$ denote the average of parameters across iterations, then by convexity of $f$, we conclude:
\begin{align*}
    \E[f(\ov{w}^T)-f(w^*)] \leq & ~ \frac{\E[\|w^0-w^*\|_2^2]}{\eta\cdot T}
\end{align*}
\end{proof}
\subsection{Non-convex \texorpdfstring{$f$}{} Convergence Analysis}
Next, we prove a version when $f$ is not even a convex function, due to loss of convexity, we can no longer bound the gap between $\E[f(w^t)]$ and $f(w^*)$, but we can instead bound the minimum (or average) expected gradient.
\begin{theorem}\label{thm:single_step_non_convex}
Let $f:\R^d\rightarrow \R$ be an $L$-smooth function (Def.~\ref{def:L_smooth}) and $\enc/\dec$ functions satisfying Theorem~\ref{thm:enc_dec}, let $w^*\in \R^d$ be the optimal solution to $f$. Suppose $\eta :=  \eta_{\mathrm{local}}\cdot \eta_{\mathrm{global}}\leq \frac{1}{(1+\alpha)L}$, then for every $T\geq 1$,
\begin{align*}
    \min_{t\in \{0,1,\ldots,T\}}~\E[\|\nabla f(w^t)\|_2^2] & \leq \frac{2}{\eta T}(\E[f(w^0)]-f(w^*))
\end{align*}
\end{theorem}
\begin{proof}
Note that the only place we used strongly-convex assumption in the proof of Theorem~\ref{thm:single_step_strong_convex} is Eq.~\eqref{eq:strong_convex_bound}, so by the same analysis, we can get for every $t\geq 1$
\begin{align*}
    \E[f(w^{t+1})-f(w^t)\mid {\cal F}_t] \leq & -\frac{\eta}{2}\cdot \|\nabla f(w^t)\|_2^2
\end{align*}

Rearranging and taking total expectation over ${\cal F}_t$, we get
\begin{align*}
    \E[\|\nabla f(w^t)\|_2^2] & \leq \frac{2}{\eta}(\E[f(w^{t})]-\E[f(w^{t+1})])
\end{align*}
Averaging over the $T$ rounds $t=1,\ldots,T$, we get
\begin{align*}
    \frac{1}{T}\sum_{t=1}^T \E[\|\nabla f(w^t)\|_2^2] & \leq \frac{2}{\eta T} \sum_{t=1}^T(\E[f(w^t)]-\E[f(w^{t+1})]) \\
    & = \frac{2}{\eta T} (\E[f(w^1)]-\E[f(w^{T+1})]) \\
    & \leq \frac{2}{\eta T} (\E[f(w^0)]-f(w^*))
\end{align*}
where the last step uses $w^1=w^0$ and $f(w^{T+1})\geq f(w^*)$. Since $w^0=w^1$, the minimum over $t\in\{0,1,\ldots,T\}$ equals the minimum over $t\in[T]$, which is at most the average above. This implies our final result:
\begin{align*}
    \min_{t\in \{0,1,\ldots,T\}}~\E[\|\nabla f(w^t)\|_2^2] & \leq \frac{2}{\eta T}(\E[f(w^0)]-f(w^*))
\end{align*}
\end{proof}
\begin{remark}
Notice due to the structure of $\enc/\dec$ functions, i.e., their variance is bounded in terms of true gradient, the convergence rate depends completely on the term $\frac{1}{(1+\alpha)L}$. If it's a constant, then we essentially recover a convergence rate of gradient descent. On the other hand, if $\frac{1}{(1+\alpha)L}\leq \frac{1}{\sqrt T}$, then we get a similar convergence rate as SGD. One clear advantage of our $\enc/\dec$ functions is they don't introduce extra noise term as in SGD, since we can choose appropriate step size to absorb the variance term.
\end{remark}
\section{\texorpdfstring{$k$}{}-step Convex \& Strongly-convex \texorpdfstring{$f_c$}{} Analysis}
\label{sec:app_kstep_convex}
\subsection{Preliminary}
In this section, we assume each $f_c$ satisfies Assumption~\ref{ass:f_ass} and $\eta_\gl = 1$. For notation simplicity, we also denote $u_c^{t,-1} = u_c^{t-1,K-1}$ for $t\geq 2$.

We use ${\cal F}_t$ to denote the $\sigma$-algebra generated by $w^0,w^1,\ldots,w^t$. Algorithm~\ref{alg:fed_learn_kstep_dp} sets $w^1=w^0$, and for $t\geq 2$, $w^t$ is a function of $w^{t-1}$ and $R_t$. Hence ${\cal F}_t$ is contained in the $\sigma$-algebra generated by $w^0$ and $R_2,\ldots,R_t$. As the sketching matrices are independent of each other and of $w^0$, $R_{t+1}$ is independent of ${\cal F}_t$.
\begin{definition}\label{def:convex_terms_2}
Let $(t,k)\in \{1,\cdots,T+1\}\times \{-1,0,1,\cdots,K-1\}$, we define the following terms for iteration $(t,k)$:
\begin{align*}
    \ov{u}^{t,k} := & ~ \frac{1}{N}\sum_{c=1}^N u_c^{t,k},\quad r^{t,k} :=  ~ \ov{u}^{t,k} - w^*
\end{align*}
to be the average of local parameters and its distance to the optimal solution, 
\begin{align*}
    g_c^{t,k} := ~ \nabla f_c(u_c^{t,k}),\quad \ov{g}^{t,k} := & ~ \frac{1}{N}\sum_{c=1}^N \nabla f_c(u_c^{t,k})
\end{align*}
to be the local gradient and its average,
\begin{align*}
    V^{t,k} := & ~ \frac{1}{N}\sum_{c=1}^N \|u_c^{t,k}-\ov{u}^{t,k}\|_2^2
\end{align*}
to be the variances of local updates,
\begin{align*}
    \sigma^2 = \frac{1}{N}\sum_{c=1}^N \|\nabla f_c(w^*)\|^2
\end{align*}
to be a finite constant that characterize the heterogeneity of local objectives. 

We also define the following indicator function: let $l\in \R$, then we define $1_{\{x=l \}}$ to be
\begin{align*}
  1_{\{x=l\}} = & ~  
   \begin{cases}
     1 & \text{if}~x=l, \\
     0 & \text{otherwise}.
   \end{cases}
\end{align*}
\end{definition}

\subsection{Unifying the Update Rule of Algorithm~\ref{alg:fed_learn_kstep_dp}}
\begin{lemma}\label{lem:update_rule}
We have the following facts for $u_c^{t,k}$ and $\ov{u}^{t,k}$:
\begin{align*}
    u_c^{t,0} = & ~ \ov{u}^{t,0}\\
    u_c^{t,k} = & ~ u_c^{t,k-1} - \eta_\lo \cdot g_c^{t,k-1}, ~\forall k\geq 1 \\
    \ov{u}^{t,k} = & ~ \ov{u}^{t,k-1} - \eta_\lo \cdot \ov{g}^{t,k-1} + 1_{\{k=0\}} \cdot \eta_\lo\cdot (I_{d} - \dec_t\circ\enc_t)(\sum_{i=0}^{K-1}\ov{g}^{t-1,i}),~\forall (t,k)\neq (1,0)
\end{align*}
where $I_d : \R^d \rightarrow \R^d$ is the identity function.
\end{lemma}
\begin{proof}
First two equation directly follows from the update rule of Algorithm~\ref{alg:fed_learn_kstep_dp}.\\
For $k=1,2,\cdots, K-1$, taking the average of the second equation we obtain:
\begin{align*}
    \ov{u}^{t,k} = & ~ \ov{u}^{t,k-1} - \eta_\lo \cdot \ov{g}^{t,k-1}
\end{align*}
For $k=0$ and $t\geq 2$, we have
\begin{align*}
    \ov{u}^{t,0} = & ~ \ov{u}^{t-1,0} - \eta_\lo \cdot \dec_t(\enc_t(\sum_{i=0}^{K-1}\ov{g}^{t-1,i}))\\
    = & ~ \ov{u}^{t-1,0} - \eta_\lo\sum_{i=0}^{K-1}\ov{g}^{t-1,i} + \eta_\lo\sum_{i=0}^{K-1}\ov{g}^{t-1,i} - \eta_\lo \cdot \dec_t(\enc_t(\sum_{i=0}^{K-1}\ov{g}^{t-1,i}))\\
    = & ~ \ov{u}^{t-1,K-1} - \eta_\lo \cdot \ov{g}^{t-1,K-1} + \eta_\lo\cdot (I_{d} - \dec_t\circ\enc_t)(\sum_{i=0}^{K-1}\ov{g}^{t-1,i})
\end{align*}
Combining above results together, we prove the third equation.
\end{proof}

\subsection{Upper Bounding \texorpdfstring{$\|\ov{g}^{t,k}\|_2^2$}{}}
\begin{lemma}\label{lem:gtk1}
Suppose for any $c\in [N]$, $f_c:\R^d\rightarrow \R$ is convex and $L$-smooth. Then
\begin{align*}
    \|\ov{g}^{t,k}\|_2^2 \leq ~ 2L^2 V^{t,k}+4L (f(\ov{u}^{t,k})-f(w^*))
\end{align*}
\end{lemma}
\begin{proof}
By triangle inequality and Cauchy-Schwartz inequality, we have
\begin{align*}
    \|\ov{g}^{t,k}\|_2^2 = & ~ \|\ov{g}^{t,k}-\nabla f(\ov{u}^{t,k})+\nabla f(\ov{u}^{t,k})\|_2^2 \\
    \leq & ~ 2\|\ov{g}^{t,k}-\nabla f(\ov{u}^{t,k})\|_2^2+2\|\nabla f(\ov{u}^{t,k})\|_2^2
\end{align*}
where the first term can be bounded as
\begin{align*}
    \|\ov{g}^{t,k}-\nabla f(\ov{u}^{t,k})\|_2^2 = & ~ \|\frac{1}{N}\sum_{c=1}^N \nabla f_c(u_c^{t,k})-\frac{1}{N}\sum_{c=1}^N \nabla f_c(\ov{u}^{t,k})\|_2^2 \\
    \leq & ~ \frac{1}{N} \sum_{c=1}^N \|\nabla f_c(u_c^{t,k})-\nabla f_c(\ov{u}^{t,k})\|_2^2 \\
    \leq & ~ \frac{L^2}{N} \sum_{c=1}^N \|u_c^{t,k}-\ov{u}^{t,k}\|_2^2
\end{align*}
and the second term can be bounded as follows:
\begin{align*}
    \|\nabla f(\ov{u}^{t,k})\|_2^2 = & ~ \|\nabla f(\ov{u}^{t,k})-\nabla f(w^*)\|_2^2 \\
    \leq & ~ 2L (f(\ov{u}^{t,k})-f(w^*))
\end{align*}
where the last step follows from that $f$ is $L$-smooth and Fact~\ref{fact:L_smooth_fact}.

Combining bounds on these two terms, we get
\begin{align*}
    \|\ov{g}^{t,k}\|_2^2 \leq & ~ \frac{2L^2}{N}\sum_{c=1}^N \|u_c^{t,k}-\ov{u}^{t,k}\|_2^2+2\|\nabla f(\ov{u}^{t,k})\|_2^2 \\
    \leq & ~ 2L^2 V^{t,k}+4L (f(\ov{u}^{t,k})-f(w^*))
\end{align*}
\end{proof}

\subsection{Lower Bounding \texorpdfstring{$\ip{\ov{u}^{t,k}-w^*}{\ov{g}^{t,k}}$}{}}
\begin{lemma}\label{lem:gtk2}
Suppose each $f_c$ satisfies Assumption~\ref{ass:f_ass} with $\mu\geq 0$, then
\begin{align*}
     \ip{\ov{u}^{t,k}-w^*}{\ov{g}^{t,k}} \geq & ~ f(\ov{u}^{t,k})-f(w^*)-\frac{L}{2}V^{t,k} + \frac{\mu}{2}\|\ov{u}^{t,k}-w^*\|_2^2
\end{align*}
\end{lemma}
\begin{proof}
We will provide a lower bound on this inner product:
\begin{align*}
    \ip{\ov{u}^{t,k}-w^*}{\ov{g}^{t,k}} = & ~ \frac{1}{N}\sum_{c=1}^N \ip{\ov{u}^{t,k}-w^*}{\nabla f_c(u_c^{t,k})} 
\end{align*}
It suffices to consider each term separately:
\begin{align*}
    \ip{\ov{u}^{t,k}-w^*}{\nabla f_c(u_c^{t,k})} = & ~ \ip{\ov{u}^{t,k}-u_c^{t,k}+u_c^{t,k}-w^*}{\nabla f_c(u_c^{t,k})} \\
    = & ~ \ip{\ov{u}^{t,k}-u_c^{t,k}}{\nabla f_c(u_c^{t,k})}+\ip{u_c^{t,k}-w^*}{\nabla f_c(u_c^{t,k})}
\end{align*}
The first term can be lower bounded via $L$-smoothness:
\begin{align*}
    \ip{\ov{u}^{t,k}-u_c^{t,k}}{\nabla f_c(u_c^{t,k})} \geq & ~ f_c(\ov{u}^{t,k})-f_c(u_c^{t,k})-\frac{L}{2}\|\ov{u}^{t,k}-u_c^{t,k}\|_2^2
\end{align*}
The second term can be lower bounded via convexity:
\begin{align*}
    \ip{u_c^{t,k}-w^*}{\nabla f_c(u_c^{t,k})} \geq & ~ f_c(u_c^{t,k})-f_c(w^*) + \frac{\mu}{2}\|u_c^{t,k}-w^*\|_2^2
\end{align*}
Combining these two bounds and average them, we get a lower bound:
\begin{align*}
    \ip{\ov{u}^{t,k}-w^*}{g^{t,k}} \geq & ~ \frac{1}{N}\sum_{c=1}^N (f_c(\ov{u}^{t,k})-f_c(w^*)-\frac{L}{2}\|\ov{u}^{t,k}-u_c^{t,k}\|_2^2 + \frac{\mu}{2}\|u_c^{t,k}-w^*\|_2^2) \\
    \geq & ~ \frac{1}{N}\sum_{c=1}^N (f_c(\ov{u}^{t,k})-f_c(w^*))-\frac{L}{2} V^{t,k} + \frac{\mu}{2}\|\ov{u}^{t,k}-w^*\|_2^2\\
    = & ~ f(\ov{u}^{t,k})-f(w^*)-\frac{L}{2}V^{t,k} + \frac{\mu}{2}\|\ov{u}^{t,k}-w^*\|_2^2
\end{align*}
\end{proof}
\subsection{Upper Bounding Variance within \texorpdfstring{$K$}{} Local Steps}
\begin{lemma}\label{lem:vtk}
Suppose each $f_c$ is convex and $L$-smooth. Assume $\eta_\lo \leq \frac{1}{8LK}$. Then for any $t\geq 1$,
\begin{align*}
    \sum_{k=0}^{K-1} V^{t,k} \leq & ~ 8\eta_{\mathrm{local}}^2L K^2 \sum_{k=0}^{K-1} (f(\ov{u}^{t,k})-f(w^*)) + 4\eta_{\mathrm{local}}^2K^3\sigma^2
\end{align*}
\end{lemma}
\begin{proof}
By Lemma~\ref{lem:update_rule}, we know $V^{t,0} = 0$ for any $t\geq 0$. Consider $k\in\{1,2,\cdots,K-1\}$, we have 
\begin{align}\label{eq:vtk_step1}
    V^{t,k} = & ~ \frac{1}{N} \sum_{c=1}^N \|u_c^{t,k}-\ov{u}^{t,k}\|_2^2 \notag\\
    = & ~ \frac{1}{N} \sum_{c=1}^N \|u_c^{t,0}-\sum_{i=0}^{k-1} \eta_{\mathrm{local}}\cdot g_c^{t,i}-\ov{u}^{t,0}+\sum_{i=0}^{k-1} \eta_{\mathrm{local}}\cdot \ov{g}^{t,i}\|_2^2 \notag\\
    = & ~ \frac{\eta_{\mathrm{local}}^2}{N} \sum_{c=1}^N \|\sum_{i=0}^{k-1} (\ov{g}^{t,i}-g_c^{t,i})\|_2^2 \notag\\
    \leq & ~  \frac{\eta_{\mathrm{local}}^2k}{N} \sum_{c=1}^N \sum_{i=0}^{k-1} \|\ov{g}^{t,i}-g_c^{t,i}\|_2^2 \notag\\
    \leq & ~ \frac{\eta_{\mathrm{local}}^2K}{N} \sum_{c=1}^N \sum_{i=0}^{k-1} \|g_c^{t,i}\|_2^2 
\end{align}
where the second step follows from Lemma~\ref{lem:update_rule}, the last step follows from $\ov{g}^{t,i}$ being the average of $g_c^{t,i}$. By Cauchy-Schwartz inequality, we further have:
\begin{align*}
    \|g_c^{t,i}\|_2^2 \leq & ~ 3\|g_c^{t,i} - \nabla f_c(\ov{u}^{t,i})\|_2^2 + 3\|\nabla f_c(\ov{u}^{t,i}) - \nabla f_c(w^*)\|_2^2 + 3\|\nabla f_c(w^*)\|_2^2 \\
    \leq & ~ 3L^2\|u_c^{t,i} - \ov{u}^{t,i}\|_2^2 + 6L(f_c(\ov{u}^{t,i})-f_c(w^*)+\ip{w^*-\ov{u}^{t,i}}{\nabla f_c(w^*)}) + 3\|\nabla f_c(w^*)\|_2^2.
\end{align*}
where the last step follows from applying $L$-smoothness to the first and second term.

Averaging with respect to $c$,
\begin{align*}
    \frac{1}{N}\sum_{c=1}^N\|g_c^{t,i}\|_2^2 \leq & ~ 3L^2 V^{t,i} + 6L(f(\ov{u}^{t,i})-f(w^*)) + 3\sigma^2.
\end{align*}
Note that the inner product term vanishes since $\frac{1}{N}\sum_{c=1}^N \nabla f_c(w^*)=\nabla f(w^*)=0$.

Plugging back to Eq.~\eqref{eq:vtk_step1}, we obtain
\begin{align*}
    V^{t,k} \leq & ~ \frac{\eta_{\mathrm{local}}^2K}{N} \sum_{c=1}^N \sum_{i=0}^{k-1} \|g_c^{t,i}\|_2^2 \\
    \leq & ~ \eta_{\mathrm{local}}^2K \sum_{i=0}^{k-1} (3L^2 V^{t,i} + 6L(f(\ov{u}^{t,i})-f(w^*)) + 3\sigma^2).
\end{align*}
Summing up above inequality as $k$ varies from $0$ to $K-1$,
\begin{align*}
    \sum_{k=0}^{K-1} V^{t,k} \leq & ~ \eta_{\mathrm{local}}^2K \sum_{k=0}^{K-1}\sum_{i=0}^{k-1} (3L^2 V^{t,i} + 6L(f(\ov{u}^{t,i})-f(w^*)) + 3\sigma^2) \\
    \leq & ~ \eta_{\mathrm{local}}^2K \sum_{k=0}^{K-1}\sum_{i=0}^{K-1} (3L^2 V^{t,i} + 6L(f(\ov{u}^{t,i})-f(w^*)) + 3\sigma^2) \\
    = & ~ 3\eta_{\mathrm{local}}^2L^2K^2 \sum_{i=0}^{K-1} V^{t,i} + 6\eta_{\mathrm{local}}^2LK^2 \sum_{i=0}^{K-1} (f(\ov{u}^{t,i})-f(w^*)) + 3\eta_{\mathrm{local}}^2K^3\sigma^2
\end{align*}
Rearranging terms we obtain:
\begin{align*}
    (1- 3\eta_{\mathrm{local}}^2L^2K^2)\sum_{k=0}^{K-1} V^{t,k} \leq & ~ 6\eta_{\mathrm{local}}^2LK^2 \sum_{i=0}^{K-1} (f(\ov{u}^{t,i})-f(w^*)) + 3\eta_{\mathrm{local}}^2K^3\sigma^2
\end{align*}
Since $\eta_{\mathrm{local}} \leq \frac{1}{8LK}$, we have $1- 3\eta_{\mathrm{local}}^2L^2K^2 \geq \frac{3}{4}$, implying
\begin{align*}
    \sum_{k=0}^{K-1} V^{t,k} \leq & ~ 8\eta_{\mathrm{local}}^2LK^2 \sum_{i=0}^{K-1} (f(\ov{u}^{t,i})-f(w^*)) + 4\eta_{\mathrm{local}}^2K^3\sigma^2
\end{align*}
\end{proof}

\subsection{Bounding the Expected Gap Between \texorpdfstring{$\ov{u}^{t,k}$}{} and \texorpdfstring{$w^*$}{}}
\begin{lemma}\label{lem:rt}
Suppose each $f_c$ satisfies Assumption~\ref{ass:f_ass} with $\mu\geq 0$. If $\enc/\dec$ satisfying Theorem~\ref{thm:enc_dec} and $\eta_\lo \leq \frac{1}{4L}$, then for any $(t,k)\neq (1,0)$, we have
\begin{align*}
     \E[\|\ov{u}^{t,k}-w^*\|_2^2] \leq & ~ (1-\mu\eta_\lo)\E[\|\ov{u}^{t,k-1}-w^*\|_2^2] + \frac{3}{2}\eta_\lo L \E[V^{t,k-1}] - \eta_\lo \E[f(\ov{u}^{t,k-1}) - f(w^*)] \\
    & + 1_{\{k=0\}} \eta_\lo^2 \alpha K \Big(2L^2\sum_{i=0}^{K-1} \E[V^{t-1,i}] + 4L\sum_{i=0}^{K-1} \E[f(\ov{u}^{t-1,i}) - f(w^*)]\Big) 
\end{align*}
\end{lemma}
\begin{proof}
By Lemma~\ref{lem:update_rule}, we have for any $(t,k)\neq (1,0)$,
\begin{align*}
    \ov{u}^{t,k} = & ~ \ov{u}^{t,k-1} - \eta_\lo \cdot \ov{g}^{t,k-1} + 1_{\{k=0\}} \cdot \eta_\lo\cdot (I_d - \dec_t\circ\enc_t)(\sum_{i=0}^{K-1}\ov{g}^{t-1,i}) 
\end{align*}
Therefore, denoting $h^{t}:= (I_d - \dec_t\circ\enc_t)(\sum_{i=0}^{K-1}\ov{g}^{t-1,i}) $, we have
\begin{align}\label{eq:rt_step1}
    \|\ov{u}^{t,k}-w^*\|_2^2 = & ~ \| \ov{u}^{t,k-1} -w^* - \eta_\lo \cdot \ov{g}^{t,k-1} + 1_{\{k=0\}}\eta_\lo\cdot h^{t} \|_2^2\notag\\
    = & ~ \| \ov{u}^{t,k-1} -w^* \|_2^2 + \eta_\lo^2 \cdot \|\ov{g}^{t,k-1}\|_2^2 - 2\eta_\lo \ip{\ov{u}^{t,k-1} -w^*}{\ov{g}^{t,k-1}}\notag\\
    & ~ + 2\eta_\lo 1_{\{k=0\}}\ip{\ov{u}^{t,k-1} -w^*}{h^{t}} - 2 \eta_\lo^2 1_{\{k=0\}}\ip{\ov{g}^{t,k-1}}{h^{t}}\notag\\
    &~ + \eta_\lo^2 1_{\{k=0\}}\cdot \|h^{t}\|_2^2 
\end{align}
Note by Theorem~\ref{thm:enc_dec}, we have:
\begin{align*}
    \E[ \dec_t( \enc_t ( h ) ) ]  = h, \qquad \E[ \| \dec_t( \enc_t ( h ) )\|_2^2] \leq   (1+\alpha) \cdot \| h \|_2^2
\end{align*}
hold for any vector $h$. Hence, by taking expectation over Eq.~\eqref{eq:rt_step1} conditioned on $\mathcal{F}_{t-1}$,
\begin{align*}
    \E[\|\ov{u}^{t,k}-w^*\|_2^2|\mathcal{F}_{t-1}] = & ~ \E[\| \ov{u}^{t,k-1} -w^* \|_2^2|\mathcal{F}_{t-1}] + \eta_\lo^2 \cdot \E[\|\ov{g}^{t,k-1}\|_2^2|\mathcal{F}_{t-1}]\notag\\
    & ~  - 2\eta_\lo \E[\ip{\ov{u}^{t,k-1} -w^*}{\ov{g}^{t,k-1}}|\mathcal{F}_{t-1}] + 1_{\{k=0\}}\cdot \eta_\lo^2 \cdot \E[\|h^{t}\|_2^2|\mathcal{F}_{t-1}]
\end{align*}
Here the two inner products involving $h^t$ vanish. They appear only for $k=0$. Then $\ov{u}^{t,-1}=\ov{u}^{t-1,K-1}$, $\ov{g}^{t,-1}=\ov{g}^{t-1,K-1}$ and $\sum_{i=0}^{K-1}\ov{g}^{t-1,i}$ are functions of $w^{t-1}$, so they are ${\cal F}_{t-1}$-measurable, while $R_t$ is independent of ${\cal F}_{t-1}$. So $\E[h^t\mid {\cal F}_{t-1}]=0$.

Since
\begin{align*}
    \E[\|h^{t}\|_2^2|\mathcal{F}_{t-1}]= &~\E[\|(I_d - \dec_t\circ\enc_t)(\sum_{i=0}^{K-1}\ov{g}^{t-1,i})\|_2^2|\mathcal{F}_{t-1}]\\
    \leq &~ \alpha\E[\|\sum_{i=0}^{K-1}\ov{g}^{t-1,i}\|_2^2|\mathcal{F}_{t-1}]\\
    \leq & ~ \alpha K\sum_{i=0}^{K-1}\E[\|\ov{g}^{t-1,i}\|_2^2|\mathcal{F}_{t-1}]
\end{align*}
Taking total expectation, we have
\begin{align*}
    & ~ \E[\|\ov{u}^{t,k}-w^*\|_2^2] \\
    \leq & ~ \E[\| \ov{u}^{t,k-1} -w^* \|_2^2] + \eta_\lo^2 \cdot \E[\|\ov{g}^{t,k-1}\|_2^2]- 2\eta_\lo \E[\ip{\ov{u}^{t,k-1} -w^*}{\ov{g}^{t,k-1}}]\\
    & ~   + 1_{\{k=0\}}\cdot \eta_\lo^2 \cdot \alpha K\sum_{i=0}^{K-1}\E[\|\ov{g}^{t-1,i}\|_2^2]\\
    \leq & ~ \E[\| \ov{u}^{t,k-1} -w^* \|_2^2] + \eta_\lo^2 \cdot \E[2L^2 V^{t,k-1}+4L (f(\ov{u}^{t,k-1})-f(w^*))]\\
    & ~ - 2\eta_\lo \E[f(\ov{u}^{t,k-1})-f(w^*)-\frac{L}{2}V^{t,k-1} + \frac{\mu}{2}\|\ov{u}^{t,k-1}-w^*\|_2^2] \\
    & ~ + 1_{\{k=0\}}\cdot \eta_\lo^2 \cdot \alpha K\sum_{i=0}^{K-1}\E[2L^2 V^{t-1,i}+4L (f(\ov{u}^{t-1,i})-f(w^*))]\\
    \leq & ~ (1-\mu\eta_\lo)\E[\|\ov{u}^{t,k-1}-w^*\|_2^2] + \eta_\lo \cdot L\cdot (1+2\eta_\lo L)\cdot  \E[V^{t,k-1}]\\
    & ~ - 2\eta_\lo\cdot (1-2\eta_\lo L) \cdot \E[f(\ov{u}^{t,k-1}) - f(w^*)] \\
    & + 1_{\{k=0\}} \cdot \eta_\lo^2 \cdot \alpha  K\cdot  \Big(2L^2\sum_{i=0}^{K-1} \E[V^{t-1,i}] + 4L\sum_{i=0}^{K-1} \E[f(\ov{u}^{t-1,i}) - f(w^*)]\Big) 
\end{align*}
where the second step follows from Lemma~\ref{lem:gtk1} and Lemma~\ref{lem:gtk2}. Since $\eta_\lo\leq \frac{1}{4L}$, we have
\begin{align*}
    \E[\|\ov{u}^{t,k}-w^*\|_2^2] \leq & ~ (1-\mu\eta_\lo)\E[\|\ov{u}^{t,k-1}-w^*\|_2^2] + \frac{3}{2}\eta_\lo L \E[V^{t,k-1}] - \eta_\lo \E[f(\ov{u}^{t,k-1}) - f(w^*)] \\
    & + 1_{\{k=0\}} \eta_\lo^2 \alpha K \Big(2L^2\sum_{i=0}^{K-1} \E[V^{t-1,i}] + 4L\sum_{i=0}^{K-1} \E[f(\ov{u}^{t-1,i}) - f(w^*)]\Big) 
\end{align*}
\end{proof}
\subsection{Main Result: Convex Case}
\begin{theorem}[Formal version of Theorem~\ref{thm:kstep_cvx}]\label{thm:kstep_kmr19}
Assume each $f_c$ is convex and $L$-smooth. If Theorem~\ref{thm:enc_dec} holds, $\eta_\gl = 1$ and $\eta_\lo \leq \frac{1}{8(1+\alpha) L K }$,
\begin{align*}
    \E[f(\ov{w}^T) - f(w^*)] \leq \frac{4\E[\|w^0 - w^*\|_2^2]}{\eta_\lo K T } + 32\eta_\lo^2  L K^2 \sigma^2,
\end{align*}
where $\ov{w}^T = \frac{1}{KT}(\sum_{t=1}^{T}\sum_{k=0}^{K-1} \ov{u}^{t,k})$ is a virtual average of the parameters of Algorithm~\ref{alg:fed_learn_kstep_dp}, used only in the analysis. When $K>1$, no party forms it, since the averages $\ov{u}^{t,k}$ with $k\geq 1$ are never communicated.
\end{theorem}
\begin{proof}
Summing up Lemma~\ref{lem:rt} as $t$ varies from $1$ to $T$ and $k$ varies from $0$ to $K-1$,
\begin{align*}
    & ~ \E[\|\ov{u}^{T+1,0}-w^*\|_2^2]  - \E[\|w^0-w^*\|_2^2]\\
    \leq & ~  \frac{3}{2}\eta_\lo L \sum_{t=1}^T\sum_{k=0}^{K-1} \E[V^{t,k}] - \eta_\lo \sum_{t=1}^T\sum_{k=0}^{K-1} \E[f(\ov{u}^{t,k}) - f(w^*)] \\
    & + \sum_{t=1}^T\sum_{k=0}^{K-1} 1_{\{k=0\}} \eta_\lo^2 \alpha K \Big(2L^2\sum_{i=0}^{K-1} \E[V^{t,i}] + 4L\sum_{i=0}^{K-1} \E[f(\ov{u}^{t,i}) - f(w^*)]\Big) \\
    = & ~ \frac{3}{2}\eta_\lo L \sum_{t=1}^T\sum_{k=0}^{K-1} \E[V^{t,k}] - \eta_\lo \sum_{t=1}^T\sum_{k=0}^{K-1} \E[f(\ov{u}^{t,k}) - f(w^*)] \\
    & + \eta_\lo^2 \alpha K \Big(2L^2 \sum_{t=1}^T\sum_{i=0}^{K-1} \E[V^{t,i}] + 4L \sum_{t=1}^T\sum_{i=0}^{K-1} \E[f(\ov{u}^{t,i}) - f(w^*)]\Big) \\
    = & ~ \eta_\lo L (\frac{3}{2}+ 2\eta_\lo \alpha L K) \sum_{t=1}^T\sum_{k=0}^{K-1} \E[V^{t,k}]\\
    & ~ - \eta_\lo (1 - 4\eta_\lo \alpha L K ) \sum_{t=1}^T\sum_{k=0}^{K-1} \E[f(\ov{u}^{t,k}) - f(w^*)] \\
    \leq & ~ 2\eta_\lo L \sum_{t=1}^T\sum_{k=0}^{K-1} \E[V^{t,k}] - \frac{1}{2}\eta_\lo  \sum_{t=1}^T\sum_{k=0}^{K-1} \E[f(\ov{u}^{t,k}) - f(w^*)] \\
    \leq & ~ 2\eta_\lo L \sum_{t=1}^T (8\eta_{\mathrm{local}}^2LK^2 \sum_{i=0}^{K-1} \E[f(\ov{u}^{t,i})-f(w^*)] + 4\eta_{\mathrm{local}}^2K^3\sigma^2) - \frac{1}{2}\eta_\lo  \sum_{t=1}^T\sum_{k=0}^{K-1} \E[f(\ov{u}^{t,k}) - f(w^*)]\\
    \leq & ~ -\frac{1}{4}\eta_\lo  \sum_{t=1}^T\sum_{k=0}^{K-1} \E[f(\ov{u}^{t,k}) - f(w^*)] + 8\eta_\lo^3 LK^3 T\sigma^2
\end{align*}
where the fourth step follows from $\eta_\lo \leq \frac{1}{8\alpha L K }$, the last step follows from $\eta_\lo \leq \frac{1}{8 L K }$. Rearranging the terms, we obtain
\begin{align*}
    \frac{1}{KT}\sum_{t=1}^T\sum_{k=0}^{K-1} \E[f(\ov{u}^{t,k}) - f(w^*)]\leq \frac{4\E[\|w^0-w^*\|_2^2]}{\eta_\lo KT} + 32\eta_\lo^2LK^2\sigma^2
\end{align*}
Finally, by the convexity of $f$ we complete the proof.
\end{proof}
Now we are ready to answer the question: how much communication cost is sufficient to guarantee $\E[f(\ov{w}^T) - f(w^*)]\leq \epsilon$? we have the following communication cost result:
\begin{theorem}[Formal version of Theorem~\ref{thm:communication_kstep_convex_main}]\label{thm:communication_kstep_convex}
Assume each $f_c$ is convex and $L$-smooth. If Theorem~\ref{thm:enc_dec} holds. With $O\left(\E[\|w^0-w^*\|_2^2]N \max\{\frac{Ld}{\epsilon}, \frac{\sigma\sqrt{ L}}{\epsilon^{3/2}}\}\right)$ words (real numbers) of communication cost,
 Algorithm~\ref{alg:fed_learn_kstep_dp} with $K=1$ and $\eta_\gl = 1$ outputs an $\epsilon$-optimal solution $\ov{w}^T$ satisfying:
\begin{align*}
    \E[f(\ov{w}^T) - f(w^*)]\leq \epsilon,
\end{align*}
where $\ov{w}^T = \frac{1}{T}\sum_{t=1}^{T} w^t$, which every client can compute. The bound is attained at $K=1$ with $b_{\sketch}=O(1)$, that is, $\alpha=\Theta(d)$.
\end{theorem}
\begin{proof}
To calculate the communication complexity, we first note communication only happens in sync steps. Specifically, in each sync step, the algorithm requires $O(Nb_{\sketch} )$ words (real numbers) of communication cost, where $b_{\sketch} $ denotes the sketching dimension. Therefore, the total cost of communication is given by $O(Nb_{\sketch} T)$.

We take $K=1$. Then $u_c^{t,0}=w^t$ for every client $c$, so $\ov{u}^{t,0}=w^t$, and the average $\ov{w}^T$ of Theorem~\ref{thm:kstep_kmr19} becomes $\frac{1}{T}\sum_{t=1}^T w^t$. Every client holds $w^1,\ldots,w^T$, so every client can compute it. The constraints of Theorem~\ref{thm:kstep_kmr19} depend on $K$ and $\eta_\lo$ only through the product $K\eta_\lo$, so taking $K=1$ does not change the optimal value below. To obtain the optimal communication cost for $\epsilon$-optimal solution, we choose $T,\eta_\lo$ and $b_{\sketch} $ by solving the following optimization problem:
\begin{align*}
    \min_{T,\eta_\lo,b_{\sketch},\alpha} & ~ Nb_{\sketch}  T\\
    \mbox{s.t.} \qquad & ~ 0<\eta_\lo\leq \frac{1}{8(1+\alpha)L}\\
    & ~ \frac{4\E[\|w^0 - w^*\|_2^2]}{\eta_\lo T } \leq \frac{\epsilon}{2}\\
    & ~ 32\eta_\lo^2  L \sigma^2 \leq \frac{\epsilon}{2}\\
    & ~ d \geq b_{\sketch}  =  O(\frac{d}{\alpha}) \geq 1
\end{align*}
where $d$ is the parameter dimension and the last constraint is due to Theorem~\ref{thm:enc_dec}. Above constraints imply:
\begin{align*}
    T\geq\frac{8\E[\|w^0 - w^*\|_2^2]}{\eta_\lo \epsilon},\quad \eta_\lo\leq\min\{\frac{1}{8(1+\alpha)L},\frac{1}{8\sigma}\sqrt{\frac{\epsilon}{L}}\}
\end{align*}
Therefore, when $\epsilon\geq \frac{\sigma^2}{(1+\alpha)^2L}$, the optimal solution is given by
\begin{align*}
    \eta_\lo = \frac{1}{8(1+\alpha)L},~T = \frac{64\E[\|w^0 - w^*\|_2^2](1+\alpha)L}{\epsilon},~b_{\sketch}  = O(\frac{d}{\alpha})
\end{align*}
and the corresponding optimal communication cost is $O(\frac{\E[\|w^0 - w^*\|_2^2]LNd}{\epsilon})$.

when $\epsilon<\frac{\sigma^2}{(1+\alpha)^2L}$, the optimal solution is given by
\begin{align*}
    \eta_\lo = \frac{1}{8\sigma}\sqrt{\frac{\epsilon}{L}},~T = \frac{64\E[\|w^0 - w^*\|_2^2]\sigma\sqrt{L}}{\epsilon^{3/2}},~b_{\sketch}  = O(\frac{d}{\alpha})
\end{align*}
and the corresponding optimal communication cost is $O(\frac{\E[\|w^0 - w^*\|_2^2]\sigma\sqrt{L }Nd}{\alpha\epsilon^{3/2}})$.

Combining above two cases, the optimal $\alpha$ is given by $\Theta(d)$, that is, $b_{\sketch}=O(1)$, and the corresponding optimal communication cost will be $O(\E[\|w^0-w^*\|_2^2]N \max\{\frac{Ld}{\epsilon}, \frac{\sigma\sqrt{ L}}{\epsilon^{3/2}}\})$.
\end{proof}
\subsection{Main Result: Strongly-convex Case}
\begin{theorem}[Formal version of Theorem~\ref{thm:kstep_kmr19_strcvx_main}]\label{thm:kstep_kmr19_strcvx}
Assume each $f_c$ is $\mu$-strongly convex and $L$-smooth. If Theorem~\ref{thm:enc_dec} holds, $\eta_\gl = 1$ and $\eta_\lo \leq \frac{1}{8(1+\alpha) L K }$,
\begin{align*}
    \E[f(w^{T+1}) - f(w^*)]  \leq \frac{L}{2}\E[\|w^0-w^*\|_2^2]e^{-\mu\eta_\lo T} + 4\eta_\lo^2 L^2 K^3\sigma^2/\mu.
\end{align*}
\end{theorem}
\begin{proof}
Summing up Lemma~\ref{lem:rt} as $k$ varies from $0$ to $K-1$, then we have for any $t \geq 1$,
\begin{align*}
    & ~ (\E[\|\ov{u}^{t+1,0}-w^*\|_2^2]+\sum_{k=1}^{K-1}\E[\|\ov{u}^{t,k}-w^*\|_2^2])  - (1-\mu\eta_\lo)\sum_{k=0}^{K-1}\E[\|\ov{u}^{t,k}-w^*\|_2^2])\\
    \leq & ~  \frac{3}{2}\eta_\lo L \sum_{k=0}^{K-1} \E[V^{t,k}] - \eta_\lo \sum_{k=0}^{K-1} \E[f(\ov{u}^{t,k}) - f(w^*)] \\
    & + \sum_{k=0}^{K-1} 1_{\{k=0\}} \eta_\lo^2 \alpha K \Big(2L^2\sum_{i=0}^{K-1} \E[V^{t,i}] + 4L\sum_{i=0}^{K-1} \E[f(\ov{u}^{t,i}) - f(w^*)]\Big) \\
    = & ~ \frac{3}{2}\eta_\lo L \sum_{k=0}^{K-1} \E[V^{t,k}] - \eta_\lo \sum_{k=0}^{K-1} \E[f(\ov{u}^{t,k}) - f(w^*)] \\
    & + \eta_\lo^2 \alpha K \Big(2L^2 \sum_{i=0}^{K-1} \E[V^{t,i}] + 4L \sum_{i=0}^{K-1} \E[f(\ov{u}^{t,i}) - f(w^*)]\Big) \\
    = & ~ \eta_\lo L (\frac{3}{2}+ 2\eta_\lo \alpha L K) \sum_{k=0}^{K-1} \E[V^{t,k}] - \eta_\lo (1 - 4\eta_\lo \alpha L K ) \sum_{k=0}^{K-1} \E[f(\ov{u}^{t,k}) - f(w^*)] \\
    \leq & ~ 2\eta_\lo L \sum_{k=0}^{K-1} \E[V^{t,k}] - \frac{1}{2}\eta_\lo  \sum_{k=0}^{K-1} \E[f(\ov{u}^{t,k}) - f(w^*)] \\
    \leq & ~ 2\eta_\lo L  (8\eta_{\mathrm{local}}^2LK^2 \sum_{i=0}^{K-1} \E[f(\ov{u}^{t,i})-f(w^*)] + 4\eta_{\mathrm{local}}^2K^3\sigma^2) - \frac{1}{2}\eta_\lo  \sum_{k=0}^{K-1} \E[f(\ov{u}^{t,k}) - f(w^*)]\\
    \leq & ~ -\frac{1}{4}\eta_\lo  \sum_{k=0}^{K-1} \E[f(\ov{u}^{t,k}) - f(w^*)] + 8\eta_\lo^3 LK^3 \sigma^2
\end{align*}
where the fourth step follows from $\eta_\lo \leq \frac{1}{8\alpha L K }$, the last step follows from $\eta_\lo \leq \frac{1}{8 L K }$. Rearranging the terms, we obtain
\begin{align*}
    \E[\|\ov{u}^{t+1,0}-w^*\|_2^2] \leq (1-\mu\eta_\lo)\E[\|\ov{u}^{t,0}-w^*\|_2^2] + 8\eta_\lo^3 LK^3\sigma^2
\end{align*}
implying
\begin{align*}
    \E[\|\ov{u}^{t+1,0}-w^*\|_2^2] - 8\eta_\lo^2 LK^3\sigma^2/\mu \leq (1-\mu\eta_\lo)(\E[\|\ov{u}^{t,0}-w^*\|_2^2] - 8\eta_\lo^2 LK^3\sigma^2/\mu).
\end{align*}
Therefore, we have
\begin{align*}
    \E[\|w^{T+1}-w^*\|_2^2] - 8\eta_\lo^2 LK^3\sigma^2/\mu
    \leq & ~ (1-\mu\eta_\lo)^T(\E[\|w^0-w^*\|_2^2] - 8\eta_\lo^2 LK^3\sigma^2/\mu)\\
    \leq & ~ \E[\|w^0-w^*\|_2^2]e^{-\mu\eta_\lo T}
\end{align*}
Finally, by $L$-smoothness of function $f$, we obtain
\begin{align*}
    \E[f(w^{T+1}) - f(w^*)] \leq \frac{L}{2}\E[\|w^{T+1}-w^*\|_2^2] \leq \frac{L}{2}\E[\|w^0-w^*\|_2^2]e^{-\mu\eta_\lo T} + 4\eta_\lo^2 L^2 K^3\sigma^2/\mu.
\end{align*}
\end{proof}

\begin{theorem}[Formal version of Corollary~\ref{thm:communication_kstep_strongly_convex_main}]\label{thm:communication_kstep_convex2}
Assume each $f_c$ is $\mu$-strongly convex and $L$-smooth. If Theorem~\ref{thm:enc_dec} holds. With $O\left(\frac{LN}{\mu} \max\{d, \sqrt{\frac{\sigma^2}{\mu\epsilon}}\}\log(\frac{L\E[\|w^0 - w^*\|_2^2]}{\epsilon})\right)$ words (real numbers) of communication cost,
 Algorithm~\ref{alg:fed_learn_kstep_dp} with $\eta_\gl = 1$ outputs an $\epsilon$-optimal solution ${w}^{T+1}$, which every client computes as $w^T+\dec_{T+1}(\Delta\wt w^T)$ from the last server message, satisfying:
\begin{align*}
    \E[f({w}^{T+1}) - f(w^*)]\leq \epsilon.
\end{align*}
The bound is attained at $K=1$ with $b_{\sketch}=O(1)$, that is, $\alpha=\Theta(d)$.
\end{theorem}
\begin{proof}
To calculate the communication complexity, we first note communication only happens in sync steps. Specifically, in each sync step, the algorithm requires $O(Nb_{\sketch} )$ words (real numbers) of communication cost, where $b_{\sketch} $ denotes the sketching dimension. Therefore, the total cost of communication is given by $O(Nb_{\sketch} T)$. To obtain the optimal communication cost for $\epsilon$-optimal solution, we choose $T,K,\eta_\lo$ and $b_{\sketch} $ by solving the following optimization problem:
\begin{align*}
    \min_{T,K,\eta_\lo,b_{\sketch},\alpha} & ~ Nb_{\sketch}  T\\
    \mbox{s.t.} \qquad & ~ 0<\eta_\lo\leq \frac{1}{8(1+\alpha)LK}\\
    & ~ \frac{L}{2}\E[\|w^0-w^*\|_2^2]e^{-\mu\eta_\lo T} \leq \frac{\epsilon}{2}\\
    & ~ 4\eta_\lo^2 L^2 K^3\sigma^2/\mu \leq \frac{\epsilon}{2}\\
    & ~ d \geq b_{\sketch}  =  O(\frac{d}{\alpha}) \geq 1
\end{align*}
where $d$ is the parameter dimension and the last constraint is due to Theorem~\ref{thm:enc_dec}. Above constraints imply: 
\begin{align*}
    T\geq\frac{1}{\mu\eta_\lo}\log(\frac{L\E[\|w^0 - w^*\|_2^2]}{\epsilon}),\quad \eta_\lo\leq\min\{\frac{1}{8(1+\alpha)LK},\frac{1}{2LK\sigma}\sqrt{\frac{\mu\epsilon}{2K}}\} 
\end{align*}
Therefore, the optimal value is given when $K =1$. For $K=1$, the first term of the minimum is at most the second one if and only if $\epsilon\geq \frac{\sigma^2}{8(1+\alpha)^2\mu}$, since
\begin{align*}
    \frac{1}{8(1+\alpha)L}\leq \frac{1}{2L\sigma}\sqrt{\frac{\mu\epsilon}{2}}
    \iff \frac{\sigma}{4(1+\alpha)}\leq \sqrt{\frac{\mu\epsilon}{2}}
    \iff \frac{\sigma^2}{16(1+\alpha)^2}\leq \frac{\mu\epsilon}{2}
    \iff \epsilon\geq \frac{\sigma^2}{8(1+\alpha)^2\mu}.
\end{align*}
When $\epsilon\geq \frac{\sigma^2}{8(1+\alpha)^2\mu}$, the optimal solution is given by
\begin{align*}
    \eta_\lo = \frac{1}{8(1+\alpha)L},~T = \frac{8(1+\alpha)L}{\mu}\log(\frac{L\E[\|w^0 - w^*\|_2^2]}{\epsilon}),~b_{\sketch}  = O(\frac{d}{\alpha})
\end{align*}
and the corresponding optimal communication cost is $O(\frac{LNd}{\mu}\log(\frac{L\E[\|w^0 - w^*\|_2^2]}{\epsilon}))$.

when $\epsilon<\frac{\sigma^2}{8(1+\alpha)^2\mu}$, the optimal solution is given by
\begin{align*}
    \eta_\lo = \frac{1}{2L\sigma}\sqrt{\frac{\mu\epsilon}{2}},~T = \frac{2L\sigma}{\mu^{3/2}}\sqrt{\frac{2}{\epsilon}}\log(\frac{L\E[\|w^0 - w^*\|_2^2]}{\epsilon}), ~b_{\sketch}  = O(\frac{d}{\alpha})
\end{align*}
and the corresponding optimal communication cost is $O(\frac{\sigma LNd}{\alpha \mu^{3/2}\sqrt{\epsilon}}\log(\frac{L\E[\|w^0 - w^*\|_2^2]}{\epsilon}))$. 

Combining above two cases, the optimal $\alpha$ is given by $\Theta(d)$, that is, $b_{\sketch}=O(1)$, and the corresponding optimal communication cost will be $O(\frac{LN}{\mu} \max\{d, \sqrt{\frac{\sigma^2}{\mu\epsilon}}\}\log(\frac{L\E[\|w^0 - w^*\|_2^2]}{\epsilon}))$.
\end{proof}
\section{\texorpdfstring{$k$}{}-step Non-convex \texorpdfstring{$f$}{} Convergence Analysis}
\label{sec:app_kstep_non}
In this section, we present convergence result for non-convex $f$ case in the $k$-local-step regime. In order for the proof to go through, we assume that there exists a universal constant $G$ such that for any $c\in[N]$ and any $w\in\R^d$,
\begin{align*}
    \|\nabla f_c(w)\|_2 \leq & ~ G.
\end{align*}

Throughout the proof, we will use ${\cal F}_t$ to denote the $\sigma$-algebra generated by $w^0$, $w^1$, $\ldots$, $w^t$. Also, we use $\eta$ as a shorthand for $\eta_{\mathrm{global}}\cdot \eta_{\mathrm{local}}$. Algorithm~\ref{alg:fed_learn_kstep_dp} sets $w^1=w^0$, and for every $t\geq 1$, round $t$ starts from $w^t=u_c^{t,0}$ and ends with
\begin{align*}
    w^{t+1} = & ~ w^t+\dec_{t+1}(\Delta \wt{w}^t), \\
    \Delta \wt{w}^t = & ~ \eta_{\mathrm{global}}\cdot\frac{1}{N}\sum_{c=1}^N\enc_{t+1}(\Delta w_c(t)), \\
    \Delta w_c(t) = & ~ -\eta_{\mathrm{local}}\sum_{k=0}^{K-1}\nabla f_c(u_c^{t,k}).
\end{align*}
Hence ${\cal F}_t$ is contained in the $\sigma$-algebra generated by $w^0$ and $R_2,\ldots,R_t$. As the sketching matrices are independent of each other and of $w^0$, $R_{t+1}$ is independent of ${\cal F}_t$. The local iterates $u_c^{t,k}$ are functions of $w^t$, so they are ${\cal F}_t$-measurable.

Note that in $k$-local-step scheme, the average of local gradients is no longer the true gradient, therefore, we can no longer bound everything using the true gradients. This means it's necessary to introduce the gradient norm upper bound assumption.

\begin{lemma}\label{lem:objective_move}
Suppose each $f_c:\R^d\rightarrow \R$ is $L$-smooth (Def.~\ref{def:L_smooth}) and $\|\nabla f_c(w)\|_2\leq G$ for all $c\in[N]$ and $w\in\R^d$. Let $f = \frac{1}{N}\sum_{c=1}^N f_c$, let $\enc/\dec$ functions satisfy Theorem~\ref{thm:enc_dec}, and let $\eta = \eta_{\mathrm{global}}\cdot \eta_{\mathrm{local}}$. Then for every $t\geq 1$,
\begin{align*}
     \E[f(w^{t+1})-f(w^t)\mid {\cal F}_t]
    \leq & ~ -\eta\cdot K\cdot \|\nabla f(w^t)\|_2^2+\frac{\eta\cdot L\cdot K^2\cdot G^2}{2}\cdot \Big(\eta_{\mathrm{local}}+\eta\cdot (1+\alpha) \Big)
\end{align*}
\end{lemma}
\begin{proof}
Since each $f_c$ is $L$-smooth, so is $f$. We start by bounding $f(w^{t+1})-f(w^t)$ without taking conditional expectation:
\begin{align*}
    & ~ f(w^{t+1})-f(w^t) \\
    \leq & ~ \ip{w^{t+1}-w^t}{\nabla f(w^t)}+\frac{L}{2}\|w^{t+1}-w^t\|_2^2 \\
    = & ~ \ip{\dec_{t+1}(\Delta \wt{w}^t)}{\nabla f(w^t)}+\frac{L}{2}\|\dec_{t+1}(\Delta \wt{w}^t)\|_2^2 \\
    = & ~ A + \frac{L}{2}B
\end{align*}
where
\begin{align*}
    A:= & ~ -\ip{\eta_{\mathrm{global}}\cdot \dec_{t+1}(\frac{1}{N}\sum_{c=1}^N\enc_{t+1}(\sum_{k=0}^{K-1}\eta_{\mathrm{local}}\cdot \nabla f_c(u_c^{t,k})))}{\nabla f(w^t)}\\
    B:= & ~ \|\eta_{\mathrm{global}}\cdot \dec_{t+1}(\frac{1}{N}\sum_{c=1}^N\enc_{t+1}(\sum_{k=0}^{K-1}\eta_{\mathrm{local}}\cdot \nabla f_c(u_c^{t,k}))) \|_2^2
\end{align*}

\paragraph{Bounding $\E[A\mid {\cal F}_t]$}
The functions $\enc_{t+1}$ and $\dec_{t+1}$ are linear, and $\E[\dec_{t+1}(\enc_{t+1}(h))]=h$ for any fixed $h$. Since $R_{t+1}$ is independent of ${\cal F}_t$ while $w^t$ and all $u_c^{t,k}$ are ${\cal F}_t$-measurable, we get
\begin{align*}
    \E[A\mid {\cal F}_t] = & ~ -\ip{\eta_{\mathrm{global}}\cdot \frac{1}{N}\sum_{c=1}^N \sum_{k=0}^{K-1} \eta_{\mathrm{local}}\cdot \nabla f_c(u_c^{t,k})}{\nabla f(w^t)} \\
    = & ~ -\eta_{\mathrm{global}}\cdot \ip{\frac{1}{N}\sum_{c=1}^N\sum_{k=0}^{K-1}\Big( \eta_{\mathrm{local}}\cdot \nabla f_c(u_c^{t,k})-\eta_{\mathrm{local}}\cdot\nabla f_c(w^t)+\eta_{\mathrm{local}}\cdot\nabla f_c(w^t)\Big)}{\nabla f(w^t)} \\
    = & ~ -\eta\cdot K\cdot \|\nabla f(w^t)\|_2^2+\eta\cdot \frac{1}{N}\sum_{c=1}^N\sum_{k=0}^{K-1}\ip{\nabla f_c(w^t)-\nabla f_c(u_c^{t,k})}{\nabla f(w^t)}
\end{align*}
where the last step uses $\frac{1}{N}\sum_{c=1}^N\sum_{k=0}^{K-1}\nabla f_c(w^t) = K\cdot \nabla f(w^t)$.

It suffices to bound the inner product, notice for $k=0$, the inner product is 0, since $u_c^{t,0}=w^t$, so assume $k\geq 1$:
\begin{align}\label{eq:ip_diff_sgd}
   & ~ \ip{\nabla f_c(w^t)-\nabla f_c(u_c^{t,k})}{\nabla f(w^t)} \notag \\
   \leq & ~ \|\nabla f_c(u_c^{t,k})-\nabla f_c(w^t) \|_2\cdot \|\nabla f(w^t) \|_2 \notag \\
   \leq & ~ L\cdot \|u_c^{t,k}-w^t\|_2\cdot \|\nabla f(w^t)\|_2
\end{align}
where the gap between $u_c^{t,k}$ and $w^t$ can be further expanded:
\begin{align}\label{eq:gap_sgd}
    \|u_c^{t,k}-w^t\|_2 = & ~ \|u_c^{t,k}-u_c^{t,0}\|_2 \notag \\
    = & ~ \|\eta_{\mathrm{local}}\sum_{i=0}^{k-1} \nabla f_c(u_c^{t,i})\|_2 \notag \\
    \leq & ~ \eta_{\mathrm{local}}\sum_{i=0}^{k-1} \|\nabla f_c(u_c^{t,i})\|_2 \notag \\
    \leq & ~ \eta_{\mathrm{local}}\cdot k\cdot G
\end{align}
Moreover, $\|\nabla f(w^t)\|_2\leq \frac{1}{N}\sum_{c=1}^N\|\nabla f_c(w^t)\|_2\leq G$.

Plug in Eq.~\eqref{eq:gap_sgd} and $\|\nabla f(w^t)\|_2\leq G$ to Eq.~\eqref{eq:ip_diff_sgd}, we get
\begin{align*}
    \ip{\nabla f_c(w^t)-\nabla f_c(u_c^{t,k})}{\nabla f(w^t)} \leq & ~ L\cdot \eta_{\mathrm{local}}\cdot k\cdot G^2
\end{align*}
Put things together, and use $\sum_{k=0}^{K-1} k = \frac{K(K-1)}{2}\leq \frac{K^2}{2}$, we finally obtain a bound on $\E[A\mid {\cal F}_t]$:
\begin{align}
    \E[A\mid {\cal F}_t] \leq & ~ -\eta\cdot K\cdot \|\nabla f(w^t)\|_2^2+\eta\cdot\eta_{\mathrm{local}} \cdot L\cdot (\sum_{k=0}^{K-1} k)\cdot G^2 \notag \\
    \leq & ~ -\eta\cdot K\cdot \|\nabla f(w^t)\|_2^2+\frac{\eta \cdot\eta_{\mathrm{local}} \cdot L\cdot K^2 \cdot G^2}{2}
\end{align}
\paragraph{Bounding $\E[B\mid {\cal F}_t]$} Using the fact that $\enc_{t+1}/\dec_{t+1}$ are linear functions and the Cauchy-Schwarz inequality, we get
\begin{align*}
    B = & ~ \eta_{\mathrm{global}}^2\cdot\eta_{\mathrm{local}}^2\cdot \frac{1}{N^2}\cdot \|\sum_{c=1}^N\sum_{k=0}^{K-1} \dec_{t+1}(\enc_{t+1}(\nabla f_c(u_c^{t,k})))\|_2^2 \\
    \leq & ~ \eta_{\mathrm{global}}^2\cdot\eta_{\mathrm{local}}^2\cdot \frac{1}{N^2}\cdot N \cdot K \sum_{c=1}^N\sum_{k=0}^{K-1} \|\dec_{t+1}(\enc_{t+1}(\nabla f_c(u_c^{t,k})))\|_2^2 \\
    = & ~ \eta^2\cdot \frac{K}{N}\cdot \sum_{c=1}^N\sum_{k=0}^{K-1}\|\dec_{t+1}(\enc_{t+1}(\nabla f_c(u_c^{t,k})))\|_2^2
\end{align*}
Using variance bound of $\dec_{t+1}(\enc_{t+1}(h))$ for the ${\cal F}_t$-measurable vectors $h=\nabla f_c(u_c^{t,k})$, we get
\begin{align}\label{eq:variance_bound}
    \E[B\mid {\cal F}_t] \leq & ~ \eta^2\cdot \frac{K}{N}\cdot (1+\alpha)\cdot \sum_{c=1}^N\sum_{k=0}^{K-1}\|\nabla f_c(u_c^{t,k})\|_2^2 \notag \\
    \leq & ~ \eta^2\cdot \frac{K}{N}\cdot (1+\alpha)\cdot \sum_{c=1}^N\sum_{k=0}^{K-1} G^2 \notag \\
    = & ~ \eta^2\cdot K^2\cdot (1+\alpha)\cdot G^2
\end{align}
\paragraph{Put things together}
Put the bound on $\E[A\mid {\cal F}_t]$ and the bound on $\E[B\mid {\cal F}_t]$, we get
\begin{align}
    & ~ \E[f(w^{t+1})-f(w^t)\mid {\cal F}_t] \notag \\
    \leq & ~ -\eta\cdot K\cdot \|\nabla f(w^t)\|_2^2+\frac{\eta\cdot\eta_{\mathrm{local}}\cdot L\cdot K^2\cdot G^2}{2}+\frac{L}{2}\cdot \eta^2\cdot K^2\cdot (1+\alpha)\cdot G^2 \notag \\
    = & ~ -\eta\cdot K\cdot \|\nabla f(w^t)\|_2^2+\frac{\eta\cdot L\cdot K^2\cdot G^2}{2}\cdot \Big(\eta_{\mathrm{local}}+\eta\cdot (1+\alpha)\Big)\notag \qedhere
\end{align}
\end{proof}

\begin{theorem}
Suppose each $f_c:\R^d\rightarrow \R$ is $L$-smooth (Def.~\ref{def:L_smooth}) and $\|\nabla f_c(w)\|_2\leq G$ for all $c\in[N]$ and $w\in\R^d$. Suppose $f = \frac{1}{N}\sum_{c=1}^N f_c$ has a global minimizer $w^*\in \R^d$, and assume $\enc/\dec$ functions satisfying Theorem~\ref{thm:enc_dec}. Let $\eta = \eta_{\mathrm{global}}\cdot \eta_{\mathrm{local}}$. Then for every $T\geq 1$,
\begin{align*}
    \min_{t\in \{0,1,\ldots,T\}}~\E[\|\nabla f(w^t)\|_2^2] \leq & ~ \frac{\E[f(w^0)]-f(w^*)}{T\cdot \eta\cdot K}+\frac{L K G^2}{2}\cdot \Big(\eta_{\mathrm{local}}+\eta\cdot (1+\alpha) \Big)
\end{align*}
\end{theorem}
\begin{proof}
By Lemma~\ref{lem:objective_move}, we know that for every $t\geq 1$,
\begin{align*}
    \E[f(w^{t+1})\mid {\cal F}_t]-f(w^t) \leq & ~ -\eta\cdot K\cdot \|\nabla f(w^t)\|_2^2+\frac{\eta\cdot L\cdot K^2\cdot G^2}{2}\cdot\Big(\eta_{\mathrm{local}}+\eta\cdot (1+\alpha) \Big)
\end{align*}
Rearranging the inequality, taking total expectation and dividing by $\eta K$, we get
\begin{align}
    \E[\|\nabla f(w^t)\|_2^2] \leq & ~ \frac{1}{\eta K}\cdot (\E[f(w^{t})]-\E[f(w^{t+1})])+\frac{L K G^2}{2}\cdot \Big(\eta_{\mathrm{local}}+\eta\cdot (1+\alpha) \Big)\notag
\end{align}
Sum over the $T$ rounds $t=1,\ldots,T$ and averaging, we arrive at
\begin{align*}
    & ~ \frac{1}{T}\sum_{t=1}^T \E[\|\nabla f(w^t)\|_2^2]\\ \leq & ~ \frac{1}{T\eta K}\cdot (\E[f(w^1)]-\E[f(w^{T+1})])+\frac{L K G^2}{2}\cdot \Big(\eta_{\mathrm{local}}+\eta\cdot (1+\alpha) \Big) \\
    \leq & ~ \frac{1}{T\eta K}\cdot (\E[f(w^0)]-f(w^*))+\frac{L K G^2}{2}\cdot \Big(\eta_{\mathrm{local}}+\eta\cdot (1+\alpha) \Big)
\end{align*}
where the last step uses $w^1=w^0$ and $f(w^{T+1})\geq f(w^*)$. Since $w^0=w^1$, the minimum over $t\in\{0,1,\ldots,T\}$ equals the minimum over $t\in[T]$, which is at most the average on the left-hand side. This proves the theorem.
\end{proof}
\section{Differential Privacy}
\label{sec:dp}
In this section, we consider a special case where each agent $c$ trying to learn upon its local dataset $\mathcal{D}_c$ with corresponding local loss $f_c(x) = \frac{1}{|\mathcal{D}_c|}\sum_{z_i\in \mathcal{D}_c}f_{c}(x, z_i)$, where we overload the notation $f_c$ to denote the local loss for notation simplicity. We assume that the per-example losses satisfy Assumption~\ref{ass:dp_diff}. This assumption holds if every per-example loss is $\ell_c$-Lipschitz, and unlike Lipschitzness, it is compatible with $\mu>0$ in Assumption~\ref{ass:f_ass}. We also assume that the dataset for each agent $c$ is disjoint. Two inputs of the algorithm are neighboring if they differ by replacing one data point of one agent. The batch $\mathcal{D}_c^{t,k}$ in Algorithm~\ref{alg:app:fed_learn_kstep_dp} is selected through a random set of indices drawn from a fixed distribution, independently of the data and of the noise vectors (for example, a uniformly random subset of a fixed size). In particular, it does not depend on the iterates.
\subsection{Differentially Private Algorithm}
\label{sec:dp_alg}
\begin{algorithm}[H]\caption{Private Iterative Sketching-based Federated Learning Algorithm with $K$ local steps}
\label{alg:app:fed_learn_kstep_dp}
\begin{algorithmic}[1]
\Procedure{\textsc{PrivateIterativeSketchingFL}}{}
\State{Each client initializes $w^0$ using the same set of random seed}
\For{$t=1 \to T$} \Comment{$T$ denotes the total number of global steps}
\State{\color{blue} /* Client */}
\ParFor{$c=1\to N$} \Comment{$N$ denotes the total number of clients}
\If{$t=1$}
\State{$u_c^{t,0}\gets w^0$} \Comment{$u_c^{t,0}\in \R^d$}
\Else
\State{$u_c^{t,0}\gets w^{t-1}+\dec_t(\Delta \wt{w}^{t-1})$} \Comment{$\dec_t: \R^{b_\text{sketch}}\rightarrow \R^d$ de-sketch the change}
\EndIf
\State {$w^t\gets u_c^{t,0}$}
\State $\sigma_{\mathrm{DP}}^2 \gets 8\log(1.25/\wh \delta)\ell_c^2/{\wh \epsilon}^2$
\For{$k=1\to K$}
\State {\color{red}$\xi_c^{t,k}\sim {\cal N}(0,\sigma_{\mathrm{DP}}^2\cdot I_{d\times d})\gets \textnormal{Independent Gaussian noise}$ }
\State ${\mathcal{D}}_{c}^{t,k}\gets \textnormal{Sample random batch of local data points}$
\State $u_c^{t,k}\gets u_c^{t,k-1}-\eta_{\mathrm{local}}\cdot (\frac{1}{|\mathcal{D}_c^{t,k}|}\cdot\sum_{z_i\in\mathcal{D}_{c}^{t,k}} \nabla f_{c}(u_c^{t,k-1}, z_i){\color{red}+\xi_c^{t,k}})$
\EndFor
\State {$\Delta w_c(t)\gets u_c^{t,K}-w^t$} 
\State {Client $c$ sends $\enc_{t+1}(\Delta w_c(t))$ to server}\Comment{$\enc_{t+1}:\R^d\rightarrow \R^{b_\text{sketch}}$ sketch the change}
\EndParFor
\State {\color{blue} /* Server */}
\State{$\Delta\wt{w}^t\gets \eta_{\mathrm{global}}\cdot \frac{1}{N}\sum_{c=1}^N \enc_{t+1}(\Delta w_c(t))$} \Comment{$\Delta \wt{w}^t\in \R^{b_\text{sketch}}$}
\State {Server sends $\Delta\wt{w}^t$ to each client}
\EndFor
\EndProcedure
\end{algorithmic}
\end{algorithm}

\subsection{Preliminary}
\label{sec:dp_preli}
We define $(\epsilon,\delta)$-differential privacy \cite{dmns06,dkmmn06} as
\begin{definition}
Let $\epsilon,\delta$ be positive real number and ${\cal M}$ be a randomized mechanism that takes a dataset as input (representing the actions of the trusted party holding the data). Let $\text{im}({\cal M})$ denote the image of ${\cal M}$. The algorithm ${\cal M}$ is said to provide $\epsilon,\delta$-differential privacy if, for all datasets $D_1$ and $D_2$ that differ on a single element (i.e., the data of one person), and all subsets $S$ of $\text{im}({\cal M})$:
\begin{align*}
    \Pr [ {\cal M}(D_1 ) \in S  ] \leq \exp(\epsilon) \cdot \Pr[ {\cal M}(D_2) \in S ] + \delta 
\end{align*}
where the probability is taken over the randomness used by the algorithm.
\end{definition}

\begin{lemma}[Parallel Composition]\label{lem:app:parallel_comp}
Let ${\cal M}_i$ be an $(\epsilon_i,\delta_i)$-DP mechanism and each ${\cal M}_i$ operates on disjoint subsets of the private database, then ${\cal M}_1\circ\ldots \circ {\cal M}_k$ is $(\max_{i\in [k]}\epsilon_i, \max_{i\in [k]}\delta_i)$-DP.
\end{lemma}

\begin{lemma}[Advanced Composition~\cite{drv10}]\label{lem:app:advanced_comp}
Let $\epsilon,\delta'\in (0,1]$ and $\delta\in [0,1]$. Consider the $k$-fold adaptive composition of mechanisms ${\cal M}_1,\ldots,{\cal M}_k$, in which each ${\cal M}_i$ may be chosen depending on the outputs of ${\cal M}_1,\ldots,{\cal M}_{i-1}$, and which releases the outputs of all $k$ mechanisms. If, for every $i\in[k]$ and every fixed value of the outputs of ${\cal M}_1,\ldots,{\cal M}_{i-1}$, the mechanism ${\cal M}_i$ is $(\epsilon,\delta)$-DP, then the $k$-fold adaptive composition is $(\epsilon',\delta'+k\delta)$-DP where
\begin{align*}
    \epsilon' = & ~ \sqrt{2k\log(1/\delta')}\cdot \epsilon+2k\epsilon^2.
\end{align*}
\end{lemma}

\begin{lemma}[Amplification via Sampling (Lemma 4.12 of~\cite{bnsv15})]\label{lem:app:amp_sample}
Let ${\cal M}$ be an $(\epsilon,\delta)$-DP mechanism where $\epsilon\leq 1$. Let ${\cal M'}$ be the mechanism that, given a database $S$ of size $n$, first constructs a database $T\subset S$ by sub-sampling with repetition $k\leq n/2$ rows from $S$ then return ${\cal M}(T)$. Then, ${\cal M'}$ is $(\frac{6\epsilon k}{n},\exp(\frac{6\epsilon k}{n})\frac{4k}{n}\cdot \delta)$-DP.
\end{lemma}

\begin{lemma}[Post-processing (Proposition 2.1 in~\cite{dr14})]\label{lem:app:post_process}
Let ${\cal M}$ be an $(\epsilon,\delta)$-DP mechanism whose image is $R$. Let $f:R\rightarrow \wt R$ be an arbitrary randomized mapping. Then $f\circ {\cal M}$ is $(\epsilon,\delta)$-DP.
\end{lemma}

As the noise we consider follows from a Gaussian distribution, it is necessary to include notions related to the Gaussian mechanism. 

\begin{definition}[$\ell_2$ Sensitivity]\label{def:app:l2_sensitivity}
Let $f:{\cal X}\rightarrow \R^d$, the $\ell_2$ sensitivity of $f$ is
\begin{align*}
    \Delta_2^{(f)} = & ~ \max_{S,S'}\|f(S)-f(S')\|_2,
\end{align*}
where $S, S'$ are neighboring databases.
\end{definition}

It is folklore that adding Gaussian noise with appropriate $\sigma^2$ will provide DP guarantee we needed.

\begin{lemma}[Gaussian Mechanism]\label{lem:app:gaussian_DP}
Let $\epsilon,\delta\in (0,1)$. Let $f:{\cal X}\rightarrow \R^d$ and let $\Delta_2\geq \Delta_2^{(f)}$ be an upper bound on its $\ell_2$ sensitivity. Suppose we define ${\cal M}(Y)=f(Y)+z$, where
\begin{align*}
z\sim {\cal N}(0,2\log(1.25/\delta)\Delta_2^2/\epsilon^2\cdot I).
\end{align*}
Then ${\cal M}$ is $(\epsilon,\delta)$-DP.
\end{lemma}

\subsection{\texorpdfstring{$\ell_2$}{} Sensitivity of the Stochastic Gradient}
\label{sec:dp_l2}
In this section, we bound the $\ell_2$ sensitivity of the stochastic gradient under Assumption~\ref{ass:dp_diff}.

\begin{lemma}\label{lem:app:l2_gradient}
Fix $t\in[T]$, $k\in[K]$, the current iterate $u_c^{t,k-1}$ and the set of indices of the data points in the batch $\mathcal{D}_c^{t,k}$. Consider the stochastic gradient $\frac{1}{|\mathcal{D}_c^{t,k}|}\cdot\sum_{z_i\in\mathcal{D}_{c}^{t,k}} \nabla f_{c}(u_c^{t,k-1}, z_i)$ as in Algorithm~\ref{alg:app:fed_learn_kstep_dp}, viewed as a function of the local dataset $\mathcal{D}_c$. Suppose Assumption~\ref{ass:dp_diff} holds. Let $\wt{\mathcal{D}}_c$ be obtained from $\mathcal{D}_c$ by replacing one data point, and let $\wt{\mathcal{D}}_c^{t,k}$ be the batch of $\wt{\mathcal{D}}_c$ with the same set of indices. Then
\begin{align*}
    \left\|\frac{1}{|\mathcal{D}_c^{t,k}|}\sum_{z_i\in\mathcal{D}_{c}^{t,k}} \nabla f_{c}(u_c^{t,k-1}, z_i) - \frac{1}{|\wt{\mathcal{D}}_c^{t,k}|}\sum_{\wt z_i\in\wt{\mathcal{D}}_{c}^{t,k}} \nabla f_{c}(u_c^{t,k-1}, \wt z_i)\right\|_2 \leq & ~ \frac{2\ell_c}{|\mathcal{D}_c^{t,k}|}\leq 2\ell_c.
\end{align*}
In particular, the $\ell_2$ sensitivity of this stochastic gradient, as a function of $\mathcal{D}_c$, is at most $2\ell_c$.
\end{lemma}

\begin{proof}
Let $q=|\mathcal{D}_c^{t,k}|=|\wt{\mathcal{D}}_c^{t,k}|\geq 1$, and suppose that $\wt{\mathcal{D}}_c$ replaces the data point $z_j$ of $\mathcal{D}_c$ by $\wt z_j$, so that $\wt z_i=z_i$ for every $i\neq j$. If $j$ is not among the indices of the batch, the two batches are equal and the difference is $0$. Otherwise, all terms with $i\neq j$ cancel, and by Assumption~\ref{ass:dp_diff},
\begin{align*}
    \left\|\frac{1}{q}\Big(\nabla f_{c}(u_c^{t,k-1}, z_j)-\nabla f_{c}(u_c^{t,k-1}, \wt z_j)\Big)\right\|_2 \leq \frac{2\ell_c}{q} \leq 2\ell_c,
\end{align*}
as desired.
\end{proof}
The proofs of Lemma~\ref{lem:app:single_agent_dp} and Theorem~\ref{thm:app:alg_dp} below use Assumption~\ref{ass:dp_diff} only through this lemma.

\subsection{Privacy Guarantee of Our Algorithm}
\label{sec:dp_thm}
In this section, we provide a formal analysis on the privacy guarantee of Algorithm~\ref{alg:app:fed_learn_kstep_dp}. We first analyze the $K$ noisy local steps of a single agent in one round. We then prove the guarantee for the whole algorithm by one application of advanced composition over all $TK$ noisy steps of the agent that holds a given data point.

For $t\in[T]$ and $k\in[K]$, we call
\begin{align*}
    \nu_c^{t,k} := \frac{1}{|\mathcal{D}_c^{t,k}|}\cdot\sum_{z_i\in\mathcal{D}_{c}^{t,k}} \nabla f_{c}(u_c^{t,k-1}, z_i)+\xi_c^{t,k}
\end{align*}
the noisy gradient of agent $c$, so that $u_c^{t,k}=u_c^{t,k-1}-\eta_{\mathrm{local}}\cdot \nu_c^{t,k}$.

\begin{lemma}[Formal version of Lemma~\ref{lem:single_agent_dp}]\label{lem:app:single_agent_dp}
Let $\wh \epsilon,\wh \delta\in (0,1)$, $\delta'\in (0,1]$ and $c\in [N]$. Fix $t\in[T]$ and the starting point $u_c^{t,0}$ of round $t$. For agent $c$, the $K$ noisy local steps of round $t$ of Algorithm~\ref{alg:app:fed_learn_kstep_dp}, that is, the mechanism that maps $\mathcal{D}_c$ to $(u_c^{t,1},\ldots,u_c^{t,K})$, are $(\epsilon_K,\delta_K)$-DP, where
\begin{align*}
    \epsilon_K = \sqrt{2K\log(1/\delta')}\cdot \wh \epsilon+2K\wh \epsilon^2, ~~~ \delta_K = \delta'+K\cdot \wh \delta.
\end{align*}
\end{lemma}

\begin{proof}
Let ${\cal M}$ denote this mechanism, and let ${\cal M}_B$ denote it with the indices of the $K$ batches $\mathcal{D}_c^{t,1},\ldots,\mathcal{D}_c^{t,K}$ fixed to $B$.

First fix $B$. We show that, for every $k\in[K]$ and every fixed value of $\nu_c^{t,1},\ldots,\nu_c^{t,k-1}$, the noisy gradient $\nu_c^{t,k}$ is $(\wh \epsilon,\wh \delta)$-DP as a function of $\mathcal{D}_c$. Given these values, the iterate $u_c^{t,k-1}=u_c^{t,0}-\eta_{\mathrm{local}}\sum_{k'<k}\nu_c^{t,k'}$ is fixed, and so are the indices of the batch $\mathcal{D}_c^{t,k}$. By Lemma~\ref{lem:app:l2_gradient}, the minibatch gradient in $\nu_c^{t,k}$ then has $\ell_2$ sensitivity at most $2\ell_c$. The noise $\xi_c^{t,k}\sim{\cal N}(0,\sigma_{\mathrm{DP}}^2\cdot I_{d\times d})$ is independent of everything else, and
\begin{align*}
    \sigma_{\mathrm{DP}}^2 = \frac{8\log(1.25/\wh \delta)\ell_c^2}{\wh \epsilon^2} = \frac{2\log(1.25/\wh \delta)(2\ell_c)^2}{\wh \epsilon^2}.
\end{align*}
So Lemma~\ref{lem:app:gaussian_DP}, applied with $\epsilon=\wh \epsilon$, $\delta=\wh \delta$ and $\Delta_2=2\ell_c$, shows that $\nu_c^{t,k}$ is $(\wh \epsilon,\wh \delta)$-DP.

Hence $(\nu_c^{t,1},\ldots,\nu_c^{t,K})$ is a $K$-fold adaptive composition of $(\wh \epsilon,\wh \delta)$-DP mechanisms. Lemma~\ref{lem:app:advanced_comp}, applied with $k=K$, $\epsilon=\wh \epsilon\in(0,1]$, $\delta=\wh \delta$ and $\delta'$, shows that it is $(\epsilon_K,\delta_K)$-DP. Since $u_c^{t,k}=u_c^{t,0}-\eta_{\mathrm{local}}\sum_{k'\leq k}\nu_c^{t,k'}$ for every $k\in[K]$, the output $(u_c^{t,1},\ldots,u_c^{t,K})$ is a deterministic function of $(\nu_c^{t,1},\ldots,\nu_c^{t,K})$, so ${\cal M}_B$ is $(\epsilon_K,\delta_K)$-DP by Lemma~\ref{lem:app:post_process}.

Finally, let $\mathcal{D}_c$ and $\wt{\mathcal{D}}_c$ be neighboring and let $S$ be a set of outputs. The distribution of the batch indices $B$ does not depend on the data, so taking the expectation over $B$ gives
\begin{align*}
    \Pr[{\cal M}(\mathcal{D}_c)\in S] = & ~ \E_B\big[\Pr[{\cal M}_B(\mathcal{D}_c)\in S]\big] \\
    \leq & ~ \E_B\big[\exp(\epsilon_K)\cdot\Pr[{\cal M}_B(\wt{\mathcal{D}}_c)\in S]+\delta_K\big] \\
    = & ~ \exp(\epsilon_K)\cdot\Pr[{\cal M}(\wt{\mathcal{D}}_c)\in S]+\delta_K.
\end{align*}
Thus ${\cal M}$ is $(\epsilon_K,\delta_K)$-DP.
\end{proof}

\begin{remark}
We want to point out that although we perform sketching on the sum of the local gradients, by Lemma~\ref{lem:app:post_process}, this does not change the privacy guarantee at all.
\end{remark}

\begin{theorem}\label{thm:app:alg_dp}
Suppose Assumption~\ref{ass:dp_diff} holds and the batch indices are drawn from a fixed distribution, independently of the data and of the noise vectors. Let $\wh \epsilon,\wh \delta\in (0,1)$ and $\delta'\in (0,1]$. Then, Algorithm~\ref{alg:app:fed_learn_kstep_dp} is $(\epsilon_{\mathrm{DP}},\delta_{\mathrm{DP}})$-DP, with
\begin{align*}
    \epsilon_{\mathrm{DP}} = \sqrt{2TK\log(1/\delta')}\cdot \wh \epsilon+2TK\wh \epsilon^2, &~ \delta_{\mathrm{DP}} = \delta'+TK\cdot \wh \delta.
\end{align*}
Here the output of the algorithm consists of the sketching matrices and all messages sent by the agents and by the server. If $\wh \epsilon\leq \sqrt{\log(1/\delta')/(2TK)}$, then the second term of $\epsilon_{\mathrm{DP}}$ is at most the first.
\end{theorem}

\begin{proof}
Fix an agent $c\in[N]$ and two neighboring inputs $D$ and $\wt D$ that differ only in one data point of $\mathcal{D}_c$. Since the local datasets are disjoint, every data point belongs to exactly one agent, so this covers all pairs of neighboring inputs. Let $\omega$ collect all randomness of Algorithm~\ref{alg:app:fed_learn_kstep_dp} other than the noise vectors $\xi_c^{t,k}$ of agent $c$: the initialization $w^0$, the sketching matrices $R_2,\ldots,R_{T+1}$, the batch indices and the noise vectors of the agents $c'\neq c$, and the batch indices of agent $c$. The distribution of $\omega$ is the same for both inputs, and $\omega$ is independent of the noise vectors $\xi_c^{t,k}$ of agent $c$. Let ${\cal M}$ denote the algorithm with its output, and let ${\cal M}_\omega$ denote it with $\omega$ fixed.

{\bf Post-processing.} Fix $\omega$. We show by induction on $t$ that $w^t$ is a deterministic function of the noisy gradients $\nu_c^{s,k}$ with $s<t$ and $k\in[K]$, and that all messages of round $t$ are deterministic functions of the noisy gradients $\nu_c^{s,k}$ with $s\leq t$. For $t=1$, $w^1=w^0$ is fixed by $\omega$. Suppose that the claim holds for $w^t$. Then $u_c^{t,k}=w^t-\eta_{\mathrm{local}}\sum_{k'\leq k}\nu_c^{t,k'}$ for every $k\in[K]$, so the message of agent $c$ is
\begin{align*}
    \enc_{t+1}(\Delta w_c(t)) = R_{t+1}(u_c^{t,K}-w^t) = -\eta_{\mathrm{local}}\cdot R_{t+1}\sum_{k=1}^K\nu_c^{t,k},
\end{align*}
a function of these noisy gradients. For $c'\neq c$, the local steps of agent $c'$ in round $t$ are a deterministic function of $w^t$, of its dataset, which is the same for both inputs, and of its batch indices and noise, which are fixed by $\omega$. So its message $\enc_{t+1}(\Delta w_{c'}(t))=R_{t+1}(u_{c'}^{t,K}-w^t)$ is a function of the noisy gradients $\nu_c^{s,k}$ with $s\leq t$. The same holds for the message $\Delta\wt w^t$ of the server, which averages these messages, and for $w^{t+1}=w^t+R_{t+1}^\top\Delta\wt w^t$. This proves the claim. Hence the output of ${\cal M}_\omega$ is $\Psi_\omega(\nu_c^{1,1},\ldots,\nu_c^{T,K})$ for a deterministic map $\Psi_\omega$ that is the same for both inputs (the sketching matrices in the output are fixed by $\omega$). Any model computed from the output, such as the final model, is a further post-processing.

{\bf Composition.} Order the noisy gradients of agent $c$ as $\nu_c^{1,1},\ldots,\nu_c^{1,K},\nu_c^{2,1},\ldots,\nu_c^{T,K}$. Fix $(t,k)$ and the values of all noisy gradients before $\nu_c^{t,k}$. Then $w^t$ is fixed by the claim above, hence so is $u_c^{t,k-1}=w^t-\eta_{\mathrm{local}}\sum_{k'<k}\nu_c^{t,k'}$, and the indices of the batch $\mathcal{D}_c^{t,k}$ are fixed by $\omega$. As in the proof of Lemma~\ref{lem:app:single_agent_dp}, Lemma~\ref{lem:app:l2_gradient} and Lemma~\ref{lem:app:gaussian_DP} with $\Delta_2=2\ell_c$ show that $\nu_c^{t,k}$ is $(\wh \epsilon,\wh \delta)$-DP. So the sequence of the $TK$ noisy gradients of agent $c$ is a $TK$-fold adaptive composition of $(\wh \epsilon,\wh \delta)$-DP mechanisms. Lemma~\ref{lem:app:advanced_comp}, applied with $k=TK$, $\epsilon=\wh \epsilon\in(0,1]$, $\delta=\wh \delta$ and $\delta'$, shows that this sequence is $(\epsilon_{\mathrm{DP}},\delta_{\mathrm{DP}})$-DP. By Lemma~\ref{lem:app:post_process}, so is $\Psi_\omega(\nu_c^{1,1},\ldots,\nu_c^{T,K})$, that is, ${\cal M}_\omega$ is $(\epsilon_{\mathrm{DP}},\delta_{\mathrm{DP}})$-DP for every fixed $\omega$.

{\bf Averaging.} Let $S$ be a set of outputs. Since $\omega$ has the same distribution for both inputs,
\begin{align*}
    \Pr[{\cal M}(D)\in S] = & ~ \E_\omega\big[\Pr[{\cal M}_\omega(D)\in S]\big] \\
    \leq & ~ \E_\omega\big[\exp(\epsilon_{\mathrm{DP}})\cdot\Pr[{\cal M}_\omega(\wt D)\in S]+\delta_{\mathrm{DP}}\big] \\
    = & ~ \exp(\epsilon_{\mathrm{DP}})\cdot\Pr[{\cal M}(\wt D)\in S]+\delta_{\mathrm{DP}}.
\end{align*}
Thus Algorithm~\ref{alg:app:fed_learn_kstep_dp} is $(\epsilon_{\mathrm{DP}},\delta_{\mathrm{DP}})$-DP.

Finally, $2TK\wh \epsilon^2\leq \sqrt{2TK\log(1/\delta')}\cdot \wh \epsilon$ holds if and only if $\wh \epsilon\leq \sqrt{2TK\log(1/\delta')}/(2TK)=\sqrt{\log(1/\delta')/(2TK)}$, which gives the last claim.
\end{proof}

\subsection{Convergence of the Private Algorithm}
\label{sec:dp_conv}
In this section, we analyze Algorithm~\ref{alg:app:fed_learn_kstep_dp} under Assumption~\ref{ass:f_ass}, with $\enc_t=R_t$ and $\dec_t=R_t^\top$ satisfying Theorem~\ref{thm:enc_dec}. For $x\in\R^d$ and a set $B$ of indices of data points of client $c$, let $m_c(x,B):=\frac{1}{|B|}\sum_{i\in B}\nabla f_c(x,z_i)$ be the minibatch gradient. Then the local step of Algorithm~\ref{alg:app:fed_learn_kstep_dp} is $u_c^{t,k}=u_c^{t,k-1}-\eta_\lo\cdot(m_c(u_c^{t,k-1},\mathcal{D}_c^{t,k})+\xi_c^{t,k})$. We write $\sigma_{\mathrm{DP},c}^2$ for the noise variance of client $c$. The results of this section hold for every choice of $\sigma_{\mathrm{DP},c}^2\geq 0$, including the value $8\log(1.25/\wh\delta)\ell_c^2/\wh\epsilon^2$ set in Algorithm~\ref{alg:app:fed_learn_kstep_dp} and the noise-free value $\sigma_{\mathrm{DP},c}=0$.

\begin{assumption}\label{ass:dp_batch}
The initialization $w^0$, the sketching matrices $R_2,R_3,\ldots$, the index sets of the batches $\mathcal{D}_c^{t,k}$ and the noise vectors $\xi_c^{t,k}$, over all $t$, $k$ and $c$, are mutually independent, and $\E[\|w^0\|_2^2]<\infty$. For every client $c$ there is $s_c\geq 0$ such that, for every $x\in\R^d$, the batch $B=\mathcal{D}_c^{t,k}$ drawn by client $c$ in every step $(t,k)$ satisfies
\begin{align*}
\E_B[m_c(x,B)]=\nabla f_c(x),\qquad \E_B[\|m_c(x,B)-\nabla f_c(x)\|_2^2]\leq s_c^2.
\end{align*}
\end{assumption}
We write
\begin{align*}
\ov\Sigma^2:=\frac{1}{N}\sum_{c=1}^N\big(s_c^2+d\,\sigma_{\mathrm{DP},c}^2\big).
\end{align*}

\begin{lemma}\label{lem:app:dp_batch_unif}
Suppose Assumption~\ref{ass:dp_diff} holds and the batch of client $c$ is a uniformly random subset of $\mathcal{D}_c$ of a fixed size $b_c\in[n_c]$, where $n_c:=|\mathcal{D}_c|$. Then for every $x\in\R^d$, $\E_B[m_c(x,B)]=\nabla f_c(x)$ and $\E_B[\|m_c(x,B)-\nabla f_c(x)\|_2^2]\leq 4\ell_c^2/b_c$, and the left side of the last inequality is $0$ if $b_c=n_c$.
\end{lemma}

\begin{theorem}[Private algorithm, $K=1$]\label{thm:app:dp_conv_k1}
Let Assumptions~\ref{ass:f_ass} and~\ref{ass:dp_batch} hold, let $K=1$, $\eta:=\eta_\gl\cdot\eta_\lo$ and $T\geq 1$, and let $w^*$ be a minimizer of $f$. Write $D_0:=\E[\|w^0-w^*\|_2^2]$ and $\Delta_0:=\E[f(w^0)]-f(w^*)$.
\begin{enumerate}
\item If $\mu>0$ and $\eta\leq\frac{1}{(1+\alpha)L}$, then
\begin{align*}
\E[f(w^{T+1})]-f(w^*)\leq(1-\mu\eta)^T\Delta_0+\frac{L\eta(1+\alpha)\ov\Sigma^2}{2\mu N}.
\end{align*}
\item If $\eta\leq\frac{1}{2(1+\alpha)L}$, then $\ov w^T:=\frac{1}{T}\sum_{t=1}^T w^t$ satisfies
\begin{align*}
\E[f(\ov w^T)]-f(w^*)\leq\frac{D_0}{\eta T}+\frac{\eta(1+\alpha)\ov\Sigma^2}{N}.
\end{align*}
\end{enumerate}
For $\ov\Sigma=0$ these are the bounds of Theorems~\ref{thm:single_step_strong_convex} and~\ref{thm:single_step_convex}.
\end{theorem}

\begin{theorem}[Private algorithm, $K\geq 1$]\label{thm:app:dp_conv_k}
Let Assumptions~\ref{ass:f_ass} and~\ref{ass:dp_batch} hold, $\eta_\gl=1$, $\eta_\lo\leq\frac{1}{16(1+\alpha)LK}$ and $T\geq 1$. Let $w^*$ be a minimizer of $f$, $D_0:=\E[\|w^0-w^*\|_2^2]$, and let $\sigma^2$ be as in Section~\ref{sec:problem_setup}.
\begin{enumerate}
\item The virtual average $\ov w^T:=\frac{1}{KT}\sum_{t=1}^T\sum_{k=0}^{K-1}\ov u^{t,k}$, where $\ov u^{t,k}:=\frac{1}{N}\sum_{c=1}^N u_c^{t,k}$, satisfies
\begin{align*}
\E[f(\ov w^T)]-f(w^*)\leq\frac{4D_0}{\eta_\lo KT}+56\eta_\lo^2LK^2\sigma^2+\frac{4(1+2\alpha)\eta_\lo\ov\Sigma^2}{N}+16\eta_\lo^2LK\ov\Sigma^2,
\end{align*}
and the average $\wt w^T:=\frac{1}{T}\sum_{t=1}^T w^t$, which every client can compute, satisfies
\begin{align*}
\E[f(\wt w^T)]-f(w^*)\leq\frac{4D_0}{\eta_\lo T}+56\eta_\lo^2LK^3\sigma^2+\frac{4(1+2\alpha)\eta_\lo K\ov\Sigma^2}{N}+16\eta_\lo^2LK^2\ov\Sigma^2.
\end{align*}
\item If $\mu>0$, then
\begin{align*}
\E[\|w^{T+1}-w^*\|_2^2]\leq\Big(1-\frac{\mu\eta_\lo K}{2}\Big)^TD_0+\frac{28\eta_\lo^2LK^2\sigma^2}{\mu}+\frac{4(1+\alpha)\eta_\lo\ov\Sigma^2}{\mu N}+\frac{8\eta_\lo^2LK\ov\Sigma^2}{\mu},
\end{align*}
and $\E[f(w^{T+1})]-f(w^*)$ is at most $\frac{L}{2}$ times the right side.
\end{enumerate}
\end{theorem}

\begin{corollary}[Privacy and convergence, $K=1$]\label{cor:app:dp_util_k1}
Let Assumptions~\ref{ass:f_ass} and~\ref{ass:dp_diff} hold, let each batch $\mathcal{D}_c^{t,k}$ be a uniformly random subset of $\mathcal{D}_c$ of a fixed size $b_c$, and let the independence condition of Assumption~\ref{ass:dp_batch} and $\E[\|w^0\|_2^2]<\infty$ hold (its batch conditions then hold with $s_c^2=4\ell_c^2/b_c$ by Lemma~\ref{lem:app:dp_batch_unif}). Let $\ov\ell^2:=\frac{1}{N}\sum_{c=1}^N\ell_c^2>0$, $\epsilon\in(0,1]$, $\delta\in(0,1)$, $T\geq 1$ and $\Lambda_m:=\log(2/\delta)\log(2.5m/\delta)$ for $m\geq 1$. Run Algorithm~\ref{alg:app:fed_learn_kstep_dp} with $K=1$, with $\wh\delta=\delta/(2T)$ and with $\wh\epsilon=\epsilon/(2\sqrt{2T\log(2/\delta)})$, let $\eta:=\eta_\gl\cdot\eta_\lo$, and let $w^*$, $D_0$, $\Delta_0$ and $\ov w^T$ be as in Theorem~\ref{thm:app:dp_conv_k1}. Then the algorithm is $(\epsilon,\delta)$-DP, by Theorem~\ref{thm:app:alg_dp} with $\delta'=\delta/2$. Moreover, $\sigma_{\mathrm{DP},c}^2=64T\Lambda_T\ell_c^2/\epsilon^2$ and $\ov\Sigma^2\leq 71\,dT\Lambda_T\ov\ell^2/\epsilon^2$, and the following hold.
\begin{enumerate}
\item If $\mu>0$, $\kappa:=L/\mu$, the number $\Delta$ does not depend on the data and satisfies $\Delta\geq\Delta_0$, $\Upsilon:=\frac{36(1+\alpha)Ld\Lambda_T\ov\ell^2}{\mu N\epsilon^2}<\mu\Delta$ and $\eta=\min\{\frac{1}{(1+\alpha)L},\frac{\log(\mu\Delta/\Upsilon)}{\mu T}\}$, then
\begin{align*}
\E[f(w^{T+1})]-f(w^*)\leq\Delta e^{-T/((1+\alpha)\kappa)}+\frac{\Upsilon}{\mu}\Big(1+\log\frac{\mu\Delta}{\Upsilon}\Big),\qquad \frac{\Upsilon}{\mu}=\frac{36\kappa(1+\alpha)d\Lambda_T\ov\ell^2}{\mu N\epsilon^2}.
\end{align*}
\item If $D>0$ is a number that does not depend on the data with $D\geq D_0$, $\Psi:=\frac{71(1+\alpha)d\Lambda_T\ov\ell^2}{N\epsilon^2}$ and $\eta=\min\{\frac{1}{2(1+\alpha)L},\frac{\sqrt{D/\Psi}}{T}\}$, then
\begin{align*}
\E[f(\ov w^T)]-f(w^*)\leq\frac{2(1+\alpha)LD}{T}+\frac{17\,\ov\ell}{\epsilon}\sqrt{\frac{(1+\alpha)d\,D\Lambda_T}{N}}.
\end{align*}
\end{enumerate}
\end{corollary}
The step size is a parameter of Algorithm~\ref{alg:app:fed_learn_kstep_dp}, and Theorem~\ref{thm:app:alg_dp} holds for each fixed value of it. Since $D_0$ and $\Delta_0$ depend on the data through $f$ and $w^*$, the step sizes above use upper bounds $D$ and $\Delta$ that do not depend on the data. For example, $\Delta=\frac{L}{2}D$ is allowed, since $\Delta_0\leq\frac{L}{2}D_0$ by smoothness.

\begin{corollary}[Privacy and convergence, $K\geq 1$]\label{cor:app:dp_util_k}
Keep the assumptions of Corollary~\ref{cor:app:dp_util_k1}, except that $K\geq1$, $\eta_\gl=1$, $\wh\delta=\delta/(2TK)$ and $\wh\epsilon=\epsilon/(2\sqrt{2TK\log(2/\delta)})$. Then Algorithm~\ref{alg:app:fed_learn_kstep_dp} is $(\epsilon,\delta)$-DP, by Theorem~\ref{thm:app:alg_dp} with $\delta'=\delta/2$, and $\ov\Sigma^2\leq 71\,dTK\Lambda_{TK}\ov\ell^2/\epsilon^2$. Let $\eta_\lo\leq\frac{1}{16(1+\alpha)LK}$, let $\tau:=\eta_\lo KT$, and let $w^*$, $D_0$ and $\sigma^2$ be as in Theorem~\ref{thm:app:dp_conv_k}. Then the virtual average $\ov w^T$ of Theorem~\ref{thm:app:dp_conv_k} satisfies
\begin{align*}
\E[f(\ov w^T)]-f(w^*)\leq\frac{4D_0}{\tau}+\frac{284(1+2\alpha)\tau d\Lambda_{TK}\ov\ell^2}{N\epsilon^2}+\frac{1}{T}\Big(\frac{1136\tau^2Ld\Lambda_{TK}\ov\ell^2}{\epsilon^2}+\frac{56\tau^2L\sigma^2}{T}\Big),
\end{align*}
and the average $\wt w^T=\frac1T\sum_{t=1}^Tw^t$, which every client can compute, satisfies the same bound multiplied by $K$. For $K>1$ the virtual average is used only in the analysis: no party forms it, since it uses the clients' intermediate local iterates. If $\mu>0$, then
\begin{align*}
\E[\|w^{T+1}-w^*\|_2^2]\leq e^{-\mu\tau/2}D_0+\frac{284(1+\alpha)\tau d\Lambda_{TK}\ov\ell^2}{\mu N\epsilon^2}+\frac{1}{\mu T}\Big(\frac{568\tau^2Ld\Lambda_{TK}\ov\ell^2}{\epsilon^2}+\frac{28\tau^2L\sigma^2}{T}\Big).
\end{align*}
\end{corollary}

\paragraph{Notation for the proofs.} For $t\in[T]$ and $k\in\{0,1,\ldots,K\}$, let $u_c^{t,k}$ be the iterates of Algorithm~\ref{alg:app:fed_learn_kstep_dp}, and let
\begin{align*}
\ov u^{t,k}:=\frac{1}{N}\sum_{c=1}^N u_c^{t,k},\quad g_c^{t,k}:=\nabla f_c(u_c^{t,k}),\quad \ov g^{t,k}:=\frac{1}{N}\sum_{c=1}^N g_c^{t,k},\\
V^{t,k}:=\frac{1}{N}\sum_{c=1}^N\|u_c^{t,k}-\ov u^{t,k}\|_2^2,\quad S_t:=R_t^\top R_t.
\end{align*}
For $k\in[K]$, let
\begin{align*}
\zeta_c^{t,k}:=m_c(u_c^{t,k-1},\mathcal{D}_c^{t,k})-g_c^{t,k-1}+\xi_c^{t,k},\qquad \ov\zeta^{t,k}:=\frac{1}{N}\sum_{c=1}^N\zeta_c^{t,k}.
\end{align*}
Then, for $k\in[K]$,
\begin{align}\label{eq:dp_conv_local}
u_c^{t,k}=u_c^{t,k-1}-\eta_\lo(g_c^{t,k-1}+\zeta_c^{t,k}),\qquad \ov u^{t,k}=\ov u^{t,k-1}-\eta_\lo(\ov g^{t,k-1}+\ov\zeta^{t,k}).
\end{align}
The algorithm sets $u_c^{t,0}=w^t$ for every $c$, so $\ov u^{t,0}=w^t$ and $V^{t,0}=0$, and it sets $w^1=w^0$. Since $\enc_{t+1}=R_{t+1}$ is linear, $\Delta\wt w^t=\eta_\gl R_{t+1}(\ov u^{t,K}-w^t)$, and hence
\begin{align}\label{eq:dp_conv_round}
w^{t+1}=w^t+\dec_{t+1}(\Delta\wt w^t)=w^t+\eta_\gl S_{t+1}(\ov u^{t,K}-w^t).
\end{align}
Let ${\cal G}^{t,k}$ be the $\sigma$-algebra generated by $w^0$, by $R_2,\ldots,R_t$, and by the batch index sets and the noise vectors of all clients in the steps $(s,i)$ with $s<t$, or with $s=t$ and $i\leq k$. By induction on $(t,k)$, using~\eqref{eq:dp_conv_local} and~\eqref{eq:dp_conv_round}, the quantities $u_c^{t,k}$, $\ov u^{t,k}$, $\ov g^{t,k}$, $V^{t,k}$ and $\zeta_c^{t,k}$ are ${\cal G}^{t,k}$-measurable, and $w^t$ is ${\cal G}^{t,0}$-measurable. By Assumption~\ref{ass:dp_batch}, the pairs $(\mathcal{D}_c^{t,k},\xi_c^{t,k})$, $c\in[N]$, are independent of each other and of ${\cal G}^{t,k-1}$, and $R_{t+1}$ is independent of ${\cal G}^{t,K}$. We use the following fact repeatedly. If $X$ is ${\cal G}$-measurable and $Y$ is independent of ${\cal G}$, then $\E[\phi(X,Y)\mid{\cal G}]=\psi(X)$ with $\psi(x):=\E[\phi(x,Y)]$.

All iterates are square integrable. This follows by induction from $\E[\|w^0\|_2^2]<\infty$, from the bound $\|\nabla f_c(x)\|_2\leq\|\nabla f_c(w^*)\|_2+L\|x-w^*\|_2$, from the finite conditional second moment of $\zeta_c^{t,k}$ in (P1) below, and from $\E[\|S_{t+1}h\|_2^2]\leq(1+\alpha)\E[\|h\|_2^2]$ in (P0) below. Since $0\leq f(x)-f(w^*)\leq\frac{L}{2}\|x-w^*\|_2^2$, all expectations below are finite.

\textbf{(P0) Sketching.} Let $h$ be a square-integrable ${\cal G}$-measurable random vector, and let $R_{t+1}$ be independent of ${\cal G}$. By items 3 and 4 of Theorem~\ref{thm:enc_dec} and the fact above, $\E[S_{t+1}h\mid{\cal G}]=h$ and $\E[\|S_{t+1}h\|_2^2\mid{\cal G}]\leq(1+\alpha)\|h\|_2^2$. Hence $\E[\|(S_{t+1}-I_d)h\|_2^2\mid{\cal G}]=\E[\|S_{t+1}h\|_2^2\mid{\cal G}]-\|h\|_2^2\leq\alpha\|h\|_2^2$.

\textbf{(P1) One perturbation.} Fix $t$ and $k\in[K]$, and let ${\cal G}={\cal G}^{t,k-1}$. Given ${\cal G}$, the point $u=u_c^{t,k-1}$ is fixed. By Assumption~\ref{ass:dp_batch}, the batch error $m_c(u,\mathcal{D}_c^{t,k})-\nabla f_c(u)$ has conditional mean $0$ and conditional second moment at most $s_c^2$, and $\xi_c^{t,k}$ has mean $0$ and second moment $d\,\sigma_{\mathrm{DP},c}^2$ and is independent of the batch. Hence
\begin{align*}
\E[\zeta_c^{t,k}\mid{\cal G}]=0,\qquad \E[\|\zeta_c^{t,k}\|_2^2\mid{\cal G}]\leq s_c^2+d\,\sigma_{\mathrm{DP},c}^2.
\end{align*}
For $c\neq c'$, the vectors $\zeta_c^{t,k}$ and $\zeta_{c'}^{t,k}$ are conditionally independent given ${\cal G}$, with conditional mean $0$. So the cross terms vanish and $\E[\|\ov\zeta^{t,k}\|_2^2\mid{\cal G}]=\frac{1}{N^2}\sum_{c=1}^N\E[\|\zeta_c^{t,k}\|_2^2\mid{\cal G}]\leq\ov\Sigma^2/N$.

\textbf{(P2) Sums within a round.} For $1\leq i<j\leq K$, $\zeta_c^{t,i}$ is ${\cal G}^{t,j-1}$-measurable, so $\E[\ip{\zeta_c^{t,i}}{\zeta_c^{t,j}}]=\E[\ip{\zeta_c^{t,i}}{\E[\zeta_c^{t,j}\mid{\cal G}^{t,j-1}]}]=0$, and the same holds for $\ov\zeta$. Hence, for $k\in[K]$,
\begin{align*}
\E\Big[\Big\|\sum_{i=1}^k\zeta_c^{t,i}\Big\|_2^2\Big]\leq k\,(s_c^2+d\,\sigma_{\mathrm{DP},c}^2),\qquad \E\Big[\Big\|\sum_{i=1}^k\ov\zeta^{t,i}\Big\|_2^2\Big]\leq\frac{k\,\ov\Sigma^2}{N}.
\end{align*}

\begin{proof}[Proof of Lemma~\ref{lem:app:dp_batch_unif}]
Fix $c$ and $x$, write $n=n_c$ and $b=b_c$, and let $a_i:=\nabla_x f_c(x,z_i)$ for $z_i\in\mathcal{D}_c$. Then $\nabla f_c(x)=\ov a:=\frac{1}{n}\sum_{i=1}^n a_i$. Let $y_i:=a_i-\ov a$, so that $\sum_i y_i=0$. By Assumption~\ref{ass:dp_diff}, $\|y_i\|_2=\|\frac{1}{n}\sum_{j=1}^n(a_i-a_j)\|_2\leq 2\ell_c$. If $b=n$, then $m_c(x,B)=\nabla f_c(x)$. Let $b<n$, so $n\geq 2$. Then $\Pr[i\in B]=b/n$ and $\Pr[i,j\in B]=\frac{b(b-1)}{n(n-1)}$ for $i\neq j$. Hence $\E_B[m_c(x,B)]=\frac{1}{b}\sum_i\frac{b}{n}a_i=\ov a$. Using $\sum_{i\neq j}\ip{y_i}{y_j}=\|\sum_iy_i\|_2^2-\sum_i\|y_i\|_2^2=-\sum_i\|y_i\|_2^2$,
\begin{align*}
\E_B[\|m_c(x,B)-\ov a\|_2^2]=\frac{1}{b^2}\Big(\frac{b}{n}\sum_i\|y_i\|_2^2+\frac{b(b-1)}{n(n-1)}\sum_{i\neq j}\ip{y_i}{y_j}\Big)=\frac{n-b}{b(n-1)}\cdot\frac{1}{n}\sum_{i=1}^n\|y_i\|_2^2\leq\frac{4\ell_c^2}{b}.
\end{align*}
\end{proof}

\begin{proof}[Proof of Theorem~\ref{thm:app:dp_conv_k1}]
Let $K=1$, ${\cal F}_t:={\cal G}^{t,0}$ and $G^t:=\nabla f(w^t)+\ov\zeta^{t,1}$. Since $u_c^{t,0}=w^t$, we have $\ov g^{t,0}=\nabla f(w^t)$, and~\eqref{eq:dp_conv_local} and~\eqref{eq:dp_conv_round} give $w^{t+1}=w^t-\eta S_{t+1}G^t$. The vector $G^t$ is ${\cal G}^{t,1}$-measurable, and $R_{t+1}$ is independent of ${\cal G}^{t,1}$. By (P0) with ${\cal G}={\cal G}^{t,1}$ and by (P1),
\begin{align}
& \E[S_{t+1}G^t\mid{\cal F}_t]=\nabla f(w^t),\notag\\
& \E[\|S_{t+1}G^t\|_2^2\mid{\cal F}_t]\leq(1+\alpha)\E[\|G^t\|_2^2\mid{\cal F}_t]\leq(1+\alpha)\Big(\|\nabla f(w^t)\|_2^2+\frac{\ov\Sigma^2}{N}\Big).\label{eq:dp_conv_A1}
\end{align}

\textbf{Part 1.} By the $L$-smoothness of $f$,
\begin{align*}
f(w^{t+1})\leq f(w^t)-\eta\ip{S_{t+1}G^t}{\nabla f(w^t)}+\frac{L\eta^2}{2}\|S_{t+1}G^t\|_2^2.
\end{align*}
Taking expectation conditioned on ${\cal F}_t$, and using~\eqref{eq:dp_conv_A1} and $L\eta(1+\alpha)\leq 1$,
\begin{align*}
\E[f(w^{t+1})\mid{\cal F}_t]\leq & ~ f(w^t)-\eta\Big(1-\frac{L\eta(1+\alpha)}{2}\Big)\|\nabla f(w^t)\|_2^2+\frac{L\eta^2(1+\alpha)\ov\Sigma^2}{2N}\\
\leq & ~ f(w^t)-\frac{\eta}{2}\|\nabla f(w^t)\|_2^2+\frac{L\eta^2(1+\alpha)\ov\Sigma^2}{2N}.
\end{align*}
The function $f$ is $\mu$-strongly convex, so Fact~\ref{fact:mu_upper_bound} gives $\|\nabla f(w^t)\|_2^2\geq 2\mu(f(w^t)-f(w^*))$. With $\phi_t:=\E[f(w^t)]-f(w^*)$, taking total expectation gives, for every $t\in[T]$,
\begin{align*}
\phi_{t+1}\leq(1-\mu\eta)\phi_t+\frac{L\eta^2(1+\alpha)\ov\Sigma^2}{2N}.
\end{align*}
Since $\mu\leq L$, we have $0\leq 1-\mu\eta<1$. Iterating from $\phi_1=\Delta_0$, which holds since $w^1=w^0$, and summing the geometric series,
\begin{align*}
\phi_{T+1}\leq(1-\mu\eta)^T\Delta_0+\frac{L\eta^2(1+\alpha)\ov\Sigma^2}{2N}\cdot\frac{1}{\mu\eta}=(1-\mu\eta)^T\Delta_0+\frac{L\eta(1+\alpha)\ov\Sigma^2}{2\mu N}.
\end{align*}

\textbf{Part 2.} Expanding,
\begin{align*}
\|w^{t+1}-w^*\|_2^2=\|w^t-w^*\|_2^2-2\eta\ip{w^t-w^*}{S_{t+1}G^t}+\eta^2\|S_{t+1}G^t\|_2^2.
\end{align*}
We use~\eqref{eq:dp_conv_A1}, the convexity of $f$, which gives $\ip{w^t-w^*}{\nabla f(w^t)}\geq f(w^t)-f(w^*)$, and Fact~\ref{fact:L_smooth_fact} with $\nabla f(w^*)=0$, which gives $\|\nabla f(w^t)\|_2^2\leq 2L(f(w^t)-f(w^*))$. Then
\begin{align*}
\E[\|w^{t+1}-w^*\|_2^2\mid{\cal F}_t]\leq & ~ \|w^t-w^*\|_2^2-2\eta\big(1-\eta(1+\alpha)L\big)(f(w^t)-f(w^*))+\frac{\eta^2(1+\alpha)\ov\Sigma^2}{N}\\
\leq & ~ \|w^t-w^*\|_2^2-\eta(f(w^t)-f(w^*))+\frac{\eta^2(1+\alpha)\ov\Sigma^2}{N},
\end{align*}
since $2\eta(1+\alpha)L\leq 1$. Taking total expectation, summing over $t\in[T]$ and using $w^1=w^0$,
\begin{align*}
\eta\sum_{t=1}^T(\E[f(w^t)]-f(w^*))\leq D_0-\E[\|w^{T+1}-w^*\|_2^2]+\frac{T\eta^2(1+\alpha)\ov\Sigma^2}{N}.
\end{align*}
Dropping the nonpositive term, dividing by $\eta T$ and using the convexity of $f$ gives Part 2. For $\ov\Sigma=0$ the two bounds are those of Theorems~\ref{thm:single_step_strong_convex} and~\ref{thm:single_step_convex}.
\end{proof}

\begin{proof}[Proof of Theorem~\ref{thm:app:dp_conv_k}]
Write $\eta:=\eta_\lo$. For $t\in[T]$ and $k\in\{0,\ldots,K\}$ let $r_{t,k}:=\E[\|\ov u^{t,k}-w^*\|_2^2]$, $\phi_{t,k}:=\E[f(\ov u^{t,k})]-f(w^*)\geq 0$, $v_{t,k}:=\E[V^{t,k}]$ and $\gamma_{t,k}:=\E[\|\ov g^{t,k}\|_2^2]$. Also let $r_{T+1,0}:=\E[\|w^{T+1}-w^*\|_2^2]$, so that $r_{t,0}=\E[\|w^t-w^*\|_2^2]$ for every $t\in[T+1]$ and $r_{1,0}=D_0$. Sums $\sum_k$ run over $k\in\{0,\ldots,K-1\}$. The step size gives $\eta LK\leq\frac{1}{16}$ and $\alpha\eta LK\leq\frac{1}{16}$. The proofs of Lemmas~\ref{lem:gtk1} and~\ref{lem:gtk2}, and of the averaged bound on $\frac{1}{N}\sum_c\|g_c^{t,i}\|_2^2$ inside the proof of Lemma~\ref{lem:vtk}, use only Assumption~\ref{ass:f_ass} and the definitions, for arbitrary points $u_c^{t,k}$. So for the iterates of Algorithm~\ref{alg:app:fed_learn_kstep_dp},
\begin{align*}
&\textup{(F1)}\quad \|\ov g^{t,k}\|_2^2\leq 2L^2V^{t,k}+4L(f(\ov u^{t,k})-f(w^*)),\\
&\textup{(F2)}\quad \ip{\ov u^{t,k}-w^*}{\ov g^{t,k}}\geq f(\ov u^{t,k})-f(w^*)-\frac{L}{2}V^{t,k}+\frac{\mu}{2}\|\ov u^{t,k}-w^*\|_2^2,\\
&\textup{(F3)}\quad \frac{1}{N}\sum_{c=1}^N\|g_c^{t,k}\|_2^2\leq 3L^2V^{t,k}+6L(f(\ov u^{t,k})-f(w^*))+3\sigma^2.
\end{align*}

\textbf{Step 1 (local steps).} Let $k\in[K]$. By~\eqref{eq:dp_conv_local}, $\ov u^{t,k}-w^*=(\ov u^{t,k-1}-w^*-\eta\ov g^{t,k-1})-\eta\ov\zeta^{t,k}$, and the first vector is ${\cal G}^{t,k-1}$-measurable. By (P1),
\begin{align*}
\E[\|\ov u^{t,k}-w^*\|_2^2\mid{\cal G}^{t,k-1}]\leq\|\ov u^{t,k-1}-w^*\|_2^2-2\eta\ip{\ov u^{t,k-1}-w^*}{\ov g^{t,k-1}}+\eta^2\|\ov g^{t,k-1}\|_2^2+\frac{\eta^2\ov\Sigma^2}{N}.
\end{align*}
By (F1) and (F2), the right side is at most $(1-\mu\eta)\|\ov u^{t,k-1}-w^*\|_2^2+\eta L(1+2\eta L)V^{t,k-1}-2\eta(1-2\eta L)(f(\ov u^{t,k-1})-f(w^*))+\eta^2\ov\Sigma^2/N$. Since $\eta L\leq\frac14$, taking total expectation gives
\begin{align*}
r_{t,k}\leq(1-\mu\eta)r_{t,k-1}+\frac{3}{2}\eta Lv_{t,k-1}-\eta\phi_{t,k-1}+\frac{\eta^2\ov\Sigma^2}{N}.
\end{align*}
Summing over $k\in[K]$,
\begin{align}\label{eq:dp_conv_S1}
r_{t,K}\leq r_{t,0}-\mu\eta\sum_k r_{t,k}+\frac{3}{2}\eta L\sum_k v_{t,k}-\eta\sum_k\phi_{t,k}+\frac{K\eta^2\ov\Sigma^2}{N}.
\end{align}

\textbf{Step 2 (sketching).} Let $\Delta^t:=\ov u^{t,K}-w^t=-\eta\sum_{k=1}^K(\ov g^{t,k-1}+\ov\zeta^{t,k})$, which is ${\cal G}^{t,K}$-measurable. By~\eqref{eq:dp_conv_round} with $\eta_\gl=1$, $w^{t+1}-w^*=(\ov u^{t,K}-w^*)+(S_{t+1}-I_d)\Delta^t$. By (P0), the cross term has conditional mean $0$ given ${\cal G}^{t,K}$, so $r_{t+1,0}\leq r_{t,K}+\alpha\E[\|\Delta^t\|_2^2]$. By $\|a+b\|_2^2\leq 2\|a\|_2^2+2\|b\|_2^2$, the Cauchy-Schwarz inequality, (P2) and (F1),
\begin{align*}
\E[\|\Delta^t\|_2^2]\leq 2\eta^2K\sum_k\gamma_{t,k}+\frac{2\eta^2K\ov\Sigma^2}{N}\leq 4\eta^2KL^2\sum_k v_{t,k}+8\eta^2KL\sum_k\phi_{t,k}+\frac{2\eta^2K\ov\Sigma^2}{N}.
\end{align*}
Hence
\begin{align}\label{eq:dp_conv_S2}
r_{t+1,0}\leq r_{t,K}+4\alpha\eta^2KL^2\sum_k v_{t,k}+8\alpha\eta^2KL\sum_k\phi_{t,k}+\frac{2\alpha\eta^2K\ov\Sigma^2}{N}.
\end{align}

\textbf{Step 3 (client drift).} Let $k\in\{0,\ldots,K-1\}$. By~\eqref{eq:dp_conv_local} and $u_c^{t,0}=\ov u^{t,0}$,
\begin{align*}
u_c^{t,k}-\ov u^{t,k}=-\eta\sum_{i=1}^k\big[(g_c^{t,i-1}-\ov g^{t,i-1})+(\zeta_c^{t,i}-\ov\zeta^{t,i})\big].
\end{align*}
By $\|a+b\|_2^2\leq 2\|a\|_2^2+2\|b\|_2^2$, the Cauchy-Schwarz inequality on the first sum, and $\frac{1}{N}\sum_c\|x_c-\ov x\|_2^2\leq\frac{1}{N}\sum_c\|x_c\|_2^2$ for both sums,
\begin{align*}
V^{t,k}\leq 2\eta^2k\sum_{i=0}^{k-1}\frac{1}{N}\sum_{c=1}^N\|g_c^{t,i}\|_2^2+\frac{2\eta^2}{N}\sum_{c=1}^N\Big\|\sum_{i=1}^k\zeta_c^{t,i}\Big\|_2^2.
\end{align*}
By (F3) and (P2),
\begin{align*}
v_{t,k}\leq 2\eta^2k\sum_{i=0}^{k-1}(3L^2v_{t,i}+6L\phi_{t,i}+3\sigma^2)+2\eta^2k\,\ov\Sigma^2.
\end{align*}
We sum over $k$. All summands are nonnegative and $\sum_k k\leq K^2/2$, so
\begin{align*}
\sum_k v_{t,k}\leq 3\eta^2L^2K^2\sum_k v_{t,k}+6\eta^2LK^2\sum_k\phi_{t,k}+3\eta^2K^3\sigma^2+\eta^2K^2\ov\Sigma^2.
\end{align*}
Since $3\eta^2L^2K^2\leq\frac12$,
\begin{align}\label{eq:dp_conv_S3}
\sum_k v_{t,k}\leq 12\eta^2LK^2\sum_k\phi_{t,k}+6\eta^2K^3\sigma^2+2\eta^2K^2\ov\Sigma^2.
\end{align}

\textbf{Step 4 (one round).} Adding~\eqref{eq:dp_conv_S1} and~\eqref{eq:dp_conv_S2}, and using $4\alpha\eta KL\leq\frac14$ and $8\alpha\eta KL\leq\frac12$,
\begin{align}\label{eq:dp_conv_S4}
r_{t+1,0}\leq r_{t,0}-\mu\eta\sum_k r_{t,k}+\frac{7}{4}\eta L\sum_k v_{t,k}-\frac{1}{2}\eta\sum_k\phi_{t,k}+\frac{(1+2\alpha)\eta^2K\ov\Sigma^2}{N}.
\end{align}

\textbf{Part 1.} Drop the term $-\mu\eta\sum_k r_{t,k}\leq 0$. By~\eqref{eq:dp_conv_S3} and $\eta LK\leq\frac{1}{16}$,
\begin{align*}
\frac{7}{4}\eta L\sum_k v_{t,k}\leq & ~ 21\eta(\eta LK)^2\sum_k\phi_{t,k}+\frac{21}{2}\eta^3LK^3\sigma^2+\frac{7}{2}\eta^3LK^2\ov\Sigma^2\\
\leq & ~ \frac{\eta}{4}\sum_k\phi_{t,k}+14\eta^3LK^3\sigma^2+4\eta^3LK^2\ov\Sigma^2.
\end{align*}
Hence
\begin{align*}
r_{t+1,0}\leq r_{t,0}-\frac{\eta}{4}\sum_k\phi_{t,k}+E_c,\qquad E_c:=14\eta^3LK^3\sigma^2+\eta^2K\Big(\frac{1+2\alpha}{N}+4\eta LK\Big)\ov\Sigma^2.
\end{align*}
Summing over $t\in[T]$ and using $r_{1,0}=D_0$ and $r_{T+1,0}\geq 0$ gives $\frac{\eta}{4}\sum_{t=1}^T\sum_k\phi_{t,k}\leq D_0+TE_c$. Dividing by $\eta KT/4$,
\begin{align*}
\frac{1}{KT}\sum_{t=1}^T\sum_k\phi_{t,k}\leq\frac{4D_0}{\eta KT}+56\eta^2LK^2\sigma^2+\frac{4(1+2\alpha)\eta\ov\Sigma^2}{N}+16\eta^2LK\ov\Sigma^2,
\end{align*}
and the bound for $\ov w^T$ follows from the convexity of $f$. Since $\ov u^{t,0}=w^t$ and $\phi_{t,k}\geq 0$, we have $\frac{1}{T}\sum_{t=1}^T\phi_{t,0}\leq K\cdot\frac{1}{KT}\sum_{t=1}^T\sum_k\phi_{t,k}$, and the convexity of $f$ gives the bound for $\wt w^T$.

\textbf{Part 2.} Let $\mu>0$. Assumption~\ref{ass:f_ass} forces $\mu\leq L$. For $k\in\{0,\ldots,K-1\}$, $\|w^t-w^*\|_2^2\leq 2\|\ov u^{t,k}-w^*\|_2^2+2\|\ov u^{t,k}-w^t\|_2^2$, so $r_{t,k}\geq\frac12 r_{t,0}-\E[\|\ov u^{t,k}-w^t\|_2^2]$. By~\eqref{eq:dp_conv_local}, $\ov u^{t,k}-w^t=-\eta\sum_{i=1}^k(\ov g^{t,i-1}+\ov\zeta^{t,i})$, so by the Cauchy-Schwarz inequality and (P2), $\E[\|\ov u^{t,k}-w^t\|_2^2]\leq 2\eta^2k\sum_{i=0}^{k-1}\gamma_{t,i}+2\eta^2k\,\ov\Sigma^2/N$. Summing over $k$ with $\sum_k k\leq K^2/2$ and using (F1),
\begin{align*}
\sum_k\E[\|\ov u^{t,k}-w^t\|_2^2]\leq\eta^2K^2\Big(2L^2\sum_k v_{t,k}+4L\sum_k\phi_{t,k}\Big)+\frac{\eta^2K^2\ov\Sigma^2}{N}.
\end{align*}
Multiplying by $\mu\eta$ and using $\mu\leq L$ and $\eta LK\leq\frac{1}{16}$,
\begin{align*}
-\mu\eta\sum_k r_{t,k}\leq-\frac{\mu\eta K}{2}r_{t,0}+\frac{\eta L}{128}\sum_k v_{t,k}+\frac{\eta}{64}\sum_k\phi_{t,k}+\frac{\eta^2K\ov\Sigma^2}{16N}.
\end{align*}
Inserting this in~\eqref{eq:dp_conv_S4}, and using $\frac74+\frac{1}{128}\leq 2$ and $\frac{17}{16}+2\alpha\leq 2(1+\alpha)$,
\begin{align*}
r_{t+1,0}\leq\Big(1-\frac{\mu\eta K}{2}\Big)r_{t,0}+2\eta L\sum_k v_{t,k}-\frac{31}{64}\eta\sum_k\phi_{t,k}+\frac{2(1+\alpha)\eta^2K\ov\Sigma^2}{N}.
\end{align*}
By~\eqref{eq:dp_conv_S3}, $2\eta L\sum_k v_{t,k}\leq\frac{3}{32}\eta\sum_k\phi_{t,k}+12\eta^3LK^3\sigma^2+4\eta^3LK^2\ov\Sigma^2$, and $\frac{3}{32}\leq\frac{31}{64}$. Hence, as $12\leq 14$,
\begin{align*}
r_{t+1,0}\leq\Big(1-\frac{\mu\eta K}{2}\Big)r_{t,0}+E_s,\qquad E_s:=14\eta^3LK^3\sigma^2+2\eta^2K\Big(\frac{1+\alpha}{N}+2\eta LK\Big)\ov\Sigma^2.
\end{align*}
Since $0<\frac{\mu\eta K}{2}\leq\frac{1}{32}$, iterating from $r_{1,0}=D_0$ gives
\begin{align*}
r_{T+1,0}\leq\Big(1-\frac{\mu\eta K}{2}\Big)^TD_0+\frac{2E_s}{\mu\eta K}=\Big(1-\frac{\mu\eta K}{2}\Big)^TD_0+\frac{28\eta^2LK^2\sigma^2}{\mu}+\frac{4(1+\alpha)\eta\ov\Sigma^2}{\mu N}+\frac{8\eta^2LK\ov\Sigma^2}{\mu}.
\end{align*}
This is the first claim of Part 2. The second follows from $f(w^{T+1})-f(w^*)\leq\frac{L}{2}\|w^{T+1}-w^*\|_2^2$, which holds since $f$ is $L$-smooth and $\nabla f(w^*)=0$.
\end{proof}

\begin{proof}[Proof of Corollary~\ref{cor:app:dp_util_k1}]
\textbf{Privacy.} Theorem~\ref{thm:app:alg_dp} applies, since Lemma~\ref{lem:app:l2_gradient} needs only Assumption~\ref{ass:dp_diff}. Its hypotheses hold, since $\wh\delta=\delta/(2T)\in(0,1)$, $\delta'=\delta/2\in(0,1]$ and $\wh\epsilon\leq 1/(2\sqrt{2\log 2})<1$. With $K=1$ it gives
\begin{align*}
\epsilon_{\mathrm{DP}}=\sqrt{2T\log(2/\delta)}\cdot\wh\epsilon+2T\wh\epsilon^2=\frac{\epsilon}{2}+\frac{\epsilon^2}{4\log(2/\delta)}\leq\epsilon,\qquad \delta_{\mathrm{DP}}=\frac{\delta}{2}+T\cdot\frac{\delta}{2T}=\delta,
\end{align*}
where the inequality uses $\epsilon\leq 1<2\log 2<2\log(2/\delta)$.

\textbf{Noise level.} The noise variance of client $c$ is $\sigma_{\mathrm{DP},c}^2=8\log(1.25/\wh\delta)\ell_c^2/\wh\epsilon^2=64T\Lambda_T\ell_c^2/\epsilon^2$. By Lemma~\ref{lem:app:dp_batch_unif}, Assumption~\ref{ass:dp_batch} holds with $s_c^2=4\ell_c^2/b_c\leq 4\ell_c^2$. Since $d,T\geq 1$, $\epsilon\leq 1$ and $\Lambda_T\geq\log 2\cdot\log 2.5>0.63$, we get $d\,\sigma_{\mathrm{DP},c}^2\geq 40\ell_c^2\geq 10s_c^2$. So $s_c^2+d\,\sigma_{\mathrm{DP},c}^2\leq 1.1\,d\,\sigma_{\mathrm{DP},c}^2\leq 71\,dT\Lambda_T\ell_c^2/\epsilon^2$, and averaging over $c$ gives the bound on $\ov\Sigma^2$.

\textbf{Part 1.} By Part 1 of Theorem~\ref{thm:app:dp_conv_k1}, the bound on $\ov\Sigma^2$, $(1-\mu\eta)^T\leq e^{-\mu\eta T}$ and $\frac{71}{2}\leq 36$,
\begin{align*}
\E[f(w^{T+1})]-f(w^*)\leq e^{-\mu\eta T}\Delta_0+\eta T\Upsilon\leq e^{-\mu\eta T}\Delta+\eta T\Upsilon,
\end{align*}
where the last step uses $\Delta_0\leq\Delta$. Let $\tau_0:=\log(\mu\Delta/\Upsilon)/\mu>0$, so that $e^{-\mu\tau_0}\Delta=\Upsilon/\mu$. If $\eta=\tau_0/T$, the bound equals $\frac{\Upsilon}{\mu}(1+\log\frac{\mu\Delta}{\Upsilon})$. Otherwise $\eta=\frac{1}{(1+\alpha)L}<\frac{\tau_0}{T}$, the first term is $\Delta e^{-T/((1+\alpha)\kappa)}$, and the second is at most $\tau_0\Upsilon=\frac{\Upsilon}{\mu}\log\frac{\mu\Delta}{\Upsilon}$. In both cases the claimed bound holds.

\textbf{Part 2.} By Part 2 of Theorem~\ref{thm:app:dp_conv_k1}, the bound on $\ov\Sigma^2$ and $D_0\leq D$, $\E[f(\ov w^T)]-f(w^*)\leq\frac{D_0}{\eta T}+\eta T\Psi\leq\frac{D}{\eta T}+\eta T\Psi$. If $\eta=\sqrt{D/\Psi}/T$, the last bound equals $2\sqrt{D\Psi}$. Otherwise $\eta=\frac{1}{2(1+\alpha)L}<\frac{\sqrt{D/\Psi}}{T}$, so $\frac{D}{\eta T}=\frac{2(1+\alpha)LD}{T}$ and $\eta T\Psi<\sqrt{D\Psi}$. In both cases the bound is at most $\frac{2(1+\alpha)LD}{T}+2\sqrt{D\Psi}$, and $2\sqrt{71}\leq 17$.
\end{proof}

\begin{proof}[Proof of Corollary~\ref{cor:app:dp_util_k}]
As in the proof of Corollary~\ref{cor:app:dp_util_k1}, with $T$ replaced by $TK$, Theorem~\ref{thm:app:alg_dp} gives $\epsilon_{\mathrm{DP}}=\frac{\epsilon}{2}+\frac{\epsilon^2}{4\log(2/\delta)}\leq\epsilon$ and $\delta_{\mathrm{DP}}=\frac{\delta}{2}+TK\cdot\frac{\delta}{2TK}=\delta$. The noise variance is $\sigma_{\mathrm{DP},c}^2=64TK\Lambda_{TK}\ell_c^2/\epsilon^2$, and $\ov\Sigma^2\leq 71\,dTK\Lambda_{TK}\ov\ell^2/\epsilon^2$. Substitute $\eta_\lo=\tau/(KT)$ and this bound into Theorem~\ref{thm:app:dp_conv_k}. In Part 1, we have $56\eta_\lo^2LK^2\sigma^2=\frac{56\tau^2L\sigma^2}{T^2}$, $\frac{4(1+2\alpha)\eta_\lo\ov\Sigma^2}{N}\leq\frac{284(1+2\alpha)\tau d\Lambda_{TK}\ov\ell^2}{N\epsilon^2}$ and $16\eta_\lo^2LK\ov\Sigma^2\leq\frac{1136\tau^2Ld\Lambda_{TK}\ov\ell^2}{T\epsilon^2}$. For $\wt w^T$, the same substitution in the second bound of Part 1 of Theorem~\ref{thm:app:dp_conv_k} multiplies each of the four terms by $K$. In Part 2, $(1-\frac{\mu\eta_\lo K}{2})^T\leq e^{-\mu\tau/2}$, $\frac{28\eta_\lo^2LK^2\sigma^2}{\mu}=\frac{28\tau^2L\sigma^2}{\mu T^2}$, $\frac{4(1+\alpha)\eta_\lo\ov\Sigma^2}{\mu N}\leq\frac{284(1+\alpha)\tau d\Lambda_{TK}\ov\ell^2}{\mu N\epsilon^2}$ and $\frac{8\eta_\lo^2LK\ov\Sigma^2}{\mu}\leq\frac{568\tau^2Ld\Lambda_{TK}\ov\ell^2}{\mu T\epsilon^2}$.
\end{proof} 

\section{Preliminary on Gradient Attack}\label{sec:conditions:app}

Throughout this section to the remainder of the paper, we use $F(x;w)$ to denote the loss function of the model, where $x\in \R^m$ is the data point and $w\in \R^d$ is the model parameter.

\subsection{Definitions}
We start with defining some conditions we will later study:

\begin{definition}
Let $F:\R^m\times \R^d\rightarrow \R$, we define the loss function $L$ to be
\begin{align*}
    L(x) := & ~ \|\nabla_w F(x;w)-g\|^2.
\end{align*}
\end{definition}

\begin{definition}[Smoothness]\label{def:smoothness:app}
We say $L:\R^m\rightarrow \R$ is $b$-smooth if for any $x,y\in \R^m$, we have
\begin{align*}
    L(y) \leq L(x) + \langle \nabla L(x), y - x \rangle + b \| y - x \|^2.
\end{align*}
\end{definition}

\begin{definition}[Lipschitz]\label{def:Lipschitz:app}
We say $L:\R^m\rightarrow \R$ is $\beta$-Lipschitz if for any $x,y\in \R^m$, we have
\begin{align*}
    \| L(x) - L(y) \|^2 \leq \beta^2 \| x - y \|^2.
\end{align*}
\end{definition}
\begin{definition}[Semi-smoothness]\label{def:semi_smooth:app}
For any $p \in [0,1]$ and any set $S\subseteq\R^m$, we say $L$ is $(a,b,p)$-semi-smooth on $S$ if for any $x\in S$ and any $y\in\R^m$,
\begin{align*}
    L(y) \le & ~ L(x) + \langle \nabla L(x), y-x \rangle  + b\|y-x\|^2  \\
    &~ + a \|x-y\|^{2-2p}L(x)^{p}.
\end{align*}
We say $L$ is $(a,b,p)$-semi-smooth if this holds with $S=\R^m$. When $a>0$ and $p>0$, the term $L(x)^p$ presumes $L\geq 0$ on $S$.
\end{definition}

\begin{definition}[Semi-Lipschitz]\label{def:semi_lipschitz:app}
For any $p\in [0,1]$, we say function $L$ is $(\alpha,\beta,p)$-semi-Lipschitz if
\begin{align*}
    (L ( x ) - L( y ) )^2  \leq  \beta^2  \| x - y \|^2  + \alpha^2 \| x  - y \|^{2-2p} \cdot L(x)^{p}.
\end{align*}
Specifically, we say function $L$ has $(\alpha,\beta,p)$-semi-Lipschitz gradient, or $L$ satisfies $(\alpha,\beta,p)$-semi-Lipschitz gradient condition, if
\begin{align*}
    \|\nabla L ( x ) - \nabla L( y ) \|^2  \leq & ~ \beta^2  \| x - y \|^2 \\
    & ~ + \alpha^2 \| x  - y \|^{2-2p} \cdot L(x)^{p}.
\end{align*}
\end{definition}

\begin{definition}[Non-critical point]\label{def:no_critical_point:app}
We say $L$ is $(\theta_1,\theta_2)$-non-critical point if 
\begin{align*}
\theta_1^2 \cdot L(x) \leq \| \nabla L(x) \|^2 \leq \theta_2^2 \cdot L(x) .
\end{align*}
\end{definition}

\begin{definition}[Pseudo-Hessian]\label{def:Phi_K:app}
Let $F:\R^m\times \R^d\rightarrow \R$, suppose $F$ is differentiable on both $x$ and $w$, then we define pseudo-Hessian mapping $\Phi : \R^{m} \times \R^d \rightarrow \R^{m \times d}$ as follows
\begin{align*}
    \Phi(x,w) = \nabla_x \nabla_w F(x;w).
\end{align*}
Thus $\Phi(x,w)^\top\in\R^{d\times m}$ is the Jacobian of the map $x\mapsto\nabla_w F(x;w)$, and the chain rule gives $\nabla L(x)=2\Phi(x,w)(\nabla_w F(x;w)-g)$.
Correspondingly, we define a pseudo-kernel $K : \R^m\times\R^d \rightarrow \R^{d \times d}$ with respect to $\nabla_x F(x;w)$ as:
\begin{align*}
    K(x, w) = \Phi(x,w)^\top \Phi(x,w).
\end{align*}
Note the weight vector $w$ is fixed in our setting, we write $K(x) = K(x,w)$ for simplicity.
\end{definition}

\subsection{Useful Lemmas}
\label{sec:semi_lipschitz_implies semi_smooth:app}
We prove two useful lemmas regarding Lipschitz gradient and smoothness, and extend this result to semi-Lipschitz gradient and semi-smoothness.
\begin{lemma}[folklore]\label{lem:folklore:app}
Suppose $L:\R^m\rightarrow \R$ has $\beta$-Lipschitz gradient, then $L$ is $b$-smooth, where $b = \beta/2$.
\end{lemma}

\begin{proof}
Suppose $L(x)$ has $\beta$-Lipschitz gradient. This means that for any $x,y\in \R^m$, we have $\|\nabla L(x)-\nabla L(y)\|\leq \beta \|x-y\|$.

By Cauchy-Schwartz, 
\begin{align*}
    \langle \nabla L(x) - \nabla L(y), x-y \rangle \le \beta\|x-y\|^2.
\end{align*} 

Hence function $G(x) = \frac{\beta}{2}\|x\|^2-L(x)$ is convex. So 
\begin{align*}
G(y) \ge G(x) + \langle \nabla G(x), y-x\rangle,
\end{align*}
which implies 
\begin{align*}
L(y)\le L(x) + \langle \nabla L(x), y-x\rangle + \frac{\beta}{2}\|y-x\|^2.
\end{align*}
Thus $L(x)$ is also $b$-smooth where $b = \frac{\beta}{2}$. 
\end{proof}

\begin{lemma}
\label{lem:semi_lip_to_semi_smooth}
Suppose $L$ satisfies $(\alpha,\beta,p)$-semi-Lipschitz gradient (Def.~\ref{def:semi_lipschitz:app}), then $L$ is also $(\alpha,\frac{\beta}{2},p/2)$-semi-smooth (Def.~\ref{def:semi_smooth:app}).
\end{lemma}

\begin{proof}
First we can bound the inner product term
\begin{align}\label{eqn:bounded_inner_prod_gradf_x}
    & ~ \langle \nabla L(x) - \nabla L(y), x-y \rangle \notag\\
    \le & ~ \|\nabla L(x) - \nabla L(y)\| \cdot \|x-y\| \notag\\
    \le & ~ \sqrt{\beta^2  \| x - y \|^2 + \alpha^2 \| x  - y \|^{2-2p} \cdot L(x)^{p}} \cdot \|x-y\| \notag\\
    \le & ~ \big(\beta \|x-y\| + \alpha \|x-y\|^{(1-p)}L(x)^{p/2}\big) \cdot \|x-y\| \notag\\
    = & ~ \beta \|x-y\|^2 + \alpha\|x-y\|^{2-p}L(x)^{p/2}.
\end{align}
The first step is Cauchy-Schwartz, the second step is the definition of $(\alpha,\beta,p)$-semi-Lipschitz, and the third step is the fact $\sqrt{a^2+b^2}\le a+b$ for non-negative $a$ and $b$.

Let $G(x) = \frac{\beta}{2}\|x\|^2- L(x)$. We could verify that
\begin{align}\label{eqn:bounded_inner_prod_gradg_x}
    & ~ \langle \nabla G(y) - \nabla G(x), y-x \rangle \notag\\
    = & ~ \langle \beta y - \nabla L(y) - \beta x + \nabla L(x), y -x \rangle \notag\\
    = & ~  \beta \|y-x\|^2 - \langle \nabla L(y) - \nabla L(x), y - x \rangle \notag\\
    \ge & ~ - \alpha\|x-y\|^{2-p}L(x)^{p/2}.
\end{align}
The first step is derived from the definition of gradient and the third step is by plugging in Eq.~\eqref{eqn:bounded_inner_prod_gradf_x}.

Let $\phi(t) = G(x+t(y-x))$. Notice that $G(y)-G(x)=\phi(1) - \phi(0) = \int_0^1 \frac{\d \phi}{\d t} \d t$, hence we have
\begin{align*}
    &~ G(y) - G(x) \\
    = & ~ \int_0^1 \langle \nabla G(x+t(y-x)), y-x \rangle \d t \notag\\
    = & ~ \int_0^1 \langle \nabla G(x+t(y-x))-\nabla G(x),y-x\rangle\d t\\
      & ~ +\int_0^1 \langle \nabla G(x),y-x\rangle \d t \notag \\
    \ge & ~ \int_0^1 \big(\langle \nabla G(x), y-x\rangle - t^{1-p}\cdot\alpha\|x-y\|^{2-p}L(x)^{p/2}\big) \d t \notag\\
    \ge & ~ \int_0^1 \big(\langle \nabla G(x), y-x\rangle - \alpha\|x-y\|^{2-p}L(x)^{p/2}\big) \d t \notag \\
    = & ~ \langle \nabla G(x), y-x\rangle- \alpha\|x-y\|^{2-p}L(x)^{p/2}.
\end{align*}
The third step follows from Eq.~\eqref{eqn:bounded_inner_prod_gradg_x} and the fourth step follows from $p\in (0,1)$.

Hence,
\begin{align}\label{eqn:bounded_diff_g}
    G(y) \ge &~ G(x) + \langle \nabla G(x), y-x\rangle \notag \\ 
    &~ - \alpha\|x-y\|^{2-p}L(x)^{p/2}.
\end{align}

Then plug in $G(x) = \frac{\beta}{2}\| x \|^2 - L(x)$, Eq.~\eqref{eqn:bounded_diff_g} implies
\begin{align*}
    \frac{\beta}{2} \|y\|^2 - L(y) \ge & ~ \frac{\beta}{2} \|x\|^2 - L(x) + \langle \nabla G(x), y-x\rangle \\
    &~ - \alpha\|x-y\|^{2-p}L(x)^{p/2}
\end{align*}
which is equivalent to
\begin{align*}
    L(y) \le & ~ L(x) + \langle \nabla L(x), y -x\rangle \\
    & ~+  \frac{\beta}{2}\big(\|y\|^2 - 2 \langle x, y\rangle + \|x\|^2\big) \\
    &~ + \alpha\|x-y\|^{2-p}L(x)^{p/2} \\
    = & ~ L(x) + \langle \nabla L(x), y-x \rangle + \frac{\beta}{2}\|y-x\|^2 \\
    &~ + \alpha\|x-y\|^{2-p}L(x)^{p/2}.
\end{align*}
\end{proof}

\begin{table*}[!ht]
\small
    \centering
        \caption{Summary of functions with different properties. We use semi-s. to denote semi-smooth, we use semi-Lg. to denote semi-Lipschitz gradient. We use semi-sc. to denote semi-strongly convex. We use N/A because such function is impossible
due to we've proved semi-Lipschitz gradient implies the semi-smoothness in Lemma~\ref{lem:semi_lip_to_semi_smooth}. $\dagger$: assume $w^\top x \ge 0$. A term $L(x)^p$ with $p>0$ and a positive coefficient requires $L\ge 0$, so a function that takes negative values can satisfy such a property only with that coefficient equal to $0$. Non-critical point means $(\theta_1,\theta_2)$-non-critical point with some $\theta_1>0$, and semi-sc. means $(c,d,p)$-semi-strongly convex with some $d>0$ and $p\in(0,1]$ (with $p=0$ every function with Lipschitz gradient would qualify). Entries are for $m\ge 2$.
    }
    \begin{tabular}{|l|p{2cm}|c|c|c|c|} \hline
        Function & assumption & semi-s. & semi-Lg. & semi-sc. & non-critical point \\ \hline
        $\|x\|_2^2$ & & $\checkmark$ & $\checkmark$ & $\checkmark$ & $\checkmark$\\ \hline \hline
         N/A & & $\times$ & \checkmark & \checkmark & \checkmark \\ \hline
        $x^\top Ax-\lambda_{\min}(A)$ & & \checkmark & \checkmark & \checkmark & $\times$ \\ \hline \hline
        $\sqrt{w^\top x+b}^\dagger$ &  $w^\top x\geq 0$& \checkmark &\checkmark & $\times$ & $\times$\\ \hline
        $\ln (1+e^x)$ & $x\in [-1,1]^\dagger$  & \checkmark & \checkmark & \checkmark & \checkmark \\ \hline
        $\mathrm{sigmoid}(w^\top x+b)$ & & \checkmark & \checkmark & $\times$ & $\times$ \\ \hline
        $\tanh(w^\top x)$ & & \checkmark & \checkmark & $\times$ & $\times$ \\ \hline
          $\ln(w^\top x)$ & $w^\top x>0$& \checkmark  &  $\times$ &  $\times$ & $\times$ \\ \hline 
          N/A & & $\times$ &  \checkmark  & \checkmark &  $\times$ \\ \hline 
         N/A & & $\times$ &  \checkmark &  $\times$ & \checkmark \\ \hline 
         $\cosh(w^\top x)$ &$\|w\|,\|x\|=1$ & $\times$  &  $\times$ &  \checkmark & $\times$ \\ \hline \hline
        N/A & & $\times$  &  \checkmark &  $\times$ & $\times$ \\ \hline 
        $\cosh(w^\top x)$ & & $\times$  &  $\times$ &  $\times$ & $\times$ \\ \hline \hline
        $\text{ReLU}(w^\top x)$ & & $\times$  &  $\times$ &  $\times$ & $\times$  \\ \hline 
        $1/\|x\|_2$ & & $\times$ & $\times$ & $\times$ & $\times$ \\ \hline
          $(w^\top x)^2\cdot \sin (1/(w^\top x))$ & $\|w\|,\|x\|=1,w^\top x\neq 0$& $\times$ & $\times$ & $\times$ & $\times$ \\ \hline
          $(w^\top x)^2\cdot \sin (1/(w^\top x))$ & & $\times$ & $\times$ & $\times$ & $\times$ \\ \hline
    \end{tabular}
    \label{tab:table1}
\end{table*}

\section{From \texorpdfstring{$F$}{} to \texorpdfstring{$L$}{}}\label{sec:from_F_to_L:app}
In this section, we show that if we impose mild conditions on $F$, it will imply certain key conditions on $L$, which is critical in proving the convergence of our loss function.

We start with a list of assumptions of $F$.
\begin{assumption}\label{ass:min_L_equal_to_zero:app}
Let $x^*$ denote the global minimum of $L(x)$ in $\R^m$. We without loss of generality assume $L(x^*) = 0$.
\end{assumption}
We give a brief justification of this assumption. Notice that
\begin{align*}
    L(x) = & ~ \|\nabla_w F(x;w) - g\|^2 \\
    = & ~\|\nabla_w F(x;w) - \nabla_w F(\tilde{x};w)\|^2 .
\end{align*}
Hence $L(x) \ge 0$ and $L(\tilde{x}) = 0$. So it is reasonable to assume $\min_{x\in\R^m} L(x) = 0$. Even if $L(x)$ has other forms and $\min_{x\in\R^m} L(x) \neq 0$, we can define a dummy objective function $L'(x)$ as
\begin{align*}
    L'(x) = L(x) - C,
\end{align*}
where $C = \min_x L(x)$. Suppose we apply gradient descent with initialization $x_0$ on $L(x)$ and apply gradient descent with initialization $y_0$ on $L'(y)$. Based on the fact that $\nabla L(x) = \nabla L'(x)$, we could show if $x_t = y_t$ then
\begin{align*}
    y_{t+1} =  y_t - \eta \cdot \nabla L'(y_t) 
    =  x_t - \eta \cdot \nabla L(x_t) 
    =  x_{t+1}.
\end{align*}
Hence by induction, for all $t$, $x_t = y_t$ when $x_0 = y_0$. Thus the convergence rate for $L(x)$ and $L'(x)$ are exactly the same as long as the initialization is the same. Since we choose $C = \min_{x\in \R^m} L(x)$, we can easily verify that
\begin{align*}
    \min_{x\in\R^m} L'(x) = \min_{x\in\R^m}L(x) - C = 0.
\end{align*}
Therefore, without loss of generality, we make Assumption~\ref{ass:min_L_equal_to_zero:app}.

The next assumption is a standard Lipschitz gradient assumption.
\begin{assumption}\label{ass:gradient_lipschitz:app}

   $\nabla_w F(x,w)$ is $\beta$-Lipschitz with respect to $x$, i.e., for any $x_1,x_2 \in \R^m$ we have
   \begin{align*}
       \| \nabla_w F(x_1;w) - \nabla_w F(x_2;w)  \| \leq \beta \cdot \| x_1 - x_2 \|.
   \end{align*}
\end{assumption}

The next assumption asks the pseudo-Hessian to be Lipschitz.
\begin{assumption}\label{ass:Phi_lipschitz:app}
Let $\Phi(x,w)$ be defined as Definition~\ref{def:Phi_K:app}. There is $M\ge 0$ such that for any $x_1,x_2\in\R^m$ we have
\begin{align*}
    \|\Phi(x_1,w)-\Phi(x_2,w)\|\le M\cdot\|x_1-x_2\|,
\end{align*}
where $\|\cdot\|$ is the spectral norm.
\end{assumption}

The next assumption is necessary to ensure $L$ has non-critical point property.
\begin{assumption}\label{ass:K_eigenvalues:app}
Let $\theta_2 \geq \theta_1 > 0$.
$\forall x\in \R^m$, let $K(x)\in\R^{d\times d}$ be defined as Definition~\ref{def:Phi_K:app}.
$K(x)$'s eigenvalues can be bounded by
\begin{align*}
    \theta_1^2 \le \lambda^2_1(x) \le \cdots \le \lambda^2_{d}(x) \le \theta_2^2.
\end{align*}
\end{assumption}
Since $K(x)=\Phi(x,w)^\top\Phi(x,w)$ has rank at most $\min(d,m)$, Assumption~\ref{ass:K_eigenvalues:app} with $\theta_1>0$ requires $d\le m$. Since $\Phi(x,w)^\top$ is the Jacobian of $x\mapsto\nabla_w F(x;w)$, Assumption~\ref{ass:gradient_lipschitz:app} gives $\|\Phi(x,w)\|\le\beta$. Hence $\theta_1\le\lambda_1(x)\le\lambda_d(x)=\|\Phi(x,w)\|\le\beta$, and the upper bound of Assumption~\ref{ass:K_eigenvalues:app} always holds with $\theta_2=\beta$, so we may take $\theta_1\le\theta_2\le\beta$.

\subsection{What \texorpdfstring{$F$}{} Implies Semi-smoothness}\label{lem:what_F_implies_smooth:app}
\begin{lemma}\label{lem:what_F_implies_L_semi_smoothness:app}

Let $\Phi(x,w)$ be defined as Def.~\ref{def:Phi_K:app}. Suppose that
\begin{itemize}
    \item (Assumption~\ref{ass:gradient_lipschitz:app}) $\nabla_w F(x;w)$ is $\beta$-Lipschitz with respect to $x$, $\forall x \in \R^m;$
    \item (Assumption~\ref{ass:Phi_lipschitz:app}) $\Phi(x,w)$ is $M$-Lipschitz with respect to $x$ in spectral norm.
\end{itemize}
Then $L(x) = \|\nabla_w F(x;w) - g\|^2$ satisfies, for all $x,y\in\R^m$,
\begin{align}\label{eq:L_semi_smooth_bound:app}
    L(y) \le & ~ L(x) + \langle \nabla L(x), y-x\rangle \notag \\
    &~ + \big(\beta^2 + M L(x)^{1/2}\big)\|y-x\|^2.
\end{align}
Consequently, for every $x_0\in\R^m$, $L$ is $(0,\beta^2+M L(x_0)^{1/2},1/2)$-semi-smooth (Def.~\ref{def:semi_smooth:app}) on $S_0:=\{x\in\R^m: L(x)\le L(x_0)\}$.
\end{lemma}

\begin{proof}
Let $G(x):=\nabla_w F(x;w)$ and $r(x):=G(x)-g$, so $L(x)=\|r(x)\|^2$. By Assumption~\ref{ass:Phi_lipschitz:app}, $\Phi$ is continuous in $x$, so $G$ is continuously differentiable with Jacobian $\Phi(x,w)^\top$ (Def.~\ref{def:Phi_K:app}), and $\nabla L(x)=2\Phi(x,w)r(x)$. Fix $x,y\in\R^m$ and let $\Delta:=G(y)-G(x)$.

{\bf An exact identity.} Since $r(y)=r(x)+\Delta$,
\begin{align*}
    L(y) = \|r(x)+\Delta\|^2 = L(x)+2\langle\Delta,r(x)\rangle+\|\Delta\|^2 .
\end{align*}
Also,
\begin{align*}
\langle\nabla L(x),y-x\rangle = & ~ 2\langle\Phi(x,w)r(x),y-x\rangle \\
= & ~ 2\langle\Phi(x,w)^\top(y-x),r(x)\rangle .
\end{align*}
Subtracting the second display from the first gives
\begin{align}\label{eq:L_exact_identity:app}
    & ~ L(y)-L(x)-\langle\nabla L(x),y-x\rangle \notag \\
    = & ~ \|\Delta\|^2+2\langle\Delta-\Phi(x,w)^\top(y-x),r(x)\rangle .
\end{align}

{\bf The first term.} By Assumption~\ref{ass:gradient_lipschitz:app}, $\|\Delta\|\le\beta\|y-x\|$, so $\|\Delta\|^2\le\beta^2\|y-x\|^2$.

{\bf The second term.} The fundamental theorem of calculus, applied to $s\mapsto G(x+s(y-x))$ on $[0,1]$, gives
\begin{align*}
    \Delta = \int_0^1\Phi(x+s(y-x),w)^\top(y-x)\,\d s ,
\end{align*}
hence
\begin{align*}
& ~ \Delta-\Phi(x,w)^\top(y-x) \\
= & ~ \int_0^1\big(\Phi(x+s(y-x),w)-\Phi(x,w)\big)^\top(y-x)\,\d s .
\end{align*}
A matrix and its transpose have the same spectral norm, so Assumption~\ref{ass:Phi_lipschitz:app} gives
\begin{align}\label{eq:Phi_taylor_remainder:app}
    & ~ \|\Delta-\Phi(x,w)^\top(y-x)\| \notag \\
    \le & ~ \int_0^1\|\Phi(x+s(y-x),w)-\Phi(x,w)\|\cdot\|y-x\|\,\d s \notag \\
    \le & ~ \int_0^1 M s\|y-x\|^2\,\d s = \frac{M}{2}\|y-x\|^2 .
\end{align}
By Cauchy-Schwarz and $\|r(x)\|=L(x)^{1/2}$,
\begin{align*}
& ~ 2\langle\Delta-\Phi(x,w)^\top(y-x),r(x)\rangle \\
\le & ~ 2\cdot\frac{M}{2}\|y-x\|^2\cdot L(x)^{1/2} = M L(x)^{1/2}\|y-x\|^2 .
\end{align*}

Putting the two bounds into Eq.~\eqref{eq:L_exact_identity:app} proves Eq.~\eqref{eq:L_semi_smooth_bound:app}. Finally, if $x\in S_0$ then $L(x)^{1/2}\le L(x_0)^{1/2}$, so Eq.~\eqref{eq:L_semi_smooth_bound:app} gives
\begin{align*}
    L(y)\le & ~ L(x)+\langle\nabla L(x),y-x\rangle \\
    & ~ +\big(\beta^2+M L(x_0)^{1/2}\big)\|y-x\|^2
\end{align*}
for every $x\in S_0$ and $y\in\R^m$. This is the inequality of Def.~\ref{def:semi_smooth:app} on $S_0$ with $a=0$, $b=\beta^2+M L(x_0)^{1/2}$ and $p=1/2$. Since $a=0$, the value of $p$ plays no role.
\end{proof}

\subsection{What \texorpdfstring{$F$}{} Implies Non-critical Point}\label{what_F_implies_non_critical_point:app}

\begin{lemma}\label{lem:what_F_implies_L_no_critical_point:app}
Let $K$ be defined as in Def.~\ref{def:Phi_K:app}. Denote the eigenvalues of $K(x)\in\R^{d\times d}$ by $\lambda^2_1(x) \le \lambda^2_2(x) \le \cdots \le \lambda^2_d(x)$. If Assumption~\ref{ass:K_eigenvalues:app} holds, i.e., for all $x \in \R^m$
\begin{itemize}
    \item $\theta_1^2 \leq \lambda^2_1(x),$
    \item $\theta_2^2 \geq \lambda^2_{d}(x),$
\end{itemize}
then $L$ satisfies $(2\theta_1, 2\theta_2)$-non-critical point condition, i.e.
\begin{align*}
    4\theta_1^2 \cdot L(x) \le \|\nabla L(x) \|^2 \le 4\theta_2^2 \cdot L(x).
\end{align*}

\end{lemma}
\begin{proof}
    Let $r(x):=\nabla_w F(x;w)-g$, so $L(x)=\|r(x)\|^2$. By Def.~\ref{def:Phi_K:app}, $\nabla_x L(x)=2\Phi(x,w)r(x)$. Hence
    \begin{align*}
        \|\nabla_x L(x) \|^2 = &~ 4\|\Phi(x,w)r(x)\|^2 \\
        = &~ 4\, r(x)^\top K(x)\, r(x).
    \end{align*}
    Given a symmetric matrix $A$ and a nonzero vector $y$, it is well-known that
    \begin{align*}
        \lambda_{\min}(A) \leq \frac{y^\top Ay}{\|y\|^2} \leq \lambda_{\max}(A).
    \end{align*}
    Applying this to $A=K(x)$, whose eigenvalues lie in $[\theta_1^2,\theta_2^2]$, gives
    \begin{align*}
        &~ \|\nabla_x L(x)\|^2 \ge 4\theta_1^2\cdot \|r(x)\|^2, \\
        &~ \|\nabla_x L(x)\|^2 \le 4\theta_2^2\cdot \|r(x)\|^2,
    \end{align*}
    where both bounds are trivial if $r(x)=0$. This is equivalent to
    \begin{align*}
        4\theta_1^2 \cdot L(x) \le \|\nabla_x L(x)\|^2 \le 4\theta_2^2\cdot L(x).
    \end{align*}
\end{proof}

\section{Converge to Optimal Solution}\label{sec:solution}
One of the important conditions we need to impose on $L$ if we want to converge to the optimal solution is $L$ has a unique minimum. In order to achieve this property, we introduce the notion of semi-strongly convex below. For the attack losses $L$ and $L_R$ of Section~\ref{sec:attack_sketched_gradients}, if $F$ satisfies the conditions of Theorem~\ref{thm:F_L_smooth_criticl}, then Theorem~\ref{thm:solution_1} can apply only with no compression ($b_{\rm sketch}=d$) and when the dimension $d$ of $w$ equals $m$. Indeed, every zero of a nonnegative differentiable $L$ is a critical point, so by Theorem~\ref{app:thm:unique_local_minima} the hypotheses of Theorem~\ref{thm:solution_1} allow at most one zero of $L$. Under the conditions of Theorem~\ref{thm:F_L_smooth_criticl}, however, the zeros of $L$ form a submanifold of dimension $m-d$, and the zeros of $L_R$ form a submanifold of dimension $m-b_{\rm sketch}\geq 1$ when $b_{\rm sketch}<d$ (Remark~\ref{rem:sketch_non_identifiable}). A submanifold of positive dimension has more than one point, so only the case $b_{\rm sketch}=d=m$ remains, that is, $d=m$ with no compression (no sketching, or an invertible $R$).
\begin{definition}[semi-strongly convex]\label{def:semi_strong_convex}
For any $p\in [0,1]$, we say function $L:\R^m\rightarrow \R$ is $(c,d,p)$-semi-strongly-convex if for any $x,y\in \R^m$, we have
\begin{align*}
    L(x) \geq &~ L(y)+\langle \nabla L(y),x-y\rangle+d\|x-y\|^2\\
    &~ -c\|x-y\|^{2-2p}\cdot L(y)^p.
\end{align*}
\end{definition}

\subsection{Conditions for Unique Minimum}\label{sec:unique_local_minima}
\begin{theorem}[Uniqueness of Critical Points]\label{app:thm:unique_local_minima}
	If $L(x)$ satisfies $(\theta_1 , \theta_2)$-non-critical point condition ($\theta_1>0$), and $(c,d,p)$-semi-strongly convex ($d>0$, $p\in(0,1]$), then $L$ has at most one point $x^*$ with $\nabla L(x^*) = 0$.
\end{theorem}

\begin{proof}
	Suppose $x_1^* \in \mathbb{R}^m$ and $x_2^* \in \mathbb{R}^m$ are two points such that
	\begin{align*}
	\nabla L(x_1^*) = \nabla L(x_2^*) = 0.
	\end{align*}
	Since $L(x)$ satisfies $(\theta_1, \theta_2)$-non-critical point condition,
	\begin{align*}
	\theta_1^2 \cdot L(x_1^*) \le \|\nabla L(x_1^*)\|^2 \le \theta_2^2 \cdot L(x_1^*).
	\end{align*}
	Therefore $L(x_1^*) = 0$ holds. Similarly $L(x_2^*) = 0$ also holds.
	By $(c,d,p)$-semi-strongly convexity of $L(x)$,
	\begin{align}\label{eqn:semi_strongly_convex_implies_unique_local_minima}
	L(x_1^*) \ge & ~ L(x_2^*) + \langle \nabla L(x_2^*), x_1^*-x_2^* \rangle \notag \\
	& ~ + d\|x_2^* - x_1^*\|^2 \notag \\
	& ~ - c\|x_2^*-x_1^*\|^{2-2p}\cdot L(x_2^*)^p .
	\end{align}
	Combining with $L(x_1^*) = L(x_2^*) = 0$, $\nabla L(x_2^*) = 0$ and $L(x_2^*)^p=0$ (which uses $p>0$), Eq.~\eqref{eqn:semi_strongly_convex_implies_unique_local_minima} implies
	\begin{align*}
	0 \ge d \|x_2^* - x_1^*\|^2.
	\end{align*}
	Hence $\|x_2^*-x_1^*\|^2 = 0$ and $x_1^* = x_2^*$.
\end{proof}

\subsection{Conditions for Convergence of \texorpdfstring{$x_t$}{}}\label{sec:solution_1}

\begin{theorem}\label{thm:solution_1}
Suppose we run gradient descent algorithm to update $x_{t+1}$ in each iteration as follows:
\begin{align*}
    x_{t+1} = x_t - \eta \cdot \nabla L(x) |_{x = x_t}
\end{align*}
Let $p\in(0,1)$ and assume that $\nabla L(x^*) = 0$. If function $L$ is
\begin{itemize}
    \item $(c,d,p)$-semi-strongly convex (Def.~\ref{def:semi_strong_convex})
    \item $(\alpha,\beta,p)$-semi-Lipschitz gradient (Def.~\ref{def:semi_lipschitz:app})
    \item $(\theta_1 , \theta_2)$-non-critical point (Def.~\ref{def:no_critical_point:app})
    \item $d > \frac{c^{1/(2p)}}{\theta_1(\theta_1-\alpha^{1/p})}\left(\beta^2 + \left( {\alpha}/{\theta_1^p}\right)^{1/(1-p)}\right)  +c^{1/(2-2p)}$
    \item $\theta_1 > \alpha^{1/p}$
\end{itemize}
by choosing
\begin{align*}
0<\eta \le \xi/(2\zeta)
\end{align*}
where
    \begin{align*}
     \zeta := \frac{\theta_1}{\theta_1-\alpha^{1/p}} \cdot \left(\beta^2 + ( {\alpha}/{\theta_1^p} )^{1/(1-p)}\right)
\end{align*}
and
\begin{align*}
        \xi := 2( d - c^{1/2p} \theta_1^{-2} \zeta - c^{1/(2-2p)} ).
\end{align*}
we have
\begin{align*}
    \| x_{t+1} - x^* \|^2 \leq (1-\gamma) \cdot \| x_t - x^* \|^2,
\end{align*}
where $\gamma = \xi\eta/2$. In particular, $\| x_{t+1} - x^* \| \le \sqrt{1-\gamma}\cdot\| x_t - x^* \| \le (1-\gamma/2)\cdot\| x_t - x^* \|$. The fourth condition on $L$ is equivalent to $\xi>0$.
\end{theorem}

\begin{proof}
We have
{
\begin{align}\label{eq:bound_x_close_eq1:app}
& ~ \| x_{t+1} - x^* \|^2 \notag\\
= &~ \| x_{t+1} - x_t + x_t - x^*\|^2 \notag \\
= & ~ \underbrace{\| x_{t+1} - x_t \|^2}_{\mathcal{A}_1} + 2 \underbrace{\langle x_{t+1} -x_t, x_t -x^* \rangle}_{ \mathcal{A}_2 }  + \| x_t - x^* \|^2.
\end{align}
}

For the first term in Eq. \eqref{eq:bound_x_close_eq1:app}, we have
\begin{align*}
\mathcal{A}_1 = \eta^2 \| \nabla L(x_t) \|^2.
\end{align*}

Consider $x_t, x^*$,  using $(\alpha,\beta,p)$-semi-Lipschitz gradient and $\nabla L(x^*) = 0$, we have
\begin{align}\label{eq:tmp_1:app}
    & ~ \| \nabla L(x_t) \|^2 \notag\\
    \leq & ~ \beta^2 \| x_t - x^* \|^2 + \alpha^2 \| x_t - x^* \|^{2-2p} \cdot L(x_t)^{p} \notag \\
    \leq & ~ \beta^2 \| x_t - x^* \|^2 + \alpha^2 \| x_t - x^* \|^{2-2p} \cdot \| \nabla L(x_t) \|^{2p} / (\theta_1^{2p}), 
\end{align}

where the second step follows from non-critical point (Definition~\ref{def:no_critical_point:app}), which gives $0\le L(x_t)\le\|\nabla L(x_t)\|^2/\theta_1^2$. Here $L\ge 0$ everywhere, because the upper bound of the non-critical point condition gives $\theta_2^2 L(x)\ge\|\nabla L(x)\|^2\ge 0$.
For the last term of the above equation, we have
\begin{align}\label{eq:tmp_2:app}
 & ~ \alpha^2 \| x_t - x^* \|^{2-2p} \cdot \| \nabla L(x_t) \|^{2p} / (\theta_1^{2p}) \notag \\
  \leq & ~  ({\alpha}/{\theta_1^p} )^{2/(2-2p)}\cdot\|x_t-x^*\|^2  +  ( {\alpha} / {\theta_1^p} )^{1/p}\|\nabla L(x_t)\|^2,
\end{align}
where the  step follows from $a^{2-2p}b^{2p} \leq (1-p)a^2 + p b^2 \le a^2 + b^2$ for $a,b\ge 0$ (weighted AM-GM), applied with $a=(\alpha/\theta_1^p)^{1/(2-2p)}\|x_t-x^*\|$ and $b=(\alpha/\theta_1^p)^{1/(2p)}\|\nabla L(x_t)\|$, so that $a^{2-2p}b^{2p}=(\alpha/\theta_1^p)^2\|x_t-x^*\|^{2-2p}\|\nabla L(x_t)\|^{2p}$.

Thus, Eq.~\eqref{eq:tmp_1:app} and \eqref{eq:tmp_2:app} imply
\begin{align*}
    \| \nabla L(x_t) \|^2 \leq & ~ \left(\beta^2 + (\alpha/\theta_1^p)^{1/(1-p)}\right)\|x_t-x^*\|^2 \\
    & ~ + (\alpha/\theta_1^p)^{1/p}\|\nabla L(x_t)\|^2 ,
\end{align*}
where we used $2/(2-2p)=1/(1-p)$. Since $(\alpha/\theta_1^p)^{1/p}=\alpha^{1/p}/\theta_1$ and $\theta_1>\alpha^{1/p}$, the coefficient $1-\alpha^{1/p}/\theta_1=(\theta_1-\alpha^{1/p})/\theta_1$ is positive, and rearranging gives
\begin{align*}
    \| \nabla L(x_t) \|^2 \leq & ~\frac{\theta_1}{\theta_1-\alpha^{1/p}} \cdot \left(\beta^2 +  ({\alpha} / {\theta_1^p} )^{1/(1-p)}\right)\\
    & ~\cdot \|x_t-x^*\|^2.
\end{align*}

For the second term in Eq. \eqref{eq:bound_x_close_eq1:app}, $x_{t+1}-x_t=-\eta\nabla L(x_t)$ gives
\begin{align*}
2\mathcal{A}_2
= & ~ -2 \eta \langle \nabla L(x_t), x_t - x^* \rangle \\
= & ~ 2 \eta \langle \nabla L(x_t), x^* - x_t \rangle \\
 \leq & ~
    2 \eta \big( \underbrace{ L(x^*) - L(x_t) }_{ \leq 0 } - d\|x_t - x^*\|^2 \\
    & ~+ c\|x_t - x^*\|^{2-2p}\cdot L(x_t)^p\big) \\
 \leq & ~ 
    2\eta(-d\|x_t - x^*\|^2 + c\|x_t - x^*\|^{2-2p}\cdot L(x_t)^p) \\
  \leq & ~ (-2\eta d + 2\eta c^{1/(2-2p)}) \|x_t - x^*\|^2 + 2\eta c^{1/(2p)} L(x_t)  \\
  \leq & ~ (-2\eta d + 2\eta c^{1/(2-2p)}) \|x_t - x^*\|^2  \\
   & ~+ 2\eta \frac{c^{1/2p}}{\theta_1^{2}} \|\nabla L(x_t)\|^2,
\end{align*}
where the third step follows from  $(c,d,p)$-semi-strongly convex (Def.~\ref{def:semi_strong_convex} with $x=x^*$ and $y=x_t$), the fourth step follows from $L(x^*) - L(x_t) \leq 0$, the fifth step follows from $a^{2-2p}b^{2p} \le a^2 + b^2$ with $a=c^{1/(4-4p)}\|x_t-x^*\|$ and $b=c^{1/(4p)}L(x_t)^{1/2}$, so that $a^{2-2p}b^{2p}=c\|x_t-x^*\|^{2-2p}L(x_t)^p$, and the last step is $L(x_t) \le (1/\theta_1^2) \|\nabla L(x_t)\|^2$. For the fourth step, $L(x^*)=0$, because the non-critical point condition and $\nabla L(x^*)=0$ give $\theta_1^2 L(x^*)\le 0$, and $L(x_t)\ge 0$.

Putting it to the Eq.~\eqref{eq:bound_x_close_eq1:app}, we have
\begin{align*}
    &~ \| x_{t+1} - x^* \|^2  \\
    = & ~ \mathcal{A}_1 + 2\mathcal{A}_2 + \|x_t - x^*\|^2 \\
    \leq & ~ \left(\eta^2 + 2\eta \frac{c^{1/2p}}{\theta_1^2}\right)\|\nabla L(x_t)\|^2 \\
    &~ + (1-2\eta d + 2\eta c^{1/(2-2p)})\|x_t - x^*\|^2 \\
    \le & ~ \|x_t - x^*\|^2  \cdot \left[ \eta^2 \cdot \frac{\theta_1}{\theta_1-\alpha^{1/p}}\left(\beta^2 + \left(\frac{\alpha}{\theta_1^p}\right)^{1/(1-p)}\right)\right.\\
    & ~  \left.- 2\eta\left(d - \frac{c^{1/2p}}{\theta_1(\theta_1-\alpha^{1/p})}\left(\beta^2 + \left(\frac{\alpha}{\theta_1^p}\right)^{1/(1-p)}\right)\right.\right.\\
    & ~ -\left.\left.c^{1/(2-2p)}\right)  + 1 \right].
\end{align*}
Let 
\begin{align*}
     \zeta = \frac{\theta_1}{\theta_1-\alpha^{1/p}}\left(\beta^2 + ( {\alpha}/{\theta_1^p} )^{1/(1-p)}\right) 
\end{align*}
and
\begin{align*}
         \xi = 2(d - c^{1/2p} \theta_1^{-2} \zeta - c^{1/(2-2p)} ). 
\end{align*}
Then we have
\begin{align*}
    & ~ \| x_{t+1} - x^* \|^2 \\
    \leq & ~ (\zeta \eta^2 - \xi \eta + 1) \|x_t - x^*\|^2 \\
    \leq & ~ (-\xi\eta/2 + 1) \|x_t-x^*\|^2 \\
    \le & ~ (1 - \gamma) \|x_t - x^*\|^2,
\end{align*}
where $\gamma = \xi\eta/2$. The second step holds because we choose $\eta \le \xi/(2\zeta)$ and hence
\begin{align*}
    \zeta \eta^2 - \xi\eta ~\le~ (\xi/2)\eta - \xi\eta ~=~ -\xi\eta/2.
\end{align*}
Finally, if $x_t\neq x^*$, the left side is nonnegative, so $0\le 1-\gamma\le 1$ (if $x_t=x^*$, then $\nabla L(x_t)=0$ and $x_{t+1}=x^*$). Taking square roots gives $\|x_{t+1}-x^*\|\le\sqrt{1-\gamma}\,\|x_t-x^*\|$, and $\sqrt{1-\gamma}\le 1-\gamma/2$ because $(1-\gamma/2)^2=1-\gamma+\gamma^2/4\ge 1-\gamma$ and $1-\gamma/2\ge 0$.
This concludes our proof.
\end{proof} 
\section{Converge to Optimal Cost}\label{sec:cost:app}
In this section, we provide the formal proof that if $L$ is semi-smooth and non-critical point, then the loss converges linearly.
 
\subsection{Conditions for Convergence of \texorpdfstring{$L(x_t)$}{}}\label{sec:cost_1}
\begin{theorem}\label{thm:cost_1}

Let $L:\R^m\rightarrow\R$ satisfy $L(x)\ge 0$ for all $x$, and let $x^*$ be a point with $L(x^*)=0$ (Assumption~\ref{ass:min_L_equal_to_zero:app}). Suppose we run gradient descent algorithm from $x_0\in\R^m$ to update $x_{t+1}$ in each iteration as follows:
\begin{align*}
    x_{t+1} = x_t - \eta \cdot \nabla L(x) |_{x = x_t}
\end{align*}
Let $S_0:=\{x\in\R^m: L(x)\le L(x_0)\}$. If we assume
\begin{itemize}
    \item $L$ is $(a,b,1/2)$-semi-smooth on $S_0$ (Def.~\ref{def:semi_smooth:app}) with $a\ge 0$ and $b>0$,
    \item $L$ is $(\theta_1,\theta_2)$-non-critical point (Def.~\ref{def:no_critical_point:app}),
    \item $\theta_1^2 > 2a\theta_2$,
\end{itemize}
using the choice
\begin{align*}
0<\eta \le 1/(2b),
\end{align*}
then every $x_t$ lies in $S_0$ and
\begin{align*}
    L(x_{t+1}) - L(x^*) \leq (1-\gamma ) (L(x_t) - L(x^*)),
\end{align*}
where $\gamma = \eta(\theta_1^2-2a\theta_2)/2$.
\end{theorem}

\begin{proof}
We prove by induction on $t$ that $x_t\in S_0$, and prove the bound for step $t$ along the way. Clearly $x_0\in S_0$. Suppose $x_t\in S_0$. Since $x_{t+1}-x_t=-\eta\nabla L(x_t)$, we have $\|x_{t+1}-x_t\|=\eta\|\nabla L(x_t)\|$. We bound the consecutive gap between $L(x_{t+1})$ and $L(x_t)$:
\begin{align*}
    &~ L(x_{t+1})-L(x_t) \\
    \leq & ~ \langle \nabla L(x_t), x_{t+1} - x_t \rangle + b \| x_{t+1} - x_t \|^2 \\
    &~ + a \| x_{t+1} - x_t \| \cdot L(x_t)^{1/2} \\
    = & ~ -\eta \| \nabla L(x_t) \|^2 + b \eta^2 \| \nabla L(x_t) \|^2 \\
    &~ + a \eta \| \nabla L(x_t) \| \cdot L(x_t)^{1/2} \\
    \leq & ~ -\frac{\eta}{2} \| \nabla L(x_t) \|^2 + a \eta \| \nabla L(x_t) \| \cdot L(x_t)^{1/2} \\
    \leq & ~ -\frac{\eta}{2} \| \nabla L(x_t) \|^2 + a \eta \theta_2 L(x_t) \\
    \leq & ~ -\frac{\eta}{2} \theta_1^2 L(x_t) + a \eta \theta_2 L(x_t) \\
    = & ~ -\gamma L(x_t),
\end{align*}
where the first step follows from $(a,b,1/2)$-semi-smoothness on $S_0$ at the point $x_t\in S_0$, the third step uses $b\eta\le 1/2$, so that $-\eta+b\eta^2=-\eta(1-b\eta)\le-\eta/2$, the fourth step uses $a\ge 0$ and the upper bound $\|\nabla L(x_t)\|\le\theta_2 L(x_t)^{1/2}$ of the non-critical point condition, the fifth step uses its lower bound $\|\nabla L(x_t)\|^2\ge\theta_1^2 L(x_t)$, and the last step is the definition of $\gamma$.

This implies that
\begin{align*}
    L(x_{t+1}) \leq (1-\gamma) L(x_t).
\end{align*}
Since $L(x^*)=0$, this is the claimed bound
\begin{align*}
    L(x_{t+1})-L(x^*) \leq (1-\gamma)\cdot(L(x_t)-L(x^*)).
\end{align*}
Moreover, $\gamma>0$ because $\theta_1^2>2a\theta_2$, and $L(x_t)\ge 0$, so $L(x_{t+1})\le L(x_t)\le L(x_0)$. Hence $x_{t+1}\in S_0$, which completes the induction.
\end{proof}

\section{Attack Sketched Gradient}\label{sec:sketching:app}
In this section, we consider the setting where the gradient is sketched, i.e., we can only observe a sketched gradient ${\cal S}(g)$ where ${\cal S}:\R^d\rightarrow \R^{b_{\rm sketch}}$ is a sketching matrix. We can also observe the sketching matrix ${\cal S}$, hence, our strategy will be solving the new sketched objective $L_{\cal S}(x)=\|{\cal S}(\nabla_w F(x,w))-{\cal S}(g) \|^2$ and optimize over the sketched objective.

Let ${\cal S}\in \R^{b_{\rm sketch}\times d}$ be a sketching matrix, popular sketching matrices are random Gaussian, Count Sketch~\cite{ccf02}, subsampled randomized Hadamard transform~\cite{ldfu13}. We view ${\cal S}$ as the linear map $u\mapsto{\cal S}(u)={\cal S}u$, so its Jacobian is ${\cal S}$ and $\|{\cal S}\|$ denotes its spectral norm. We impose following assumptions on ${\cal S}$.

\begin{assumption}\label{ass:S_preserve_dist:app}
Let $\tau>0$, for any in $u,v\in \R^d$,
$$\|{\cal S}(u)-{\cal S}(v)\|\le\tau\|u-v\|.$$
\end{assumption}
Since ${\cal S}$ is linear, this assumption holds with $\tau=\|{\cal S}\|$. We keep the symbol $\tau$ only for notation.
\begin{assumption}\label{ass:S_eigenvalue:app}
The sketching matrix ${\cal S}\in \R^{b_{\rm sketch}\times d}$ satisfies
\begin{align*}
   0 < \gamma_1\leq \sigma_1({\cal S}^\top)\leq \ldots \leq \sigma_{b_{\rm sketch}}({\cal S}^\top)\leq \gamma_2.
\end{align*}
\end{assumption}
Assumption~\ref{ass:S_eigenvalue:app} means $\gamma_1\|u\|\le\|{\cal S}^\top u\|\le\gamma_2\|u\|$ for every $u\in\R^{b_{\rm sketch}}$, and in particular $\|{\cal S}\|=\|{\cal S}^\top\|\le\gamma_2$. Assumption~\ref{ass:S_preserve_dist:app} with $v=0$ gives $\|{\cal S}\|\le\tau$, so $\gamma_1\le\|{\cal S}\|\le\tau$.
For many sketching matrices, such as a random Gaussian matrix with $b_{\rm sketch}\leq d$, ${\cal S}^\top$ has full column rank almost surely, hence, $\gamma_1>0$ is a reasonable assumption.

\subsection{What Sketching Implies Semi-smoothness}\label{sec:what_S_implies_semi_smooth:app}
\begin{lemma}\label{lem:sketching_semi_smooth:app}
If the sketching matrix ${\cal S}$ satisfies $\|{\cal S}(u)-{\cal S}(v)\| \le \tau\|u-v\|$ for all $u,v\in\R^d$ and $\|{\cal S}\|\le \gamma_2$, and $F$ satisfies the conditions of Lemma~\ref{lem:what_F_implies_L_semi_smoothness:app},
then $L(x):=\|\mathcal{S}(\nabla_w F(x;w)) -  \mathcal{S}(g)\|^2$ satisfies, for all $x,y\in\R^m$,
\begin{align}\label{eq:LS_semi_smooth_bound:app}
    L(y)\le & ~ L(x)+\langle\nabla L(x),y-x\rangle \notag \\
    & ~ +\big(\tau^2\beta^2+\gamma_2 M L(x)^{1/2}\big)\|y-x\|^2 .
\end{align}
Consequently, for every $x_0\in\R^m$, $L$ is $(0,B,1/2)$-semi-smooth (Def.~\ref{def:semi_smooth:app}) on $S_0:=\{x\in\R^m: L(x)\le L(x_0)\}$, where $B=\tau^2\beta^2+\gamma_2 M L(x_0)^{1/2}$.
\end{lemma}

\begin{proof}
Let $G(x) := \nabla_w F(x;w)$ and $r(x):=G(x)-g$. Since ${\cal S}$ is linear, $L(x)=\|{\cal S}r(x)\|^2$. As in the proof of Lemma~\ref{lem:what_F_implies_L_semi_smoothness:app}, $G$ has Jacobian $\Phi(x,w)^\top$, so the chain rule gives $\nabla L(x)=2\Phi(x,w){\cal S}^\top{\cal S}r(x)$. Fix $x,y\in\R^m$ and let $\Delta:=G(y)-G(x)$.

{\bf An exact identity.} Since ${\cal S}r(y)={\cal S}r(x)+{\cal S}\Delta$,
\begin{align*}
    L(y) = & ~ \|{\cal S}r(x)+{\cal S}\Delta\|^2 \\
    = & ~ L(x)+2\langle{\cal S}\Delta,{\cal S}r(x)\rangle+\|{\cal S}\Delta\|^2 .
\end{align*}
Also,
\begin{align*}
\langle\nabla L(x),y-x\rangle = & ~ 2\langle\Phi(x,w){\cal S}^\top{\cal S}r(x),y-x\rangle \\
= & ~ 2\langle{\cal S}\Phi(x,w)^\top(y-x),{\cal S}r(x)\rangle .
\end{align*}
Subtracting the second display from the first gives
\begin{align*}
    & ~ L(y)-L(x)-\langle\nabla L(x),y-x\rangle \\
    = & ~ \|{\cal S}\Delta\|^2+2\langle{\cal S}(\Delta-\Phi(x,w)^\top(y-x)),{\cal S}r(x)\rangle .
\end{align*}

{\bf The first term.} By Assumption~\ref{ass:S_preserve_dist:app} with $u=G(y)$ and $v=G(x)$, and by Assumption~\ref{ass:gradient_lipschitz:app},
\begin{align*}
    \|{\cal S}\Delta\|^2\le\tau^2\|\Delta\|^2\le\tau^2\beta^2\|y-x\|^2 .
\end{align*}

{\bf The second term.} Moving ${\cal S}$ to the other side of the inner product and using Cauchy-Schwarz, Eq.~\eqref{eq:Phi_taylor_remainder:app} and $\|{\cal S}^\top\|=\|{\cal S}\|\le\gamma_2$,
\begin{align*}
    & ~ 2\langle{\cal S}(\Delta-\Phi(x,w)^\top(y-x)),{\cal S}r(x)\rangle \\
    = & ~ 2\langle\Delta-\Phi(x,w)^\top(y-x),{\cal S}^\top{\cal S}r(x)\rangle \\
    \le & ~ 2\cdot\frac{M}{2}\|y-x\|^2\cdot\gamma_2\|{\cal S}r(x)\| \\
    = & ~ \gamma_2 M L(x)^{1/2}\|y-x\|^2 .
\end{align*}

Adding the two bounds proves Eq.~\eqref{eq:LS_semi_smooth_bound:app}. If $x\in S_0$ then $L(x)^{1/2}\le L(x_0)^{1/2}$, so Eq.~\eqref{eq:LS_semi_smooth_bound:app} gives the inequality of Def.~\ref{def:semi_smooth:app} on $S_0$ with $a=0$, $b=B$ and $p=1/2$.
\end{proof}

\subsection{What Sketching Implies Non-critical Point}\label{sec:what_S_implies_non_critical_point:app}
\begin{lemma}\label{lem:sketching_non_critical_point:app}
If the sketching matrix ${\cal S}$ satisfies Assumption~\ref{ass:S_eigenvalue:app} and $F$ satisfies Assumption~\ref{ass:K_eigenvalues:app}, then $L(x):=\|\mathcal{S}(\nabla_w F(x;w)) -  \mathcal{S}(g)\|^2$ is $(2\theta_1\gamma_1,2\theta_2\gamma_2)$-non-critical-point.
\end{lemma}

\begin{proof}
Let $G(x):=\nabla_w F(x;w)$ and $r(x):=G(x)-g$, so $L(x)=\|{\cal S}r(x)\|^2$. Since $G$ has Jacobian $\Phi(x,w)^\top$ and ${\cal S}$ is linear, the chain rule gives
\begin{align*}
    \nabla L(x) = 2\Phi(x,w){\cal S}^\top{\cal S}r(x) = 2\Phi(x,w)v ,
\end{align*}
where $u:={\cal S}r(x)\in\R^{b_{\rm sketch}}$ and $v:={\cal S}^\top u\in\R^d$. Note $\|u\|^2=L(x)$. First, as in the proof of Lemma~\ref{lem:what_F_implies_L_no_critical_point:app}, Assumption~\ref{ass:K_eigenvalues:app} gives
\begin{align*}
    \|\Phi(x,w)v\|^2 = v^\top K(x)v\in[\theta_1^2\|v\|^2,\theta_2^2\|v\|^2] .
\end{align*}
Second, Assumption~\ref{ass:S_eigenvalue:app} gives
\begin{align*}
    \|v\|=\|{\cal S}^\top u\|\in[\gamma_1\|u\|,\gamma_2\|u\|] .
\end{align*}
Combining the two,
\begin{align*}
    \|\nabla L(x)\|^2 = & ~ 4\|\Phi(x,w)v\|^2 \ge 4\theta_1^2\|v\|^2 \\
    \ge & ~ 4\theta_1^2\gamma_1^2\|u\|^2 = (2\theta_1\gamma_1)^2 L(x),
\end{align*}
and
\begin{align*}
    \|\nabla L(x)\|^2 \le & ~ 4\theta_2^2\|v\|^2 \\
    \le & ~ 4\theta_2^2\gamma_2^2\|u\|^2 = (2\theta_2\gamma_2)^2 L(x).
\end{align*}
Hence we conclude that
\begin{align*}
    (2\theta_1\gamma_1)^2 L(x) \le \|\nabla L(x)\|^2 \le (2\theta_2\gamma_2)^2 L(x).
\end{align*}
\end{proof}
\section{Recovery from Several Rounds: Proofs}\label{sec:multi_round:app}

In this section, we prove the results of Section~\ref{sec:attack_multi_round}. Throughout, $\|\cdot\|$ is the Euclidean norm of vectors and the spectral norm of matrices, and $r(x):=H(x)-y$.

\begin{lemma}[Gradient descent on a least-squares loss]\label{lem:multi_gd}
Let $H:\R^m\rightarrow\R^n$ be differentiable with $\|DH(x)\|\le\beta_H$ for all $x$, and let $DH$ be $M_H$-Lipschitz. Let $y\in\R^n$, $\mathcal L(x):=\|H(x)-y\|^2$, $x_0\in\R^m$ and $S_0:=\{x:\mathcal L(x)\le\mathcal L(x_0)\}$. Suppose $\|DH(x)^\top r(x)\|\ge\mu\|r(x)\|$ for all $x\in S_0$, where $0<\mu\le\beta_H$. Let $B:=\beta_H^2+M_H\mathcal L(x_0)^{1/2}$, $0<\eta\le1/(2B)$ and $x_{k+1}=x_k-\eta\nabla\mathcal L(x_k)$. Then every $x_k$ lies in $S_0$, $\mathcal L(x_{k+1})\le(1-2\eta\mu^2)\mathcal L(x_k)$, and $0\le1-2\eta\mu^2<1$.
\end{lemma}
\begin{proof}
We have $\nabla\mathcal L(x)=2DH(x)^\top r(x)$. Expanding $\mathcal L(z)=\|r(x)+(H(z)-H(x))\|^2$ gives the identity
\begin{align*}
\mathcal L(z)-\mathcal L(x)-\langle\nabla\mathcal L(x),z-x\rangle=\|H(z)-H(x)\|^2+2\langle H(z)-H(x)-DH(x)(z-x),\,r(x)\rangle .
\end{align*}
By the mean value inequality, $\|H(z)-H(x)\|\le\beta_H\|z-x\|$. Since $DH$ is continuous, $H(z)-H(x)-DH(x)(z-x)=\int_0^1\big(DH(x+s(z-x))-DH(x)\big)(z-x)\,ds$, and its norm is at most $\int_0^1M_Hs\|z-x\|^2\,ds=\frac{M_H}{2}\|z-x\|^2$. With the Cauchy--Schwarz inequality, this gives
\begin{align*}
\mathcal L(z)\le\mathcal L(x)+\langle\nabla\mathcal L(x),z-x\rangle+\big(\beta_H^2+M_H\mathcal L(x)^{1/2}\big)\|z-x\|^2 .
\end{align*}
Suppose $x_k\in S_0$. Then the constant in this bound at $x=x_k$ is at most $B$, and $z=x_{k+1}$ gives
\begin{align*}
\mathcal L(x_{k+1})\le\mathcal L(x_k)-\eta(1-\eta B)\|\nabla\mathcal L(x_k)\|^2\le\mathcal L(x_k)-\frac{\eta}{2}\|\nabla\mathcal L(x_k)\|^2 .
\end{align*}
In particular $\mathcal L(x_{k+1})\le\mathcal L(x_k)\le\mathcal L(x_0)$, so $x_{k+1}\in S_0$, and induction on $k$ shows that every iterate lies in $S_0$. Since $\|\nabla\mathcal L(x_k)\|^2=4\|DH(x_k)^\top r(x_k)\|^2\ge4\mu^2\mathcal L(x_k)$, we get $\mathcal L(x_{k+1})\le(1-2\eta\mu^2)\mathcal L(x_k)$. Finally, $0<2\eta\mu^2\le\mu^2/B\le\mu^2/\beta_H^2\le1$.
\end{proof}

\begin{lemma}\label{lem:multi_blocks}
Under Assumption~\ref{ass:multi_round}, $\|J(x)\|\le\Gamma\beta$ and $\|J(x)-J(z)\|\le\Gamma M\|x-z\|$ for all $x,z\in\R^m$. The same holds after deleting rows of the $\bar R_j$, with $\Gamma$ computed from the remaining rows.
\end{lemma}
\begin{proof}
$\Phi(x,w^{t_j})^\top$ is the Jacobian of the $\beta$-Lipschitz map $x\mapsto\nabla_wF(w^{t_j},x)$, so $\|\Phi(x,w^{t_j})\|\le\beta$. For $v\in\R^m$,
\begin{align*}
\|J(x)v\|^2=\sum_{j=1}^T\|\bar R_j\Phi(x,w^{t_j})^\top v\|^2\le\sum_{j=1}^T\|\bar R_j\|^2\beta^2\|v\|^2=\Gamma^2\beta^2\|v\|^2 ,
\end{align*}
and in the same way $\|(J(x)-J(z))v\|^2\le\sum_j\|\bar R_j\|^2M^2\|x-z\|^2\|v\|^2$. The computation is the same for any subset of rows.
\end{proof}

\begin{proof}[Proof of Theorem~\ref{thm:multi_square}]
By Lemma~\ref{lem:multi_blocks}, Lemma~\ref{lem:multi_gd} applies to $H$ with $\beta_H=\Gamma\beta$ and $M_H=\Gamma M$, which gives the $B$ of the theorem. Since $J(x)$ is square, $\|J(x)^\top r\|\ge\sigma_{\min}(J(x))\|r\|\ge\mu\|r\|$ for every $r\in\R^m$. Also $\mu\le\sigma_{\min}(J(x))\le\|J(x)\|\le\Gamma\beta$. So Lemma~\ref{lem:multi_gd} gives the contraction and $0\le1-2\eta\mu^2<1$.

$H$ is continuously differentiable, because $J$ is Lipschitz. For every $x$, $J(x)$ is invertible with $\|J(x)^{-1}\|=1/\sigma_{\min}(J(x))\le1/\mu$. By Hadamard's global inverse function theorem, in the form of Hadamard and L\'evy~\cite{p74} used after Corollary~\ref{cor:attack_rate}, $H$ is a bijection of $\R^m$ onto $\R^m$ with a continuously differentiable inverse, and $DH^{-1}(H(x))=J(x)^{-1}$. For $u,v\in\R^m$, $H^{-1}(u)-H^{-1}(v)=\int_0^1DH^{-1}(v+s(u-v))(u-v)\,ds$, so $\|H^{-1}(u)-H^{-1}(v)\|\le\|u-v\|/\mu$. Since $y=H(\tilde x)$,
\begin{align*}
\|x_k-\tilde x\|=\|H^{-1}(H(x_k))-H^{-1}(y)\|\le\frac{\|H(x_k)-y\|}{\mu}=\frac{\mathcal L(x_k)^{1/2}}{\mu}\le\frac{(1-2\eta\mu^2)^{k/2}\mathcal L(x_0)^{1/2}}{\mu}.
\end{align*}
Injectivity of $H$ shows that $H(x)=y$ only for $x=\tilde x$. If $n>m$, the kept rows define a map of the same form with $m$ coordinates, whose observed value consists of the corresponding coordinates of $y$. Lemma~\ref{lem:multi_blocks} covers this map, and the argument above applies to it verbatim.
\end{proof}

\begin{proof}[Proof of Theorem~\ref{thm:multi_flat}]
Let $\mathcal E(x)$ be the vector with blocks $\bar R_je_j(x)$ and $\mathbf q$ the vector with blocks $\bar R_jq_j$. Then $H(x)=\mathcal A\psi(x)+\mathbf q+\mathcal E(x)$ and $J(x)=\mathcal AD\psi(x)+D\mathcal E(x)$. As in Lemma~\ref{lem:multi_blocks}, $\|D\mathcal E(x)\|\le\Gamma\varepsilon_0=\varepsilon$, and $D\mathcal E$ is $\Gamma M_e$-Lipschitz. Hence $\|J(x)\|\le\ell\|\mathcal A\|+\varepsilon$, and $J$ is $(\|\mathcal A\|M_\psi+\Gamma M_e)$-Lipschitz. So Lemma~\ref{lem:multi_gd} applies with the $B$ of (b), and in case (a), where $\varepsilon=M_e=0$ may be taken, with the $B$ of (a).

$\psi$ is continuously differentiable with $\|D\psi(x)^{-1}\|\le1/\nu$. So by the Hadamard--L\'evy theorem, as above, $\psi$ is a bijection of $\R^m$ with a $(1/\nu)$-Lipschitz inverse, and $\|\psi(x)-\psi(\tilde x)\|\ge\nu\|x-\tilde x\|$. Also $\nu\le\|D\psi(x)\|\le\ell$ and $\alpha\le\|\mathcal A\|$, so $\alpha\nu\le\ell\|\mathcal A\|$.

Fix $x$, and put $u:=\psi(x)-\psi(\tilde x)$ and $e:=\mathcal E(x)-\mathcal E(\tilde x)$, so that $r(x)=\mathcal Au+e$. We use two facts about $\mathcal A$. Write a thin singular value decomposition $\mathcal A=U\Sigma V^\top$, where $\Sigma\in\R^{m\times m}$ is diagonal with entries at least $\alpha$ and $V$ is orthogonal. Then $\|\mathcal A^\top\mathcal Au\|=\|\Sigma^2V^\top u\|\ge\alpha\|\Sigma V^\top u\|=\alpha\|\mathcal Au\|$. Also $\|\mathcal Au\|\ge\alpha\|u\|\ge\alpha\nu\|x-\tilde x\|$.

(a) Here $e=0$ and $J(x)^\top r(x)=D\psi(x)^\top\mathcal A^\top\mathcal Au$. Since $D\psi(x)$ is square with smallest singular value at least $\nu$,
\begin{align*}
\|J(x)^\top r(x)\|\ge\nu\|\mathcal A^\top\mathcal Au\|\ge\alpha\nu\|\mathcal Au\|=\alpha\nu\|r(x)\| .
\end{align*}
Lemma~\ref{lem:multi_gd} with $\mu=\alpha\nu\le\ell\|\mathcal A\|$ gives the contraction and the range of the factor. For recovery, $\|r(x)\|=\|\mathcal Au\|\ge\alpha\nu\|x-\tilde x\|$, so $\|x_k-\tilde x\|\le\mathcal L(x_k)^{1/2}/(\alpha\nu)$, and $r(x)=0$ forces $x=\tilde x$.

(b) Let $\delta_1:=\varepsilon\ell\|\mathcal A\|/(\nu^2\alpha^2)\le1/4$. Since $\alpha\nu\le\ell\|\mathcal A\|$, also $\varepsilon/(\alpha\nu)\le\alpha\nu/(4\ell\|\mathcal A\|)\le1/4$. By the mean value inequality,
\begin{align*}
\|e\|\le\varepsilon\|x-\tilde x\|\le\frac{\varepsilon}{\nu}\|u\|\le\frac{\varepsilon}{\alpha\nu}\|\mathcal Au\|\le\frac14\|\mathcal Au\| .
\end{align*}
With $D\psi=D\psi(x)$ and $D\mathcal E=D\mathcal E(x)$,
\begin{align*}
J(x)^\top r(x)=D\psi^\top\mathcal A^\top\mathcal Au+D\psi^\top\mathcal A^\top e+D\mathcal E^\top\mathcal Au+D\mathcal E^\top e .
\end{align*}
The first term has norm at least $\alpha\nu\|\mathcal Au\|$, as in (a). The other three have norms at most
\begin{align*}
\ell\|\mathcal A\|\frac{\varepsilon}{\alpha\nu}\|\mathcal Au\|=\delta_1\,\alpha\nu\|\mathcal Au\|\le\frac14\alpha\nu\|\mathcal Au\|,\quad
\varepsilon\|\mathcal Au\|\le\frac14\alpha\nu\|\mathcal Au\|,\quad
\varepsilon\cdot\frac{\varepsilon}{\alpha\nu}\|\mathcal Au\|\le\frac1{16}\alpha\nu\|\mathcal Au\| .
\end{align*}
Hence $\|J(x)^\top r(x)\|\ge(1-\frac14-\frac14-\frac1{16})\alpha\nu\|\mathcal Au\|=\frac7{16}\alpha\nu\|\mathcal Au\|$. Since $\|r(x)\|\le\|\mathcal Au\|+\|e\|\le\frac54\|\mathcal Au\|$, we get $\|J(x)^\top r(x)\|\ge\frac7{20}\alpha\nu\|r(x)\|\ge\frac{\alpha\nu}{3}\|r(x)\|$. Lemma~\ref{lem:multi_gd} with $\mu=\alpha\nu/3\le\ell\|\mathcal A\|+\varepsilon$ gives the contraction and the range of the factor. For recovery, $\|r(x)\|\ge\|\mathcal Au\|-\|e\|\ge\frac34\|\mathcal Au\|\ge\frac34\alpha\nu\|x-\tilde x\|$. So $\|x_k-\tilde x\|\le\frac{4}{3\alpha\nu}\mathcal L(x_k)^{1/2}$, and $r(x)=0$ forces $x=\tilde x$.
\end{proof}

\begin{proof}[Proof of Proposition~\ref{prop:multi_random} and of the claims after it]
Let $S_j:=\bar R_j^\top\bar R_j$, $X_j:=A_j^\top S_jA_j\succeq0$ and $P_j:=A_j^\top A_j$. Then $\mathcal A^\top\mathcal A=\sum_jX_j$ and $\mathcal K_T=\sum_jP_j$. Let $Z_j:=P_j-X_j$ and $Y_k:=\sum_{j\le k}Z_j$, with $Y_0:=0$.

\emph{Martingale.} $Z_j$ is $\mathcal G_j$-measurable, because $A_j$ is $\mathcal G_{j-1}$-measurable and $\bar R_j$ is $\mathcal G_j$-measurable. Since $\bar R_j$ is independent of $\mathcal G_{j-1}$, $\E[X_j\mid\mathcal G_{j-1}]=A_j^\top\E[S_j]A_j=P_j$. All terms are bounded, since $\|X_j\|\le\rho a^2$. So $(Y_k)$ is a matrix martingale with respect to $(\mathcal G_k)$, with values in symmetric $m\times m$ matrices and differences $Z_j$.

\emph{Bounds.} Since $X_j\succeq0$, $\lambda_{\max}(Z_j)\le\|P_j\|\le a^2$. For $v\in\R^m$,
\begin{align*}
v^\top X_j^2v=\|A_j^\top S_jA_jv\|^2\le a^2\,v^\top A_j^\top S_j^2A_jv\le\rho a^2\,v^\top X_jv ,
\end{align*}
because $S_j^2\preceq\|S_j\|S_j\preceq\rho S_j$. So $X_j^2\preceq\rho a^2X_j$. Expanding $Z_j^2$ and using $\E[X_j\mid\mathcal G_{j-1}]=P_j$ gives $\E[Z_j^2\mid\mathcal G_{j-1}]=\E[X_j^2\mid\mathcal G_{j-1}]-P_j^2\preceq\rho a^2P_j$. Hence the predictable quadratic variation satisfies $W_T:=\sum_j\E[Z_j^2\mid\mathcal G_{j-1}]\preceq\rho a^2\mathcal K_T$.

\emph{Tail.} On the event $\mathfrak E$ of the proposition, Weyl's inequality applied to $\mathcal K_T=\mathcal A^\top\mathcal A+Y_T$ gives
\begin{align*}
\lambda_{\max}(Y_T)\ge\lambda_{\min}(\mathcal K_T)-\lambda_{\min}(\mathcal A^\top\mathcal A)>\kappa-\kappa/2=\kappa/2 ,
\end{align*}
and $\|W_T\|\le\rho a^2\bar K$. By the matrix Freedman inequality (\cite{t11freedman}, Theorem~1.2), with $t=\kappa/2$, $R=a^2$ and $\sigma^2=\rho a^2\bar K$,
\begin{align*}
\Pr[\mathfrak E]\le\Pr\big[\exists k:\lambda_{\max}(Y_k)\ge\kappa/2\text{ and }\|W_k\|\le\rho a^2\bar K\big]\le m\exp\Big(-\frac{\kappa^2/8}{\rho a^2\bar K+a^2\kappa/6}\Big).
\end{align*}
We have $1=\|\E[S_j]\|\le\E\|S_j\|\le\rho$, and $\kappa\le\bar K$. So $a^2\kappa/6\le\rho a^2\bar K/6$, and the exponent is at least $\frac{3\kappa^2}{28\rho a^2\bar K}\ge\frac{\kappa^2}{10\rho a^2\bar K}$.

\emph{Algorithm~\ref{alg:fed_learn_kstep_dp}.} Let $\mathcal G_{j-1}:=\sigma(w^0,R_2,\ldots,R_{t_j})$ (that is, $\sigma(w^0)$ when $t_j=1$) for $j\in[T]$ and $\mathcal G_T:=\sigma(w^0,R_2,\ldots,R_{t_T+1})$. These $\sigma$-algebras increase with $j$. With $K=1$ and no noise, Algorithm~\ref{alg:fed_learn_kstep_dp} sets $w^1=w^0$ and $w^{t+1}=w^t+R_{t+1}^\top\Delta\wt w^t$, where $\Delta\wt w^t$ is computed from $w^t$, $R_{t+1}$ and the fixed data. So the weights $w^{t_j}$ are computed from $w^0$, $R_2,\ldots,R_{t_j}$ and the fixed data, hence they are $\mathcal G_{j-1}$-measurable, and so is $A_j=A(w^{t_j})$. Next, $\bar R_j=R_{t_j+1}$ is $\mathcal G_j$-measurable since $t_{j+1}\ge t_j+1$. It is independent of $\mathcal G_{j-1}$, since the $R_t$ are independent of each other and of $w^0$.

\emph{Sketches.} Write $b:=b_{\rm sketch}$, and write $\mathsf H$ for the Walsh--Hadamard matrix of Definition~\ref{def:srht}, since $H$ denotes the stacked map in this section. For SRHT, $R^\top R=\frac{d}{b}D\mathsf H^\top S^\top S\mathsf HD$. For uniform sampling, $R^\top R=\frac{d}{b}DS^\top SD=\frac{d}{b}S^\top S$, because $S^\top S$ and $D$ are diagonal and $D^2=I_d$. Here $S^\top S$ is the diagonal $0/1$ matrix that selects $b$ of the $d$ coordinates, chosen uniformly without replacement. So $S^\top S$ is an orthogonal projection of rank $b$ with $\E[S^\top S]=\frac{b}{d}I_d$, since each coordinate is selected with probability $b/d$. Since $D$ and $\mathsf H$ are orthogonal and $S$ is independent of $D$, in both cases $\E[R^\top R]=I_d$ and $\|R^\top R\|=d/b$.

\emph{Round counts.} If $\kappa I_m\preceq\mathcal K_T$ holds for some outcome, then $\kappa\le\lambda_{\min}(\mathcal K_T)\le{\rm tr}(\mathcal K_T)/m$, and ${\rm tr}(A_j^\top A_j)\le\rank(A_j)\|A_j\|^2\le\min(d,m)a^2$ gives $\kappa\le T\min(d,m)a^2/m$. Since $\kappa\le\bar K$, the condition $\kappa^2\ge10\rho a^2\bar K\ln(m/\delta)$ gives $\kappa\ge10\rho a^2\ln(m/\delta)$. With $\rho=d/b$, the two bounds on $\kappa$ give $Tb\ge10\,m\ln(m/\delta)$ when $d\le m$, and $Tb\ge10\,d\ln(m/\delta)$ when $d>m$. When $\kappa=\bar K=Tda^2/m$, the condition is $\kappa\ge10\rho a^2\ln(m/\delta)$, that is, $Tb\ge10\,m\ln(m/\delta)$. With $\kappa=Tda^2/(c_1m)$ and $\bar K=c_2Tda^2/m$ for $c_1,c_2\ge1$, which satisfy $0<\kappa\le\bar K$, the condition reads $T^2d^2a^4/(c_1^2m^2)\ge10c_2(d/b)(Tda^4/m)\ln(m/\delta)$, that is, $Tb\ge10c_1^2c_2\,m\ln(m/\delta)$. Finally, let $d=m$, $\sigma_{\min}(A(w))\ge\theta$ and $\|A(w)\|\le a$ for all $w$. Then $\lambda_{\min}(\mathcal K_T)\ge\sum_j\lambda_{\min}(A_j^\top A_j)\ge T\theta^2$ and $\|\mathcal K_T\|\le Ta^2$ surely, and the condition with $\kappa=T\theta^2$ and $\bar K=Ta^2$ reads $T^2\theta^4\ge10\rho a^4T\ln(m/\delta)$, that is, $T\ge10(d/b)(a/\theta)^4\ln(m/\delta)$. The event that $\mathcal K_T$ violates its bounds is then empty, so $\sigma_{\min}(\mathcal A)^2\ge T\theta^2/2$ with probability at least $1-\delta$.
\end{proof}

\begin{proof}[Proof of Proposition~\ref{prop:multi_necessary}]
(i) Let $\mathbf m(x)\in\R^{Tb_{\rm sketch}}$ be the messages that client $c$ sends in rounds $1,\ldots,T$ when its data are $x$. We have $w^1=w^0$, each local step $u\mapsto u-\eta_{\rm local}\nabla_wF(u,x)$ is continuous in $(u,x)$, and $w^{t+1}=w^t+R_{t+1}^\top\Delta\wt w^t$ is computed from $w^t$ and the messages of round $t$. By induction on $t$, the weights and all messages of round $t$ are continuous functions of $x$. The same induction shows that the whole view is a function of $\mathbf m(x)$. Indeed, the messages of the other clients in round $t$ are functions of $w^t$ and of their fixed data, and $w^{t+1}$ is a function of $w^t$ and the messages of round $t$, among which those of client $c$ are entries of $\mathbf m(x)$. Suppose $\mathbf m$ were injective, and let $\iota(z):=(z,0)\in\R^m$. Then $\iota\circ\mathbf m:\R^m\rightarrow\R^m$ would be continuous and injective, so by Brouwer's invariance of domain its image would be open in $\R^m$. But the image is nonempty and lies in $\R^{Tb_{\rm sketch}}\times\{0\}$, which has empty interior when $Tb_{\rm sketch}<m$. So $\mathbf m(x)=\mathbf m(x')$ for some $x\ne x'$, and $x$ and $x'$ give the same view.

(ii) Write $\Phi(x)$ for $\Phi(x,w)$. Since $\frac{d}{ds}\nabla_wF(w,sx)=\Phi(sx)^\top x$, we have $\nabla_wF(w,x)=\nabla_wF(w,0)+g_0(x)$ with $g_0(x):=\int_0^1\Phi(sx)^\top x\,ds$, which does not depend on $w$. Hence $\nabla f_c(u)=\nabla_wF(u,0)+g_0(\tilde x)$ for every $u$. The induction of (i) then shows that the view of the first $T$ rounds is a function of $g_0(\tilde x)\in\R^d$. The map $g_0$ is continuous, so for $d<m$ it is not injective, by the argument of (i).

(iii) Here $\nabla_wF(w,x)=\sin(x-w)$ and $\Phi(x,w)=\cos(x-w)$, both $1$-Lipschitz in $x$. So $H(x)=(\sin x,-\cos x)$ and $J(x)=(\cos x,\sin x)^\top$, and $\sigma_{\min}(J(x))=1$. $H$ is $2\pi$-periodic, and $\mathcal L(x)=(\sin x-\sin\tilde x)^2+(\cos x-\cos\tilde x)^2=2-2\cos(x-\tilde x)$. Its derivative $2\sin(x-\tilde x)$ vanishes at $x=\tilde x+\pi$, where $\mathcal L=4$.
\end{proof}

\begin{proof}[Proof of Remark~\ref{rem:multi_one_round}]
With $b_{\rm sketch}=d=m$, $R$ is square, and $\sigma_{\min}(R)=\sigma_{\min}(R^\top)\ge\gamma_1$. The eigenvalues of $K(x)=\Phi(x,w)^\top\Phi(x,w)$ are at least $\theta_1^2$, so $\sigma_{\min}(\Phi(x,w)^\top)\ge\theta_1$, and therefore $\sigma_{\min}(R\Phi(x,w)^\top)\ge\gamma_1\theta_1$. The conditions of Theorem~\ref{thm:F_L_smooth_criticl} give Assumption~\ref{ass:multi_round}. With $T=1$ we have $\mathcal L=L_R$ and $\Gamma=\|R\|$, so the $B$ of Theorem~\ref{thm:multi_square} is at most the $B$ of Corollary~\ref{cor:sketch_attack_rate}, because $\|R\|\le\tau$ and $\|R\|\le\gamma_2$.
\end{proof}

\begin{proof}[Proof of Remark~\ref{rem:multi_gaussian}]
(1) For fixed $A_j$, $\det\mathcal A$ is a polynomial in the entries of the kept rows of $\bar R_1,\ldots,\bar R_T$, which have a joint density. If the polynomial is not identically zero, its zero set has Lebesgue measure zero, and $\mathcal A$ is invertible almost surely. Otherwise $\mathcal A$ is always singular. The polynomial is not identically zero if and only if some choice of the $\bar R_j$ makes $\mathcal A$ invertible. Given such a choice, let $V_j:={\rm row}(\bar R_jA_j)\subseteq{\rm row}(A_j)$. Then $\dim V_j\le b_j$ and $\sum_jV_j={\rm row}(\mathcal A)=\R^m$, and $\sum_jb_j=m$ forces $\dim V_j=b_j$ and a direct sum. Conversely, given the $V_j$, pick a basis $v_{j,1},\ldots,v_{j,b_j}$ of $V_j$ and vectors $r_{j,i}$ with $A_j^\top r_{j,i}=v_{j,i}$, which exist since $V_j\subseteq{\rm row}(A_j)$. Use the $r_{j,i}^\top$ as the rows of $\bar R_j$. Then the rows of $\mathcal A$ form a basis of $\R^m$.

(2) Process the kept rows of $\mathcal A$ one at a time, round by round. A row of $\bar R_j$ is $r^\top$ with $r\sim\N(0,I_d/b_{\rm sketch})$, and the corresponding row of $\mathcal A$ is $(A_j^\top r)^\top$. Given $\mathcal G_{j-1}$ (as in the proof of Proposition~\ref{prop:multi_random}) and the earlier rows of $\bar R_j$, the vector $r$ is still $\N(0,I_d/b_{\rm sketch})$ and $A_j$ is fixed. So $A_j^\top r\sim\N(0,A_j^\top A_j/b_{\rm sketch})$, which is nondegenerate on $\R^m$ because $A_j$ is invertible. The span of the rows processed before is a fixed subspace of dimension less than $m$, and a nondegenerate Gaussian vector lies in it with probability zero. Hence, almost surely, each new row lies outside the span of the earlier ones, and the $m$ kept rows are linearly independent.
\end{proof}

\end{document}